%% file: main.tex
\documentclass[acmlarge,screen,nonacm]{acmart}
\AtBeginDocument{%
  \providecommand\BibTeX{{%
    \normalfont B\kern-0.5em{\scshape i\kern-0.25em b}\kern-0.8em\TeX}}}

\usepackage{pdfpages}
\usepackage{listings} 
\usepackage{xcolor} 
\usepackage{xspace} 
\usepackage{graphicx} 
\usepackage{caption} 
\usepackage{mdframed} 
\usepackage{lipsum} 
\usepackage{subcaption}
\usepackage{ifthen}
\usepackage{wrapfig}
\usepackage{enumitem}
\usepackage{placeins}
\usepackage{afterpage}
\usepackage{subfiles}
\usepackage{adjustbox}
\usepackage{multirow}

\begin{document}

\include{macros}

\include{snippet-styling2}

\title{How Can Large Language Models Help Humans in Design And Manufacturing?}



\newcommand{\mitAffil}{
 \institution{MIT}
 \streetaddress{77 Massachusetts Ave}
 \city{Cambridge}
 \state{MA}
 \postcode{02139}
 \country{USA}
}

\newcommand{\harvardAffil}{
 \institution{Harvard University}
 \streetaddress{Massachusetts Hall}
 \city{Cambridge}
 \state{MA}
 \postcode{02138}
 \country{USA}
}

\newcommand{\uwAffil}{
 \institution{University of Washington}
 \streetaddress{1410 NE Campus Parkway}
 \city{Seattle}
 \state{WA}
 \postcode{98195}
 \country{USA}
}

\author{Liane Makatura}
\orcid{0000-0003-4804-2173}
\email{makatura@mit.edu}
%
\author{Michael Foshey}
\orcid{0000-0002-9047-8387}
\email{mfoshey@mit.edu}
%
\author{Bohan Wang}
\orcid{0000-0003-1439-1455}
\email{bohanw@mit.edu}
\affiliation{\mitAffil}

\author{Felix H\"{a}hnlein}
\orcid{0000-0002-3484-4004}
\email{fhahnlei@cs.washington.edu}
\affiliation{\uwAffil}

\author{Pingchuan Ma}
\orcid{0000-0002-5698-9503}
\email{pcma@csail.mit.edu}
%
\author{Bolei Deng}
\orcid{0000-0003-2589-2837}
\email{boleiden@mit.edu}
%
\author{Megan Tjandrasuwita}
\orcid{0000-0002-4950-8679}
\email{megantj@mit.edu}
\affiliation{\mitAffil}

\author{Andrew Spielberg}
\orcid{0000-0002-6937-6204}
\email{aespielberg@seas.harvard.edu}
\affiliation{\harvardAffil}

\author{Crystal Elaine Owens}
\orcid{0000-0002-2433-7025}
\email{crystalo@mit.edu}
%
\author{Peter Yichen Chen}
\orcid{0000-0003-1919-5437}
\email{pyc@csail.mit.edu}
%
\author{Allan Zhao}
\orcid{0000-0002-9162-6716}
\email{azhao@csail.mit.edu}
\affiliation{\mitAffil}

\author{Amy Zhu} 
\orcid{0000-0001-5766-7090}
\email{amyzhu@cs.washington.edu}
\affiliation{\uwAffil}

\author{Wil J Norton}
\orcid{0000-0001-9465-9751}
\email{wn1024@mit.edu}
%
\author{Edward Gu}
\orcid{0009-0006-0641-7757}
\email{egu@mit.edu}
%
\author{Joshua Jacob}
\orcid{0009-0001-3165-1012}
\email{jmjacob@csail.mit.edu}
%
\author{Yifei Li}
\orcid{0000-0002-3770-0575}
\email{liyifei@csail.mit.edu}
\affiliation{\mitAffil}

\author{Adriana Schulz}
\orcid{0000-0002-2464-0876}
\email{adriana@cs.washington.edu}
\affiliation{\uwAffil}

\author{Wojciech Matusik}
\orcid{0000-0003-0212-5643}
\email{wojciech@csail.mit.edu}
\affiliation{\mitAffil}

\renewcommand{\shortauthors}{Makatura et al.}

\begin{abstract}
  The advancement of Large Language Models (\llms), including \gpt, provides exciting new opportunities for generative design. We investigate the application of this tool across the entire design and manufacturing workflow. Specifically, we scrutinize the utility of \llms in tasks such as: converting a text-based prompt into a design specification, transforming a design into manufacturing instructions, producing a design space and design variations, computing the performance of a design, and searching for designs predicated on performance. Through a series of examples, we highlight both the benefits and the limitations of the current \llms. By exposing these limitations, we aspire to catalyze the continued improvement and progression of these models.
\end{abstract}





\maketitle

\input{sections/introduction}
\input{sections/related}

\input{sections/overview}

\input{sections/allGraphArrows}

\input{sections/results}
\input{sections/discussion}

\begin{acks}
This material is based upon work supported in part by Defense Advanced Research Projects Agency (DARPA) Grant No. FA8750-20-C-0075.
\end{acks}

\bibliographystyle{ACM-Reference-Format}
\bibliography{sources}

\appendix
\include{appendix}

\end{document}

%% file: macros.tex
\lstset{
basicstyle=\small\ttfamily,
columns=flexible,
breaklines=true
}

\definecolor{BrewerOrange}{HTML}{d95f02}
\definecolor{BrewerGreen}{HTML}{1b9e77}
\definecolor{BrewerPurple}{HTML}{7570b3}
\definecolor{BrewerPink}{HTML}{e7298a}

\newcommand{\question}[1]{\textcolor{red}{(Question: #1)}}
\newcommand{\todo}[1]{\textcolor{blue}{(TODO: #1)}}
\newcommand{\toclarify}[2]{\textcolor{orange}{#1 (CLARIFY: #2)}}
\newcommand{\team}[1]{\textcolor{red}{(Lead: \textbf{#1})}}
\newcommand{\ignore}[1]{}

\newcommand{\wojciech}[1]{\textcolor{cyan}{(WM: #1)}}
\newcommand{\FH}[1]{\textcolor{purple}{(FH: #1)}}
\newcommand{\liane}[1]{\textcolor{teal}{(LM: #1)}}
\newcommand{\bolei}[1]{\textcolor{blue}{(BD: #1)}}
\newcommand{\yifei}[1]{\textcolor{orange}{(YL: #1)}}
\newcommand{\adriana}[1]{\textcolor{violet}{(AS: #1)}}
\newcommand{\amy}[1]{\textcolor{green}{(AZ: #1)}}
\newcommand{\egu}[1]{\textcolor{lime}{(EG: #1)}}
\newcommand{\ceo}[1]{\textcolor{brown}{(CO: #1)}}
\newcommand{\mt}[1]{\textcolor{magenta}{(MT: #1)}}

\newcommand{\gpt}{GPT-4\xspace} 
\newcommand{\llm}{LLM\xspace}
\newcommand{\llms}{LLMs\xspace}
\newcommand{\llmgraph}{LLM graph\xspace}
\newcommand{\jscad}{OpenJSCAD\xspace}
\newcommand{\scad}{OpenSCAD\xspace}
\newcommand{\cdam}{CDaM\xspace}

\newcommand{\sref}[1]{Section~\ref{#1}} 
\newcommand{\dref}[1]{Definition~\ref{#1}} 
\newcommand{\fref}[1]{Figure~\ref{#1}} 
\newcommand{\aref}[1]{Algorithm~\ref{#1}} 
\newcommand{\eref}[1]{Equation~\ref{#1}} 
\newcommand{\lref}[1]{Lemma~\ref{#1}} 
\newcommand{\pref}[1]{Proposition~\ref{#1}} 
\newcommand{\coref}[1]{Corollary~\ref{#1}} 
\newcommand{\tref}[1]{Table~\ref{#1}} 
\newcommand{\appref}[1]{Appendix~\ref{#1}} 
\newtheorem{remark}{Remark}

\makeatletter
\DeclareRobustCommand\onedot{\futurelet\@let@token\@onedot}
\def\@onedot{\ifx\@let@token.\else.\null\fi\xspace}

\def\eg{e.g\onedot} \def\Eg{E.g\onedot}
\def\ie{i.e\onedot} \def\Ie{I.e\onedot}
\def\cf{c.f\onedot} \def\Cf{C.f\onedot}
\def\etc{etc\onedot} \def\vs{vs\onedot}
\def\wrt{w.r.t\onedot} \def\dof{d.o.f\onedot} \def\WLOG{w.l.o.g\onedot}
\def\etal{et al\onedot}
\makeatother

%% file: snippet-styling2.tex
\definecolor{codegreen}{rgb}{0,0.6,0}
\definecolor{codeblue}{rgb}{.11,.56,1}
\definecolor{codegray}{rgb}{0.5,0.5,0.5}
\definecolor{codepurple}{rgb}{0.58,0,0.82}

\definecolor{codeKeyword}{RGB}{211	54	130}
\definecolor{codeComment}{RGB}{42	161	152}
\definecolor{codeOmitted}{RGB}{108	113	196}
\definecolor{codeNumbers}{rgb}{0.5,0.5,0.5}
\definecolor{codeString}{RGB}{128, 161, 16}

\definecolor{textusercolor}{RGB}{40 20 10}
\definecolor{textgptcolor}{RGB}{62, 65, 115}

\definecolor{codebackcolour}{RGB}{	253	246	227}
\definecolor{backuserprompt}{RGB}{
253, 250, 250}

\definecolor{backgptresponse}{RGB}{226 228 255}

\newcommand{\gpticon}{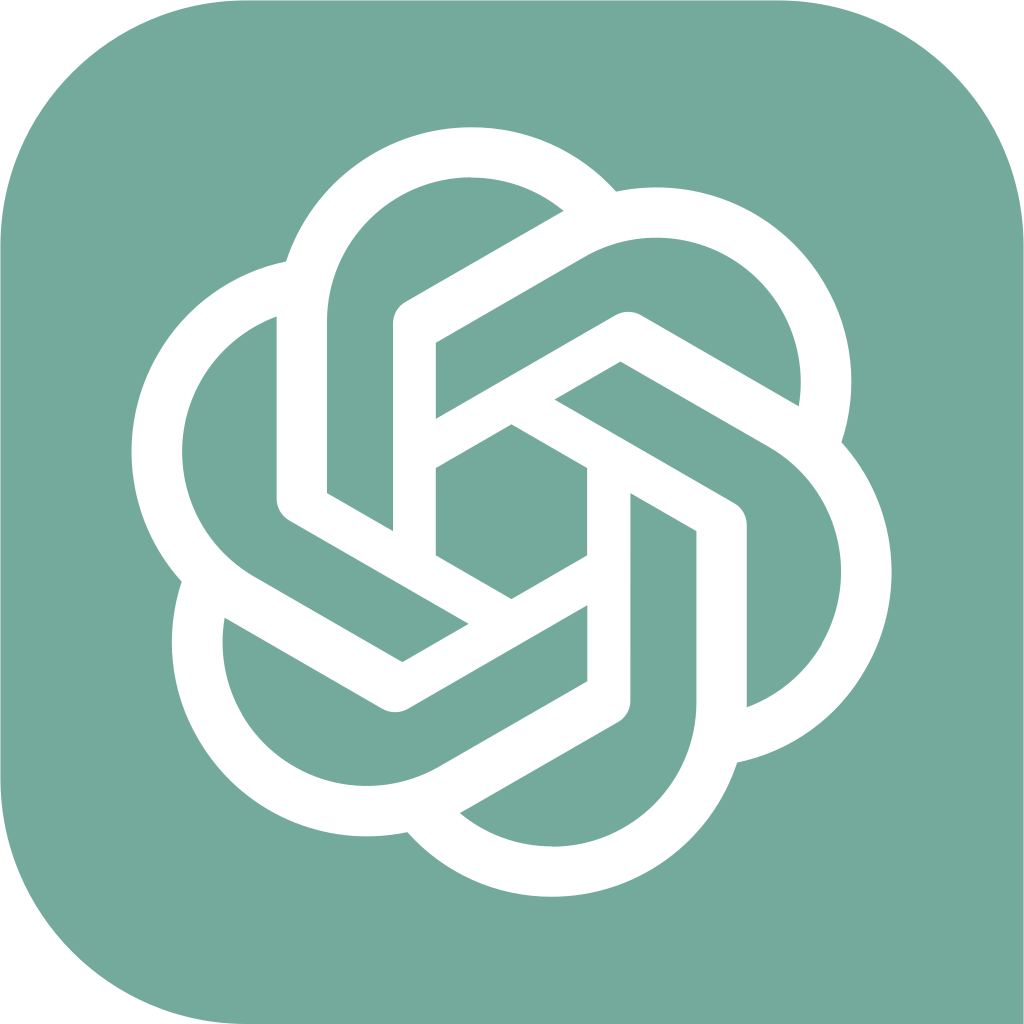}
\newcommand{\usericon}{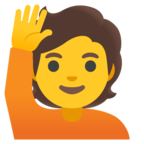}

\lstdefinelanguage{JavaScript}{
  keywords={typeof, new, true, false, catch, function, return, null, catch, switch, var, if, in, while, do, else, case, break, const},
  ndkeywords={class, export, boolean, throw, implements, import, this, require},
  sensitive=false,
  comment=[l]{//},
  morecomment=[s]{/*}{*/},
  morestring=[b]',
  morestring=[b]"
}

\lstdefinestyle{codestyle}{
    commentstyle=\color{codeComment},
    keywordstyle=\color{codeKeyword},
    numberstyle=\tiny\color{codeNumbers},
    stringstyle=\color{codeString},
    basicstyle=\linespread{0.85}\footnotesize,
    columns=flexible,
    breakatwhitespace=false,         
    breaklines=true,                 
    captionpos=b,                    
    showspaces=false,
    showstringspaces=false,
    showtabs=false,
    tabsize=2,
    escapeinside={\$}{\$},
}

\surroundwithmdframed[
  hidealllines=true,
  backgroundcolor=codebackcolour,
  innerleftmargin=0pt,
  innertopmargin=0pt,
  innerbottommargin=0pt]{gptcodeblock}

\newcommand\colboxcolor{codeComment} 
\newsavebox{\savedcolorbox}
\newenvironment{colbox}[2]
  {\renewcommand\colboxcolor{#1}%
   \begin{lrbox}{\savedcolorbox}%
    \begin{minipage}{\dimexpr\columnwidth-2\fboxsep\relax}

   \footnotesize
   \bgroup\color{#2}
   }
  {\egroup\end{minipage}\end{lrbox}%
   \begin{center}
   \colorbox{\colboxcolor}{\usebox{\savedcolorbox}}
   \end{center}
}

\newsavebox{\savedfigurebox}
\newenvironment{blurbwithfig}[5]
{
    \newcommand{\figurewidth}{#1}
    \newcommand{\iconwidth}{0.025\textwidth}
    \newcommand{\blurbwidth}{0.982\textwidth - \figurewidth - \iconwidth}
    \newcommand{\imagetoshow}{#2}
    \newcommand{\backgroundcolor}{#3}
    \newcommand{\boxtextcolor}{#4}
    \newcommand{\icontoshow}{#5}

    \begin{lrbox}{\savedfigurebox}%
    \begin{minipage}[t]{\figurewidth}
        \vspace{3pt}
        \ifthenelse{\equal{\imagetoshow}{}}{}{\includegraphics[width=\textwidth]{\imagetoshow}}
    \end{minipage}\end{lrbox}%

    \noindent
    \begin{minipage}[t]{\iconwidth}
    \vspace{2pt}
    \centering
    \includegraphics[width=\textwidth]{\icontoshow}
    \end{minipage}
    %
    \noindent
    \begin{minipage}[t]{\blurbwidth}
    \vspace{0pt}
    \begin{colbox}{\backgroundcolor}{\boxtextcolor}
}
{
    \end{colbox}
    \end{minipage}
    \hfill
    \usebox{\savedfigurebox}
}

\newenvironment{userprompt}[2]
{
    \begin{blurbwithfig}{#1}{#2}{backuserprompt}{textusercolor}{\usericon}
}
{
    \end{blurbwithfig}
}

\newenvironment{gptresponse}[2]
{
    \begin{blurbwithfig}{#1}{#2}{backgptresponse}{textgptcolor}{\gpticon}
}
{

    \end{blurbwithfig}
}

\lstnewenvironment{gptcodeblock}[1]
{
    \lstset{style=codestyle} 
    \lstset{language=#1}
}
{}

\newenvironment{chat}[1]
{
    \newcommand{\preventbreaks}{#1}
    \begin{center}
    \mdfsetup{nobreak=\preventbreaks}
    \begin{mdframed}[
        linecolor=black,
        innerleftmargin=0.04cm,
        innerrightmargin=0cm
        innertopmargin=0cm
        innerbottommargin=0cm
    ]{}

}
{ 
    \end{mdframed}
    \end{center}
}


\newcommand{\authorremark}[1]{\footnotesize\textit{(Author remark: #1)}}

\newcommand{\omitted}[2]{
    \ifthenelse{\equal{#1}{}}{\textit{(... content omitted by authors ...)}} 
                            {\textit{(... omitted by authors: #1 ...)}}
}



\newcommand{\omittedCode}[2]{
    \textcolor{codeOmitted}{
    \ifthenelse{\equal{#1}{}}{\textit{(... code omitted by authors ...)}} 
                            {\textit{(... omitted by authors: #1 ...)}}
    }
}

%% file: sections/introduction.tex
\section{Introduction}

Advances in computational design and manufacturing (\cdam) have already permeated and transformed numerous industries, including aerospace, architecture, electronics, dental, and digital media, among others. 
Nevertheless, the full potential of the \cdam workflow is still limited by a number of barriers, such as the extensive domain-specific knowledge that is often required to use \cdam software packages or integrate \cdam solutions into existing workflows.
Generative AI tools such as Large Language Models (\llms) have the potential to remove these barriers, by expediting the \cdam process and providing an intuitive, unified, and user-friendly interface that connects each stage of the pipeline. 
However, to date, generative AI and \llms have predominantly been applied to non-engineering domains. 
In this study, we show how these tools can also be used to develop new design and manufacturing workflows. 

Our analysis examines the standard \cdam workflow to identify opportunities for \llm-driven automation or acceleration. 
Specifically, we break the \cdam workflow into five phases, and then assess whether and how the efficiency and quality of each phase could be improved by integrating \llms. 
The components under investigation include (1) generating a design, (2) constructing a design space and design variations, (3) preparing designs for manufacturing, (4) evaluating a design's performance, and (5) discovering high-performing designs based on a given performance and design space.

Although it is feasible to create specialized \llms for design and manufacturing, we demonstrate the opportunities offered by generic, pre-trained models. 
To this end, we conduct all of our experiments using \gpt \cite{openai2023gpt4}\footnote{We use the OpenAI ChatGPT interface to interact with the \gpt versions released between May 24, 2023 and July 19, 2023}, a state-of-the-art general-purpose \llm. 
Our \gpt-augmented \cdam workflows demonstrate how \llms could be used to simplify and expedite the design and production of complex objects. 
Our analysis also showcases how \llms can can leverage existing solvers, algorithms, tools, and visualizers to synthesize an integrated workflow. 
Finally, our work demonstrates current limitations of \gpt in the context of design and manufacturing, which naturally suggests a series of potential improvements for future \llms and \llm-augmented workflows.

%% file: sections/related.tex
\section{Background \& Related Work}
To contextualize our work, we briefly describe the state of the art for generative \llms and various aspects of \cdam. 

\subsection{LLMs for Generative Modeling}
\label{sec:related_llm}
Large Language Models (\llms) have garnered significant interest in the research community and beyond, as a result of both their already-demonstrated generative capabilities and their seemingly unbounded promise. 
Although these models are recognized primarily for their influence on text generation \cite{radford2019language}, their reach has been extended to impact various other domains, including image generation \cite{ramesh2021zero}, music generation \cite{dhariwal2020jukebox}, motion generation \cite{jiang2023motiongpt}, code generation \cite{chen2021evaluating}, 3D model creation \cite{liu2023zero}, and robotic control \cite{mirchandani2023large}. Notable foundational models include OpenAI's GPT series, ranging from GPT-2 to the more recent GPT-4 \cite{openai2023gpt4}. These models have showcased progressive improvements in fluency, coherence, and generalization capabilities. Meta AI's LLaMa model has further extended the reach of LLMs by demonstrating proficiency in both text and image synthesis \cite{touvron2023llama}. The Falcon LLM \cite{penedo2023refinedweb}, trained exclusively on properly filtered and deduplicated web data, has exhibited comparable performance to models trained on meticulously curated datasets. These models have been utilized in conjunction with Reinforcement Learning from Human Feedback (RLHF) to improve the quality of the generated content \cite{ouyang2022training}. This is done by incorporating human feedback into the training process, where humans rate the quality of the generated outputs and provide examples of ideal outputs for a given input \cite{christiano2017deep}. In parallel, domain-specific LLMs have also been trained for performance within a specific subject area. For example, ProtGPT2 specializes in predicting protein folding structures \cite{ferruz2022protgpt2}, while Codex has been specifically tailored to understand and generate code \cite{chen2021evaluating}. 
In this work, we investigate the generative capabilities of generic, pre-trained \llms within \cdam.

\subsection{Computational Design and Manufacturing}
\label{sec:related_cdam}
The \cdam workflow is often decomposed into a series of steps including (1) representing a design, (2) representing and exploring a design space, (3) preparing a design for manufacturing, (4) computing the performance of a design, and (5) finding a design with optimal performance.
For each phase, we provide a brief overview of the relevant work, with a focus on aspects that offer the best opportunities for \llm integration.

\paragraph{Design Representations.} 
The cornerstone of computational design is the capacity to digitally represent and manipulate the salient aspects of a given design -- such as geometry, articulated joints, material composition, \etc. 
There are many ways to represent such aspects, but we focus on focus on design representations that are compact, understandable, and editable. 
For example, modern CAD systems represent a shape as a sequence of operations such as 2D sketches, extrusions and Boolean operations~\cite{willis2021fusion}. 
These can be represented as compact programs written in domain specific languages (DSLs) such as OnShape's FeatureScript~\cite{featurescript}. 
Designs can also be represented compactly as a graph~\cite{prusinkiewicz2012algorithmic, zhang2018graphit}, in which the nodes typically represent individual components, while edges represent multi-component interactions. 
Such graphs have been used to efficiently and hierarchically represent CAD models~\cite{du2018inversecsg}, robots~\cite{zhao2020robogrammar}, metamaterials~\cite{makatura2023procedural}, architecture~\cite{muller2006procedural}, and chemical molecules~\cite{guo2022data}. 
To represent even more complex designs -- such as a quadcopter with a physical design and a software controller -- multiple DSLs may be used simultaneously.
For example, the copter's physical design may be encoded using CAD, while its software is coded using a control-specific DSL.

\paragraph{Design Space Representations} A design space represents an entire family of designs -- rather than a single instantiation -- which allows for design exploration, customization, and performance-driven design optimization. 
One of the most popular design space representations is parametric design, in which  a few exposed parameters are used to control a design.
This is commonly used in CAD systems, where \eg a bookshelf may be parametrized by its height, width, depth, and number of shelves.
Another popular option is formal languages such as L-systems~\cite{rozenberg1980mathematical} or shape-grammars~\cite{stiny1980introduction,ozkar2009shape}, which generate design variations by manipulating a set of terminal and non-terminal symbols according to given rewrite rules. 
Formal languages have been used in domains such as architecture~\cite{muller2006procedural}, robotics~\cite{zhao2020robogrammar}, and chemistry~\cite{guo2022data}.

\paragraph{Design for Manufacturing} 
Design for Manufacturing (DfM) is a planning process used to generate designs that can be fabricated with maximal efficiency and minimal cost.
One prominent aspect of this is Computer-Aided Manufacturing (CAM), which transforms a digital design into a viable fabrication plan for some manufacturing process, such as 3D printing, 3- or 5-axis CNC milling, or sheet-metal stretching. 
CAM also extends to multi-process representations such as STEP-NC, which abstracts away from machine-specific G-code in favor of tool-type-specific machining operations that are interpretable on different hardware.
Because all of these fabrication plans can also be described as a program in some DSL,
CAM can be interpreted as a \textit{translation} from a design DSL to a manufacturing-oriented DSL.
DfM also includes many other aspects, such as selecting an appropriate manufacturing method, optimizing manufacturing process parameters \cite{erps2021processOpt}, sourcing parts and materials, or modifying a design in light of manufacturing constraints \cite{Koo2017zeroWaste}.

\paragraph{Performance Prediction} Before manufacturing a design, engineers typically want to understand its predicted performance.
For example, automobile engineers may wish to evaluate and iteratively refine a candidate design's efficiency, safety, and aesthetics.
To do this, engineers frequently make use of numerical simulation methods such as general-purpose finite element analysis (FEA)~\cite{du2021diffpd} or more domain-specific approaches for \eg acoustics~\cite{o2002synthesizing}, robotics~\cite{erez2015simulation}, and electromagnetism~\cite{sullivan2013electromagnetic}. 
Commercial CAD systems (e.g., Autodesk~\cite{AutodeskSimulation} and Dassault Systèmes~\cite{DassaultSimulation}) integrate simulation into their ecosystem. 
Since engineers are primarily interested in the performance of the design's manufactured counterpart, it is crucial to minimize the gap between an object's performance in simulation versus reality.

\paragraph{Performance Optimization:} Given a design space and a way to predict performance, it is natural to seek designs that perform best with respect to a particular metric. 
Although this search could be performed via manual trial and error, it is more efficient and effective to use automated exploration tools.
One process known as \textit{inverse design} can automatically search (or optimize) over a given design space to find a design that exhibits some target performance~\cite{ma2021diffaqua}. 
Inverse design has already been applied to many problem domains. 
For example, a parametric design space can be searched for designs that have the best value of a simulated metric~\cite{xu2021end}. 
Topology optimization has been applied to problems such as minimum compliance. 
In addition, designs can be optimized for metrics such as weight, cost, and manufacturing time.

%% file: sections/overview.tex
\section{Overview}
\label{sec:overview}

\begin{figure}[tb]
\centering
\includegraphics[width=0.92\textwidth]{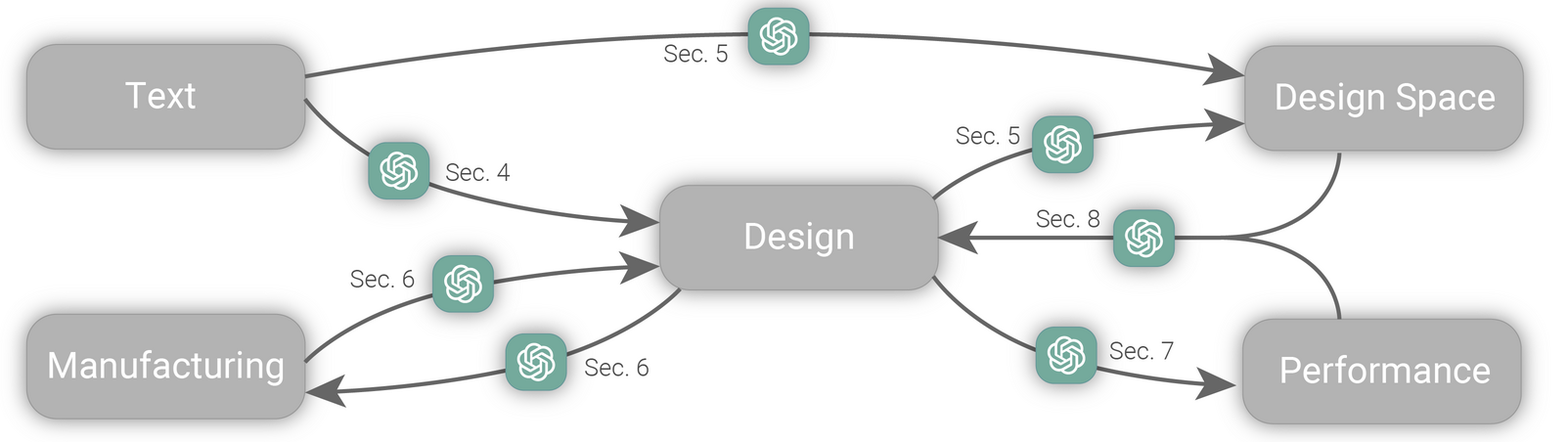}
\caption{\textbf{Opportunities for \llm Integration within the \cdam Workflow.} Each technical section of our paper covers opportunities for \llm integration in one of the tasks depicted above: text to design, text/design to design space, bi-directional design for manufacturing, design to performance, and inverse design (from performance and design space to an optimized design).}
\label{fig:llm_graph}
\end{figure}

The fundamental aim of this study is to conduct an in-depth exploration of the opportunities and challenges of applying contemporary \llms  within the landscape of the \cdam workflow described in \sref{sec:related_cdam}. 
Driven by this objective, we propose a thorough and wide-ranging exploration that is independent of any predefined or proposed framework.

To apply \llms coherently across such diverse tasks, we leverage the insight that all building blocks in the \cdam workflow (design, design spaces, manufacturing instructions, and performance metrics) can be represented by compact programs.
Thus, at a high level, every phase of the \cdam workflow can be seen as a translation layer between an input DSL and an output DSL. 
The fact that \llms excel at such symbolic manipulations suggests that \llms have the potential to address these tasks while simultaneously leveraging and improving upon our traditional solutions.

To achieve comprehensive coverage and uncover the different facets of \llm-assisted \cdam, we have undertaken an extensive suite of experiments, incorporating a broad variety of design representations, manufacturing processes, and performance metrics. These are detailed further in \sref{sec:domains}.

\subsection{Methodology}
\label{sec:methodologyAndCLD}

\begin{table}[tb]
    \centering
    \resizebox{\textwidth}{!}{\begin{tabular}{c|c|l|l}
        Category & Code & Title & Summary\\
        \hline
        \multirow{ 2}{*}{Capabilities} 
         & C.1 & Extensive Knowledge Base in Des.\&Mfg. & \gpt has a broad knowledge of design and mfg. considerations\\
        &  C.2 & Iteration Support & \gpt attempts (and often succeeds) to iterate and recitfy errors when prompted\\
        & C.3 & Modularity Support & \gpt can reuse or adapt previous/provided designs or solutions \\
        \hline
        \multirow{ 2}{*}{Limitations} 
        &L.1 & Reasoning Challenges & \gpt struggles with spatial reasoning, analytical reasoning, and computations\\
        & L.2 & Correctness and Verification & \gpt produces inaccurate results or justifications for its solutions\\
        & L.3 & Scalability & \gpt struggles to respect multiple requests concurrently\\
        & L.4 & Iterative Editing & \gpt forgets/introduces errors when modifying previously-generated designs\\
        \hline
        \multirow{ 2}{*}{Dualisms} 
        & D.1 & Context Information & \gpt's performance depends on the amount of context provided\\
        & D.2 & Unprompted Responses & \gpt makes inferences/suggestions beyond what is specified in the prompt
    \end{tabular}}
    \caption{\textbf{\gpt's key properties for \cdam} To facilitate discussion of \gpt's applicability for design and manufacturing (Des.\&Mfg.), we have identified 9 key observations about \gpt that persist across several aspects of the \cdam workflow. This includes 3 powerful capabilities, 4 limitations, and 2 dualisms (so named because they may manifest either as an opportunity or a drawback, depending on the context). We use these observations to frame our discussions about \gpt's suitability for each stage of the \cdam workflow.}
    \label{tab:key_properties}
\end{table}

Our methodology is crafted to provide a comprehensive inspection of the opportunities for and efficacy of various interfaces between \gpt and the \cdam workflow. 
We investigate each of the five stages of the design and manufacturing pipeline individually.
As illustrated in \fref{fig:llm_graph}, these stages include: 
design generation (\sref{sec:text_to_design}), 
design space generation (\sref{sec:text_to_design_space}), 
design for manufacturing (\sref{sec:design_for_manufacturing}), 
performance prediction (\sref{sec:design_to_perf}), 
and inverse design (\sref{sec:inverse}).

In each of these stages, we pose fundamental questions about ways in which \gpt may offer some benefit, and then conduct a series of experiments to answer these questions.
For each query, we investigate aspects such as 
(1) strategies for engineering effective prompts,
(2) strategies for integrating human feedback, expertise, or preferences into the \llm-assisted design process, and
(3) tasks that \gpt can accomplish natively versus tasks that are better completed by asking \gpt to leverage external tools.

After a detailed examination of each stage, we sought to understand the implications of incorporating \gpt within an end-to-end \cdam process. 
To this end, we designed and fabricated two practical examples (a cabinet and a quadcopter) with \gpt's support.
The end-to-end design process for each example is detailed in \sref{sec:examples}.

Beyond these individual questions, our comprehensive investigation has also exposed several key insights about \gpt's general capabilities and limitations with respect to \cdam. 
We have also observed a group of properties that we term 'dualisms', because they may manifest either as an opportunity or a drawback, depending on the situation. 
Our findings are summarized in \tref{tab:key_properties}, with a full description in \sref{sec:capabilities}. 
To emphasize the pervasive nature of these properties, we also use these labels as a framework for our discussions and takeaways at the end of each section. 
Specifically, we draw on each section's findings and examples in order to illustrate the manifestation and impact of various properties in \tref{tab:key_properties} throughout the \cdam workflow.

\input{sections/domains}


\ignore{

}

%% file: sections/domains.tex
\subsection{Scope of Evaluation}
\label{sec:domains}

To conduct a holistic survey of \gpt-assisted \cdam, our experiments span a number of different 
design domains (\sref{sec:overview_domains_design}), 
performance metrics (\sref{sec:overview_domains_perf}) and 
manufacturing methods (\sref{sec:overview_domains_manufacturing}).
Here, we briefly describe each domain of interest, along with the specific challenges they pose and the sort of representative, transferable insight we hope to glean by studying each domain in connection with \llms. 

\subsubsection{Target Design Domains} 
\label{sec:overview_domains_design}
Our experiments are concentrated in three main design domains, including 2d vector graphic design, 3D parametric modeling, and articulated robotics problems.  

Vector graphics use a series of text-based commands to represent paths and areas that form a given design.
Vector image formats are an important part of \cdam, as they can be used as both a design specification and a manufacturing specification for \eg laser cutters. 
Despite their simplicity, vector graphics can represent a wide range of 2D and 3D objects, such as artistic engravings or flat-pack furniture.
We examine \llms' capacity to generate two popular vector formats: SVG and DFX.
These formats present several challenges: 
they contain boilerplate formatting that \gpt may struggle to reproduce; 
it may be difficult to layout individual pieces on the canvas; and finally, it may be difficult to decompose higher-dimensional designs into 2d.
Thus, vector graphics will test \gpt's spatial reasoning and ability to respect highly-constrained syntax, either on its own or with the use of external libraries.

Parametric modeling languages generate 3D geometry through a sequence of constructive instructions. The term ``parametric modeling'' reflects how each constructive operator exposes a set of parameters, such as the radius of a circle.
We explore two distinct approaches that are powerful, widely-used, and well-documented online.
The first is rooted in classic Constructive Solid Geometry (CSG), which constructs shapes by successively deploying boolean operations (union, intersection, subtraction) over basic shapes or primitives (such as cuboids, spheres, cylinders, and so forth) that can undergo transformations such as translations, rotations, and scaling.
The CSG approach is intended to test the \textit{global} spatial reasoning capacity of \gpt, as every CSG operation/transformation occurs \wrt a shared coordinate space.
The second representation relies on the contemporary B-rep format used by modern CAD systems.
Here, geometry is built through a sequence of operations like sketching, extruding, and filleting. 
Each operation in this context is parametric and uses references to previously created geometry to \eg, select a plane for a sketch design or select a sketch for an extrusion. 
Sketch-based CAD will test \gpt's ability to effectively switch between and reason over multiple relative, local coordinate frames.


Robotics offers a particularly rich design domain, as \gpt must coordinate a set of \textit{articulated} and \textit{actuated} geometries to form complex objects such as open chain robot arms, wheeled robots, copters/UAVs, and robot grippers.
Robotics representations must describe not only the high-level geometry of each part, but also their properties and relationships -- including the joints between parts, the degrees of freedom that those joints exhibit, and dynamics information such as the inertia of a given part. 
Several existing formats support these tasks, but we primarily use the XML-based language known as the Universal Robot Description Format (URDF). 
We also investigate the use of a more general graph-based robot representation.
These formats test \gpt's ability to simultaneously reason about multiple aspects of design, such as static geometric bodies and dynamic articulation constraints.

\subsubsection{Target Performance Domains} 
\label{sec:overview_domains_perf}
Diverse performance domains within engineering design require evaluation of aspects such as structural and material properties, mechanical integrity, geometry-based functionality, materials use, electromechanical integration, and subjective features. The results of such evaluation allow us to (dis)qualify a design for use and to use the evaluation to further understand and improve the design. Using \gpt, we focus on assessing mechanical and structural properties through generating first-order analysis equations for input designs of standard objects like chairs, cabinets, and a quadcopter, which test the ability of \gpt to sufficiently understand a given input design in text form or through a DSL and to evaluate criteria for functionality and failure. Mechanical properties assessed include weight, size, load capacity, storage capacity, and stability. Analysis of electromechanical functionality include battery life and quadcopter travel distance. Further use of \gpt aims to streamline the computationally intensive process of Finite Element Analysis (FEA), a crucial tool for understanding structural behavior in detail under various conditions, and we apply this to the case of a load on a set of chairs. 

In addition to these technical aspects, our investigation extends into the subjective domains of sustainability and aesthetics, which cannot be strictly quantified. The inherent complexity and qualitative nature of these areas present unique challenges in evaluation. While it is well-known that computational systems can compute quantitative features, machine learning systems are becoming more sophisticated in artistic domains, and so we seek to leverage the capacity of \llms for lexical analysis to aid more holistically in the more ambiguous realms of the design process and to find its limits. For example, could an \llm reasonably address whether a piece of furniture of a given size is ``large'', or if a shoe of a given design is ``comfortable,'' or can it only handle classically quantifiable features? Can it even help us to reason more objectively about what aspects delineate these properties? To this end, we test evaluation of subjective domains and use \gpt to generate a scoring system and functions for quantifying the sustainability of a chair, the classification of chairs based on categories of aesthetic influence, and the appropriate distribution of a set of chairs into a set of rooms in a house, among other examples. 

We further combine these performance metric evaluations with the principles of inverse design. Inverse design entails setting desired performance attributes and employing computational methodologies to deduce design parameters that satisfy these attributes, both by generating areas for improvement within a design domain and by testing the effects of implementing improvements suggested by \gpt or target design goals of our own interest,
as well as selecting appropriate methods of optimization. In this case, given a design/decision space for an object, we use \gpt to generate and implement methods to computationally improve or optimize qualifying designs to satisfy designated performance goals. This methodical approach evaluates if \llms can apply constructive logic for design enhancement and innovation.

\subsubsection{Target Manufacturing Domains} 
\label{sec:overview_domains_manufacturing}

 Leveraging language models like \gpt in DfM context can yield more consistent and scalable decision-making, potentially augmenting human expertise and reducing our reliance on CAD software usage. Potential applications of \gpt include the selection of optimal manufacturing techniques, suggestion of design modifications that would enable easier production, identification of potential suppliers, and creation of manufacturing instructions. The approach is aimed to alleviate many of the bottlenecks caused by the designers' lack of knowledge and experience in DfM. 

In a set of experiments, we've explored \gpt's capabilities across various tasks. Firstly, \gpt was used to identify the optimal manufacturing process for a given part, considering factors such as part geometry, material, production volume, and tolerance requirements. Next, \gpt was tasked with optimizing a component design for CNC machining. Given the geometry of the component, \gpt identified potential manufacturing difficulties and modified the design to address these. We also leveraged \gpt's extensive dataset knowledge to identify parts needed for manufacturing.

In addition to these, \gpt was used to create manufacturing instructions for both additive and subtractive design processes. Additive design can be challenging due to the need for spatial reasoning, precision, and meticulous planning, and often requires many iterations. We've explored the generation of fabrication instructions using subtractive manufacturing techniques for a cabinet design. We also investigated \gpt's potential in generating machine-readable instructions for robot assembly tasks and converting those into human-readable standard operating procedures. This allowed for effective communication and collaboration between robots and human operators.

%% file: sections/allGraphArrows.tex
\input{sections/GraphArrows/TextToDesign.tex}

\input{sections/GraphArrows/TextToDesignSpace.tex}

\input{sections/GraphArrows/DesignForManufacturing}

\input{sections/GraphArrows/DesignToPerformance.tex}

\input{sections/GraphArrows/PerformanceAndDesignSpaceToDesignv2.tex}

%% file: sections/GraphArrows/TextToDesign.tex
\section{Text-To-Design} 
\label{sec:text_to_design}




For our first line of inquiry, we explore the extent to which \gpt is able to generate designs across a variety of domains. 
Even within the specific context of manufacturable design, the concept of a ``design'' is quite broad, and exists at many scales.
For example, we may want to specify a single self-contained part, or a sizable hierarchical assembly containing several levels of sub-assemblies and/or other individual component modules. 
Such assemblies may be completely customized/self-contained, with all parts designed simultaneously, or they may be hybrid designs that integrate existing, pre-manufactured elements such as brackets or motors.
In many cases, our target design tasks also include dynamic considerations such as assembly mating or articulated joints.

Although these complex tasks may initially seem out-of-scope for lexical models such as {\llm}s, there are many modeling and design paradigms that can be expressed in terms of potentially-\llm-compatible language.
To guide our exploration of \gpt's ability to interface with each of these models, we pose the following questions:

\begin{itemize}
\item \textbf{Q1} Can \gpt generate a meaningful design when provided with a high-level description of the goal and a given modeling language?
\item \textbf{Q2} To what extent is the user able to control the designs created by \gpt? Is \gpt able to interpret and respect user-defined constraints, such as spatial relationships between objects or integration of standard pre-fabricated parts?
\item \textbf{Q3} Is \gpt able to incorporate high-level abstractions used by human designers, such as modular (de)composition?
\end{itemize}


\subsection{Simple, self-contained designs from high-level input (Q1)}


To explore \gpt's capacity for design, we first test its ability to do one- (or few-) shot generation of an object from a minimal high-level text description as input. 
Ideally, we would like to understand \gpt's ability to complete design tasks independent of any particular modeling paradigm.
However, it is not immediately clear how much dependence there may be on the specific representation that is chosen, because the variation in possible language-based modeling paradigms is significant. 
Some languages are very general and versatile, with a wide variety of features and capabilites, while others may be highly-specialized for a specific set of tasks or outcomes. 
Similarly, some languages are well-established with plentiful online documentation or examples, while others may be custom-defined, poorly documented, or otherwise underrepresented in \gpt's training repository. 
Finally, some languages are fairly streamlined, while others may be syntactically complex and/or require the use/coordination of many modules.
Each possibility offers unique capabilities and challenges. 
Thus, we set out to test a wide variety of them, in an effort to determine 
\llms' ability to use each representation;
whether there are any conclusions that seem to span across different representations;
and
whether any particular representations seem uniquely well- or poorly-suited for \llm integration.


\subsubsection{Vector Graphics with SVG/DXF} 

\input{sections/GraphArrows/TextToDesignExperiments/VectorGraphics_basic}

\subsubsection{CSG with self-defined primitives}
\label{sec:textToDesign_CSG_boxes}
\input{sections/GraphArrows/TextToDesignExperiments/CSG_customPrimivites_basicDesigns}

\subsubsection{CSG with PyVista}
\input{sections/GraphArrows/TextToDesignExperiments/CSG_PyVista_basicDesigns}

\subsubsection{CSG with \jscad}
\label{sec:textToDesign_JSCAD_basic}
\input{sections/GraphArrows/TextToDesignExperiments/CSG_JSCAD_basicDesigns}

\subsubsection{Sketch-based CAD with OnShape}
\label{sec:textToDesign_OnShape_basic}
\input{sections/GraphArrows/TextToDesignExperiments/CAD_OnShape_basicDesigns}

\subsubsection{URDF}
\label{sec:textToDesign_urdf}
\input{sections/GraphArrows/TextToDesignExperiments/Robots_URDF_basicDesigns}

\subsubsection{Graph-based DSL}
\input{sections/GraphArrows/TextToDesignExperiments/Robots_graph_basicDesigns}


\subsubsection{Summary Discussion}
In light of these experiments, we conclude that \gpt is capable of generating designs based on high-level text input, even across a wide variety of representations and problem domains. 
We note that several of \gpt's capabilites and limitations remain consistent independent of the representation.
For example, in all cases, \gpt is able to generate sensible, well-structured code with semantically meaningful variables and comments.
Moreover, independent of the representation or the problem domain, \gpt consistently shows superior performance with respect to the high-level, \textit{discrete} elements of a problem (\eg, identifying the correct type and quantity of each primitive/operation) as opposed to the lower-level continuous parameter assignments (\eg, correctly positioning the primitives relative to one another). 
A more detailed discussion of capabilities, limitations and opportunities will follow in \sref{sec:textToDesign_discussion}.
For now, we rely on the similarities between various representations to justify a reduced scope for our future experiments.
In particular, moving forward, we study each question with respect to only a subset of the design representations and domains introduced above.

\subsection{Interpreting and Respecting User Control (Q2)}

The above examples demonstrate \gpt's ability to generate a design based on very high-level semantic input.
However, we also wanted to test its ability to generate designs that adhere to a specific user-given intent.
This section also tests whether \gpt is able to overcome its own potential biases induced by the training data, in order to generate something that truly adheres to a user's specified constraints -- whether or not those constraints match the ``common'' form of a given design target. 
In particular, we choose to study whether \gpt is able to 
(1) understand and respect semantically meaningful spatial constraints, and
(2) incorporate specific pre-fabricated elements into a design.

\subsubsection{Spatial Constraints}
\label{sec:textTODesign_spatial_constraints}
Through the general experiments above, \gpt has already shown some capacity to respect high-level spatial constraints, such as a design element's absolute size or its position relative to another element of the design. 
\gpt's compliance with such requests was frequently flawed at the outset, but the results were generally workable after some amount of interactive feedback. 
This section aims to explore the types of constraints \gpt is able to natively understand, and how we might best interact with \gpt in order to improve the chance of successful compliance with such constraints.

\input{sections/GraphArrows/TextToDesignExperiments/CSG_JSCAD_constraints}


%

\subsubsection{Incorporating pre-fabricated elements}
\label{sec:prefabbed-ele}

It's also common to design an object around specific pre-manufactured elements, such as hinges, brackets, or motors. 
We explore the possibility of using \gpt to source the parts in \sref{sec:part_sourcing} -- at that time, we explore whether \gpt can identify the required part categories, provide options, and/or select a set of options that are compatible with one another and the intended overall design. 

For now, we assume that the user has a specific (set of) part(s) in mind that they would like to incorporate into their design.
Then we investigate whether, given these components, \gpt is able to (1) build a reasonable proxy of this design, then (2) effectively use it as a module within a larger assembly.

\paragraph{Cabinet with Standard Hardware} 
\input{sections/GraphArrows/TextToDesignExperiments/CSG_JSCAD_integratingPrefabbedParts}

\paragraph{Quadcopter}
\input{sections/GraphArrows/TextToDesignExperiments/Robots_customPrimitives_integratingPrefabbedParts}

\subsection{Incorporating Abstractions such as Modular/Hierarchical Designs (Q3)}
\label{sec:textTODesign_abstractions}
\input{sections/GraphArrows/TextToDesignExperiments/CSG_JSCAD_modules}

\subsection{Discussion}
\label{sec:textToDesign_discussion}

In this section, we elaborate on the key capabilities (C), limitations (L), and dualisms (D) previously outlined, particularly as they relate to the domain of text-to-design.

\noindent \textbf{C.1 Extensive Knowledge Base in Design and Manufacturing:} Within the text-to-design space, \gpt exhibited proficiency in supporting high-level structure and discrete composition. For instance, \gpt consistently generated the correct primitives (type and quantity) for a given task, regardless of the specific design language it was using. 
\gpt also demonstrated a capacity for interpreting and auto-completing under-specified prompts, as in the case of the CSG table example, where \gpt inferred and provided reasonable values for a set of missing parameters (see \sref{sec:textToDesign_CSG_boxes}).
Finally, \gpt generated readable, explainable, and maintainable code that contained descriptive variable names and comments, along with appropriate modularity and other high-level structural elements.

\noindent \textbf{C.2 Iteration Support:} Even when \gpt did not immediately arrive at a suitable design solution, it often succeeded in rectifying errors after a reasonably small number of user interactions. For example, it was able to successfully adjust the placement of the cabinet handle after a handful of additional prompts. The ability to engage in iterative design is also very helpful when building up complex structures such as the wheeled robot from \sref{sec:textToDesign_urdf} or the L-bracket proxy discussed in \sref{sec:prefabbed-ele}, because users can start with a simple prompt, then iteratively increase the complexity to arrive at a suitable result.

\noindent \textbf{C.3 Modularity Support:} \gpt effectively incorporates modules and hierarchical structures, using natural language as a powerful tool for conceptualization and orientation.

\noindent \textbf{L.1 Reasoning Challenges:} Spatial reasoning posed a significant challenge for \gpt. Well-crafted domain-specific languages (DSLs) may be able to mitigate this issue. We noted specific difficulties with constructive solid geometry (CSG) due to the computational requirements for object placement. Sketch and extrude languages that utilize reference points can minimize this challenge to an extent, as they offload the computation to reference resolution. This approach is effective for simpler designs but falters when managing complex sequences of transformations. As discussed in the sketch-based car example from \sref{sec:textToDesign_OnShape_basic}, we found that DSLs that balance the benefits of reference-based language with global positioning information may be more effective. 

\gpt's lack of spatial awareness also created difficulties with constraint handling, such as when \gpt was asked to ensure that elements were non-overlapping. We found that iterative refinements and careful prompting often provided a workaround for these issues. For example, \gpt typically failed to respect ``non-overlapping'' constraints, but it generally responded well to the instruction that some element should be ``in contact with (but not protruding into)'' another element.

\noindent \textbf{L.2 Correctness and Verification:} \gpt is not able to reliably verify its own output, and it frequently makes contradictory claims. For example, when asked to place a handle on the right side of the cabinet structure, \gpt frequently placed the handle on the left-hand side of the cabinet, then immediately declared its design a success, because the handle was on the right, as requested. This seems to suggest that external verification tools may be helpful, particularly in cases where the contradictions are less obvious.

\noindent \textbf{L.3 Scalability:} \gpt's success seems to decline as the number of simultaneous requests increases. For example, it is best to issue 1-2 constraints or correct 1-2 issues at a time, rather than trying to issue several constraints or correct several issues at once. 
Similarly, \gpt encountered challenges when interpreting high-level information to build proxies for more complex designs all at once; instead, the models must be built iteratively, with gradually increasing complexity.
This iterative modeling was most effective when the user provides explicit instructions about both the aspects that should change, as well as the aspects that should remain unaltered (either because they are already correct, or because they will be addressed later). 
Despite \gpt's initial difficulty creating complex models, \gpt is able to effectively use and combine existing modules to create more intricate models. 

\noindent \textbf{L.4 Iterative Editing:}
As discussed in \sref{sec:prefabbed-ele}, \gpt seems to exhibit limited memory and attention span. In particular, it often ``forgets'' things from previous messages. We address this by occasionally reminding \gpt of its previous input/output, either by asking it to summarize a previous interaction/finding, or by explicitly including a prior result as a starting point in our prompt.

\noindent \textbf{D.2 Unprompted Responses:} \gpt is frequently able to recognize and address under-specified problem statements.
For example, in the CSG table specification (\sref{sec:textToDesign_CSG_boxes}), \gpt correctly inferred the need to assign a tabletop thickness value.
Similarly, when augmenting the cabinet with a door and a handle in \sref{sec:textTODesign_spatial_constraints}, \gpt responded with several distinct approaches for handle design.
This can be powerful, as it may alert the user to parameters or variations which may otherwise have gone overlooked; then, users have an explicit opportunity to consider and refine the specification accordingly.
Moreover, it allows users to undertake a design process and begin receiving feedback without first needing to craft a perfect specification or prompt.
However, if \gpt confidently hallucinates a particular solution to an under-specified aspect of a design problem -- rather than explicitly prompting the user to consider a range of options -- it may limit and/or bias their exploration in unexpected ways.

\ignore{

- Text-To-Design Takeaways
	- good at high level structure, bad at placement
	- can use a wide range of DSLs, but best if they're intrinsically consistent, seems to have trouble reasoning about local coordinate systems (needs to be reminded about e.g. centered positioning, sketch based CAD hard,)
	- better to build it up gradually, has trouble with large requests all at once
	- constraints are difficult -- has trouble reasoning about them, but can usually be accomplished after some back and forth / playing with the word choice or prompt style 

	- has difficulty directly interpreting/building proxy from high level information, but can use it once it exists to create more complex designs.
	-
As part of this exploration, we also examine \gpt's ability to provide semantically-meaningful design specifications -- \eg, scripts including variables rather than hard-coded ``magic'' numbers, and reusable functions or modules where appropriate.

\begin{itemize}
    \item \textbf{Q} Is there a universally effective style of prompt that seems to generate the desired design quickly, with few errors (wrt design or boilerplate code)?
    \item \textbf{Q} Is there a limit to the number of primitives, constraints, modules, levels of hierarchy that \gpt seems to encounter? 
    \item \textbf{Q} (How) can we iteratively evaluate validity and the correctness of the design, and use this evaluation either to improve the model through feedback or to automatically post-process the design so that it is valid and correct.?
\end{itemize}

When specifying geometric constraints, effectiveness of that constraint being reflected in the final design varies based on the adjectives used to describe the constraint. For example, asking \gpt to make two primitives "non-overlapping" may still produce a result where the two bodies intersect, while asking \gpt to make it so that a body "does not protrude" into the other is effective in preventing intersections. So far, a pattern as to which adjectives are interpreted correctly by \gpt has not been observed. 

Specifying constraints via numeric means --- i.e. offset the second cube so that it's top face is aligned with z = 0 --- is almost always effective, whereas asking for geometric relations --- i.e. offset the second cube so its top face is aligned with the bottom face of the first cube --- does not guarantee consistent interpretation.

Introducing multiple constraints in a single prompt often results in some of the constraints being ignored, whereas introducing the same constraints in sequential prompts results in the desired behaviour. 

}

\ignore{
\paragraph{Problem setup}

There are three aspects to a prompt for specifying a design task: The objective, the primitives, and the additional constraints.  We describe in each turn, and then provide an example of a complete prompt and its output.

\includepdf[pages=1]{figures/BohanExample.pdf}

Function box(x1,x2,x3, dx1, dx2, dx3) creates a box centered at point (x1, x2, x3) with dimensions (dx1, dx2, dx3). Create a capital letter T from non-overlapping boxes.

This prompt has three components. Specifying a primitive, specifying the objective/goal, and constraints.

\paragraph{Design Representation}
Design can be expressed code, i.e. formally (e.g., CAD, OpenSCAD, SVG, STL, etc.). Sequential List of commands. 
Design can be represented using graph. Nodes (components). Edges connections (e.g., connectivity) between components.

\paragraph{Design specifications}
Describe a text example of a goal.

\paragraph{Design primitive}
How do specify design primitives.
Can you specify different primitives? How many? Can you mix different primitives. 

\paragraph{Constraints}
Describe different types of constraints.
Describe which constraints work and which do not work.
Symmetries. Spatial constraints (on the top, on the side, colinear, at the distance from X, parallel, perpendicular, at an an angle, reflective, minimum, maximum size, setting something to a value)

\paragraph{Hierarchy}
Describe how to introduce hierarchy.
E.g,, Specify in detail a leg for a table and a tabletop and then how to make the whole table. A leg for a table is made of a few components. a tabletop is made from a few components.
Check if you can make multiple levels of hierarchy.
Check if constraints still hold across multiple levels of hierarchy.

}

%% file: sections/GraphArrows/TextToDesignExperiments/VectorGraphics_basic.tex
Our initial focus in the design domain is on 2D vector graphics.
Vector formats such as SVGs or DXFs are prevalently utilized in manufacturing processes,
like those for laser or waterjet cutting.
The goal of our investigation was to ascertain whether \gpt could empower designers to 
transform their text directly into these vector formats. 
To evaluate this, we conducted experiments to determine 
if \gpt is capable of generating a valid SVG file and converting the design into DXF format.

\begin{figure}[tb]
\begin{chat}{false} 

\begin{userprompt}{0.03\textwidth}{} 
I would like to build a cabinet; the cabinet will have 3 shelves 6 feet high, 1 foot deep, and 4 feet wide. 
Please ensure that you account for the thickness of the wood. I want to use 1/2 inch plywood.
Can you label the parts?
Create a script that creates an SVG file. Use lines to depict the edges of the parts.
\end{userprompt}

\begin{gptresponse}{0.2\textwidth}{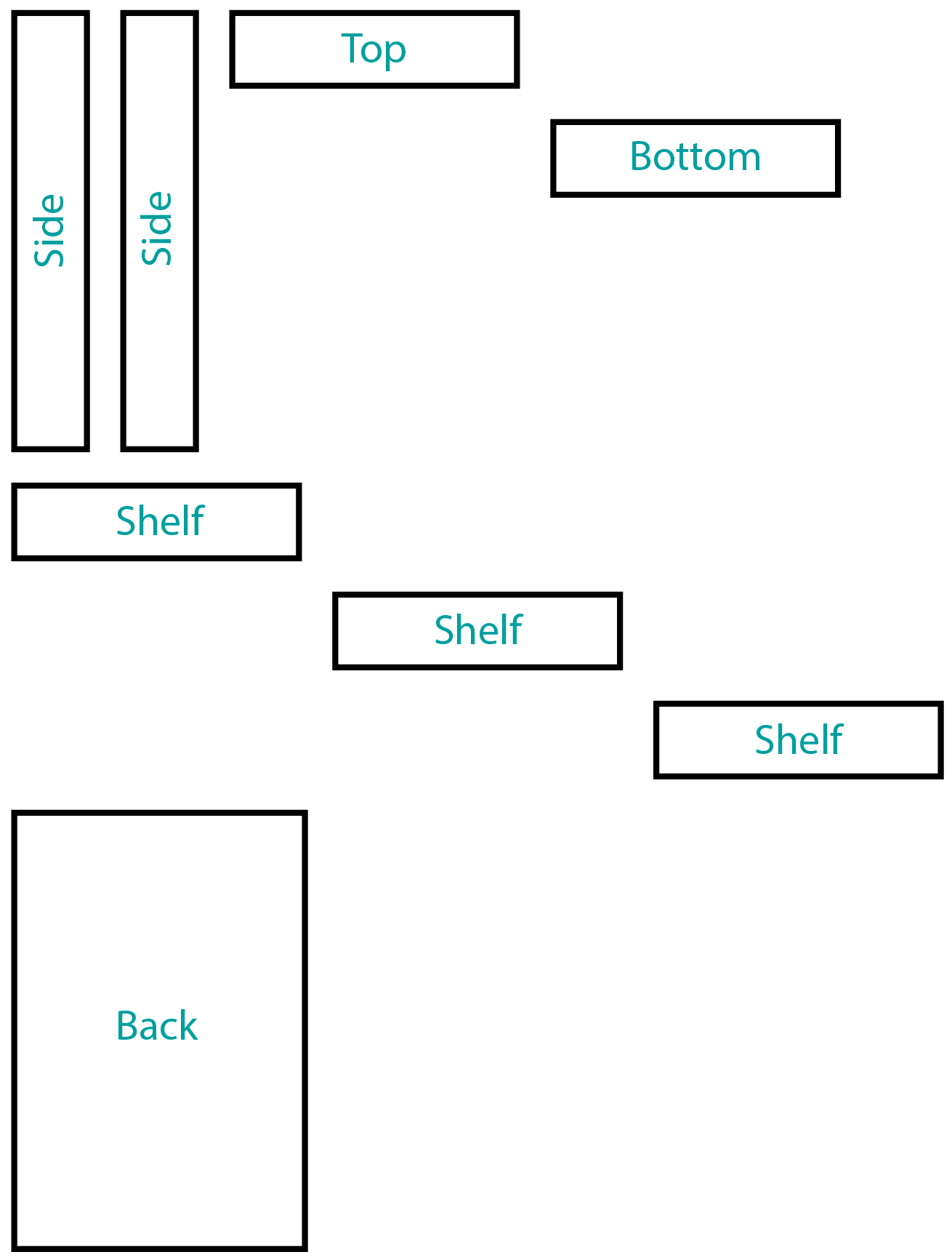} 
\begin{gptcodeblock}{Python}
import svgwrite
$\omittedCode{dimension conversion from feet to inches}{
# Convert feet to inches for better granularity
def ft_to_in(ft):
    return ft * 12

# Dimensions in feet
side_panel_dims = (1, 6)
wood_thickness = ft_to_in(1/2 * 1/12) # Converted to inches
top_bottom_panel_dims = (4 - 2*(1/2 * 1/12), 1) # Accounting for the wood thickness
shelf_dims = (4 - 2*(1/2 * 1/12), 1) # Accounting for the wood thickness
back_panel_dims = (4, 6)

# Convert to inches
side_panel_dims = (ft_to_in(side_panel_dims[0]), ft_to_in(side_panel_dims[1]))
top_bottom_panel_dims = (ft_to_in(top_bottom_panel_dims[0]), ft_to_in(top_bottom_panel_dims[1]))
shelf_dims = (ft_to_in(shelf_dims[0]), ft_to_in(shelf_dims[1]))
back_panel_dims = (ft_to_in(back_panel_dims[0]), ft_to_in(back_panel_dims[1]))
}$
dwg = svgwrite.Drawing('cabinet.svg', profile='tiny')

$\omittedCode{variable setup}{
# Space between panels
space = 6

# Variables to hold the cumulative width and the max height for the next row
cumulative_width = 0
max_height = 0
}$
# Side panels
for i in range(2):
    dwg.add(dwg.rect((cumulative_width, 0), side_panel_dims))
    cumulative_width += side_panel_dims[0] + space
    max_height = max(max_height, side_panel_dims[1])
$\omittedCode{similar for loop to write the top and bottom panels}{
# Top and bottom panels
for i in range(2):
    dwg.add(dwg.rect((cumulative_width, i*(top_bottom_panel_dims[1]+space)), top_bottom_panel_dims))
    cumulative_width += top_bottom_panel_dims[0] + space
    max_height = max(max_height, top_bottom_panel_dims[1]*(i+1) + space*i)
}$

# Reset cumulative_width for next row and update max_height
cumulative_width = 0
max_height += space
$\omittedCode{for loop to write the shelves}{
# Shelves
for i in range(3):
    dwg.add(dwg.rect((cumulative_width, max_height + i*(shelf_dims[1]+space)), shelf_dims))
    cumulative_width += shelf_dims[0] + space
}$

$\omittedCode{reset cumulative width and height variables, then write back panel}{
# Reset cumulative_width for next row and update max_height
cumulative_width = 0
max_height += 3*shelf_dims[1] + 2*space

# Back panel
dwg.add(dwg.rect((cumulative_width, max_height + space), back_panel_dims))
}$

dwg.save()
\end{gptcodeblock}
\end{gptresponse}

\begin{userprompt}{0.03\textwidth}{} 
Can you use the same method you used in this to generate the placement? Except generate a DXF rather than an SVG? 
\end{userprompt}

\begin{gptresponse}{0.2\textwidth}{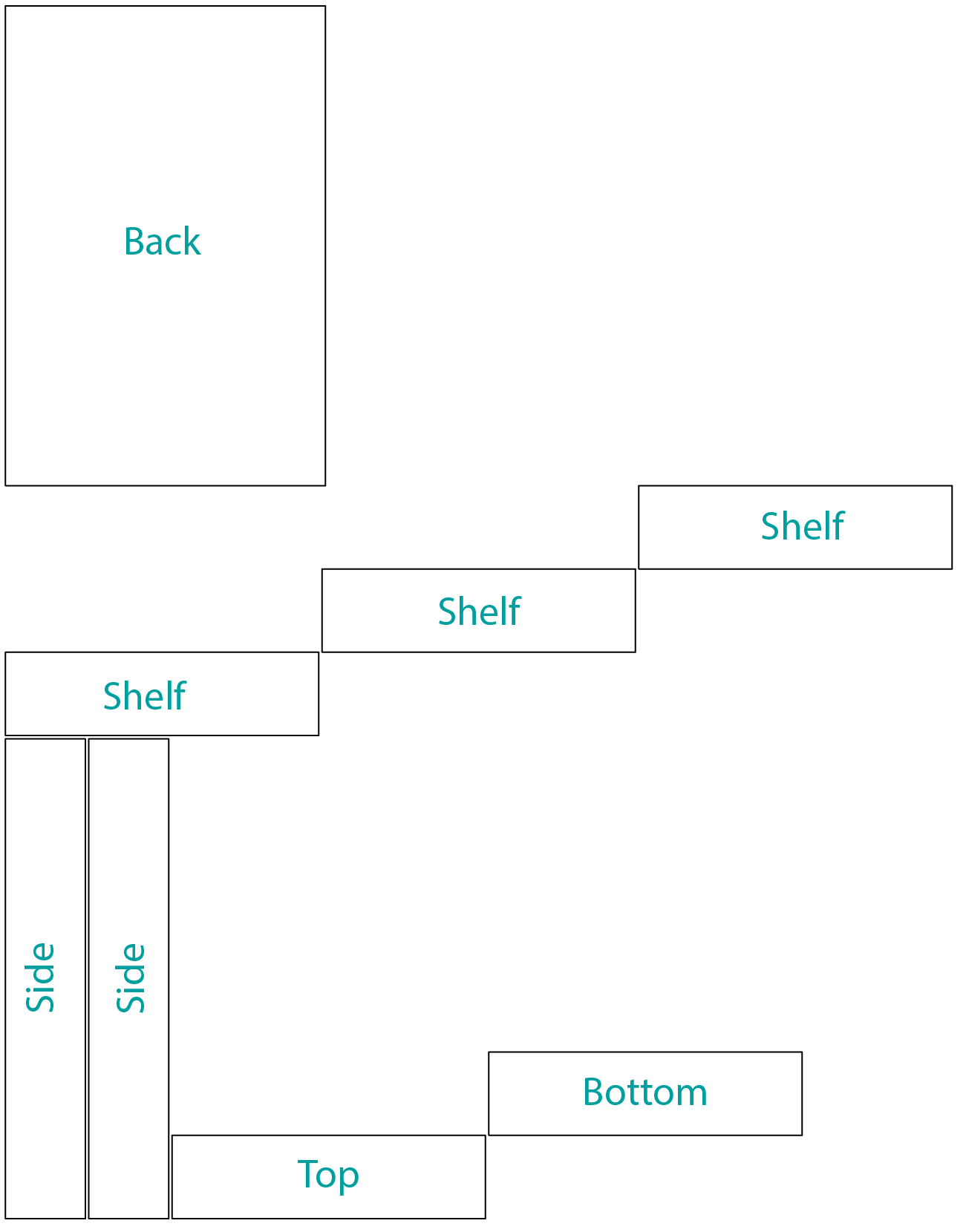} 
\begin{gptcodeblock}{Python}
import ezdxf
$\omittedCode{variable setup and dimension conversion from feet to inches}{
# Convert feet to inches for better granularity
def ft_to_in(ft):
    return ft * 12

# Dimensions in feet
side_panel_dims = (1, 6)
wood_thickness = ft_to_in(1/2 * 1/12)  # Converted to inches
top_bottom_panel_dims = (4 - 2*(1/2 * 1/12), 1)  # Accounting for the wood thickness
shelf_dims = (4 - 2*(1/2 * 1/12), 1)  # Accounting for the wood thickness
back_panel_dims = (4, 6)

# Convert to inches
side_panel_dims = (ft_to_in(side_panel_dims[0]), ft_to_in(side_panel_dims[1]))
top_bottom_panel_dims = (ft_to_in(top_bottom_panel_dims[0]), ft_to_in(top_bottom_panel_dims[1]))
shelf_dims = (ft_to_in(shelf_dims[0]), ft_to_in(shelf_dims[1]))
back_panel_dims = (ft_to_in(back_panel_dims[0]), ft_to_in(back_panel_dims[1]))
}$
doc = ezdxf.new("R2010") # Create a new DXF document
msp = doc.modelspace() # Create a new layout

$\omittedCode{variable setup}{
# Space between panels
space = wood_thickness

# Variables to hold the cumulative width and the max height for the next row
cumulative_width = 0
max_height = 0
}$ 
# Side panels
for i in range(2):
    msp.add_lwpolyline([(cumulative_width, 0), (cumulative_width, side_panel_dims[1]), (cumulative_width + side_panel_dims[0], side_panel_dims[1]), (cumulative_width + side_panel_dims[0], 0), (cumulative_width, 0)])
    cumulative_width += side_panel_dims[0] + space
    max_height = max(max_height, side_panel_dims[1])
$\omittedCode{similar for loop to write the top and bottom panels}{
# Top and bottom panels
for i in range(2):
    msp.add_lwpolyline([(cumulative_width, i*(top_bottom_panel_dims[1] + space)),
                        (cumulative_width, (i+1)*(top_bottom_panel_dims[1] + space)),
                        (cumulative_width + top_bottom_panel_dims[0], (i+1)*(top_bottom_panel_dims[1] + space)),
                        (cumulative_width + top_bottom_panel_dims[0], i*(top_bottom_panel_dims[1] + space)),
                        (cumulative_width, i*(top_bottom_panel_dims[1] + space))])
    cumulative_width += top_bottom_panel_dims[0] + space
    max_height = max(max_height, top_bottom_panel_dims[1]*(i+1) + space*i)
}$

$\omittedCode{reset cumulative width and heigh variables, then for loop to write the shelves}{
# Reset cumulative_width for the next row and update max_height
cumulative_width = 0
max_height += space

# Shelves
for i in range(3):
    msp.add_lwpolyline([(cumulative_width, max_height + i*(shelf_dims[1] + space)),
                        (cumulative_width, max_height + (i+1)*(shelf_dims[1] + space)),
                        (cumulative_width + shelf_dims[0], max_height + (i+1)*(shelf_dims[1] + space)),
                        (cumulative_width + shelf_dims[0], max_height + i*(shelf_dims[1] + space)),
                        (cumulative_width, max_height + i*(shelf_dims[1] + space))])
    cumulative_width += shelf_dims[0] + space
}$
$\omittedCode{reset cumulative width and height variables, then write back panel}{
# Reset cumulative_width for next row and update max_height
cumulative_width = 0
max_height += 3*shelf_dims[1] + 2*space

# Back panel
msp.add_lwpolyline([(cumulative_width, max_height + space),
                    (cumulative_width, max_height + back_panel_dims[1] + space),
                    (cumulative_width + back_panel_dims[0], max_height + back_panel_dims[1] + space),
                    (cumulative_width + back_panel_dims[0], max_height + space),
                    (cumulative_width, max_height + space)])
}$

doc.saveas("cabinet.dxf") # Save the DXF file
\end{gptcodeblock}
\end{gptresponse}

\end{chat}
\caption{\textbf{The prompts used for generating files in the format of SVG and DXF using \gpt.}
The visualization of the graphs are shown on the right side. 
It is clear that \gpt can accomplish this task after several iterations.}
\label{fig:SVG_DXF_Gen}
\vspace{-3mm}
\end{figure}

The primary aim of our experiment was to design an SVG file for a cabinet, 
with predetermined dimensions, to be constructed from 1/2 inch plywood. 
This implies that the thickness of each wall, a preset parameter, is 0.5 inches.
The experimental setup involved the design of a cabinet comprising three shelves, 
with overall dimensions measuring 6 feet in height, 1 foot in depth, and 4 feet in width.
A crucial aspect of the investigation was to see if \gpt could accurately account for this wall thickness 
during the design of the cabinet, appropriately adjusting the dimensions of its various components.
\gpt was able to design the specified cabinet and 
subsequently generated a Python script to create an SVG file reflecting the cabinet's layout.
The script considered the necessary clearances for the thickness and 
accurately positioned the side panels, top and bottom panels, shelves, and back panel.
Moreover, it factored in the prescribed spacing between parts and 
leveraged `svgwrite' to generate the SVG file. 
The resulting SVG file provided a visual depiction of the cabinet's design.
We also replicated the experiment to create a DXF file, 
where \gpt utilized `ezdxf' to generate the file.
The results of these experiments are depicted in Figure \ref{fig:SVG_DXF_Gen}.

In conclusion, \gpt demonstrated its capability to employ the APIs for 
generating the vector file in the correct format without any simplifications.
Nevertheless, it was necessary to perform several iterations to 
ensure \gpt did not cause any overlap among the cabinet parts.

\afterpage{\FloatBarrier}

%% file: sections/GraphArrows/TextToDesignExperiments/CSG_customPrimivites_basicDesigns.tex

The next design domain we are investigating is CSG.
As outlined in Sec.\ref{sec:overview_domains_design}, CSG languages generally operate by building up a collection of primitives that have been altered or combined via linear transformations and Boolean operations. 
Because the associated design logic can be quite complex, it was not immediately clear that \gpt should be able to generate designs using these languages.
Thus, to progressively test \gpt's modeling capabilities, we begin by exploring a very simple, custom CSG language based on a single primitive: a box.

Boxes are one of the most common primitives seen in manufacturing. 
Moreover, many shapes can be considered as a combination of boxes with different sizes.
Because of the simplicity of a box or any shape formed by the boxes, 
we would like to see if \gpt is able to generate designs of such kind of simple shapes, 
such as tables and chairs.

\begin{figure}[tb]
    \centering
    \includegraphics[width=0.65\textwidth]{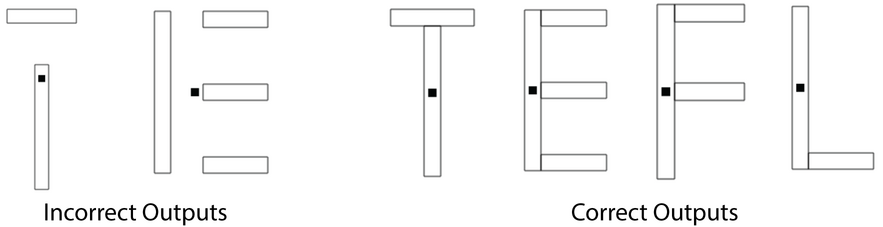}
    \caption{\textbf{Failed and Successful Cases of Letter Creation Using \gpt.} The solid square is the origin of the 2D coordinate system.}
    \label{fig:letter}
\end{figure}

Our initial approach to this task is performed in 2D.
We provide a function, foo(x, y, w, h), 
which forms a box of dimensions $w \times h$ centred at the point $(x, y)$. 
We subsequently employ this function to generate letters composed of axis-aligned bars, 
such as `F' and `E'.
During the testing phase, we observed that 
while the system understands the requirement of 2D boxes, 
it struggles with their accurate placement. 
A particularly prominent issue is the collision problem. 
More specifically, the \gpt system fails to determine 
whether two boxes are overlapping or whether there is a vacant space between them. 
This issue is observable when creating letters like `T' and `E'.
Using three to five targeted prompts enabled \gpt to ascertain the correct positions.
However, these prompts had to be granular and often involved providing the direct solution. 
The outcomes of these attempts are demonstrated in Figure~\ref{fig:letter}.
Interestingly, after addressing this issue, \gpt appears to retain the corrections. 
This is evidenced by its successful generation of the new letters `F' and `L' in a single attempt. 
These letters share a similar structure to `T' and `E', 
and the results can be seen in Figure~\ref{fig:letter}.

\begin{figure}[tb]
    \centering
    \includegraphics[width=\textwidth]{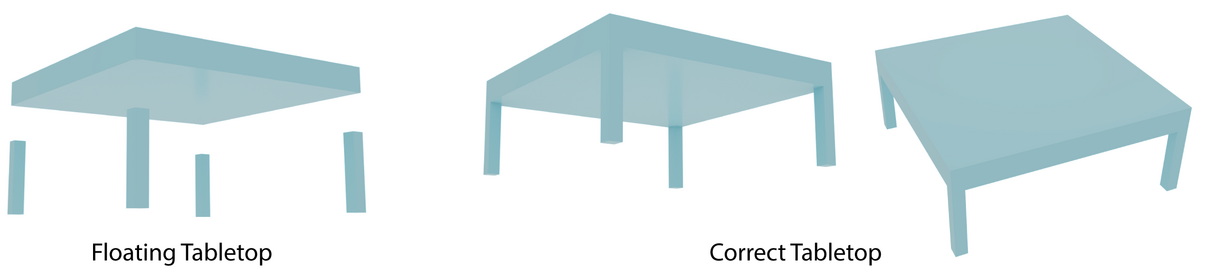}
    \caption{\textbf{Failed and Successful Cases of Table Creation Using \gpt.} 
    The table consists of five parts: 4 legs and a tabletop. 
    Although \gpt successfully gives a correct composition of the table,
    \gpt outputs a floating tabletop without any human intervention.}
    \label{fig:tabletable}
\end{figure}

Our next step involved venturing into 3D, 
which holds more practical values. 
Analogous to the 2D scenarios, 
we inform \gpt of a pre-established function, box(x, y, z, w, h, d), 
which generates a 3D box of dimensions $w \times h \times d$ centred at the 3D coordinates $(x, y, z)$.
We then tested if \gpt could write a program to produce a simple box of specified dimensions, 
for instance, $100 \times 100 \times 40$, utilizing function `box'. 
\gpt successfully accomplished this task, and the resulting text explanation 
illustrates its understanding of the box concept and the usage of our predefined function.
Next, we presented a more complex challenge: having \gpt design a simple table, 
typically consisting of four legs and a tabletop in the real world. 
We posed the question of whether \gpt could craft a program to
generate such a table with a provided size using solely our box function.
The output text explanation revealed that 
\gpt accurately comprehends the structure of a basic table.
Given that we only provide the overall table size, 
\gpt lacks information about individual leg lengths or tabletop thickness. 
Yet, it was able to identify these missing parameters and make reasonable assumptions. 
Consequently, \gpt succeeded in writing a program to represent 
the table by creating five boxes using our predefined function.
Upon visualizing the 3D table, however, 
the relative positioning of each pair of boxes was not always accurate.
We noticed that the tabletop appeared to be suspended in the air, 
not in contact with the legs, as shown in Figure~\ref{fig:tabletable}. 
This difficulty, also observed in our 2D tests (Figure~\ref{fig:letter}), 
pertains to \gpt's understanding of mathematical concepts.
In this instance, we expedited the process by directly providing \gpt with the solution.
We indicated the necessary translations for the misplaced boxes, 
acknowledging that it would take several prompts to rectify the issue otherwise. 
After correcting the floating tabletop, the table appeared as intended, as demonstrated in Figure~\ref{fig:tabletable}.
Therefore, to create a table, it only required two prompts, 
significantly streamlining the procedure for generating a basic table.

\begin{figure}[tb]
    \centering
    \includegraphics[width=\textwidth]{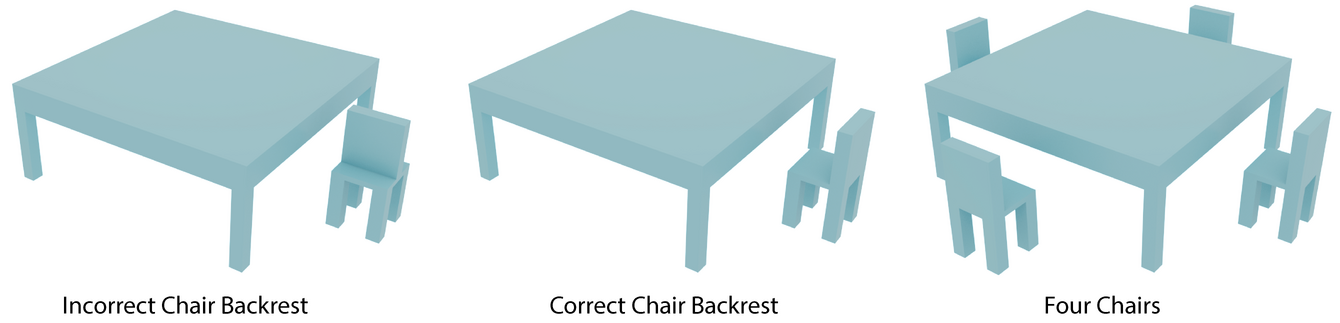}
    \caption{\textbf{Failed and Successful Cases of Chair Creation using \gpt.}
    \gpt successfully gives a correct composition of the chair. 
    In the incorrect version (left), the dimension of the backseat is
    wrong and it looks like the orientation is wrong.}
    \label{fig:chairchair}
\end{figure}

Once we successfully generate the table, 
our next more challenging goal is to design a few accompanying chairs.
We tasked \gpt with creating a chair compatible with the table, using only our predefined function. Similar to its approach with the table, 
\gpt successfully deduced the basic structure of a simple chair, 
comprising the seat, four legs, and a backrest.
Unlike the table instance, we didn't observe any `floating' issues in this scenario.
It appears that \gpt might have indeed gleaned some insights from previous experiences,
as we also observed when creating 2D letters. 
After we rectified the letters `T' and `E', 
there were no issues with the remaining letters.
Additionally, \gpt demonstrated comprehension of the concept of 
compatibility by outputting a chair of an appropriate size.
However, it was not successful in all aspects, as depicted in Figure~\ref{fig:chairchair}. 
We attempted to correct the backrest but were unable to do so.
As a result, we had to manually adjust the position,
directing \gpt to the specific lines that needed modification to correct the structure.
The final result can be seen in Figure~\ref{fig:chairchair}.
We believe the root of these issues lies in \gpt's struggles to comprehend geometric concepts,
a difficulty also observed in previous examples.
Despite these hurdles, the process for creating a basic table and chairs
has been considerably simplified.

Our final objective was to position four identical chairs around the table. 
Although theoretically feasible without invoking rotation, 
\gpt failed to generate the chairs with the correct orientations.
We believe this failure stems from the same root cause we've encountered previously, 
namely, \gpt's difficulty in handling mathematical and geometric concepts.
Creating four chairs with correct orientations without the support of rotation 
entails complex geometric transformations. 
\gpt must comprehend that a box rotated 90 degrees 
around its center is equivalent to a swap of its width and depth dimensions.
To alleviate this issue, we expanded our `box` function to include an additional input argument, `angle`, corresponding to a rotation angle around the vertical axis.
With this extension, \gpt was able to create a program using solely the `box` function 
that successfully positioned four chairs around the table with correct orientations, 
as displayed in Figure~\ref{fig:chairchair}. 
We surmise that the introduction of `angle` considerably simplifies the logic behind chair placement, 
enabling \gpt to create such a program.

In conclusion, \gpt exhibits strong understanding of posed questions 
and excels at analyzing requested objects to determine their composition.
However, it demonstrates a weakness in handling geometric and mathematical concepts.
While it can provide nearly accurate solutions when mathematics is involved,
it struggles to comprehend the underlying mathematical principles and, 
as a result, cannot independently correct math-related issues when they arise.

%% file: sections/GraphArrows/TextToDesignExperiments/CSG_PyVista_basicDesigns.tex

Building on \gpt's success generating CSG-like models with boxes, we set out to explore \gpt's capacity to use a larger suite of primitives.
For this, we used an existing 3D visualization library, PyVista, which allows us to create and place a variety of 3D primitives such as spheres and cones.
Thanks to the library's documentation, \gpt is able to automatically assemble a functional python program using PyVista's primitive functions.

\begin{figure}[tb]
    \centering
    \begin{subfigure}[b]{0.25\textwidth}
        \centering
        \includegraphics[width=.99\linewidth]{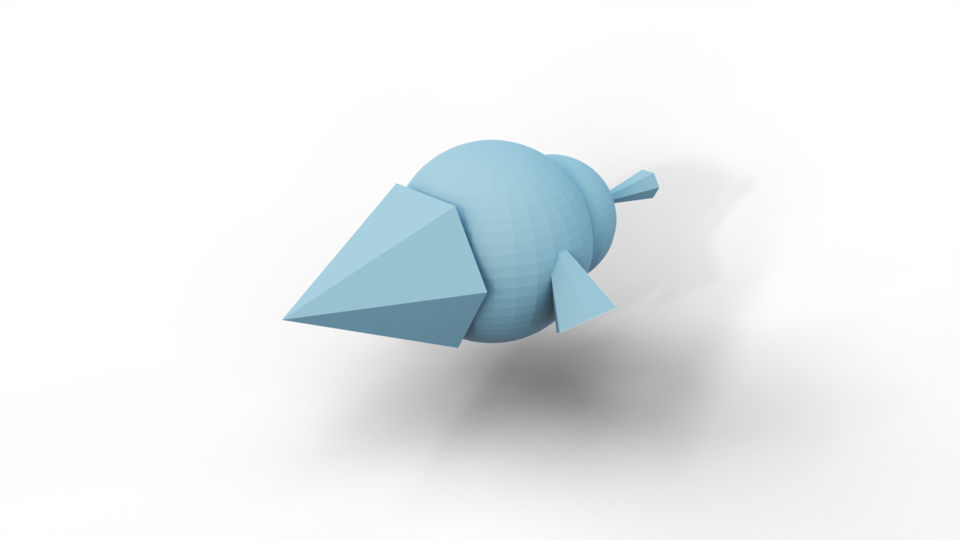}
        \caption{Generic fish}
        \label{fig:fish-orig}
    \end{subfigure}%
    \begin{subfigure}[b]{0.25\textwidth}
        \centering
        \includegraphics[width=.99\linewidth]{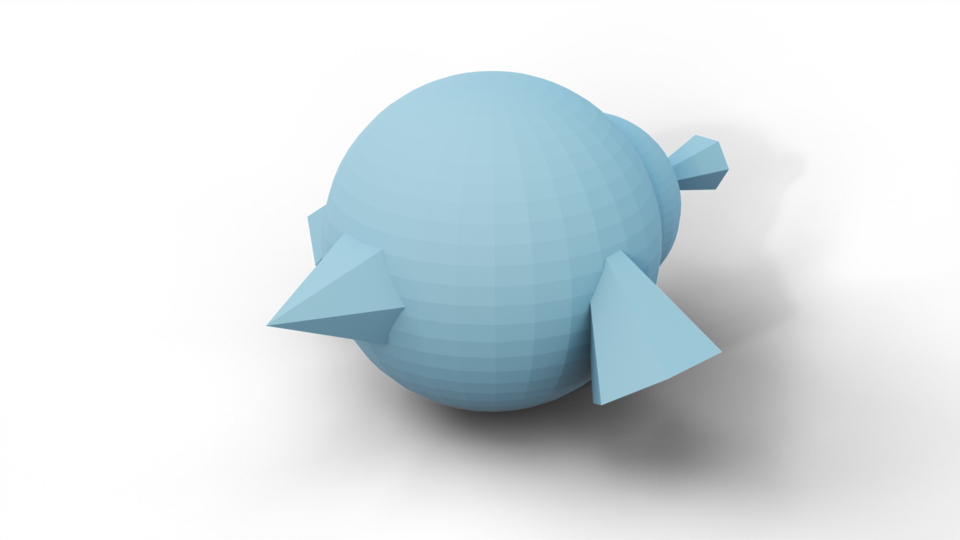}
        \caption{Goldfish}
        \label{fig:fish-goldfish}
    \end{subfigure}%
    \begin{subfigure}[b]{0.25\textwidth}
        \centering
        \includegraphics[width=.99\linewidth]{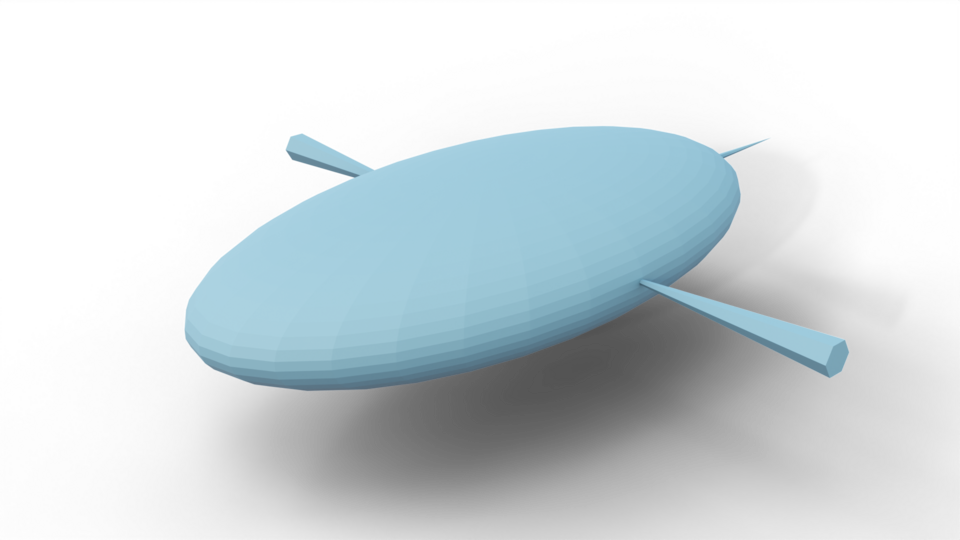}
        \caption{Manta ray}
        \label{fig:fish-mentaray}
    \end{subfigure}%
    \begin{subfigure}[b]{0.25\textwidth}
        \centering
        \includegraphics[width=.99\linewidth]{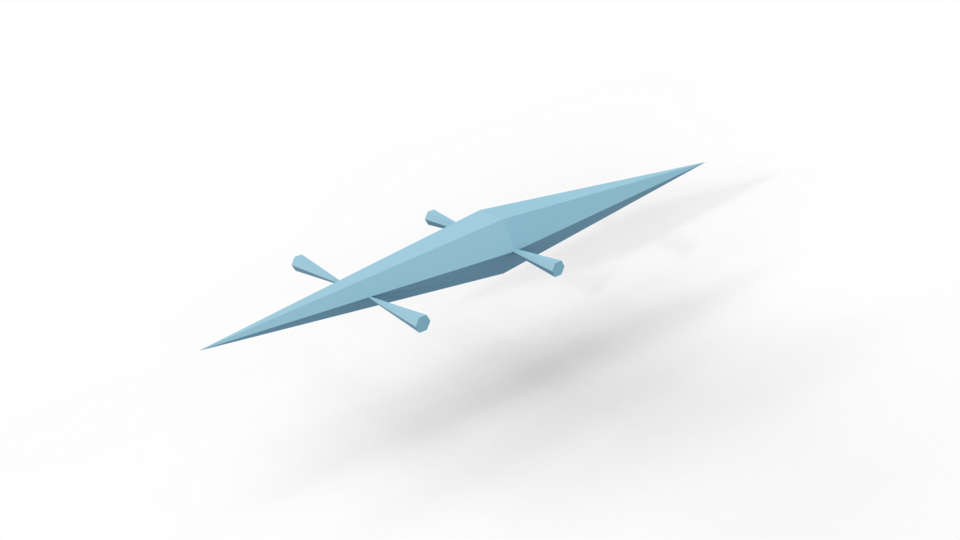}
        \caption{Loach}
        \label{fig:fish-loach}
    \end{subfigure}
    \caption{\textbf{Aquatic Creatures Generated by \gpt} \gpt successfully generated variations of aquatic creatures automatically using primitives from the PyVista package in Python. }
    \label{fig:fishes}
    \vspace{-5mm}
\end{figure}

We asked \gpt to use PyVista's primitives to model several variations of a fish, including specific bio-inspirations such as goldfish, a manta ray, and a loach (\fref{fig:fishes}). 
\gpt successfully selected and scaled an appropriate set of primitives for each example, and provided sound bio-inspired rationale for its decisions.
In particular, although most of the fish are composed using a sphere for the body, \gpt intuits that a loach would be most effectively approximated by using \lstinline{two cones for the body to give it an elongated shape}.

One area in which \gpt struggled was the determination of the primitives' orientations. 
It often produced results that indicated an internal confusion of some of the axes, or an otherwise flawed approximation of the orientation that would be required to achieve a desired effect.
After engaging in a dialogue with \gpt, it was able to rectify the orientations of the primitives to more closely resemble the target creatures.
While promising, these tests reiterate \gpt's seemingly limited capacity to account for local coordinate frames.



\ignore{
\FH{start of old text}

In this conversation, we explored using 3D mesh library to generate and visualize various designs based on constructive solid geometry (CSG) principles  with the aid of \gpt. 
The integration of PyVista, a 3D visualization library, with large language models facilitates an interactive, real-time computational design process. 
This approach simplifies 3D operations and democratizes design by allowing natural language inputs, fostering broader accessibility. 
The dynamic interaction enables efficient exploration of design parameters and serves as an instructive tool for understanding 3D modeling and computational design. 

We started by creating simple 3D primitives such as spheres. 
\gpt is able to assemble a python program automatically using PyVista's primitive functions that meets our design goals.  
Then, we expanded upon this concept to model fish (Fig.~\ref{fig:fish_and_variation}) and its variations with different bio-inspirations such as goldfish (Fig.~\ref{fig:fish_goldfish_variation}). 

We initially attempted to use boolean operations like union to combine different primitives. 
However, this process encountered some issues as PyVista's boolean operations require all the meshes to be composed of triangles, which wasn't the case with some of the primitives used by \gpt. 
This could result from the model being trained on outdated library version.

As a workaround, instead of using boolean operations, we explored a different approach where we directly added all the part meshes to the PyVista plotter for visualization. 
This method effectively bypassed the requirement for the meshes to be triangulated, enabling us to create more complex shapes from primitive elements.

Using this method, we generated a series of parametric designs, demonstrating the flexibility of this approach. We experimented with a variety of aquatic creatures, including a generic fish, a goldfish, a manta ray~\ref{fig:fish-mentaray}, and a loach~\ref{fig:fish-loach}. During the process, we discovered that \gpt has confusion in terms of how to correctly orient individual primitives to achieve desired looks, and we have to manually tweak orientations of the primitives to closely resemble the target creatures.

In conclusion, our exploration highlights the potential of large language models in aiding computational design. 
With an intuitive and interactive dialogue, we can generate a wide range of designs from basic primitives using libraries like PyVista. 
While certain limitations exist, such as \gpt's failure to meet the requirement for triangulated meshes in boolean operations (which could result from being trained from old codebases) as well as confusion in coordinate systems, creative solutions can be found to bypass these and generate complex and versatile designs. 
This shows that with the right tools and approach, large language models can significantly aid in the computational design process, making it more accessible and intuitive.
}

\ignore{
\begin{figure}[h]
\begin{chat}{false} 
\begin{userprompt}{0.2\textwidth}{}
Now fix the primitives and the boolean operations you use in the code. Try creating different variations of the fish by simply changing the parameters of the primitives. Can you make the fish more look like a goldfish? 
\end{userprompt}

\begin{gptresponse}{0.2\textwidth}{figures/fish-goldfish.png} 

 \begin{gptcodeblock}{Python}
import pyvista as pv
import numpy as np

# Create main body (larger for a goldfish)
body = pv.Sphere(center=(0, 0, 0), radius=1.0)
# Create front body as cone (smaller and repositioned for a goldfish)
front_body = pv.Cone(center=(1.2, 0, 0), direction=(1, 0, 0), height=0.8, radius=0.4)
# Triangulate the front body
front_body = front_body.triangulate()
# Create back body (smaller and repositioned for a goldfish)
back_body = pv.Sphere(center=(-0.8, 0, 0), radius=0.6)
# Create tail (smaller and repositioned for a goldfish)
tail = pv.Cone(center=(-1.6, 0, 0), direction=(1, 0, 0), height=0.8, radius=0.2)
# Triangulate the tail
tail = tail.triangulate()
# Create fins (larger and repositioned for a goldfish)
fin1 = pv.Cone(center=(0, 1.0, 0), direction=(0, -1, np.sqrt(2)), height=1.0, radius=0.4)
fin2 = pv.Cone(center=(0, -1.0, 0), direction=(0, 1, np.sqrt(2)), height=1.0, radius=0.4)
# Triangulate the fins
fin1 = fin1.triangulate()
fin2 = fin2.triangulate()
# Combine body parts
fish = body.boolean_union(front_body)
fish = fish.boolean_union(back_body)
fish = fish.boolean_union(tail)
fish = fish.boolean_union(fin1)
fish = fish.boolean_union(fin2)
# Create a plotter object
plotter = pv.Plotter()
# Add the fish to the plotter
plotter.add_mesh(fish)
# Show the fish
plotter.show()
\end{gptcodeblock}
This code creates a fish that has a larger, more rounded body and smaller, more forward-positioned head, which are more characteristic of a goldfish. The fins are also larger and positioned higher and lower on the body to mimic the large, flowing fins of some types of goldfish.
\end{gptresponse}

\end{chat} 
\caption{\label{fig:fish_goldfish_variation}\gpt modifies the previous fish example to create a parametric goldfish variation by fixing the functions used and only changing the parameters to the function calls.}
\end{figure}
\afterpage{\FloatBarrier}
}

%% file: sections/GraphArrows/TextToDesignExperiments/CSG_JSCAD_basicDesigns.tex

To explore a full-fledged approach for \llm-aided CSG, we test \gpt's ability to generate meaningful designs using the open source javascript-based CSG library, \jscad \citep{jscad}.
\jscad has extensive documentation available online, and we found that \gpt natively possesses a good grasp of the API, its components, and the required code structure.
In particular, it understood that it needed to import each function from the corresponding modules, and that it needed to define and export a function named \lstinline{main}.
For our experiments, we provided \gpt with access to the full API, and generally allowed it to select the appropriate primitives and operations without user interference. 

\begin{figure}[tb]
\begin{chat}{false} 
    \begin{userprompt}{0\textwidth}{} 
    Let's use \jscad to design a cabinet ... with 3/4" wood sheets, and final exterior dimensions of 30 inches tall, 20 inches wide, and 18 inches deep. The stationary part of the cabinet should be comprised of 6 boards: bottom, top, back, two sides, and one shelf centered inside the cabinet. 
    \omitted{\jscad modeling hints; see \appref{sec:appx-jscad-hints}}{The components ... should be returned as an array of individual components. Use the colorize function to give each component a unique color for visualization.}
    \end{userprompt}
    

    \begin{gptresponse}{0.16\textwidth}{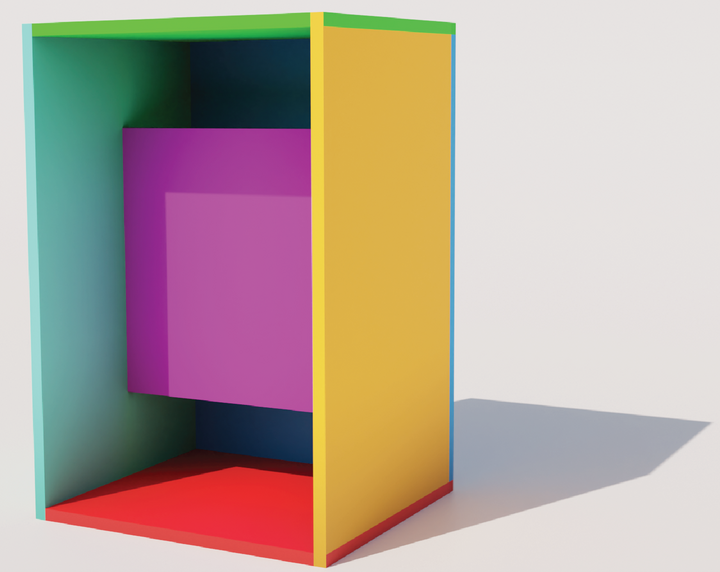} 

\begin{gptcodeblock}{JavaScript}
$\omittedCode{import statements and variable setup}{
const { colorize } = require('@jscad/modeling').colors
const { cuboid } = require('@jscad/modeling').primitives
const { translate } = require('@jscad/modeling').transforms

const thickness = 0.75
const height = 30
const width = 20
const depth = 18

function main() [IF COMMENTED BACK IN, REPLACE opening CURLY BRACE]
  const innerWidth = width - (2 * thickness)
  const innerHeight = height - (2 * thickness)
  const innerDepth = depth - thickness
  const pieces = []
}$
  pieces.push(colorize([1, 0, 0], translate([0, 0, -height/2 + thickness/2], 
                                    cuboid({size: [width, depth, thickness]})))) // bottom
  pieces.push(colorize([0, 1, 0], translate([0, 0, height/2 - thickness/2], 
                                    cuboid({size: [width, depth, thickness]})))) // top
  pieces.push(colorize([0, 0, 1], translate([0, -depth/2 + thickness/2, 0], 
                                    cuboid({size: [width, thickness, height]})))) // back
  pieces.push(colorize([1, 1, 0], translate([-width/2 + thickness/2, 0, 0], 
                                    cuboid({size: [thickness, depth, height]})))) // left
  pieces.push(colorize([0, 1, 1], translate([width/2 - thickness/2, 0, 0], 
                                    cuboid({size: [thickness, depth, height]})))) // right
  pieces.push(colorize([1, 0, 1], translate([0, -thickness/2, 0], 
                                    cuboid({size: [innerWidth, thickness, innerDepth]})))) // shelf
$\omittedCode{rest of design}{
  return pieces
[IF COMMENTED BACK IN, REPLACE closing CURLY BRACE]
module.exports = { main }
}$
\end{gptcodeblock}

    \end{gptresponse}

    \begin{userprompt}{0\textwidth}{} 
This is a great first guess! However, there are two issues. 
First, the shelf is oriented vertically, rather than horizontally. 
Second, there are many cases where the boards are protruding into one another. 
\omitted{description of the desired relative extent of each board}{
Can you adjust the design such that: 
1) the side and back boards are fully contained between the top and bottom boards; 
2) the back panel extends across the full width of the top/bottom boards; 
3) the side panels only extend until they contact the front face of the back panel
}
    \end{userprompt}
    
    
\begin{gptresponse}{0.16\textwidth}{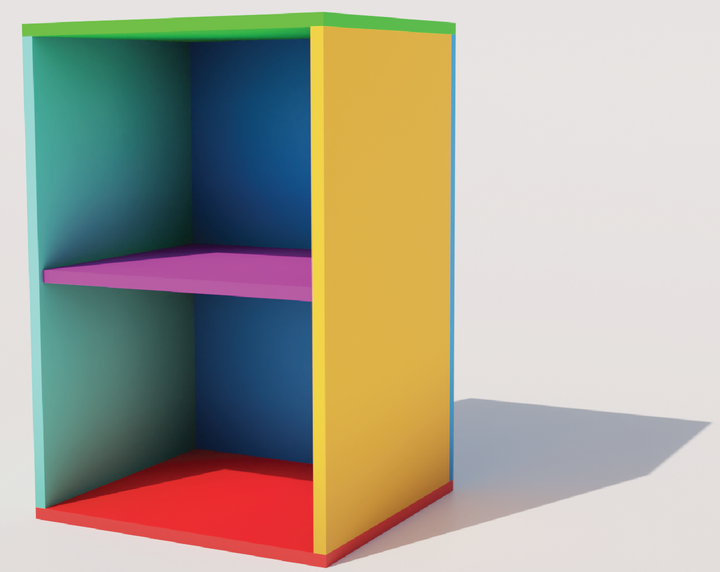}
\begin{gptcodeblock}{JavaScript}
$\omittedCode{preceding code was unchanged}{
  pieces.push(colorize([1, 0, 0], translate([0, 0, -height/2 + thickness/2], 
                                    cuboid({size: [width, depth, thickness]})))) // bottom
  pieces.push(colorize([0, 1, 0], translate([0, 0, height/2 - thickness/2], 
                                    cuboid({size: [width, depth, thickness]})))) // top
}$
  pieces.push(colorize([0, 0, 1], translate([0, -depth/2 + thickness/2, 0], 
                                    cuboid({size: [width, thickness, innerHeight]})))) // back
  pieces.push(colorize([1, 1, 0], translate([-width/2 + thickness/2, thickness/2, 0], 
                                    cuboid({size: [thickness, innerDepth, innerHeight]})))) // left
  pieces.push(colorize([0, 1, 1], translate([width/2 - thickness/2, thickness/2, 0], 
                                    cuboid({size: [thickness, innerDepth, innerHeight]})))) // right
  pieces.push(colorize([1, 0, 1], translate([0, 0, 0], 
                                    cuboid({size: [innerWidth, innerDepth, thickness]})))) // shelf
$\omittedCode{subsequent code was unchanged}{}$
\end{gptcodeblock}
\end{gptresponse}
    
\end{chat}
\caption{\textbf{\gpt's cabinet design process.} Simple cabinet design, from initial prompt to the final fabricable result, in which all elements have the correct orientation and no components overlap. \gpt reached the final result in 8 messages (4 each of prompt/response).}
\label{fig:textToDesign_simpleCabinet}
\end{figure}

To test \gpt's design abilities, we ask it to design a simple cabinet with one shelf, as shown in \fref{fig:textToDesign_simpleCabinet}.
\gpt reliably selects and instantiates the required primitives, along with intuitive naming conventions and structure within the \jscad code. 
\gpt's initial orientation of the parts was also generally reasonable, but the specific positioning of each part was often incorrect. 
Despite multiple attempts, \gpt was unable to generate any fully-correct cabinet in a single shot, with no subsequent user intervention.

Moreover, \gpt frequently produced highly disparate results from one run to the next. Even when using an identical prompt on fresh chat environments, \gpt's responses varied widely in terms of their overall code structure, design accuracy, and the specific errors or oversights made. 
\fref{fig:cabinet_vertical_explosion} shows one example of a drastically different design process, even when seeded with the same initial prompt as \fref{fig:textToDesign_simpleCabinet}.

Throughout our experiments, we found that \gpt encountered a few common pitfalls when generating designs in \jscad.
Occasionally, \gpt made small syntatic errors such as generating incorrect boilerplate, importing functions from incorrect modules, or making ``typos'' in API calls -- \eg, trying to import from the \lstinline{boolean} module rather than the correct \lstinline{booleans} module, or calling the \lstinline{cube()} function with parameters that were intended to generate a \lstinline{cuboid()}. 
In an attempt to avoid these pitfalls, we created a small list of ``hints''/``reminders'' for best practices when working with \jscad; this short list was always passed in alongside our initial prompt. See \appref{sec:appx-jscad-hints} for a full listing of these reminders.
Although these reminders seemed to help mitigate these issues, we were unable to eradicate them entirely. 
However, \gpt can easily correct the majority of these issues when they were pointed out by the user.
Often, the process of correcting the issue through prompts and responses was faster than actually adjusting the code manually, making {\llm}s a useful design partner.

One pervasive issue that seemed more difficult to correct was the fact that \gpt had issues positioning the primitives in 3D space.
In particular, \gpt frequently seemed to forget that \jscad positions elements relative to the \textit{center} of a given primitive, rather than an external point on the primitive (\eg, the lower left corner). 
\gpt's arrangements were frequently incorrect due to this issue.
When \gpt is reminded of this convention, it does generally alter the design, but it is not always able to correct the issue. 
If sufficiently many rounds of local edits prove unable to address the alignment issues, we found that it was generally more effective to direct \gpt to disregard all existing measurements, and re-derive the elements' positions from scratch (see \fref{fig:cabinet_vertical_explosion}).

Overall, we find that \gpt is able to generate reasonable \jscad models from high-level input.
However, the design specifications that emerge on the first attempt are rarely fully correct, so users should expect to engage in some amount of corrective feedback or iteration in order to attain the desired result.

\newcommand{\cabVertExpImHeight}{2.2cm}
\begin{figure}[tb]
    \centering
    \includegraphics[height=\cabVertExpImHeight]{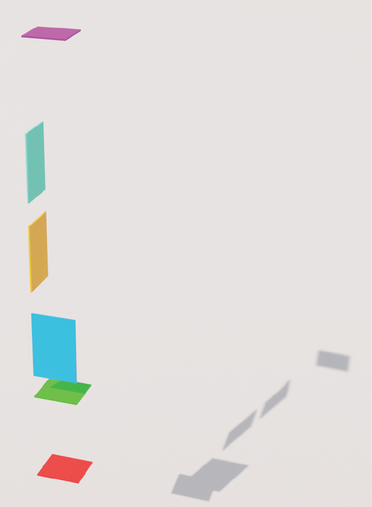}
    \includegraphics[height=\cabVertExpImHeight]{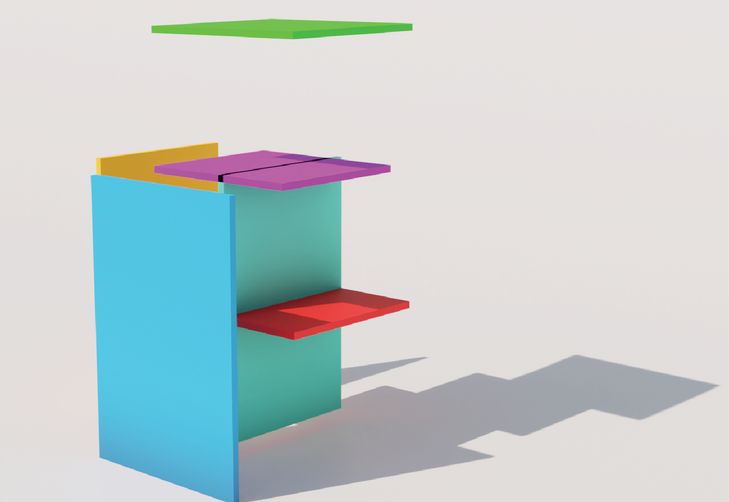}
    \includegraphics[height=\cabVertExpImHeight]{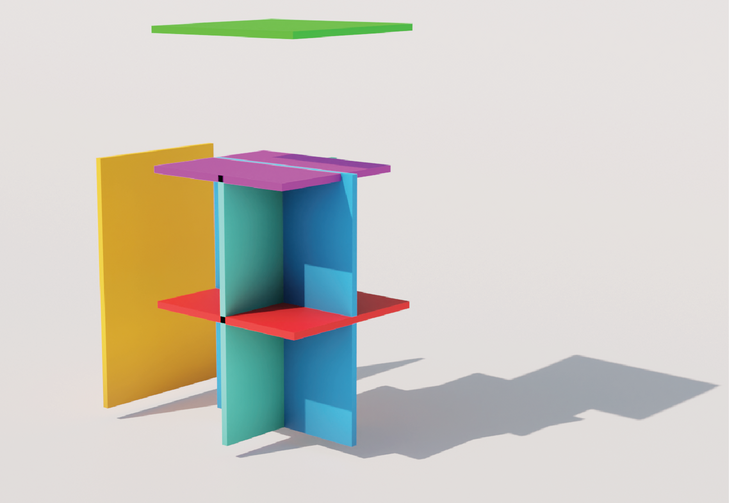}
    \includegraphics[height=\cabVertExpImHeight]{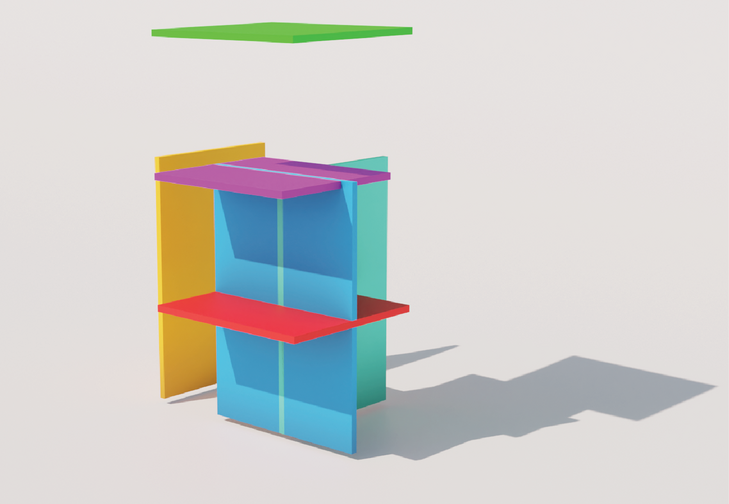}
    \includegraphics[height=\cabVertExpImHeight]{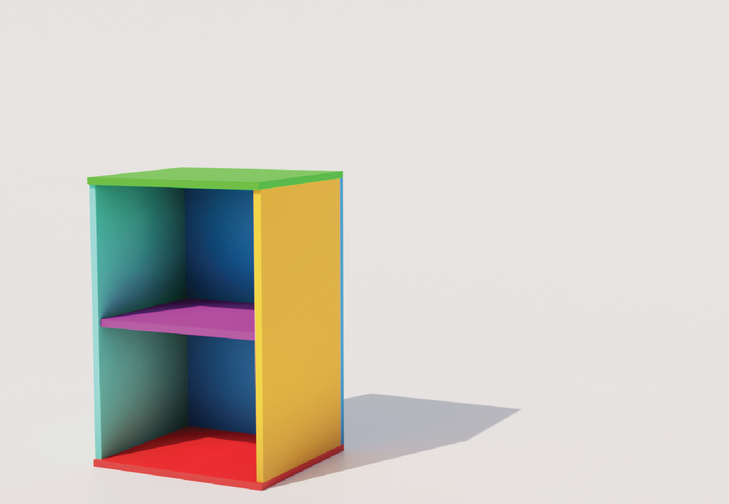}
    \caption{\textbf{Alternate cabinet design process by \gpt.} From left to right, these images show a second example of \gpt's design process for the simple cabinet, beginning from the same prompt as \fref{fig:textToDesign_simpleCabinet}. This example highlights the widely varying paths that \gpt may take to construct a design. \gpt reached the final result after 16 messages (8 each of prompt/response). Up until the fourth displayed result, the user prompts' phrasing focused on improving upon the previous result. However, after failing to make progress, we asked \gpt to disregard the previous attempts and recompute the elements' positions from scratch; this quickly resulted in a valid cabinet, as shown in the final image.}
    \label{fig:cabinet_vertical_explosion}
\end{figure}

%% file: sections/GraphArrows/TextToDesignExperiments/CAD_OnShape_basicDesigns.tex

Another popular method for 3D shape modeling comes from contemporary computer-aided design (CAD) software.
Rather than directly constructing and modifying solid primitives (as in the CSG approaches discussed above), modern parametric CAD systems generally work by lifting planar sketches into 3D and subsequently modifying the 3D geometry.
These sketches are placed on planes, which can be offsetted construction planes, or planar faces of the current 3D model.
The selected sketching plane serves as a local coordinate system in which the sketch primitives are defined.
In graphical user interfaces, this change of coordinate systems is accounted for by letting the user easily align their camera view to a top down view onto the sketch plane.
This change of view effectively comes back to drawing sketches in 2D, removing the cognitive burden of having to think about sketches in 3D.
Despite the lack of graphical assistance, we want to investigate whether \gpt is able to design objects using a sketch-based modeling language.

However, since the graphical assistance is very prevalent in this modeling paradigm, CAD models are mostly constructed via a GUI and not via textual programming, even though textual APIs exist, e.g. Onshape's Featurescript \cite{featurescript}.
Therefore, documentation and examples are less available than for the modeling paradigms from the previous sections.
And indeed, \gpt performs poorly when trying to generate Featurescript code directly, which is why we decided to provide a simplified DSL.

For our experiments, we constructed a single prompt containing the following DSL description:
Our DSL exposes two operators, \lstinline{createSketch} and \lstinline{extrude}, and two sketch primitives, \lstinline{circle} and \lstinline{rectangle}.
Additionally, we provide a construction example using this language of a single leg round table.
Lastly, we also add some hints about how to write the program, e.g. to explicitly use design variables and to write in \lstinline{syntactically correct python}.
All of the output designs generated by \gpt in this section are automatically translated into Onshape PartStudios.
The full prompt can be found in the supplemental material. 
\begin{figure}[tb]
\begin{chat}{false} 
\begin{userprompt}{0.2\textwidth}{} 
Can you design a chair with 4 legs, a rectangular back and a rectangular backrest?
\end{userprompt}


\begin{gptresponse}{0.2\textwidth}{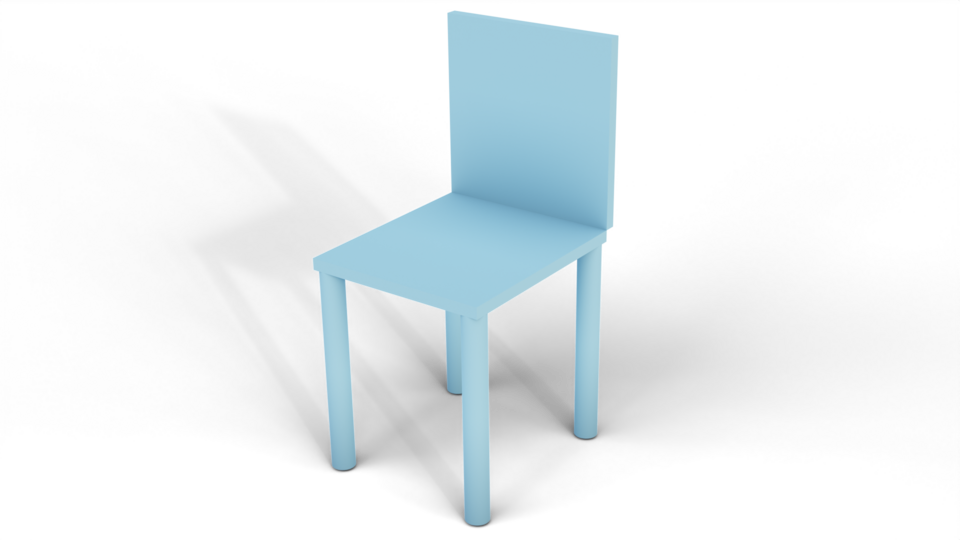} 
\begin{gptcodeblock}{Python}
$\omittedCode{variable setup}{}$
seat_sketch = createSketch(rectangle(0, 0, seat_length, seat_width), XY_PLANE)
seat_solid = extrude(seat_sketch, seat_thickness)

leg_sketch_1 = createSketch(circle(-3.5, -3.5, leg_radius), cap(seat_solid, "min_z"))
leg_solid_1 = extrude(leg_sketch_1, leg_length)
$\omittedCode{rest design}{}$
\end{gptcodeblock}
\end{gptresponse}

\begin{gptresponse}{0.2\textwidth}{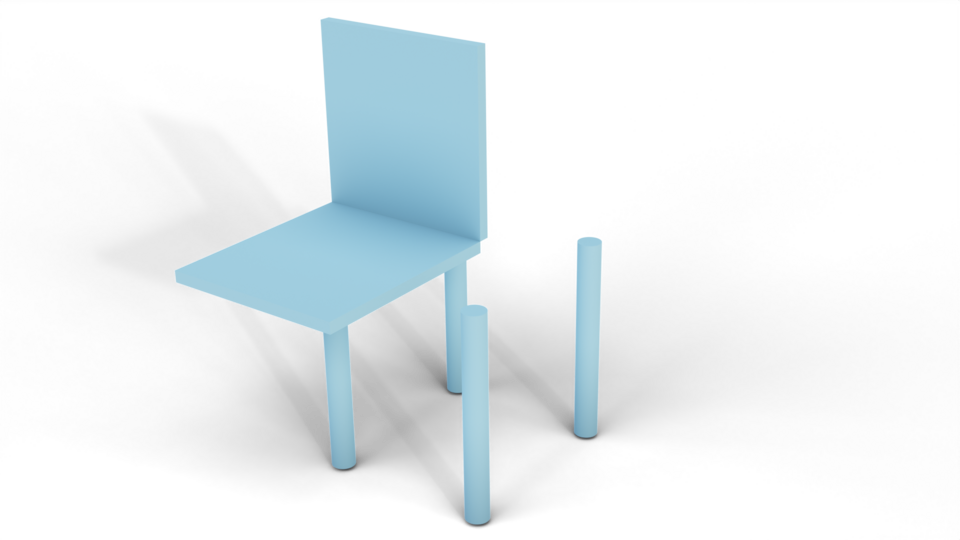} 
    \begin{gptcodeblock}{Python}
$\omittedCode{variable setup}{}$
leg_sketch_1 = createSketch(circle(-3.5, -3.5, leg_radius), XY_PLANE)
leg_solid_1 = extrude(leg_sketch_1, leg_length)
// other legs
seat_sketch = createSketch(rectangle(0, 0, seat_length, seat_width), cap(leg_solid_1, "max_z"))
seat_solid = extrude(seat_sketch, seat_thickness)
$\omittedCode{rest design}{}$
\end{gptcodeblock}
\end{gptresponse}

\end{chat}
\caption{\label{fig:cad_chair_design}\textbf{Different Chair Design Strategies.} \gpt designs a chair in two different ways. When starting with the seat, it does not have to change coordinate systems when specifying sketch coordinates of the legs. However, sketching a seat on the top face of one of the legs requires a coordinate change which \gpt does not manage to do.}
\vspace{-5mm}
\end{figure}

Our first task is the design of a \lstinline{chair with 4 legs, a rectangular seat and a rectangular back}, see Fig.~\ref{fig:cad_chair_design}.
We asked \gpt to perform this task several times and observed the following.
\begin{itemize}
    \item The design sometimes includes cylindrical legs, sometimes rectangular legs.
    \item The design is always constructed in a single direction, the $Z$ direction.
    Our input example of the round table only used the $Z$ direction to select sketching planes, but the description of our language documented the use of other plane directions.
    \item We observe mainly two types of designs: (i) designs which are constructed in both the negative and positive $Z$ direction starting from the seat, see first answer in Fig.~\ref{fig:cad_chair_design}, and (ii) designs which start from a leg, see the second response in Fig.~\ref{fig:cad_chair_design}.
    We observe that the first type of designs has a higher chance of being correct, whereas the second type fails more often.
    The failures are due to changes in the coordinate system.
    For example, when selecting the top plane of the first leg as a sketch plane for the seat, the sketch plane's origin will be in the center of the leg.
    \gpt will often ignore this or won't be able to account for it when pointed out.
    Conversely, when starting with the seat and choosing the lower seat plane as a sketch plane for the legs, it can specify the leg sketch coordinates in global coordinates, since the global origin coincides with the seat's origin.
    The same is true for the backrest.
    
\end{itemize}

From this test, we can observe that \gpt seems to have difficulties translating the coordinate system's origin on the XY plane.


\begin{figure}[tb]
\begin{chat}{false} 
\begin{userprompt}{0.2\textwidth}{} 
Can you design a car?
\end{userprompt}


\begin{gptresponse}{0.2\textwidth}{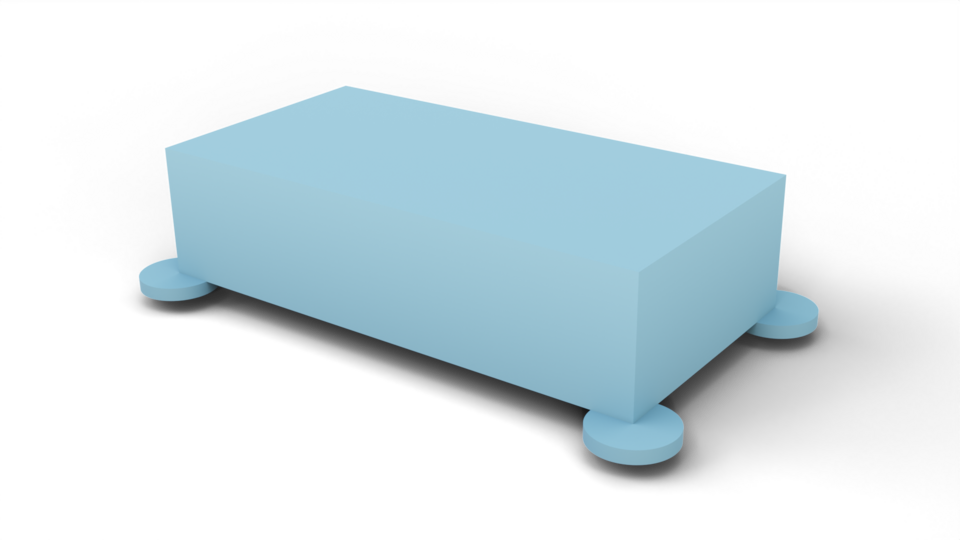} 
\begin{gptcodeblock}{Python}
$\omittedCode{variable setup}{}$
body_sketch = createSketch(rectangle(0, 0, bodyLength, bodyWidth), XY_PLANE)
body_solid = extrude(body_sketch, bodyHeight)

wheel_sketch1 = createSketch(circle(-bodyLength/2, -bodyWidth/2, wheelRadius), cap(body_solid, "min_z"))
wheel_solid1 = extrude(wheel_sketch1, wheelWidth)
$\omittedCode{other wheels}{}$
\end{gptcodeblock}
\end{gptresponse}

\begin{gptresponse}{0.2\textwidth}{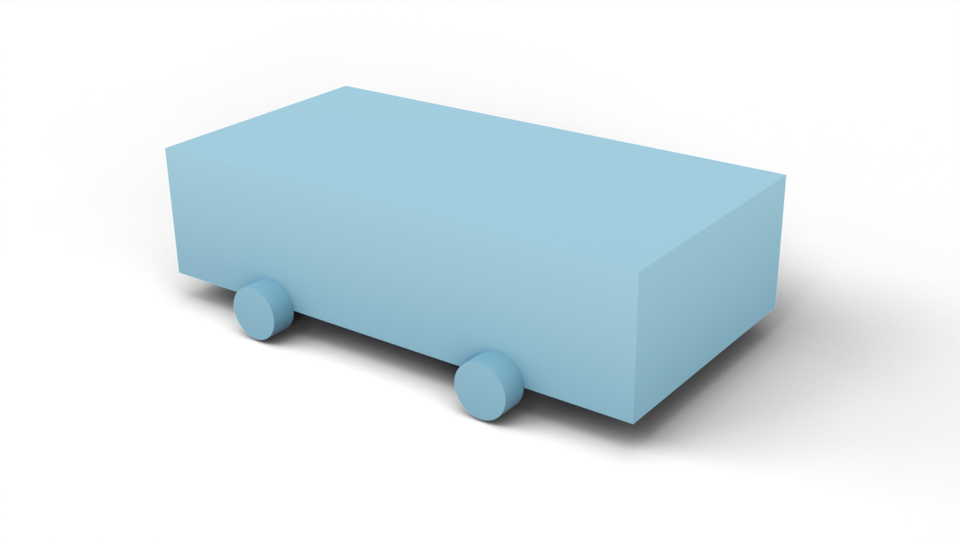} 
    \begin{gptcodeblock}{Python}
$\omittedCode{variable setup}{}$
body_sketch = createSketch(rectangle(0, 0, 0, bodyLength, bodyWidth), XY_PLANE)
body_solid = extrude(body_sketch, bodyHeight)

wheel_sketch1 = createSketch(circle(-bodyLength/4, -bodyWidth/2 - wheelWidth/2, 0, wheelRadius), cap(body_solid, "min_y"))
wheel_solid1 = extrude(wheel_sketch1, wheelWidth)
$\omittedCode{other wheels}{}$
\end{gptcodeblock}
\end{gptresponse}

\end{chat}
\caption{\label{fig:cad_car_design}\textbf{Local vs Global Coordinate Systems.} \gpt does not manage to rotate the wheel sketch planes. If we allow it to express sketch primitives in global coordinates and select a plane to orient the extrude operation, it places them correctly.}
\vspace{-3mm}
\end{figure}

Next, we want to see if \gpt can account for rotating sketch planes.
To test this, we ask it to design a car.
\gpt always suggests a simple car shape, composed out of 4 cylindrical wheels and a rectangular car body, see Fig.\ref{fig:cad_car_design}.
The difficulty with this shape is that the cylinder sketches of the wheels have to be extruded on the side planes of the car body.
There are a couple different modeling strategies to achieve this, but we observe that \gpt has difficulties coming up with these designs without any further indication.
Instead, it often extrudes the car body along its \lstinline{height}, starting from the ground plane, and then places the wheel circles on the bottom plane of the car, which is also the ground plane.
This has the effect that the car wheels will be extruded vertically.
Although we were able to correct this design in an iterative prompt-based fashion, we had little success engineering the initial prompt in such a way that we could effectively prevent this behavior.

Note that intuitively placing wheels at the bottom of a car body makes sense and that without any graphical feedback, humans could also easily make this mistake.
From this test, we can observe that \gpt is struggling to rotationally change coordinate systems.

To address this, we changed our design language description to allow \gpt to specify sketch primitive coordinates directly in a single global coordinate system. 
Now, a sketch primitive center takes as input three coordinates, which we project in post-processing directly on the selected sketch plane.
The extrude direction is still defined by the sketch plane's normal vector.
This means that \gpt does not have to take coordinate translations into account anymore.
We observe that this change in the DSL led to a higher success rate in generated designs, see second answer in Fig.~\ref{fig:cad_car_design}.

In conclusion, \gpt is able to design models in a sketch-based parametric CAD framework. 
However it is not successful at changing coordinate systems.
In this case, our backup strategy is to use a single global coordinate system.
One possible future direction is to let \gpt communicate with a geometric solver and create a feedback loop.

\afterpage{\FloatBarrier}

%% file: sections/GraphArrows/TextToDesignExperiments/Robots_URDF_basicDesigns.tex


\begin{figure}[tb]
\begin{chat}{false} 

\begin{userprompt}{0.03\textwidth}{} 
Generate a urdf for a two link robot. The end of the second link should connect to the end of the first link with a revolute joint. The origin of the joint should be a the end of the first link
\end{userprompt}

\begin{gptresponse}{0.2\textwidth}{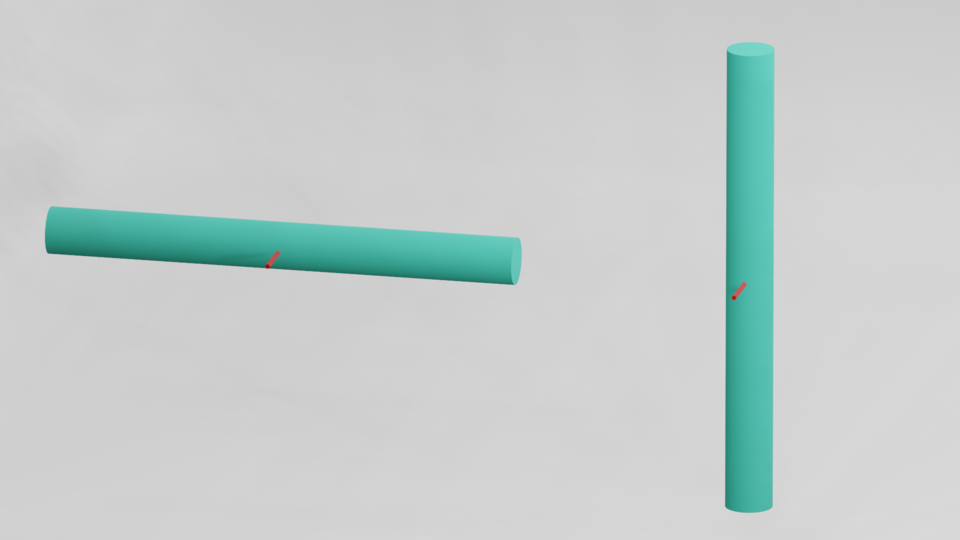} 
\begin{gptcodeblock}{XML}
  <joint name="joint1" type="revolute">
        <axis xyz="0 0 1" />
        <parent link="link1" /><child link="link2" />
        <origin xyz="1 0 0" rpy="0 0 0" />
  </joint>
  
  <link name="link1"><visual><geometry><box size="0.1 1 0.1" /></geometry></visual></link>
  <link name="link2"><visual><geometry><box size="0.1 1 0.1" /></geometry></visual></link>
\end{gptcodeblock}
\end{gptresponse}

\begin{userprompt}{0.03\textwidth}{} 
urdf link origins are in the center of the link. To make a joint origin at the end of a link, it must be half the link's length past the link origin. 
Generate a urdf for a 2 link robot. The end of the second link should connect to the end of the first link with a revolute joint. The origin of the joint should be at the end of the first link. Both links should be lined up along the z axis.
\end{userprompt}

\begin{gptresponse}{0.2\textwidth}{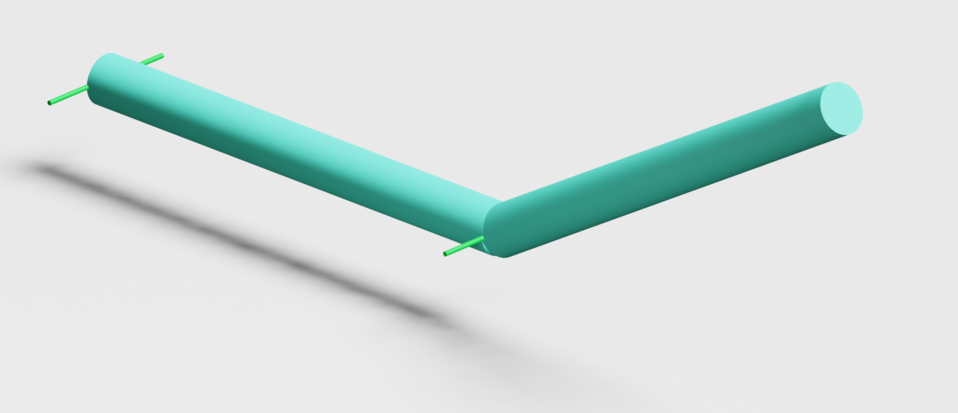}
\omitted{URDF description of an appropriately connected two-link robot}{}
\end{gptresponse}

\begin{userprompt}{0.03\textwidth}{}
    urdf link origins are in the center of the link. To make a joint origin at the end of a link, it must be half the link's length past the link origin.
    Generate a urdf for a three link robot. The end of the second link should connect to the end of the first link with a revolute joint about the x axis, the origin of the first joint should be a the end of the first link. The other end of the second link should connect to the end of the third link via a revolute joint about the x axis with an origin at the end of the third link. All three links should be lined up along the z axis
\end{userprompt}

\begin{gptresponse}{0.2\textwidth}{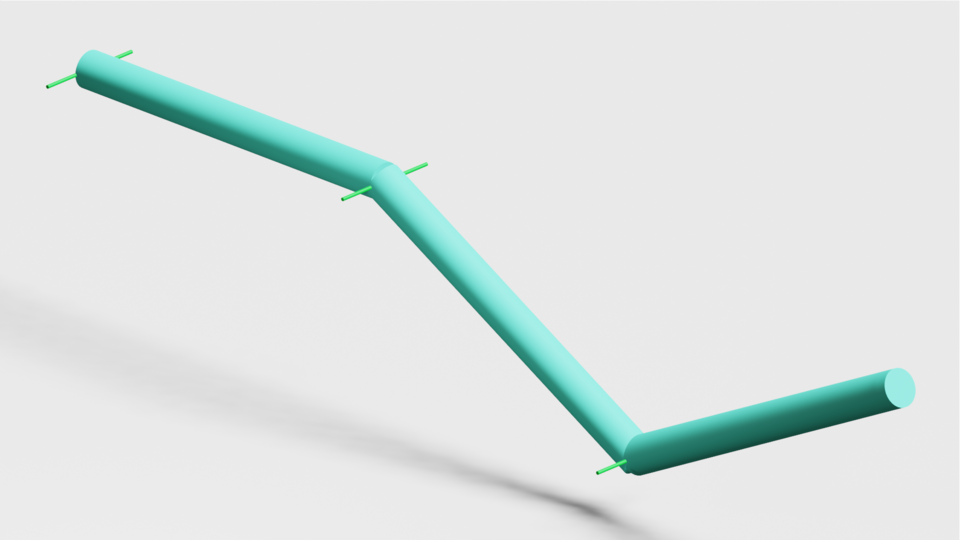}
\omitted{URDF description of an appropriately connected three-link robot}{}
\end{gptresponse}

\end{chat}
\caption{\label{fig:multi-link-urdf} \textbf{\gpt's Process for Designing N-Link Robot Arms.} \gpt is unable to correctly interpret the initial high-level specification, but when provided with more explicit detail, \gpt is able to generate correct URDF specifications for N-link robots.}
\vspace{-5mm}
\end{figure}

The Universal Robot Description Format (URDF) is a common XML-based language for describing articulated structures in robotics. 
URDF files specify a robot's structure (including both visual and collision geometry), joint locations, and dynamics information. 
The URDF format appears well-suited for potential \llm design because it is human-readable and heavily documented online.

\paragraph{Open Chain Robot Arms}
Initially, we asked \gpt to generate simple open chain robots (commonly called ``arms'') with a particular number of links. 
However, when we used the word ''arm'' to prompt \gpt to generate a robot, \gpt was unable to determine that the links should connect at the end.
Most often, \gpt placed the joints such that each link revolved about its center, and the links were not connected to each other (\fref{fig:multi-link-urdf}, initial prompt).
As shown in the subsequent prompts of \fref{fig:multi-link-urdf}, to achieve an arm with two connected links, it was necessary to describe both the joint position relative to the link 
(\lstinline{``the joint origin must be half the link's length past the link origin''}, rather than \lstinline{``the joint origin should be at the end of the link''} ) 
as well as the joint axis (\lstinline{``a revolute joint about the x axis''}).
Given this prompt pattern, \gpt was easily able to generate proper N-link robots. 

\begin{figure}[tb]
\centering
\includegraphics[width=0.6\textwidth]{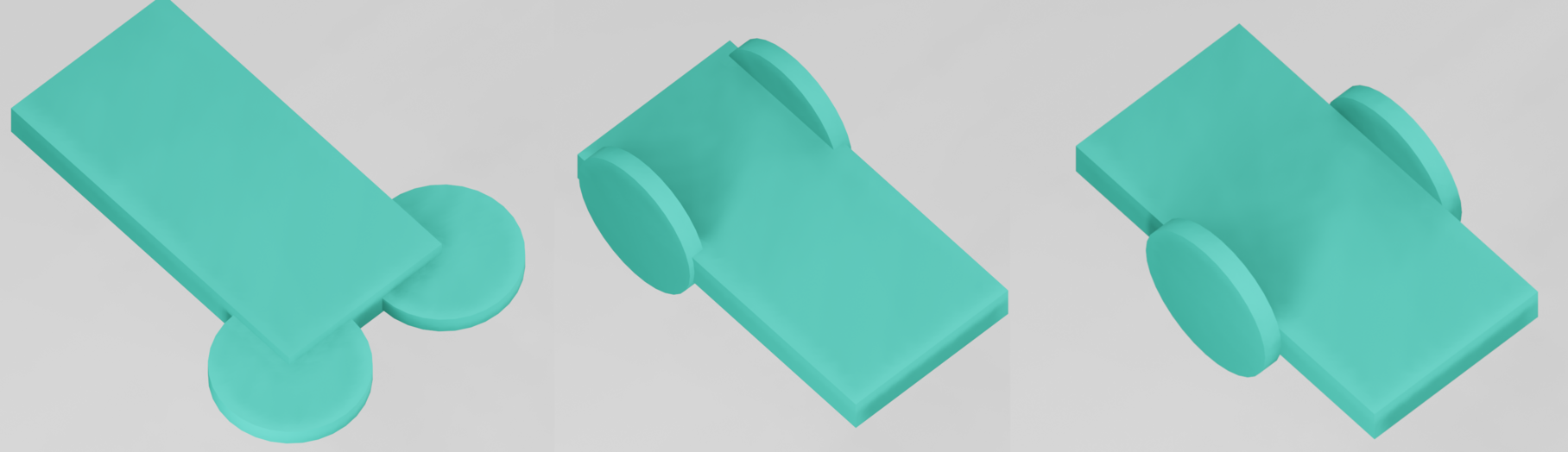}
\caption{\textbf{Two Wheeled Robot Attempts by \gpt.} \gpt's attempts to generate a two-wheeled robot with varying types of prompt constraints on geometry. \textbf{Left:} no constraints \textbf{Center:} language constraints \textbf{Right:} numeric constraints.}
\label{fig:urdf-wheeled-constraints}
\end{figure}

\paragraph{Wheeled Robots}
Next, we asked \gpt to generate wheeled robots composed of N wheels attached to a central rectangular platform.
A proper design of this type must have wheels that 
(1) are aligned to share an axis of rotation normal to and through the center of their circular faces;
(2) have circular faces displaced along said axis of rotation, and 
(3) contact, but do not intersect, either side of the center platform.
The combination of non-intersection and geometry relation constraints prove challenging for \gpt, which seems to exhibit limited geometric reasoning. 
Initially, we tried to specify these using language-based constraints (i.e. ``the wheels should touch, but not intersect, either side of the platform'').
These proved ineffective, as shown in \fref{fig:urdf-wheeled-constraints} (middle).
To overcome these challenges, we crafted prompts with very explicit numeric constraints (i.e. ``wheels should be offset on the global y axis by half the width of the platform plus half the height of the wheel cylinder''). 
This style of prompt successfully generated a viable result, as shown in \fref{fig:urdf-wheeled-constraints} (right).

\begin{figure}[tb]
\begin{chat}{false} 
\begin{userprompt}{0\textwidth}{} 
I want you to generate a urdf for a robot with two wheels. Follow the below steps:
1. generate a center rectangular platform
2. add two cylinders for wheels. The wheels should be oriented so that the global y axis is normal to the center of their circular faces, and should rotate about their cylinder's z axis
3. The wheels should be offset on the global y axis so that they are on either side of the center rectangular platform. They should be offset by half the width of the platform plus half the height of the wheel cylinder.
\end{userprompt}
\authorremark{The above generates a two wheel robot. The following prompt extends this to create a four wheeled robot}
\begin{userprompt}{0\textwidth}{} 
Now, would you be able to modify this urdf to be a four wheel robot? There should be two more wheels, in an identical configuration to the first two.
Shift back two of the wheels on the x axis to be at the back of the platform. Also move the front two wheels to the front of the platform by shifting them forwards on the x axis
\end{userprompt}

\begin{gptresponse}{0.2\textwidth}{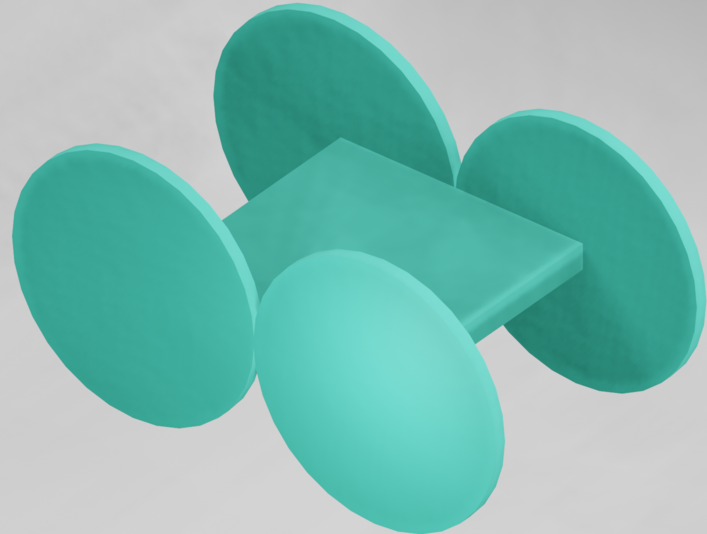} 
\begin{gptcodeblock}{XML}
$\authorremark{inertial and collision parameters were also generated correctly, but omitted below for brevity}$
  <!-- Center rectangular platform link -->
  <link name="base_link"><visual><geometry>
        <box size="0.1 0.1 0.02"/>
  </geometry></visual></link>

  <!-- Front Left wheel link -->
  <link name="front_left_wheel"><visual><geometry>
        <cylinder radius="0.05" length="0.01"/>
   </geometry></visual></link>
 $\omittedCode{other wheels}{}$

  <!-- Front Left wheel joint -->
  <joint name="front_left_wheel_joint" type="continuous">
    <parent link="base_link"/><child link="front_left_wheel"/>
    <origin xyz="0.05 0.055 0" rpy="1.570796 0 0"/><axis xyz="0 0 1"/>
  </joint>
 $\omittedCode{other wheel joints}{}$
\end{gptcodeblock}
\end{gptresponse}

\end{chat}
\caption{\label{fig:urdf-four-wheel} \textbf{URDF of Four Wheeled Robot.} A correct four wheeled robot generated by \gpt by modifying a previously generated URDF for a two wheeled robot}
\end{figure}

As in the case of robot arms, we find that \gpt is immediately able to generalize a successful two-wheeled design into a four-wheeled robot. 
We achieve this by asking for a duplicate, shifted version of the existing wheel configuration, as shown in \fref{fig:urdf-four-wheel}.
However, we were unable to directly generate a successful four-wheel robot; in general, we found that as the number of constraints in a prompt increases, it becomes increasingly likely that \gpt will ignore any individual constraint. 
Thus, rather than directly requesting a four-wheeled robot in a single prompt, we found greater success by first generating a two-wheeled robot and then prompting \gpt to modify the URDF by adding additional wheels than placing the text in a single prompt.

\begin{figure}[tb]
\centering
\includegraphics[width=0.5\textwidth]{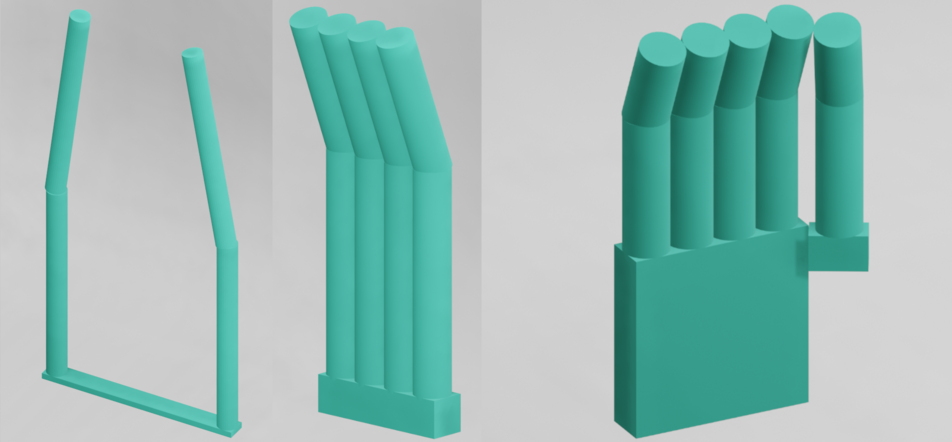}
\caption{\textbf{URDF Grippers Generated by \gpt.} \textbf{Left:} Two fingered Gripper. \textbf{Center:} Four fingered gripper. \textbf{Right:} Five Finger Hand}
\label{fig:urdf-hands}

\end{figure}

\paragraph{Robot Grippers}
To test the effectiveness of our iterative, multi-prompt approach for building robots of increasing complexity, we seeded \gpt with a successful two-link open chain URDF, then asked it to modify this design into a collection of multi-finger robot grippers. As shown in \fref{fig:urdf-hands}, we were able to build two-, four-, and five-finger grippers using a sequence of prompts to add features and change proportions. 
To create a two-finger gripper, we asked \gpt to use two of the previously generated two-link open chain robots as fingers, separated by a distance equal to half the height of the finger, and connected by a rectangular platform on the base.
The four-finger gripper was similarly derived from the two-link arm by specifying that the hand should consist of four two-link robots right next to each other on a rectangular platform. To specify a five finger hand, we requested a rectangular link that hinges as a base for the thumb, then prompted \gpt to add another finger on that link and to adjust the hand proportions.

\ignore{

\begin{figure}
\begin{chat}{false} 
\begin{userprompt}{0\textwidth}{} 
urdf link origins are in the center of the link. To make a joint origin at the end of a link, it must be half the link's length past the link origin. Generate a urdf for a 2 link robot. The end of the second link should connect to the end of the first link with a revolute joint. The origin of the joint should be at the end of the first link. Both links should be lined up along the z axis... 
\\
\authorremark{Modifications were made for the generation of a three link robot, with additional sentences to specify the joint location for the third link}
\end{userprompt}

\includegraphics[width=\textwidth]{figures/multi-link.png}
\end{chat}
\caption{\label{fig:one-and-two-link-urdf} Two and Three Link URDF \wojciech{Expand Caption, Why one prompt and two figures?}}
\end{figure}

\begin{figure}
\begin{chat}{false} 
\begin{userprompt}{0.03\textwidth}{} 
Generate a urdf for a two link robot. The end of the second link should connect to the end of the first link with a revolute joint. The origin of the joint should be a the end of the first link
\end{userprompt}

\begin{gptresponse}{0.4\textwidth}{figures/implicit_joint.png} 
\begin{gptcodeblock}{XML}
  <joint name="joint1" type="revolute">
    <axis xyz="0 0 1" />
    <parent link="link1" /><child link="link2" />
    <origin xyz="1 0 0" rpy="0 0 0" />
  </joint>

  <link name="link1"><visual><geometry>
        <box size="0.1 1 0.1" />
  </geometry></visual></link>

  <link name="link2"><visual><geometry>
        <box size="0.1 1 0.1" />
  </geometry></visual></link>
\end{gptcodeblock}
\end{gptresponse}

\end{chat}
\caption{\label{fig:unspecified-joint-position-urdf} URDF when joint position left implicit}
\end{figure}

}

%% file: sections/GraphArrows/TextToDesignExperiments/Robots_graph_basicDesigns.tex


While designing an entire robot end-to-end using LLMs may not be feasible, we find that \gpt has the ability to reason about the spatial layout of robot components. These spatial layouts are naturally represented as graphs where the nodes are components and edges are connections between them. Unlike URDF, this representation is more general and is applicable in domains outside of simulation.

To generate robot design graphs using \gpt, we first need a text-based graph representation. Our first approach involved asking \gpt to output the popular GraphViz format. While convenient, this format makes it difficult for \gpt to provide metadata for each part (such as motor torque, size) in a format usable by downstream applications. Instead, we take advantage of \gpt's ability to generate Python code that conforms to a provided domain-specific language (DSL). The full DSL is detailed in \appref{sec:appx_graph_robots}. 

When prompted with a small DSL embedded in Python, \gpt is able to write code that selects and places robot components at a high level of abstraction. By supplying a function that translates components in three-dimensional space, we can extract \gpt's concept of each component's position relative to the others. 

\begin{figure}[tb]
\begin{chat}{false}

\begin{userprompt}{0\textwidth}{}

$\omittedCode{problem setup with available Python functions}{}$

Write a function to construct a humanoid robot.
\end{userprompt}

\begin{gptresponse}{0.4\textwidth}{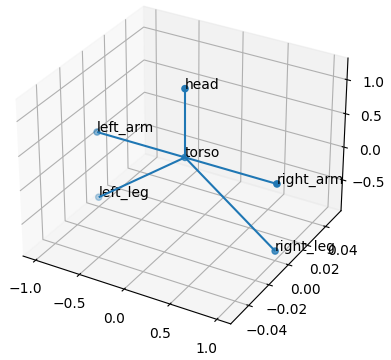}
To construct a humanoid robot, we can define the different parts or links of the robot, such as the head, torso, arms, and legs. We'll then connect these links together using joints. Here's an example Python function to construct a basic humanoid robot:

\begin{gptcodeblock}{Python}
def construct_humanoid_robot():
    # Create links for the head, torso, arms, and legs
    head_link = add_link("head")
    $\omittedCode{adding other parts}{}$

    # Connect the head to the torso
    add_joint(torso_link, head_link)
    $\omittedCode{connecting other parts}{}$

    # Translate the links to their respective positions
    translate(head_link, "up")
    translate(left_arm_link, "left")
    $\omittedCode{positioning other parts}{}$
\end{gptcodeblock}

$\omittedCode{summary of code, disclaimer}{}$
\end{gptresponse}

\end{chat}
\label{fig:robot_graph}
\caption{\label{fig:graph-humanoid} \textbf{Graph of Humanoid Robot.} Graph generated by \gpt describing the high-level components of a humanoid robot as nodes and connections between them as edges}
\end{figure}

In this example, we ask \gpt to generate a humanoid robot using the provided functions. \gpt makes appropriate calls to \texttt{add\_link} to create nodes in the design graph, \texttt{add\_joint} to create edges between them, and \texttt{translate} to establish their relative positions.

We manually implement the functions described in the prompt in order to visualize the resulting robot topology. The arms are positioned beside the torso, the legs are positioned below, and the head rests on top as expected for a humanoid robot.

We saw similar success when asking \gpt to construct a snake robot, car robot, and scorpion robot. When requesting a robot dog, however, \gpt only adds two legs initially. Specifying a ``robot dog with four legs'' was necessary to obtain the expected behavior. We also encountered difficulties when attempting to obtain a more detailed design for the robot dog. Asking for a ``robot dog with four legs, two links per leg'' produced a graph with two nodes per leg, but \gpt did not position them relative to each other.

%% file: sections/GraphArrows/TextToDesignExperiments/CSG_JSCAD_constraints.tex
\begin{figure}[tb]
\begin{minipage}{0.21\textwidth}
\includegraphics[width=\textwidth]{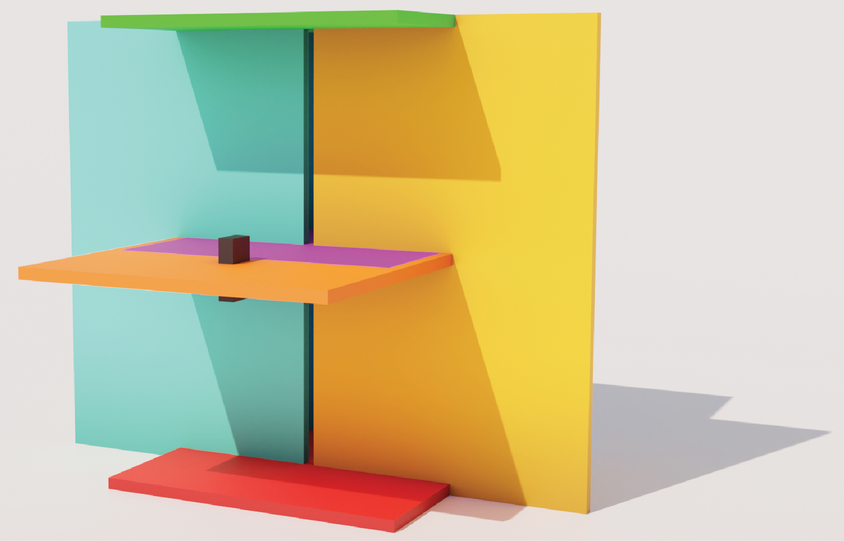}
\end{minipage}
\hfill
\begin{minipage}{0.19\textwidth}
\includegraphics[width=\textwidth]{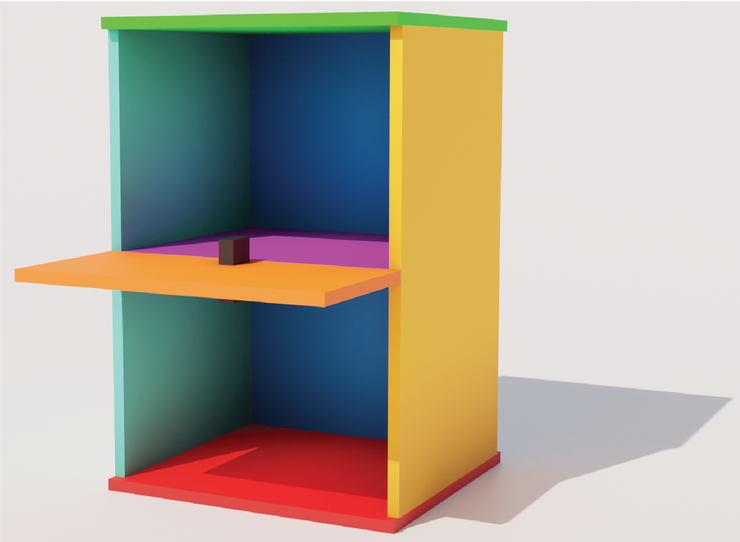}
\end{minipage}
\hfill
\begin{minipage}{0.19\textwidth}
\includegraphics[width=\textwidth]{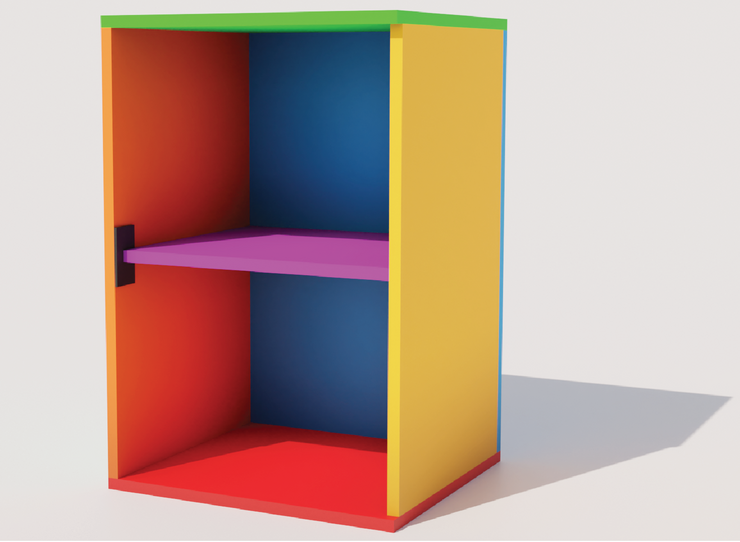}
\end{minipage}
\hfill
\begin{minipage}{0.19\textwidth}
\includegraphics[width=\textwidth]{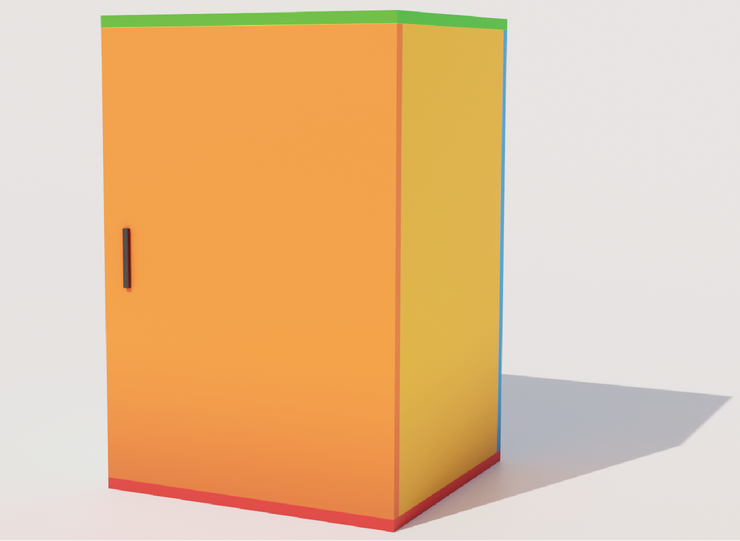}
\end{minipage}
\hfill
\begin{minipage}{0.19\textwidth}
\includegraphics[width=\textwidth]{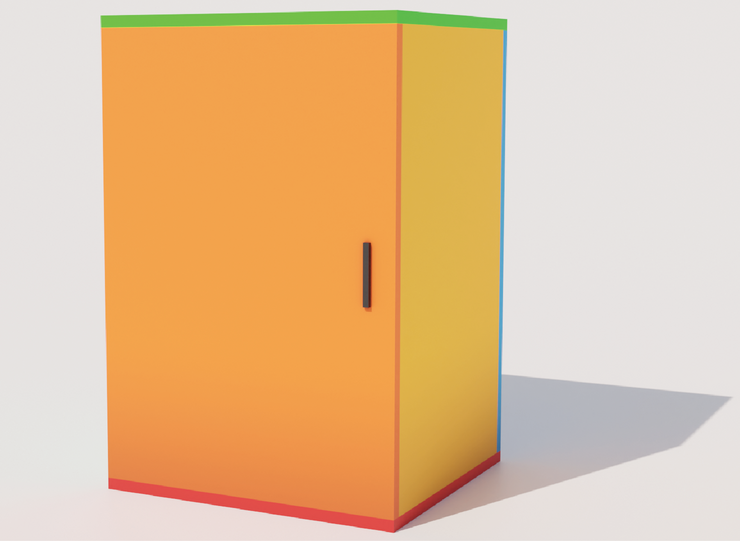}
\end{minipage}
\caption{\textbf{Building a cabinet with a door.} \gpt's attempt to build a cabinet similar to that from \sref{sec:textToDesign_JSCAD_basic}, with the addition of a simple door (orange) that has a handle (dark grey) on the right-hand side. \gpt quickly fixes the position of the cabinet's primary pieces (\eg, the yellow and cyan side panels), but it struggles to correct the door. \gpt must be iteratively prompted to fix the door orientation, the relative door placement, and the handle's placement and protrusion into the door.
\gpt is able to arrive at a suitable design after several iterations of user feedback.} 
\label{fig:cabinet_with_handle_start}
\end{figure}

As an initial experiment, we explored whether \gpt is able to construct a version of the previous cabinet design that includes a door and a handle (see \fref{fig:cabinet_with_handle}). 
We started from a fresh chat, and provided \gpt with a prompt similar to the one described in \sref{sec:textToDesign_JSCAD_basic}, asking for a cabinet to be built from scratch.
However, this time, we also request a door at the front of the cabinet, with a handle on the right hand side of its outward-facing face.
As shown in \fref{fig:cabinet_with_handle_start}, \gpt initially struggled to position several of the cabinet elements -- particularly the side panels and the door.
Although \gpt corrected the position of the side boards immediately, \gpt continued to have trouble placing the door, as it was oriented incorrectly relative to the rest of the design. 
When reminded that the door should be oriented vertically, \gpt was able to comply with the request, but the corrected position was still not fully suitable, as the door coincided with the cabinet's side panel.
After another reminder that the door should reside at the front of the cabinet, with the handle on the right so it could be attached with hinges on the left, \gpt was able to place the door correctly. 
However, the handle remained ill-positioned as it was located on the left-hand side, and was protruding into the door panel. 
After 2 additional prompts, \gpt was able to correct the position to the left hand side. To correct the protrusion issues, \gpt needed 3 more prompts. During these iterations, \gpt moved the handle fully to the \textit{inside} of the door; it needed explicit reminder that the handle should be placed on the \textit{outside} of the door.

\begin{figure}[tb]
    \centering
    \begin{minipage}{0.19\textwidth}
        \includegraphics[width=\textwidth]{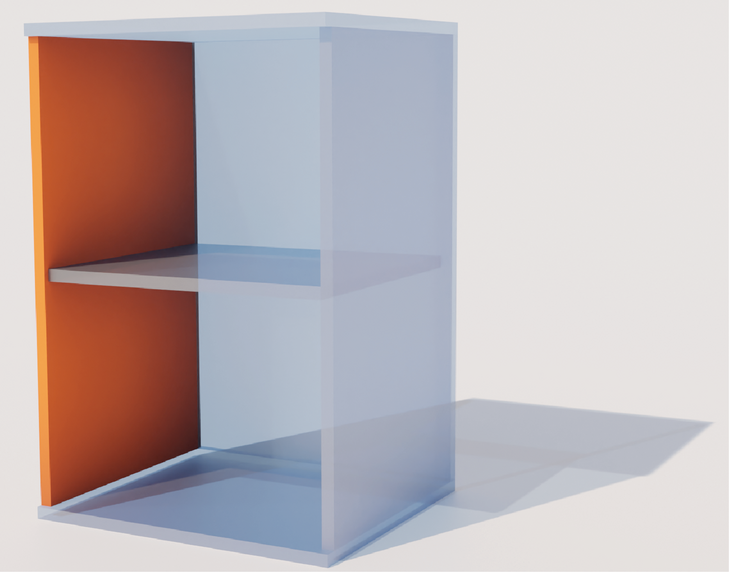}
    \end{minipage}
    \begin{minipage}{0.19\textwidth}
        \includegraphics[width=\textwidth]{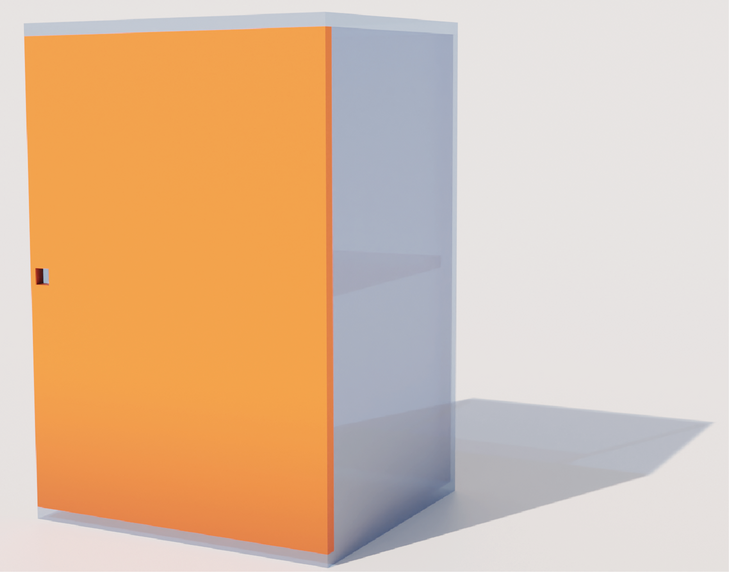}
    \end{minipage}
    \begin{minipage}{0.19\textwidth}
        \includegraphics[width=\textwidth]{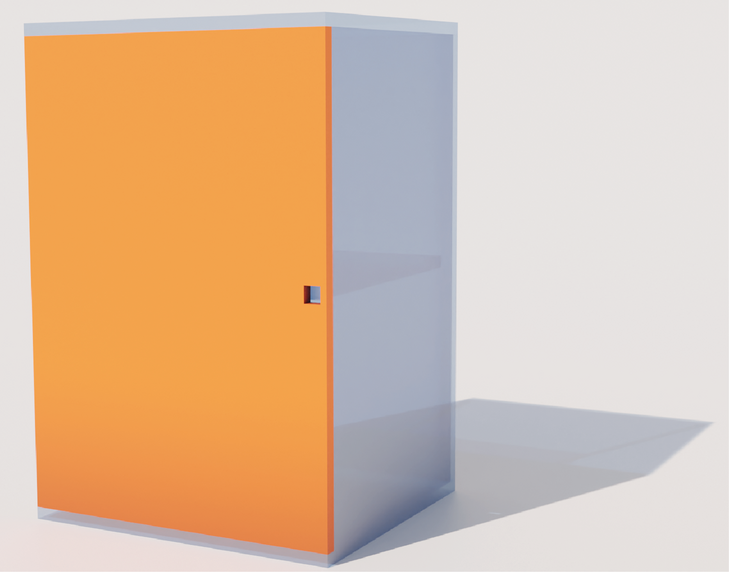}
    \end{minipage}
    \caption{\textbf{Adding a door to an existing cabinet.} We provide \gpt with the initial cabinet design from \sref{sec:textToDesign_JSCAD_basic} (semi-transparent blue), then ask it to add a door (orange) with a handle on the right-hand side. Despite beginning from a largely-complete model, \gpt still has difficulty placing the door and handle correctly.}
    \label{fig:cabinet_with_door_add}
    \vspace{-2mm}
\end{figure}

With a fresh \gpt session, we also tried providing the previous \jscad specification of the cabinet as part of our input prompt, then asking \gpt to modify the existing design such that it contained a door and a handle, as before. 
Despite the different starting points, \gpt followed a similar trajectory, as shown in \fref{fig:cabinet_with_door_add}: the door was initially aligned incorrectly, as it coincided with one of the side panels; after 1 prompt, \gpt was able to correct the door placement. However, despite \gpt's explicit assertion that \lstinline{the handle is also placed on the right side of the door's exterior face}, the handle remained on the left. Finally, after another prompt, \gpt was able to correct the handle position such that it was on the right rather than the left. 

The way in which \gpt dealt with the under-specified handle request also proved interesting. 
In \fref{fig:cabinet_with_handle_start}, \gpt opted for an additional cuboid that would be unioned into the final design. 
By contrast, in \fref{fig:cabinet_with_door_add}, \gpt opted to create the handle by subtracting a small cuboid from the door panel. 
In still other examples, \gpt refused to add the handle, and instead offered the following disclaimer: \lstinline{Note that the handle for the door is not included in this script, as its size, shape, and position would depend on additional details not provided. This would likely require additional modules, such as cylinder from @jscad/primitives, and might be added as an eighth component in the main function.}

These interactions provide a promising basis for interactive user control of the design, but the process is somewhat tedious at the moment, as \gpt requires very explicit instructions about the design or correction intent. 
The addition of highly-detailed user constraints also seems to confuse \gpt to an extent, as it seems to ``forget'' the larger context of the design in the process, so it must be frequently reminded. 

%% file: sections/GraphArrows/TextToDesignExperiments/CSG_JSCAD_integratingPrefabbedParts.tex
\newcommand{\bracketImHeight}{1cm}

\begin{figure}[tb]
\begin{minipage}{0.19\textwidth}
\includegraphics[width=\textwidth]{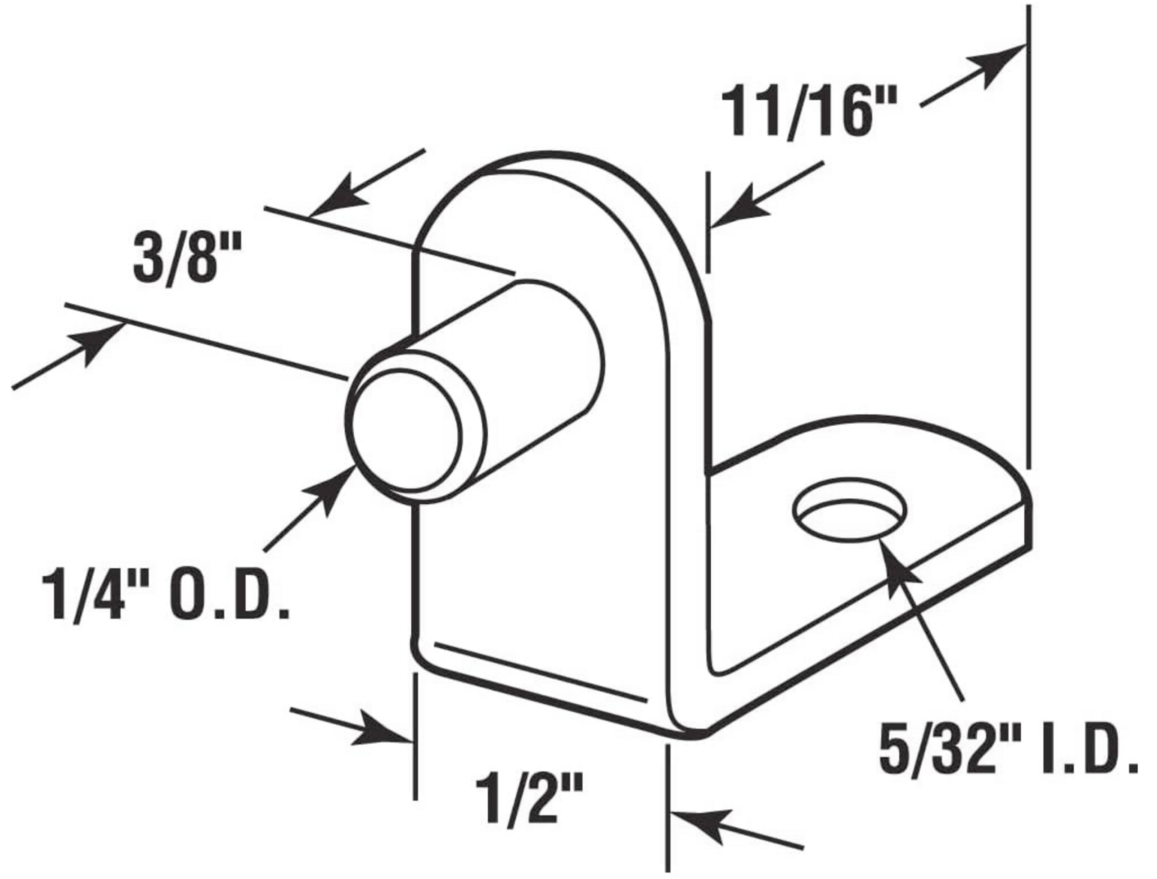}
\end{minipage}
\hfill
\begin{minipage}{0.8\textwidth}
    \begin{minipage}{\textwidth}
        \hspace{2pt}
        \includegraphics[height=\bracketImHeight]{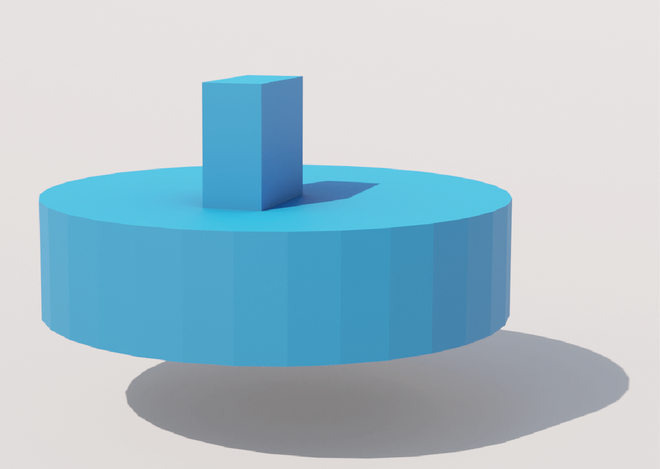}
        \hspace{-5pt}
        \includegraphics[height=\bracketImHeight]{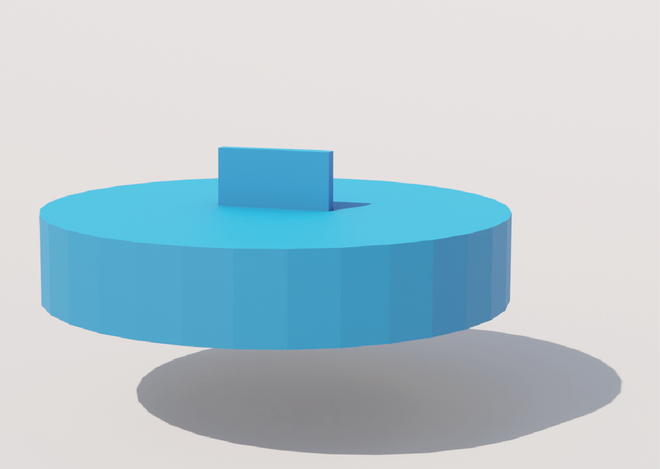}
        \hspace{-5pt}
        \includegraphics[height=\bracketImHeight]{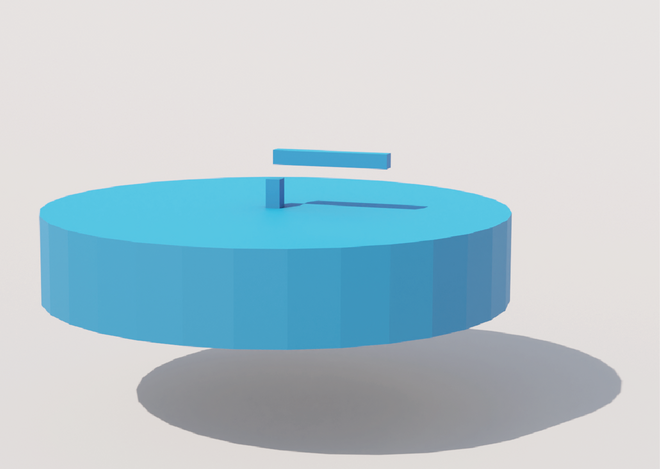}
        \hspace{-5pt}
        \includegraphics[height=\bracketImHeight]{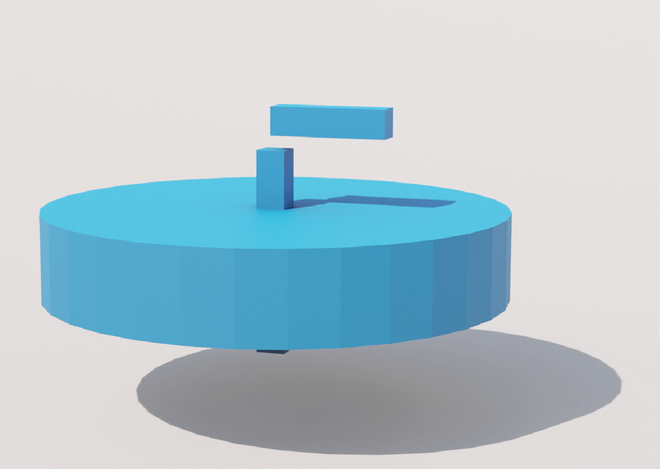}
        \hspace{-5pt}
        \includegraphics[height=\bracketImHeight]{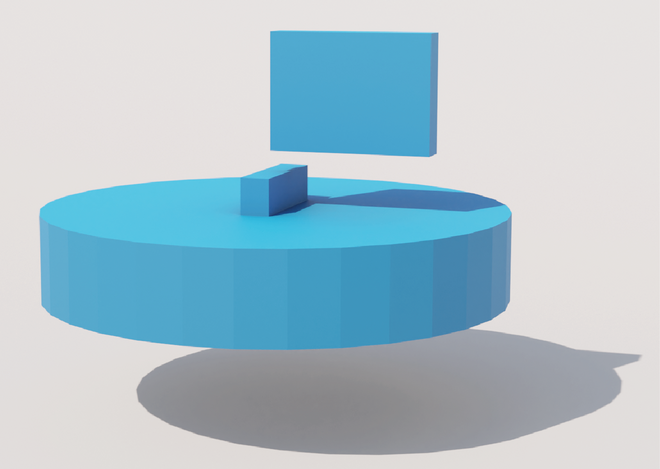}
        \hspace{-5pt}
        \includegraphics[height=\bracketImHeight]{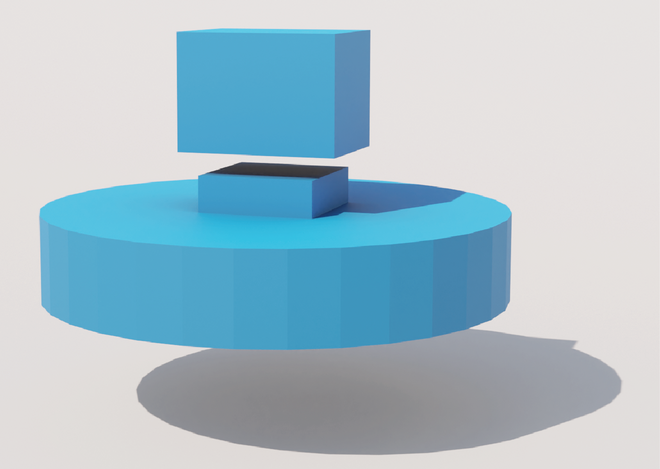}
        \hspace{-5pt}
        \includegraphics[height=\bracketImHeight]{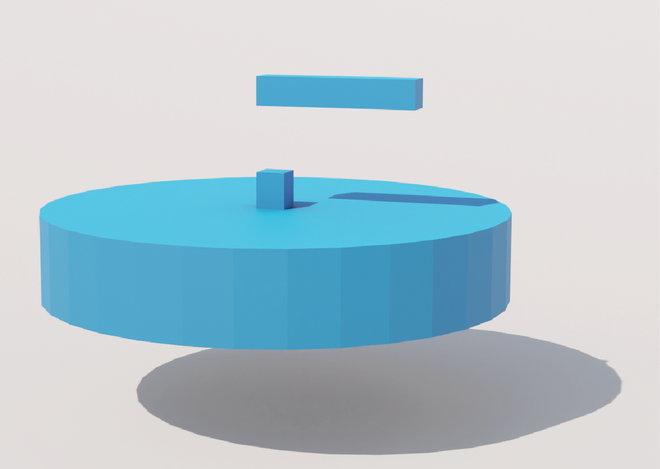}
    \end{minipage}
    \\ \vspace{5pt}
    \begin{minipage}{\textwidth}
        \includegraphics[height=\bracketImHeight]{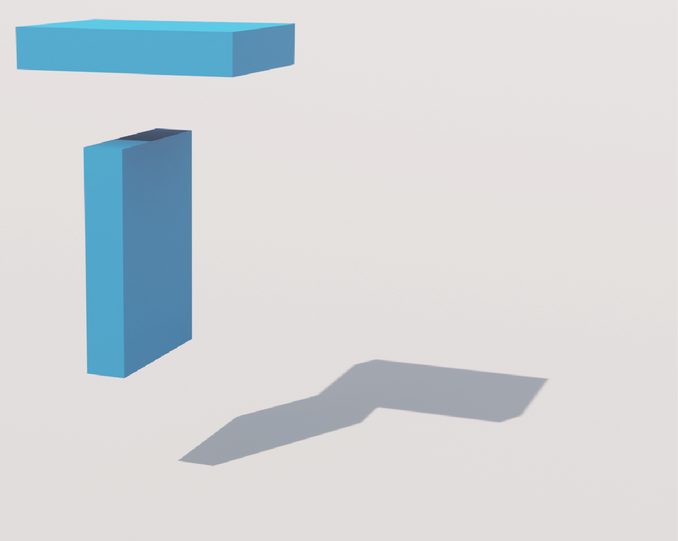}
        \hspace{-5pt}
        \includegraphics[height=\bracketImHeight]{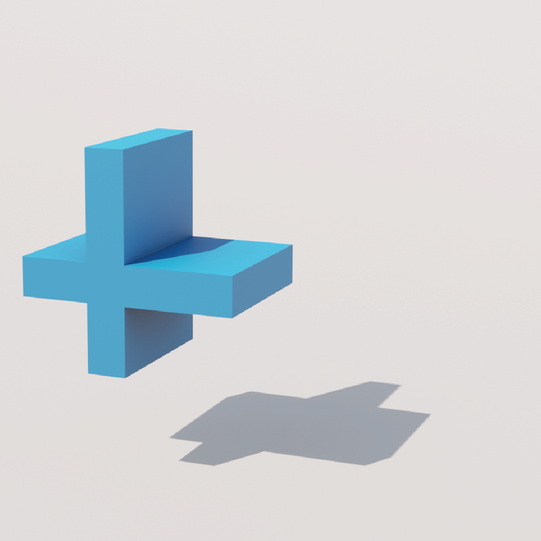}
        \hspace{-5pt}
        \includegraphics[height=\bracketImHeight]{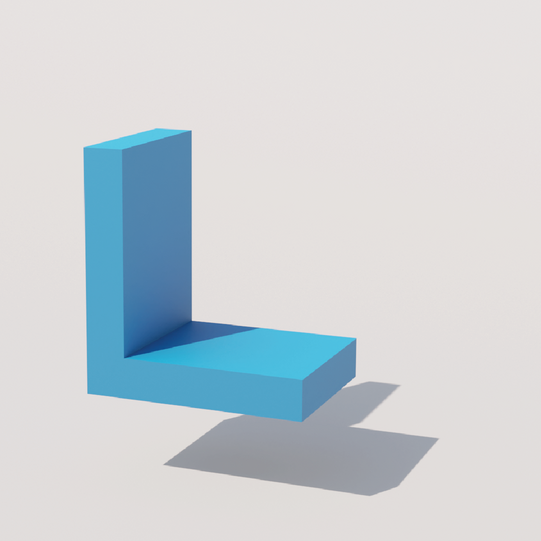}
        \hspace{-5pt}
        \includegraphics[height=\bracketImHeight]{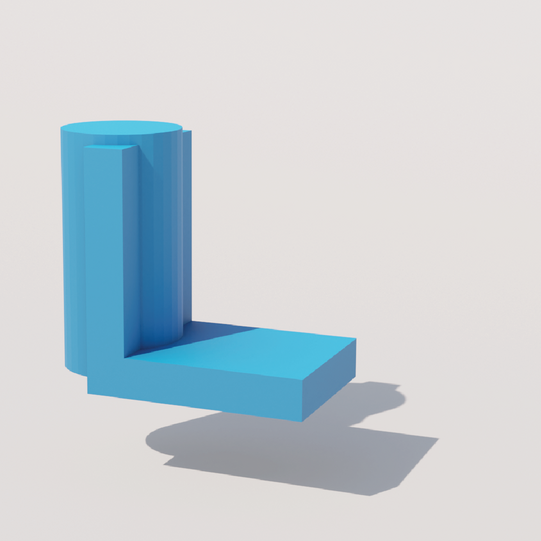}
        \hspace{-5pt}
        \includegraphics[height=\bracketImHeight]{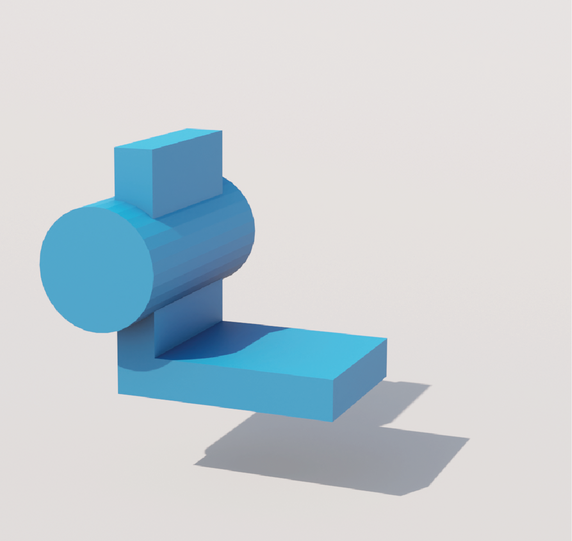}
        \hspace{-5pt}
        \includegraphics[height=\bracketImHeight]{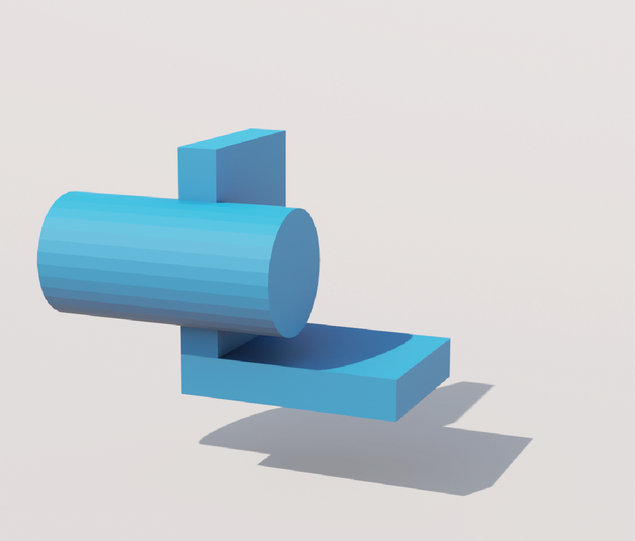}
        \hspace{-5pt}
        \includegraphics[height=\bracketImHeight]{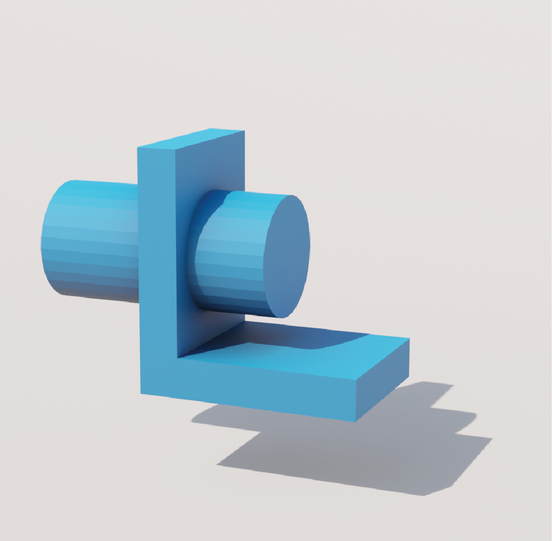}
        \hspace{-5pt}
        \includegraphics[height=\bracketImHeight]{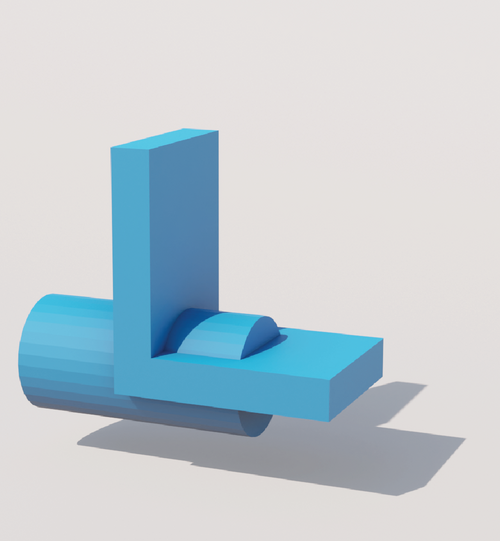}
        \hspace{-5pt}
        \includegraphics[height=\bracketImHeight]{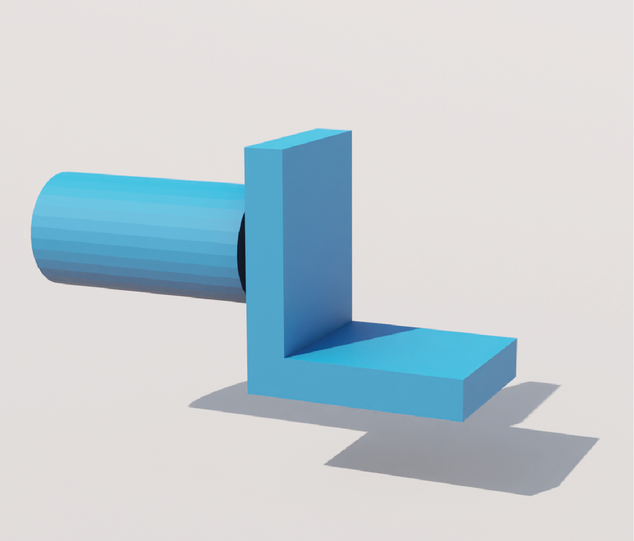}
        \hspace{-5pt}
        \includegraphics[height=\bracketImHeight]{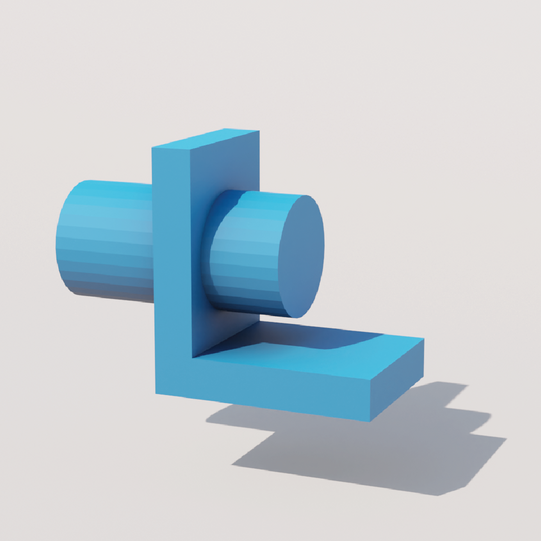}
        \hspace{-5pt}
        \includegraphics[height=\bracketImHeight]{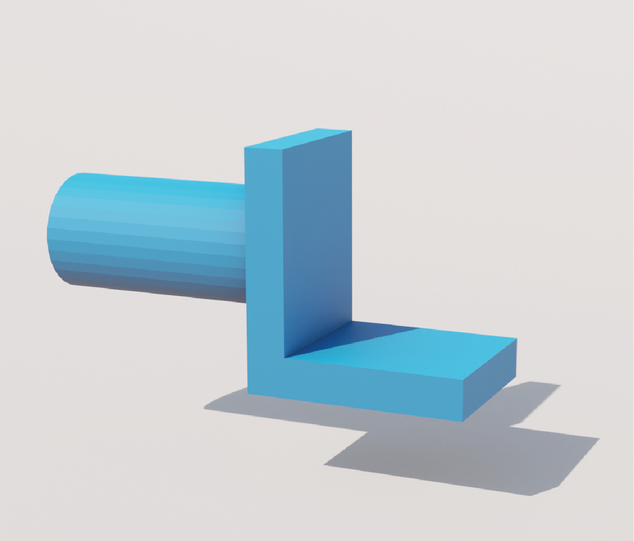}
    \end{minipage}
\end{minipage}

\caption{\textbf{\gpt's Attempts to Create a Proxy for an L-bracket.} \textbf{Left:} Image of the desired pre-fabricated part, to which \gpt was provided a link. \textbf{Right, Top:} \gpt's attempt to design a proxy based on the knowledge it gleaned from the provided product webpage, with iterative high-level user feedback. Although \gpt identified the primary structures (two cuboids for the L and a cylinder for the peg), it was unable to arrive at a proper design in this manner. \textbf{Right, Bottom:} \gpt's process for designing a proxy for the part from scratch with explicit user guidance about the structure and its dimensions.}
\label{fig:bracket-buildup}
\end{figure}

\newcommand{\bracketPlacingImWidth}{0.162\textwidth}
\begin{figure}[tb]
\begin{chat}{true}
\begin{userprompt}{0\textwidth}{}
Now, we are going to try and integrate these bracket supports into our cabinet design from before. 
You can directly use the createBracketWithPeg() function, without regenerating it each time. Now, please go back to the most recent cabinet design, and add 4 of these brackets underneath the middle shelf. There should be two brackets supporting the left side of the shelf, and two brackets supporting the right side of the shelf. Each bracket should be positioned such that the top of the horizontal face is in contact with the bottom of the shelf; the back of the vertical segment is in contact with the innermost side of the cabinet's side wall; and the peg protrudes into the side wall of the cabinet. 
\end{userprompt}
\footnotesize\omitted{33 total messages, used to iteratively position the brackets; process summarized visually below}{}
\begin{userprompt}{0\textwidth}{}
Excellent! The bracket positions are suitable now. Now, we will move onto the shelf: at the moment, the shelf is protruding into the vertical part of the brackets. We need to reduce the width of the shelf in the left-right direction such that it leaves space for the brackets. Can you adjust the shelf size accordingly? 
\end{userprompt}
\begin{minipage}{0.99\textwidth}
    \includegraphics[width=\bracketPlacingImWidth]{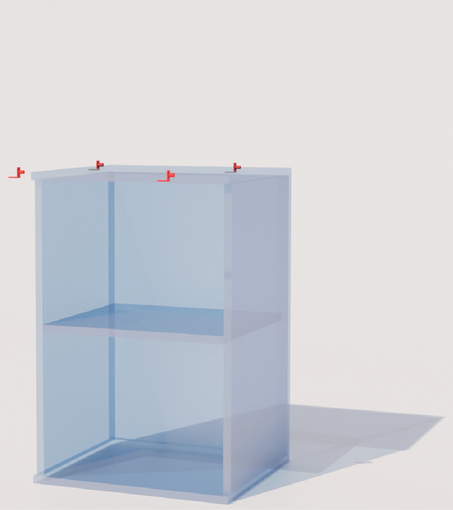}
    \includegraphics[width=\bracketPlacingImWidth]{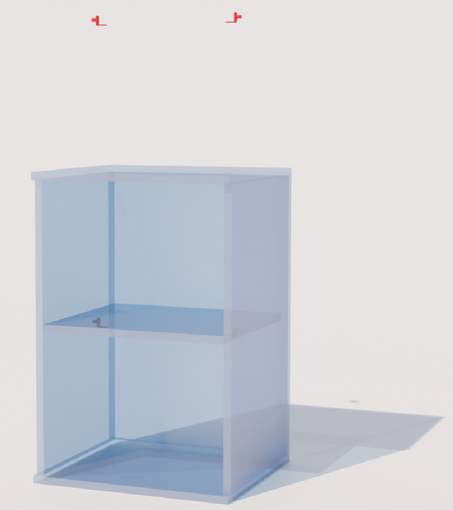}
    \includegraphics[width=\bracketPlacingImWidth]{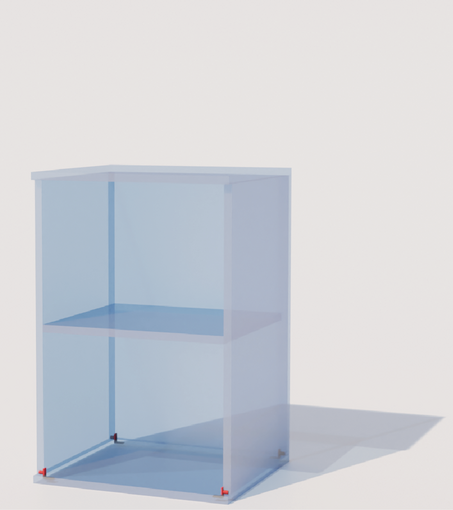}
    \includegraphics[width=\bracketPlacingImWidth]{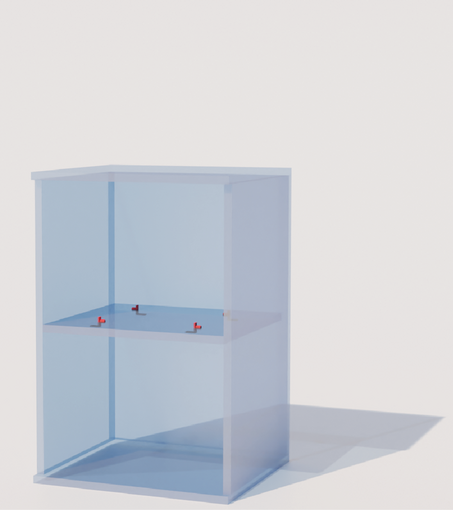}
    \includegraphics[width=\bracketPlacingImWidth]{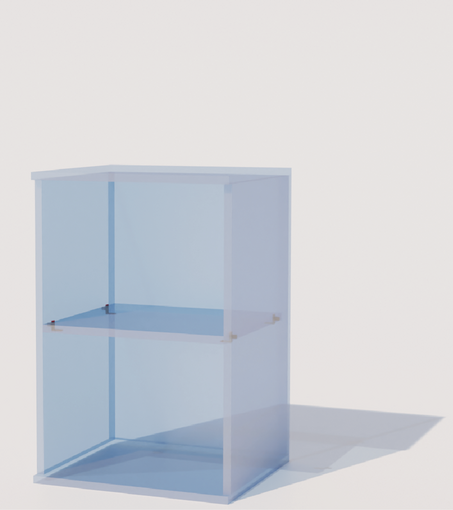}
    \includegraphics[width=\bracketPlacingImWidth]{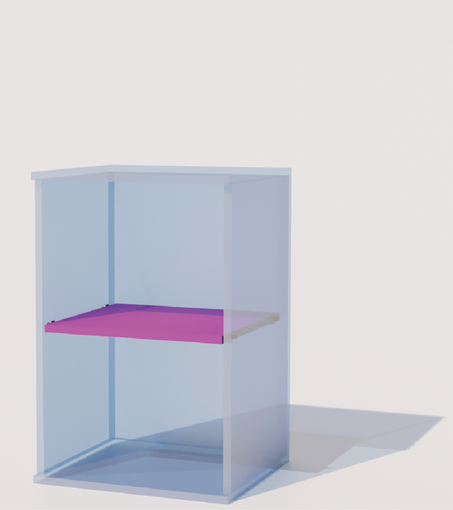}
\end{minipage}

\end{chat}

\caption{\textbf{Process for Integrating L-brackets (red) into an Existing Cabinet Design (semi-transparent blue) Using \gpt.} It takes 34 messages to position the brackets appropriately (17 each of prompt/response), but once this is done, \gpt is able to efficiently generate a modified shelf (pink) to accommodate the placed brackets (6 messages; 3 each of prompt/response).}
\label{fig:cabinet_with_handle}
\end{figure}
\afterpage{\FloatBarrier}

To make the cabinet design more stable, a designer may wish to include extra support brackets to work with. 
Many pre-fabricated variations of these brackets exist, and they are inexpensive and readily available.
Given this, it does not make sense to design or manufacture these parts via \gpt.
Rather, we'd like to incorporate instances of a pre-fabricated version.
To do this, \gpt must first build a proxy of the part, place the proxies throughout the design appropriately, and adjust the remaining elements of the design to accommodate these components.

For our first experiment, we chose to incorporate the Prime-Line 1/4 in. Nickel-Plated Shelf Support Pegs from Home Depot into our design.
We provided \gpt with a URL to this part's listing on the Home Depot website, which contained a text description of the item and the schematic diagram pictured in \fref{fig:bracket-buildup}(left).
We then asked \gpt to build a simple geometric proxy that we could incorporate into our design as a placeholder.
As shown in \fref{fig:bracket-buildup}(right, top), \gpt was able to infer and generate the appropriate primitives (one cylinder for the peg and two cuboids for the L bracket).
However, it was not able to correctly scale, orient, or position the elements. 
In an effort to test \gpt's understanding of the structure, we asked it to describe the structure in its own words. 
Although it gave a reasonable description of the bracket, there was little improvement in the result when it was asked to improve the script accordingly.
Thus, even with several iterations of user feedback, \gpt was unable to construct this shape from high-level third-party (URL) or user input.

Ultimately, we had to provide \gpt with an explicit description of the structure that we wanted.
Moreover, we found that even with an explicit description, \gpt was unable to generate the correct shape when provided with all directions at once.
Instead, we had to create the shape in an iterative fashion, beginning with the L bracket and then adding in the peg, as shown in \fref{fig:bracket-buildup}(right, bottom).
Eventually, it was able to generate the structure and consolidate the instructions into a high-level module called \lstinline{createBracketWithPeg}, as desired.

We then provided the module \lstinline{createBracketWithPeg} as an input to \gpt, and asked it to incorporate these structures into the design, as detailed in \fref{fig:cabinet_with_handle}. In particular, we asked for four brackets under each shelf, with the pegs protruding into the cabinet's side walls, the back face of the bracket's vertical leg in contact with (but not protruding into) the side wall, and the top face of the bracket's horizontal leg in contact with (but not protruding into) the bottom face of the shelf. 
We initially tried to complete this experiment in a single continuous chat that (1) designed the cabinet, (2) designed the L-bracket, and then (3) incorporated the brackets into the cabinet. 
However, we found that after the extended discussion regarding the L-bracket design, \gpt seemed to have completely forgotten its cabinet specification.
Despite multiple prompts, it was unable to recover the previous design. 
Instead, we directly provided \gpt with the L-bracket module and its prior cabinet design, and then asked for a modification.
This approach was far more successful.
Overall, we found that \gpt was able to instantiate the correct number of brackets, but it struggled to rotate and position them appropriately. 
After several user prompts, \gpt was able to successfully place the brackets in their locations.
Finally, we asked \gpt to adjust the shelf in order to (1) not protrude into the brackets, and (2) incorporate some additional allowance so the shelf could easily fit between the supporting brackets in a physical assembly.
\gpt was able to complete these requests without issue.

Overall, although \gpt initially struggled to build a proxy of the pre-fabricated part we had in mind, \gpt seemed quite capable of incorporating the completed proxy into a given design, as desired.

%% file: sections/GraphArrows/TextToDesignExperiments/Robots_customPrimitives_integratingPrefabbedParts.tex



\begin{figure}[tb]
\includegraphics[width=\textwidth]{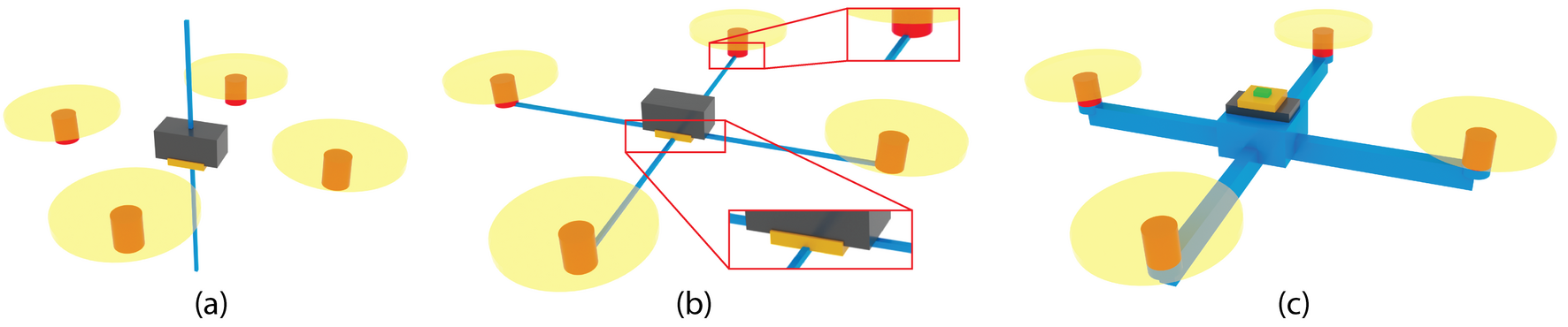}
\caption{\textbf{A Quadcopter Designed with the Aid of \gpt.} The motors are colored in red.
The propellers are in yellow. The battery is in dark gray. The frame is in blue.
The dark yellow box is the controller and the green box is the receiver.}
\label{fig:quadcopter-stage}
\vspace{-5mm}
\end{figure}

Designing a quadcopter involves integrating pre-built elements like the motor, propeller, and battery. Detailed sourcing of these parts will be addressed in the later section (Section~\ref{sec:part_sourcing}). Once these components are sourced, the frame must be designed to accommodate their dimensions. We'll explore how \gpt can assist with this task.

However, enabling \gpt to accurately represent these parts isn't straightforward. To simplify the task, parts are represented as either a box of dimensions $w \times h \times d$ or a cylinder with radius $r$ and height $h$. \gpt can handle these representations well as demonstrated in Section~\ref{sec:textToDesign_CSG_boxes}. Rather than having a single function which creates a primitive and translates it as in Section~\ref{sec:textToDesign_CSG_boxes}, we introduce three functions for ease of design: createBox(w, h, d), createCylinder(r, h), and place(item, x, y, z, a). The first two functions generate a box or a cylinder at origin (0,0,0), while the third rotates and moves the item to desired coordinates.

Subsequently, we task \gpt with creating a design that integrates these parts using only the above functions. The primary element \gpt must design is the frame, which should hold the selected components. Initially, \gpt produced a correct textual design, but struggled with the geometric representation, similar to Section~\ref{sec:textToDesign_CSG_boxes}. It understood the quadcopter structure, but had issues with part positioning and orientation (Figure~\ref{fig:quadcopter-stage}(a)). Problems included incorrect frame orientation and part intersections. By guiding \gpt in correcting these issues, we achieved a near-correct quadcopter design (Figure~\ref{fig:quadcopter-stage}(b)).

The initial frame design wasn't practical because it was directly attached to the motor cylinder and insufficient to hold components like the battery, controller, and signal receiver. To address this, we asked \gpt to incrementally implement specific solutions, such as adding a cylinder base under each motor and a box body to reinforce the frame bars and house remaining parts. After minor adjustments, we arrived at a valid design, which will undergo further testing in a simulator or real world conditions (Figure~\ref{fig:quadcopter-stage}(c)).

Throughout the design process, \gpt demonstrated proficiency in textual design analysis but struggled with mathematical and physical concepts such as collision and structural integrity. Thus, human guidance remains crucial in these areas.

%% file: sections/GraphArrows/TextToDesignExperiments/CSG_JSCAD_modules.tex

As we have seen from previous examples, \gpt is inclined to use some abstractions like variables by default.
It is also clear that \gpt is well suited to the use of modular or hierarchical design, as in the case of the pre-fabricated L-brackets that it was able to instantiate several copies of, and distribute throughout a design. 
However, there are often instances where a user might want to impose their own specific modules -- for example, a certain hierarchical grouping may facilitate easier debugging or cleaner code.

To test \gpt's abilities in this area, we revisit the cabinet example, and try to modify it such that it contains multiple shelves. Because we have already incorporated pre-fabricated brackets, this modification is non-trivial, as \gpt must instantiate and position the appropriate number of shelves \textit{and} all associated support brackets. 
We began by directly asking \gpt to make this modification on top of the existing code, by generating two evenly spaced shelves within the cabinet instead of one. 
\gpt correctly identifies the elements which must be duplicated, and it instantiates the correct number of them. 
However, it is unable to correctly adjust the position of each module; after the initial request, neither the shelves nor the brackets were in reasonable locations. 
It took 4 additional user prompts to correct the relative positions of these components. 
After this correction, \gpt did seem able to generalize its logic directly to generate cabinets with a varying number of shelves. 
However, the code itself is fairly convoluted.

To avoid these issues, it may be more natural to consider a shelf with its appropriate supporting brackets as a single module.
This way, the entire ``subassembly'' could be instantiated and positioned as a unit on future calls. 
We asked \gpt to implement this plan, by requesting the creation of a module named \lstinline{supportedShelves()}, which instantiates and appropriately positions a shelf and its associated support brackets within the design.
Then, we asked \gpt to refactor the original script such that it used the new module to generate a cabinet with two evenly-spaced shelves.
The initial response had a minor compilation error, a shelf tolerance issue, and a bracket alignment issue, as before, but each of these issues were immediately corrected after a single user prompt. 

Overall, the approaches resulting from both experiments seem equally effective and flexible once they have been fine-tuned. 
Thus, we conclude that \gpt is able to effectively create and use modules, whether they are explicit (\eg, in the form of a function, as in the second experiment) or implicit (\eg, in the form of a for-loop, as in the first experiment).
However, it seems as if the explicit module made it slightly easier for \gpt to reason about a challenging alignment problem.
Moreover, it is useful to know that users can effectively request this kind of hierarchical refactoring, as most human programmers/designers would generally find it easier to reason over a function in this scenario. 

%% file: sections/GraphArrows/TextToDesignSpace.tex
\section{Text-To-Design-Space}
\label{sec:text_to_design_space}


\hfill

\begin{figure}[tb]
\begin{minipage}{0.3\textwidth}
\includegraphics[width=\textwidth]{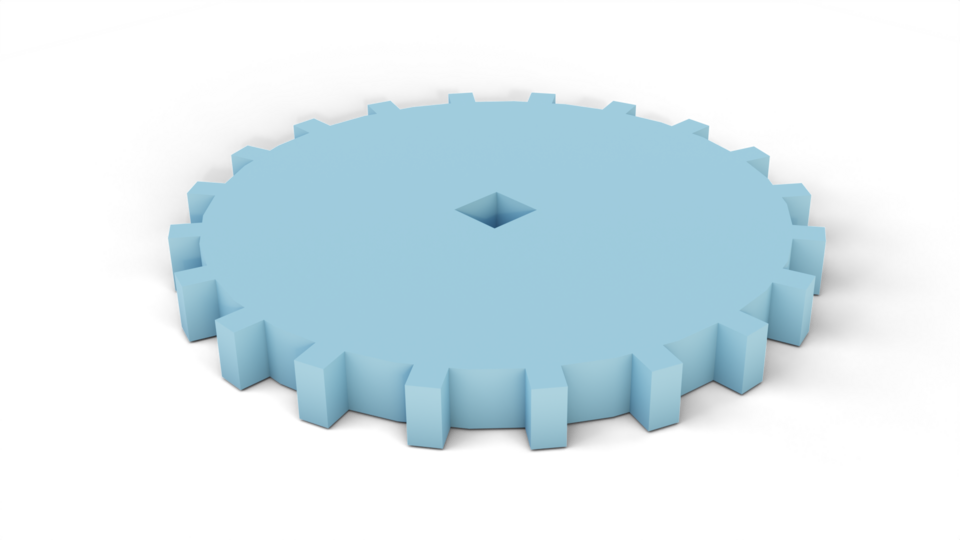}
\end{minipage} 
\begin{minipage}{0.69\textwidth}
\includegraphics[width=\textwidth]{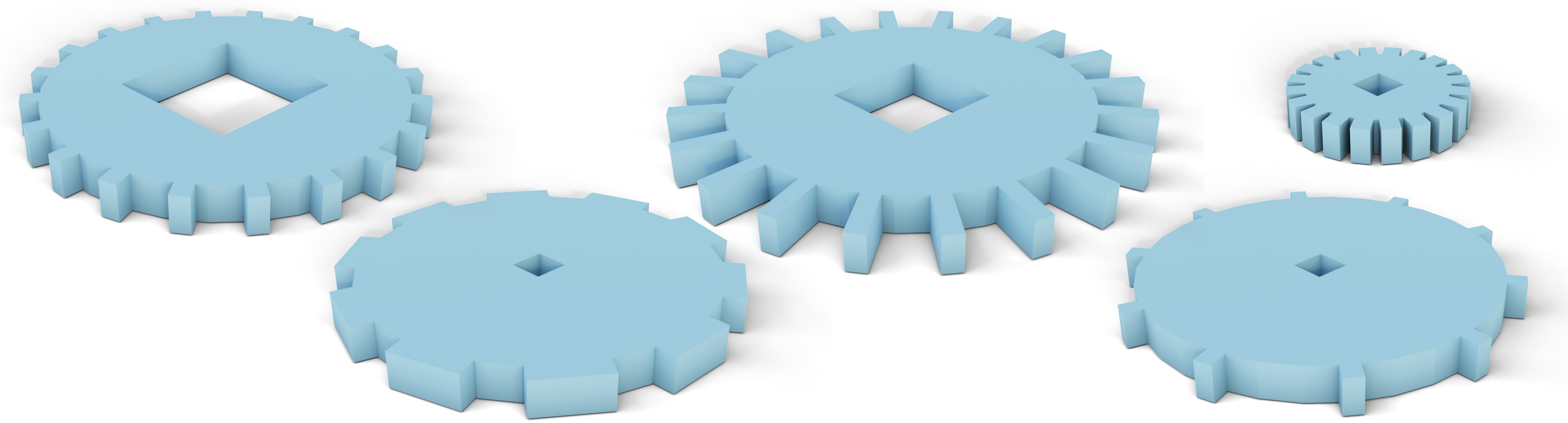}
\end{minipage} 

\begin{minipage}{0.3\textwidth}
\footnotesize
\begin{lstlisting}
difference() {
    union() {
        translate([0, 0, -10/2])
        cylinder(10, 50, 50);
        
        for (i = [0:20-1])
        rotate([0, 0, i*360/20])
        translate([50, 0, 0])
        cube([5*2, 5, 10], true);
    }
    cube([10, 10, 10*2], true);
}
\end{lstlisting}
\end{minipage}
\begin{minipage}{0.69\textwidth}
\footnotesize
\begin{lstlisting}
difference() {
    union() {
        translate([0, 0, -gear_thickness/2])
        cylinder(gear_thickness, gear_rad, gear_rad);
        
        for (i = [0:tooth_count-1])
        rotate([0, 0, i*360/tooth_count])
        translate([gear_rad, 0, 0])
        cube([tooth_prot*2, tooth_width, gear_thickness], true);
    }
    cube([center_hole_width, center_hole_width, gear_thickness*2], true);
}
\end{lstlisting}
\end{minipage}
\caption{\label{fig:gear_model} \textbf{Gear Design Space.} The same gear (\textbf{Top left}) can be constructed with different design representations. Here, we have an OpenSCAD design (\textbf{Bottom left}) and a parametric OpenSCAD design (\textbf{Bottom right}) where the degrees of freedom (\textbf{Top right}) have been exposed via design variables.}
\end{figure}


A design is a sequence of construction operations which take input values and which modify the current state of the design.
These input values can directly be represented as numbers. For example in Fig.~\ref{fig:gear_model} (left), the design of a 3D gear is constructed by directly using 3D coordinates and dimensions.
While this representation has the merit of being direct, without any references to previous code, it does not expose the degrees of freedom of a design.
To modify the thickness of the gear, we have to modify several input values at once to obtain the desired 3D model.
The introduction of \emph{design parameters} in Fig.~\ref{fig:gear_model} (right) makes this change easier by modifying a single variable, namely \lstinline{gear_thickness}.
We call this representation a \emph{parametric design}.
Note that design parameters can be continuous or discrete, e.g. \lstinline{gear_thickness} or \lstinline{tooth_count} respectively.

To explore different design variations, either manually or automatically, having a parametric design is not enough.
We still need to know which specific values we can assign to the design parameters.
For this, we introduce lower and upper \emph{bounds} for each design parameter.
Each design parameter can take any value within its specific bounds.
Together, a parametric design and parameter bounds define a \emph{design space} which is the set of all possible design variations.

Design spaces are an import tool to understand what a design can accommodate for.
This is important for both the manual and automatic optimization of designs.
With this in mind, we want to investigate the following questions:
\begin{itemize}
    \item \textbf{Q1} Can GPT-4 create a design space from text?
    \item \textbf{Q2} Can GPT-4 create a design space from an existing design?
    \item \textbf{Q3} Can GPT-4 create a design space from multiple designs?
    \item \textbf{Q4} Can GPT-4 explore a given design space?
\end{itemize}

For each of these questions, we want to find out what is currently possible and what seems to be beyond its capabilities.

\subsection{Generating a Design Space from Text (Q1)}
In Sec.\ref{sec:text_to_design}, we showed that GPT-4 is capable of generating designs.
The next step towards generating a design space is to test if it can also generate \emph{parametric} designs.
To enforce the generation of parametric designs in our prompts, we ask it to \lstinline{explicitly use high-level design parameters} and to use \lstinline{as few variables as possible}.
It should be noted that GPT-4 often introduces variables to improve readability by itself, without explicitly being asked to do so.
However, we found that including this in our prompts \emph{always} resulted in parametric designs.

\begin{figure}[tb]
\begin{chat}{false} 

\begin{userprompt}{0.2\textwidth}{} 
Can you design a chair with 4 legs, a rectangular seat and a rectangular back?
\end{userprompt}

\begin{gptresponse}{0.2\textwidth}{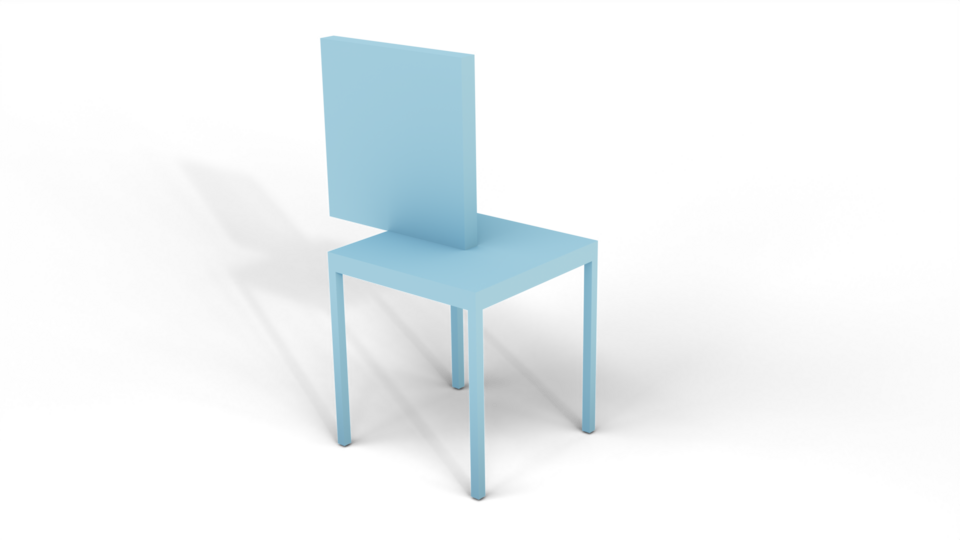} 
    \begin{gptcodeblock}{Python}
$\omittedCode{variable setup and legs construction}{}$
# Create the seat
seat_sketch = createSketch(rectangle(0, 0, 16, 18), XY_PLANE)
seat_solid = extrude(seat_sketch, seat_height)

# Create the back
back_sketch = createSketch(rectangle(0, 8.5, 16, 1), cap(seat_solid, "max_z"))
back_solid = extrude(back_sketch, back_height)
\end{gptcodeblock}
\end{gptresponse}

\begin{userprompt}{0.2\textwidth}{} 
Can you design a chair with 4 legs, a rectangular seat and a rectangular back with explicitly using exposed high-level parameters?
\end{userprompt}

\begin{gptresponse}{0.2\textwidth}{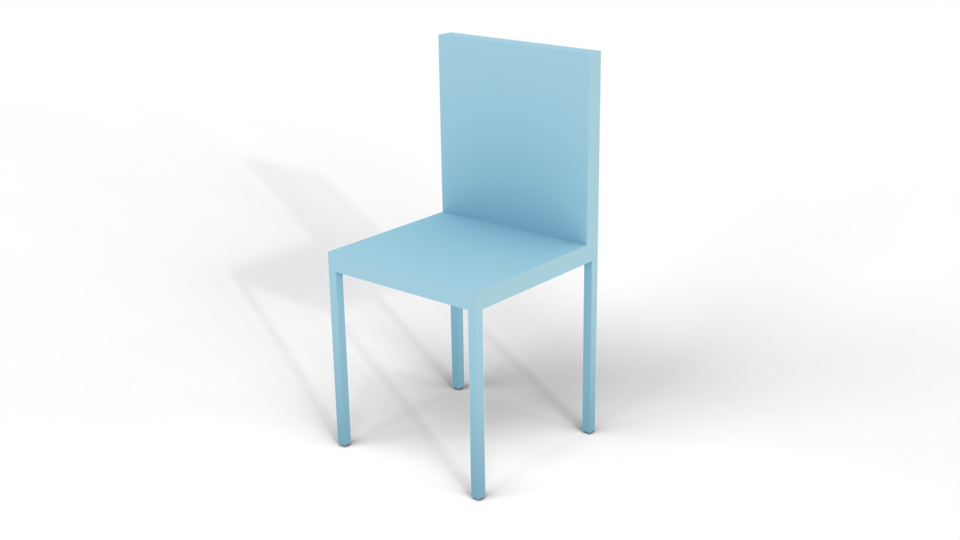} 
\begin{gptcodeblock}{Python}
$\omittedCode{variable setup and legs construction}{}$
# Create the seat
seat_sketch = createSketch(rectangle(0, 0, seat_length, seat_width), XY_PLANE)
seat_solid = extrude(seat_sketch, seat_height)
# Create the back
back_sketch = createSketch(rectangle(0, seat_width/2 + back_thickness/2, seat_length, back_thickness), cap(seat_solid, "max_z"))
back_solid = extrude(back_sketch, back_height)
\end{gptcodeblock}
\end{gptresponse}

\end{chat}
\caption{\label{fig:chair_dimensions} \textbf{Advantage of Parametric Modeling.} Without using explicit variables, \gpt does the computation for sketch coordinates by itself and is more likely to confuse variables and to make mistakes.}
\end{figure}

We also notice that when asking for a simple design and asking for a parametric design of the same object, there are generally fewer mistakes in the reuse of certain dimensions.
For example, in Fig.~\ref{fig:chair_dimensions}, at first, \gpt positions the backrest on top of the seat using the correct numerical values, but not for the correct dimensions.

Whereas when asked for a parametric design, the use of \lstinline{width} and \lstinline{length} suffixes in the parameter names seem to be more consistently associated with the corresponding 3D axis.


\begin{figure}[tb]
\includegraphics[width=0.5\textwidth]{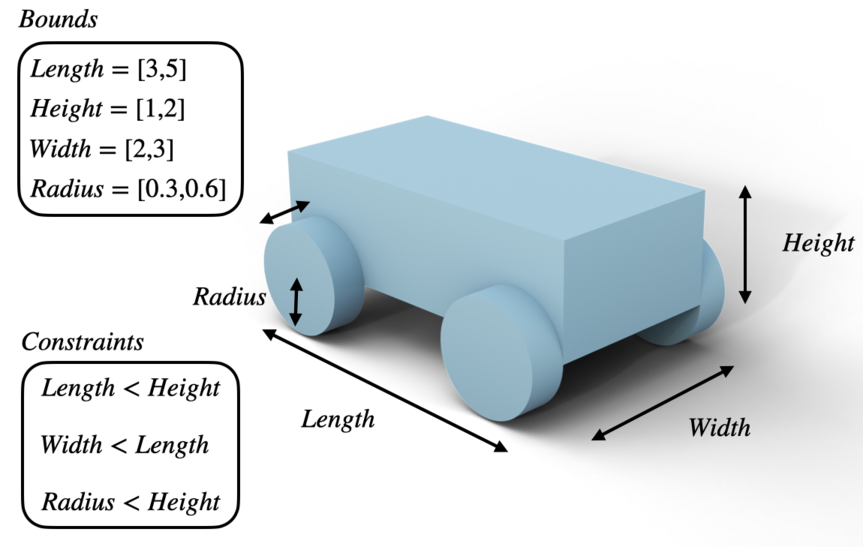}
\caption{\textbf{Car Parameter Bounds.} \gpt generates semantically based parameter bounds and constraints based on this simplified car design.}
\label{fig:car_bounds}
\end{figure}
To generate a design space, we need parameter bounds.
When asked for lower and upper bounds for parameters, GPT-4 proposes bounds that are \lstinline{based on typical proportions} of the designed object.
This implies that the scale is often arbitrary but that bounds are semantically reasonable relative to each other.
For example, when asked to design a parametric car with exposed parameter bounds, GPT-4 returns lower and upper bounds and arguments for these bounds in terms of inequalities, see Fig.~\ref{fig:car_bounds}.
According to GPT-4, the width of the car body \lstinline{should be less than the length but larger than the height} and the radius for the cylindrical wheels \lstinline{should be less than the height of the car's body so the wheels don't exceed the height of the body}.
These constraints between design parameters can also be queried in the form of actual inequalities, which is useful for downstream optimization when combined with parameter bounds.

However, these bounds are based on semantic knowledge about the object and not on the geometric design sequence.
For example, for a pen holder, the angle of a rotated cylinder will get a lower bound of $45^{\circ}$ to prevent any pen from falling out, but not to prevent the 3D object from creating unwanted intersections with other parts.
Constraints in real-world design sequences often need to also consider purely geometric aspects of a design.


%
%
%
%

\subsection{Generate a Design Space from an existing design (Q2)}
Given the current limitations of creating designs and design spaces from text prompts alone, it is interesting to understand how \gpt can create design spaces from existing designs, made by human designers.
Just as regular code, input designs for \gpt can vary in quality of semantic annotations and comments about what is being constructed.
For all of these inputs, we are interested in how easy it is for \gpt to create a design space, i.e., a parametric design with parameter bounds. 
We investigate how helpful semantic context is to parametrize designs.
For the prompts of the following experiments, we have found that we get more consistently a good parametrization when we include that it should \lstinline{expose high-level design parameters} while using \lstinline{as few variables as possible} and that it should \lstinline{keep the same program structure and the resulting input values to modeling functions}.
These constraints prevent it from slightly modifying operator input values to extract fewer design parameters.

\begin{figure}[tb]
\begin{chat}{false} 

\begin{userprompt}{0.3\textwidth}{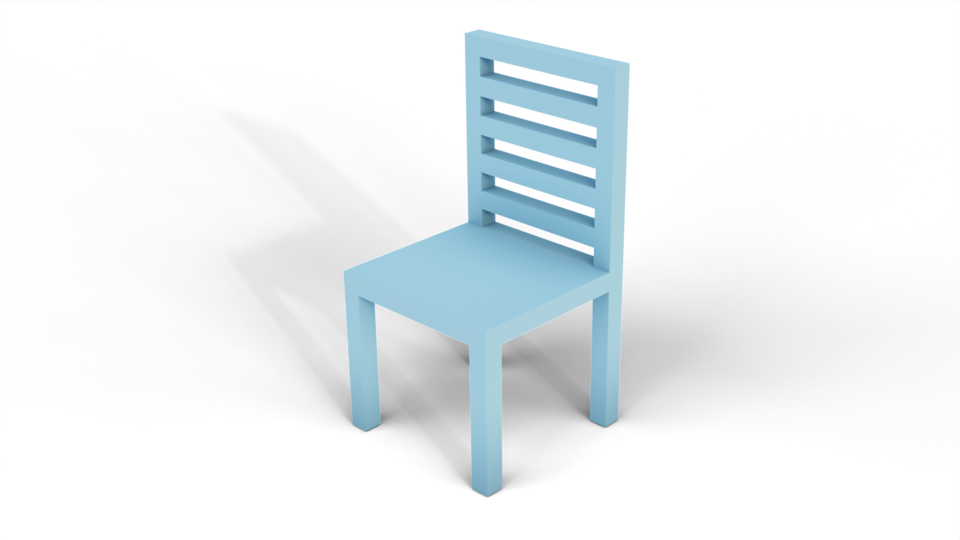} 
Can you expose high-level design parameters for this openscad design? Use as few variables as possible
    \begin{gptcodeblock}{Javascript}
union() {
    difference() {
        cube([19, 19, 15 + 2]);
$\omittedCode{intermediate construction operations}{}$
    for (i = [0:5-1])
$\omittedCode{rest design}{}$
    \end{gptcodeblock}
\end{userprompt}

\begin{gptresponse}{0\textwidth}{} 
    \begin{gptcodeblock}{Javascript}
// High-level design parameters
cube_size = 19;
cube_height = 15;
extra_height = 2;
num_layers = 5;
// Other parameters ...
union() {
    difference() {
        cube([cube_size, cube_size, cube_height + extra_height]);
$\omittedCode{intermediate construction operations}{}$
    for (i = [0:num_layers-1])
$\omittedCode{rest design}{}$
\end{gptcodeblock}
\end{gptresponse}

\begin{userprompt}{0.03\textwidth}{} 
Can you expose high-level design parameters for this openscad design of a chair with slats? Use as few variables as possible
\end{userprompt}

\begin{gptresponse}{0\textwidth}{}
    \begin{gptcodeblock}{Javascript}
length = 19; // Length of the chair
width = 19; // Width of the chair
height = 15; // Height of the chair base
slat_count = 5; // Number of slats
union() {
    difference() {
        cube([length, width, height + thickness]);
$\omittedCode{intermediate construction operations}{}$
    for (i = [0:slat_count-1])
$\omittedCode{rest design}{}$
    \end{gptcodeblock}
\end{gptresponse}

\end{chat}
\caption{\textbf{Semantic Cues for Chair Design Parameters.} When asked to parametrize a chair design, it \gpt gives different parametrization with or without context cue.}
\label{fig:chair_parametrization}
\vspace{-5mm}
\end{figure}

\begin{figure}[tb]
\begin{chat}{false} 

\begin{userprompt}{0.3\textwidth}{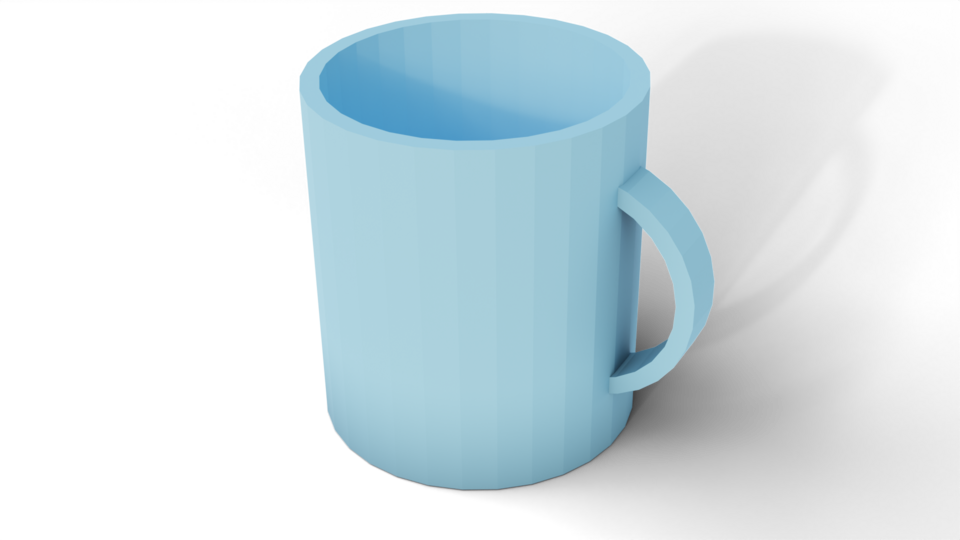} 
Can you expose high-level design parameters for this openscad design? Use as few variables as possible
    \begin{gptcodeblock}{Javascript}
union() {
    difference() {
        cylinder(30, 13.5, 13.5);
        translate([0, 0, 1.5])
        cylinder(30, 12, 12);
    }
$\omittedCode{rest design}{}$
    \end{gptcodeblock}
\end{userprompt}

\begin{gptresponse}{0\textwidth}{} 
To expose high-level design parameters, we can identify the key variables that control the size and position of these shapes. Here's a version of your code with these parameters exposed:
    \begin{gptcodeblock}{Javascript}
// High-level design parameters
heightOfMainCylinder = 30;
outerRadiusOfMainCylinder = 13.5;
innerRadiusOfMainCylinder = 12;
$\omittedCode{other parameters}{}$
union() {
    difference() {
        cylinder(heightOfMainCylinder, outerRadiusOfMainCylinder, outerRadiusOfMainCylinder);
        translate([0, 0, 1.5])
        cylinder(heightOfMainCylinder, innerRadiusOfMainCylinder, innerRadiusOfMainCylinder);
    }
$\omittedCode{rest design}{}$
\end{gptcodeblock}
\end{gptresponse}

\begin{userprompt}{0.03\textwidth}{} 
Can you expose high-level design parameters for this openscad design of a mug? Use as few variables as possible
\end{userprompt}

\begin{gptresponse}{0\textwidth}{} 
Sure, let's define some high-level design parameters for this mug model.
    \begin{gptcodeblock}{Javascript}
module mug(mug_height = 30, mug_outer_diameter = 13.5, mug_wall_thickness = 1.5, handle_thickness = 1.5, handle_length = 10, handle_height = 4) {
union() {
    difference() {
        cylinder(mug_height, mug_outer_diameter, mug_outer_diameter);
        translate([0, 0, mug_wall_thickness])
        cylinder(mug_height, mug_outer_diameter - mug_wall_thickness*2, mug_outer_diameter - mug_wall_thickness*2);
    }
$\omittedCode{rest design}{}$
    \end{gptcodeblock}
\end{gptresponse}

\end{chat}
\caption{\textbf{Semantic Cues for Mug Design Parameters.} When asked to parametrize a mug design, it \gpt gives different parametrization with or without context cue.}
\label{fig:mug_parametrization}
\vspace{-5mm}
\end{figure}

First, when given a design with no semantic context, we observe that \gpt exposes design parameters based on equivalence between numerical values and based on which design operators these values were used in.
For example, in Fig.\ref{fig:chair_parametrization}, it introduces a variable \lstinline{cube_size} which replaces the value \lstinline{19} which was used for both the chair's width and length.
For the mug in Fig.\ref{fig:mug_parametrization}, we can observe that the exposed variables also stay close to their original usage for a given geometric operator.

Second, we repeat the previous experiment with additional semantic context.
Providing \gpt with the name of the object that is being modeled proves useful for generating a parametric design.
We can see that now, design parameters get exposed which are semantically more useful for modifying the design.
For example, the cylinder radii in Fig.~\ref{fig:mug_parametrization} gets replaced for a parameter \lstinline{mug_wall_thickness} which controls the thickness of the mug by considering both radii jointly.
Also, some ambiguity caused by numerical equivalence can be resolved and produce more useful parametrizations.
In Fig.~\ref{fig:chair_parametrization}, the \lstinline{cube_size} from the previous parametrization without any semantic context, gets disentangled into a \lstinline{length} and a \lstinline{width} parameter, allowing to have more control over the shape.
This might prove especially useful in this case, since all the slats are associated to the chair's width and not its length.

\begin{figure}[tb]
\begin{chat}{false} 
\begin{userprompt}{0pt}{} 
Can you give bounds for each design parameter? 
\end{userprompt}
\begin{gptresponse}{0.5\textwidth}{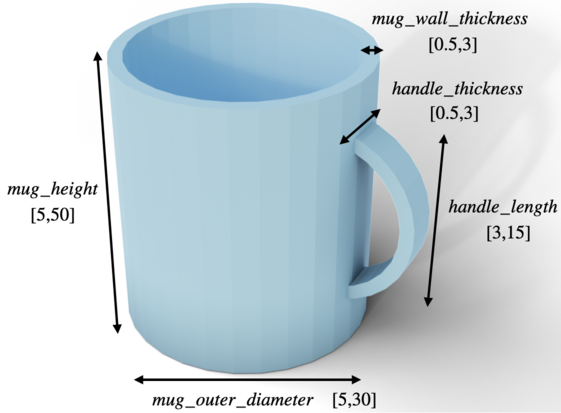}
Bounds for each parameter could be something like this:
\\
\\
    \lstinline{handle_thickness}: This parameter could range from $0.5$ (for a thin handle) to $3$ (for a thick handle). So the bounds are $[0.5, 3]$.
\\
\omitted{explanation of other bounds}
\\
\\
Note that these bounds are somewhat arbitrary and based on common mug sizes. You could adjust them based on your specific needs or the scale of your 3D printing.
\end{gptresponse}
\end{chat}
\caption{\textbf{Mug Design Parameter Bounds and Constraints.} \gpt gives parameter bounds based on common mug sizes.}
\label{fig:mug_bounds}
\end{figure}

Once parametrized, we can complete the design space by asking for parameter bounds, see Fig.~\ref{fig:mug_bounds}.
Again, notice how these bounds are \lstinline{somewhat arbitrary} and not based on the 3D design sequence.

While these results are encouraging, GPT-4 is easily confused by the final effect of a series of geometric transformations.
An example for this is the generated parameter \lstinline{handle_thickness} in Fig.~\ref{fig:mug_parametrization} which actually modifies the $y$ position of the handle.
Once again, it is limited by cases where geometric computation prevails over semantic reasoning.

%

\subsection{Can GPT-4 create a design space from multiple designs? (Q3)}
Design spaces based on a single design are useful to explore the family of possible shapes generated by varying the design parameters.

However, sometimes a designer might want to make more structural changes, inspired by another design of the same object class, they want to interpolate them.
Interpolating two designs can be difficult to achieve and there are a number of difficult questions which arise:
Are two designs modeled in a similar way?
Do they have the same dimensions and if not, how do you match the dimensions between two sub-designs?
Do you have to add extra operations to combine two parts?
Can you actually extract a subpart of an object from a design?
If you cannot exactly extract a sub-design, can you design something which is \textit{inspired} by two design sequences?
How do you accurately refer to two sub-designs in a text prompt?

To investigate if \gpt can help with design interpolation, we test three different design scenarios.
All of the designs were presented to \gpt in our sketch-based parametric CAD DSL, explained in Sec.~\ref{sec:text_to_design}.

\begin{figure}[tb]
\includegraphics[width=\textwidth]{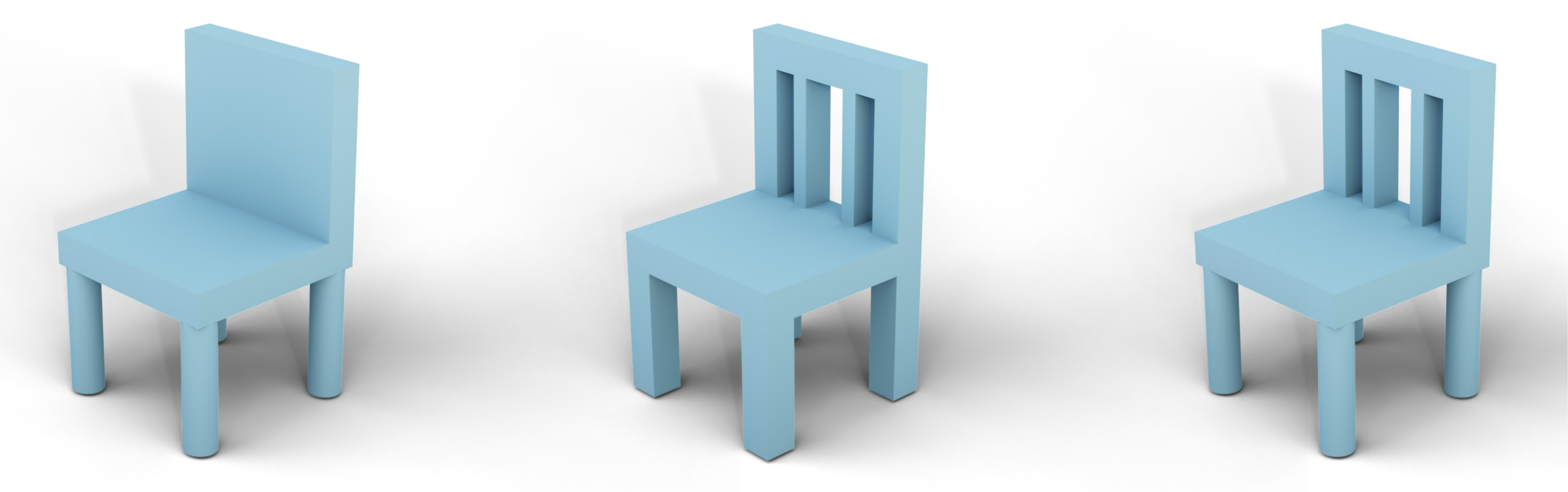}
\caption{\textbf{Chair Interpolation.}  \textbf{Left}: Input chair design with cylindrical legs. \textbf{Middle}: Input chair design with splats. \textbf{Right}: Interpolated output chair design with cylindrical legs and splats.}
\vspace{-5mm}
\label{fig:interpolation_chairs}
\end{figure}

%
%
%
%
%
%

First, we present it with two chairs which are modeled similarly, but the first chair has cylindrical legs and the second chair has a backrest with splats, see Fig.~\ref{fig:interpolation_chairs}.
In our prompt, we ask if it can \lstinline{mix these two designs to create a chair with cylindrical legs and splats in the back}.
The result can be seen in Fig.~\ref{fig:interpolation_chairs} (c).
It should be noted that variables in the code are descriptive, e.g. \lstinline{leg4_solid} and \lstinline{splat_3_sketch}, which helps provide semantic cues.
Also, in our designs, the first half of the code describes the construction of the seat and the legs and the second part describes the construction of the backrest.
This means that mixing these two designs comes down to replacing the second half of the first design with the second half of the second design.

\begin{figure}[tb]
\includegraphics[width=\textwidth]{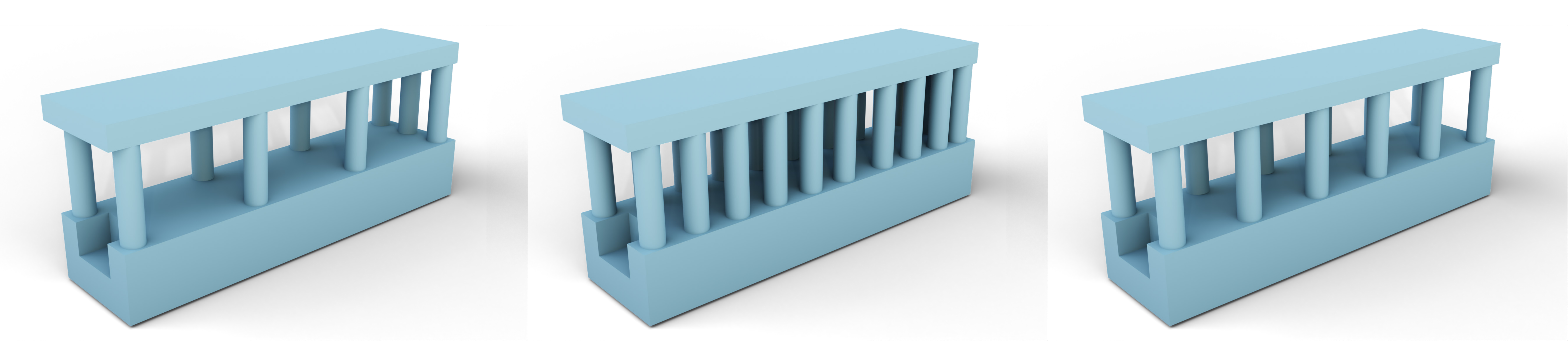}
\caption{\textbf{Temple Design Interpolation.} \textbf{Left}: Input temple design with 4 pillars on each side. \textbf{Middle}: Input temple design with 10 pillars on each side. \textbf{Right}: Interpolated output temple design with 6 pillars on each side.}
\label{fig:temple_interpolation}
\end{figure}
Next, we present \gpt with two designs of a temple, involving a different number of pillars, one with 4 pillars and one with 10 pillars on each side, see Fig.~\ref{fig:temple_interpolation}.
In our prompt, we ask it to \lstinline{design a temple with steps, a roof and 6 pillars on the left and right side}.
For this, \gpt has to find how these pillars have been modeled and how to model a varying number of pillars, given the two input examples.
The code for the design of the pillars did not contain any looping structures nor variables and it was more spread out throughout the program than in the chair example, to make it more challenging.
Despite these challenges, \gpt manages to extract the construction logic of the pillars and introduces variables and a looping structure to place them correctly, see Fig.~\ref{fig:temple_interpolation} (c).
Note that we have mentioned the steps and the roof in the prompt. We have noticed that without this reminder, it would solely focus on the construction of the pillars and forget about the rest of the design.

\begin{figure}[tb]
\includegraphics[width=\textwidth]{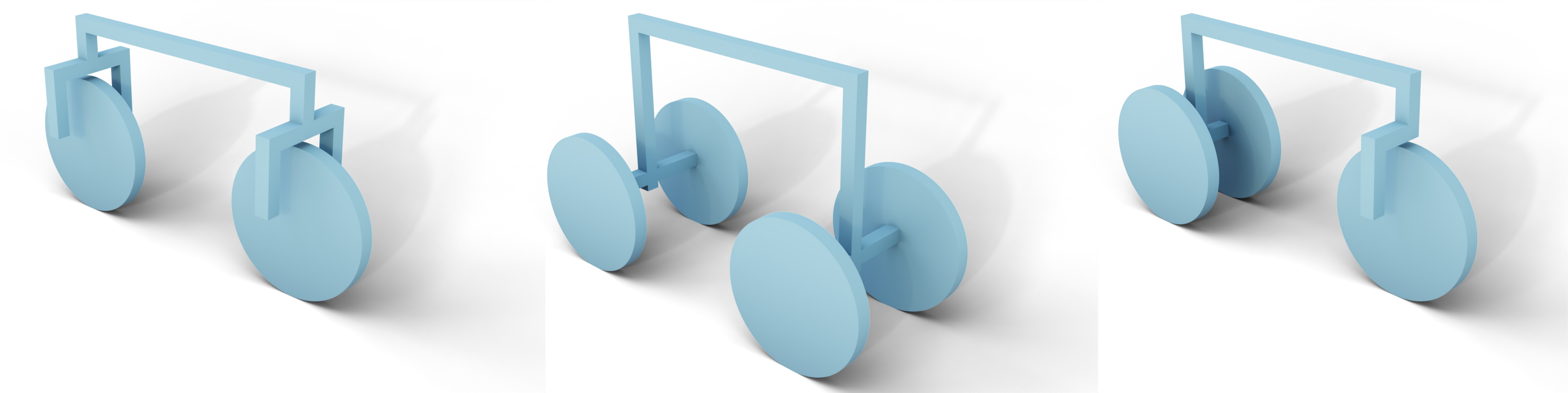}
\caption{\textbf{Bike Design Interpolation} \textbf{Left}: Input bicycle design . \textbf{Middle}: Input quad-bike design. \textbf{Right}: Interpolated output tricycle design.}
\label{fig:bicycle_interpolation}
\end{figure}
Our last test is structurally more challenging.
We present \gpt with a design of a bicycle and a design of a quad-bike, see Fig.~\ref{fig:bicycle_interpolation}. 
The two designs differ not only by the number of wheels in the front and the back, but also by the construction of the bike forks.
In the case of the bicycle, the fork surrounds the wheel and in the case of the quad-bike, the wheels are connected by a horizontal bar to the vertical bar of the frame.
This makes the mixing of sub-designs more complex.
And indeed, when asked to design a tricycle, \gpt reasons correctly about the number of wheels in the front and the back, and where to find these structures.
It also adjusts the size of the quad-bike's vertical bar such that the two back wheels and the front wheel are on the same plane.
This was not the case for the quad-bike and the bicycle in the input designs.
But it does not succeed at extracting the complete fork from the bicycle design, as can be seen in Fig.~\ref{fig:bicycle_interpolation} (c). 
Note that this experiment was performed via a single prompt and \gpt would likely be able to copy the missing part via further interaction with the user.

We find these examples promising, as they show how \gpt manages to combine its general knowledge about part relationships and its coding abilities.
One of the observed limitations is the ability to extract long sub-sequences and to detect which other parts are still important for plausible interpolation.

%

\subsection{Exploration of a given Design Space (Q4)}
A design space is conceptually useful to reliably generate variations of a given design.
However, coming up with parameters which represent \textit{meaningful} design variations can be a time-consuming iterative process.

\begin{figure}[tb]
\begin{chat}{false} 

\begin{userprompt}{0.25\textwidth}{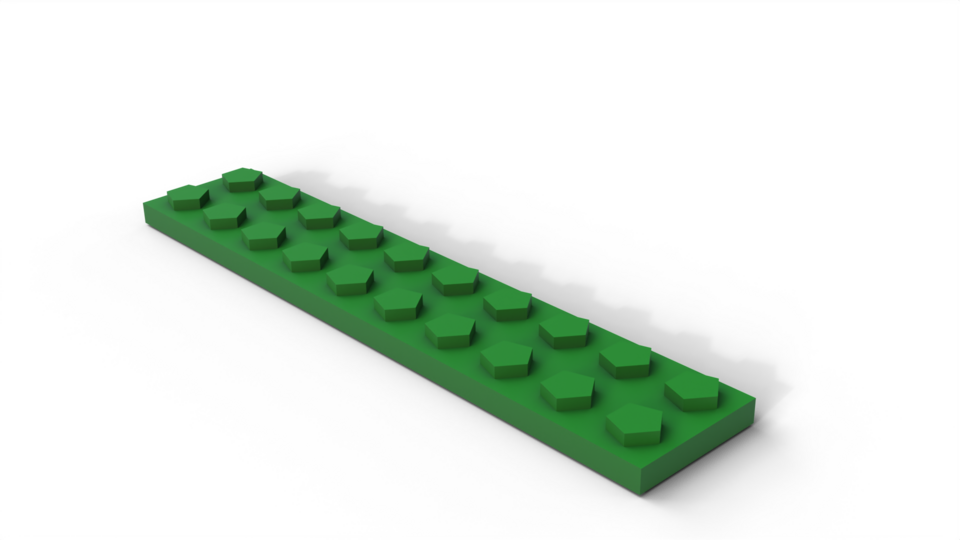} 
This is an openscad design for a lego brick:
\begin{gptcodeblock}{Javascript}
module lego_brick(brick_length=30, brick_width=6, brick_height=1) {
    union() {
        cube([brick_length, brick_width, brick_height]);
        for (i = [0:brick_length/3-1])
        for (j = [0:brick_width/3-1])
        translate([1.5 + 3*i, 1.5 + 3*j, brick_height-1/2])
        cylinder(1, 1, 1);
    }
}
\end{gptcodeblock}
Can you give parameter bounds (in form of intervals) and parameter constraints (in form of inequalities)?
\end{userprompt}

\begin{gptresponse}{0pt}{} 
Parameter bounds could be like this:\\
\lstinline{brick_length: [3, Infinity)} - In this design, the length of each stud is represented as 3 units, therefore the minimum is 3. // ...\\
\lstinline{brick_width: [3, Infinity)} \omitted{explanation of bounds}\\
\\
\lstinline{brick_height: [3, Infinity)} \omitted{explanation of bounds}\\
\\
Parameter constraints in the form of inequalities:\\
\omitted{other constraints}
\\
\\
\lstinline{brick_length \% 3 = 0} - The length of the Lego brick should be a multiple of 3 to make sure studs are equally spaced and the brick ends with the complete stud.
\end{gptresponse}

\end{chat}
\caption{\textbf{Lego Design Space.} We ask \gpt if it can generate parameter bounds and constraints for a parametric Lego brick design.}
\label{fig:lego_parametric_design}
\vspace{-5mm}
\end{figure}

\begin{figure}[tb]
\includegraphics[width=0.7\textwidth]{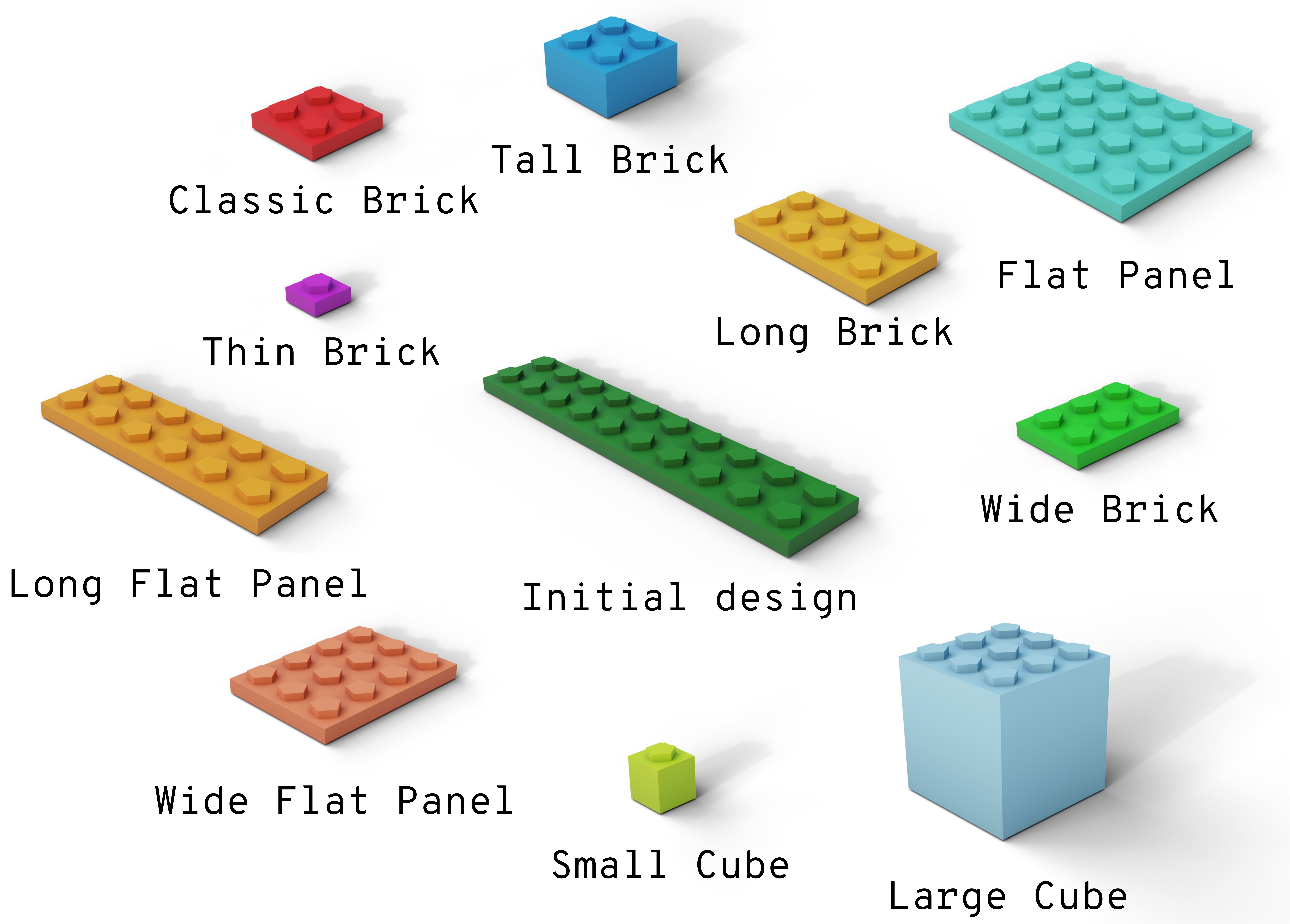}
\caption{\textbf{Lego Design Space Exploration.} \gpt generates 10 different design variations for the parametric \lstinline{Initial design}. The label under each model corresponds to the name given by \gpt.}
\label{fig:lego_exploration}
\vspace{-5mm}
\end{figure}

To investigate if \gpt can help with this task, we perform the following experiment.
We present it with a parametric design of a Lego brick, see Fig.~\ref{fig:lego_parametric_design}.
Then, we ask it to generate \lstinline{parameter bounds} and \lstinline{parameter constraints}.
Interestingly, \gpt generated the non-trivial constraint that the length and width of the brick should be multiples of 3.
We ask it to use the design space to \lstinline{come up with 10 different parameter settings which correspond to meaningful lego bricks}.
Finally, it should \lstinline{give each variation a name}, see Fig.~\ref{fig:lego_exploration}.

We can observe that the proposed parameter settings respect the previously generated bounds and constraints and that they lead to distinct 3D models, for which it generates plausible semantic labels.

\subsection{Discussion}

In this section, we summarize the key capabilities (C), limitations (L), and dualisms (D) specific to the creation and manipulation of design spaces.

\noindent \textbf{C.1 Extensive Knowledge Base in Design and Manufacturing:} 
We observe that we can leverage \gpt's semantic knowledge base to create parameters, bounds and constraints for text-based designs and already existing designs.
Additionally, \gpt can be useful for finding semantically meaningful design variations in a given design space.

\noindent \textbf{C.3 Modularity Support:} 
We observe that \gpt can interpolate existing designs by extracting and adapting sub-designs based on their program representations.
Interestingly, even when designs are not presented in a modular fashion, it tries to recognize and abstract sub-modules in input designs.

\noindent \textbf{L.1 Reasoning Challenges:} 
The design spaces created by \gpt are based both on semantic knowledge and on code interpretation. 
However, it does not take into account geometric considerations, such as intersecting or non-connecting parts.
As a result, generated parameter bounds can create non-valid geometry and it has proven difficult to make \gpt correct these.
However, in general the generation of valid parameter bounds and constraints is a difficult problem for which mainly approximations have been proposed \cite{mathur2021constraint}.

\noindent \textbf{L.3 Scalability:} 
The interpolation task revealed that \gpt has limited capabilities to infer what parts of a design should be linked to a semantic part specified in a prompt. 
One promising future direction to manage increasingly complex designs is to make them increasingly modular by adding intermediate levels of abstraction.

\noindent \textbf{D.1 Context Information:}
We observe that the generation of correct parametric designs and the reparametrization of already existing designs can be improved by providing semantic cues, such as the name of the modeled object.
As seen in Sec.\ref{sec:text_to_design}, \gpt creates designs which contain a lot of semantic information and it generally performs even better when using meaningful variable names.
Leveraging this aspect in the generation of design spaces and throughout other aspects in the design process should prove extremely useful.

%% file: sections/GraphArrows/DesignForManufacturing.tex
\section{Design-For-Manufacturing}
\label{sec:design_for_manufacturing}

The utilization of \llms in the context of Design for Manufacturing (DfM) provides a broad range of applications that have the potential to enhance the design and manufacturing process of different parts and assemblies. One useful application of \llms involves leveraging their pattern identification and language interpretation capabilities to imitate a manufacturing expertise bank that can be tapped into during various parts of the design and manufacturing stages. Furthermore, because \llms such as \gpt have the ability to create programs and find and interpret patterns in text, it can potentially be used to generate and alter design and manufacturing files. Currently, DfM is often accomplished by human expertise with the aid of CAD software. Engineers and designers review design plans and use their industry experience to suggest alterations that would improve manufacturability. The CAD software then allows these alterations to be modeled. The replacement of human manufacturing knowledge with \gpt in this context could streamline the design for manufacturing process, offering more consistent, scalable, and efficient decision-making, which is not limited by individual human capacity.

\begin{figure}[tb]
\includegraphics[width=\textwidth]{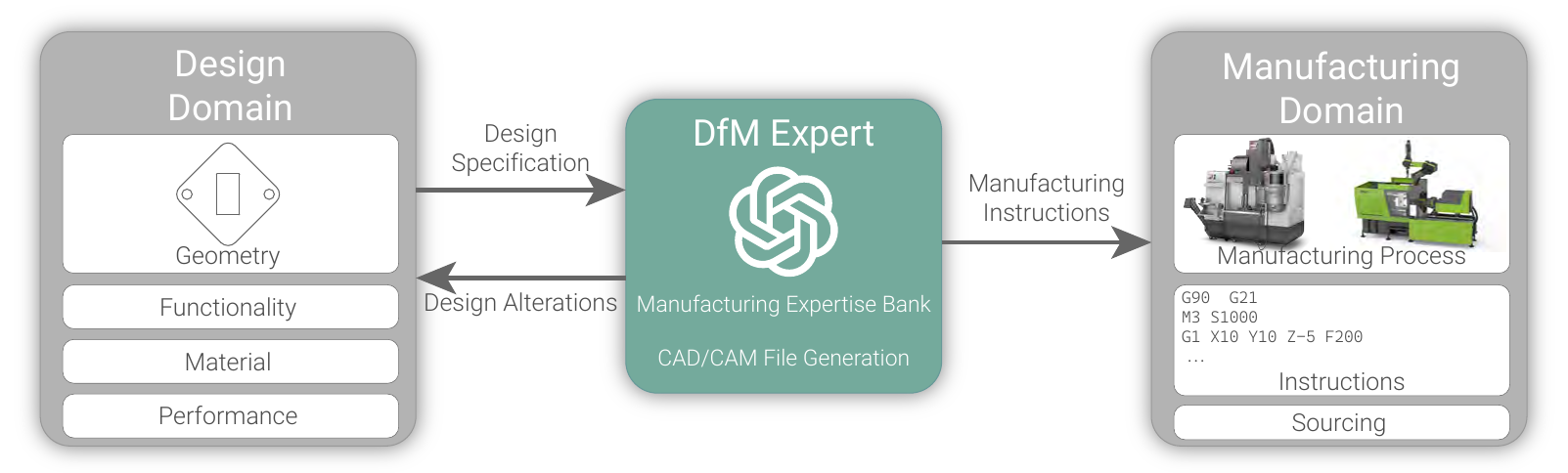}
\caption{\textbf{Integration of GPT-4 into Design for Manufacturing(DfM):} \gpt can be used to augment the DfM Process when designing a part.}
\label{fig:DfM_Overview_Figure}
\end{figure}

In this section, we propose multiple ways that this new manufacturing expertise bank could be used in design and manufacturing, as shown in \fref{fig:DfM_Overview_Figure}. \gpt can be used to select optimal manufacturing techniques based on a part's features. Furthermore, it can propose and implement modifications to a design to improve its manufacturability, ultimately leading to more efficient production processes. Additionally, this idea can be extended to part sourcing by leveraging the model's reasoning capabilities to identify potential suppliers based on the part's desired function and performance. Finally, it could be used to develop manufacturing instructions for various processes. To understand \gpt's ability to alter designs based on manufacturing/sourcing constraints, we pose the following questions:

\begin{itemize}
    \item \textbf{Q1} Given a part geometry, production run and other desired outcomes, can \gpt select optimal manufacturing processes? 
    \item \textbf{Q2} Given a manufacturing process, can \gpt directly suggest and make design alterations to a parts file based on constraints driven by the process capabilities? 
    \item \textbf{Q3} Given a desired functionality and geometric specifications, can an LLM find a source for a part that fits those specifications? 
    \item \textbf{Q4} Given a design can an LLM create a set of manufacturing and assembly instructions? 

\end{itemize}

\subsection{Finding Optimal Manufacturing Process (Q1)}
\begin{figure}[tb]
\includegraphics[width=\textwidth]{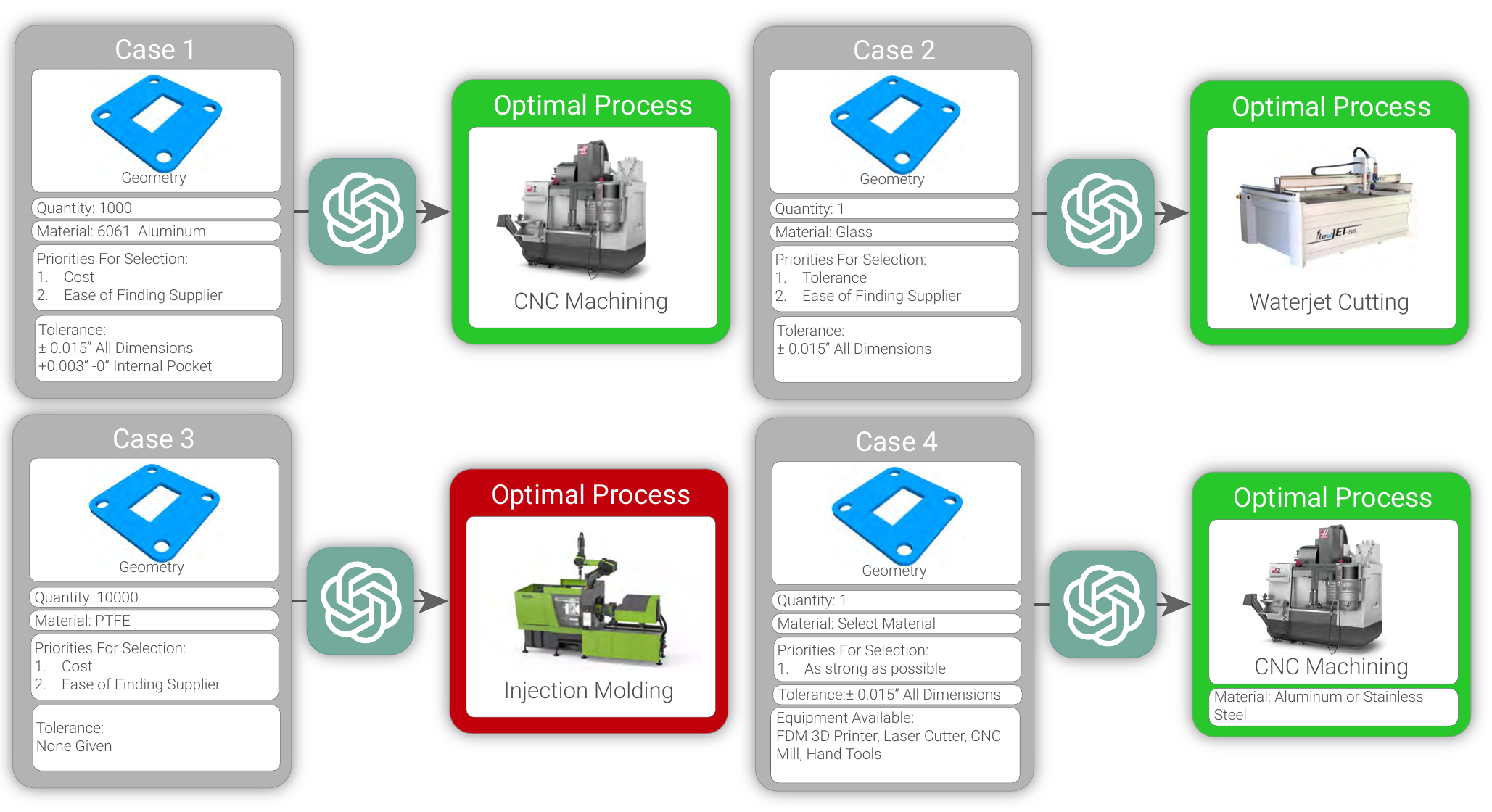}
\caption{\textbf{Using \gpt to select optimal manufacturing process:} Inputs and selected optimal manufacturing processes using \gpt.}
\label{fig:Optimal_Manufacturing_Process}
\end{figure}

To test these capabilities, we tasked \gpt with advising on identifying an optimal manufacturing process for a part with the geometry shown in \fref{fig:Optimal_Manufacturing_Process}. We tested it with four different cases where in each case the geometry, material, tolerance requirements, and quantity were varied. We described the part's geometry as an \jscad file. Finally, given a set of priorities, we task \gpt to select an optimal manufacturing process. In Case four, we provided a finite list of manufacturing processes to evaluate the effectiveness of the selection process under the constraint of a limited set of options. The goal was to determine how well the process could choose the appropriate manufacturing processes to meet the specified priorities.

\begin{figure}[tb]
\begin{chat}{true} 

\begin{userprompt}{0.12\textwidth}{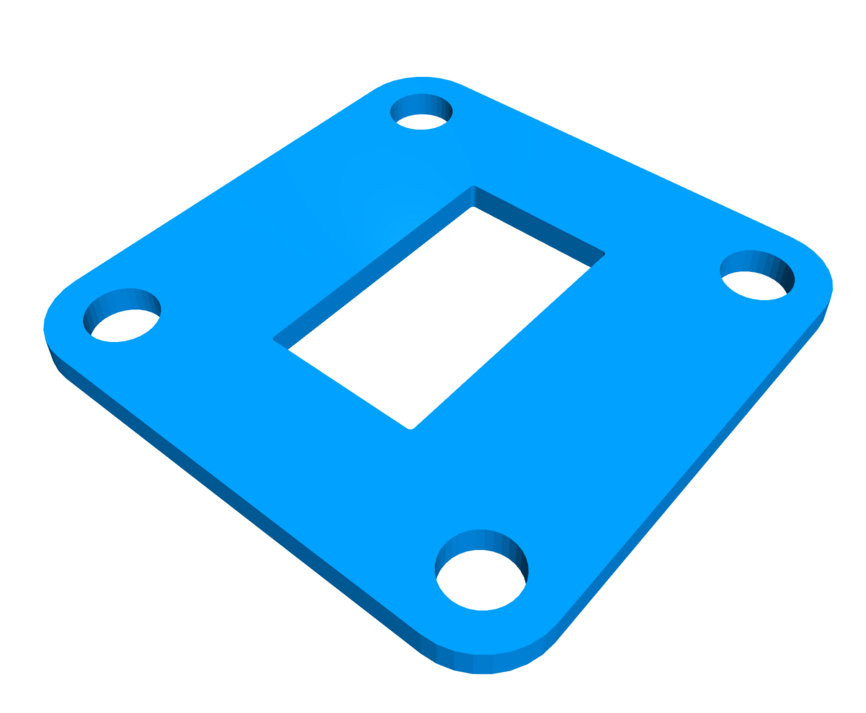} 
...  You are trying to select an optimal set of manufacturing process to make the part. 

...The parts must be manufactured to the tolerance of plus or minus 0.015 inches. 

You need to make 10000 part out of PTFE. 

... priorities for selecting a manufacturing process are the following: 
1. Cost
2. Ease of finding multiple suppliers.
\end{userprompt}

\begin{gptresponse}{0\textwidth}{} 
Given the description of the part provided, the best method to produce this component from PTFE (Polytetrafluoroethylene) in high volume, with a tight tolerance of +/- 0.015 inches would be injection molding.

...
\textbf{}
Regarding the material choice, PTFE is a common material for injection molding due to its high melting point, excellent chemical resistance, and excellent electrical insulating properties. However, it's worth noting that PTFE can be challenging to mold due to its high melt viscosity and tendency to warp as it cools. Proper mold design and processing conditions are critical to achieving the desired part dimensions and properties.

...
\end{gptresponse}

\end{chat}
\caption{\label{fig:PTFE_Example} \textbf{Prompt and Response for Selecting an Optimal Manufacturing Process for a PTFE Part.} \gpt gives the response that PTFE is commonly injection molded, which is incorrect. Typically, it is not suggested that PTFE be injection molded.}
\end{figure}

\gpt was successful at selecting an optimal manufacturing process for three out the four cases. For cases one, two, and four, \gpt selected the optimal process that was approved by an expert. However, in case three, shown in \fref{fig:PTFE_Example}, \gpt suggested an injection molding process, which is not suitable for processing a Polytetrafluoroethylene (PTFE) material. In all cases, \gpt initially only provided a range of manufacturing options; it required additional prompts to arrive at the optimal manufacturing process selection.

\subsection{Design Alterations for Manufacturability (Q2)}
\begin{figure}[h]
\includegraphics[width=\textwidth]{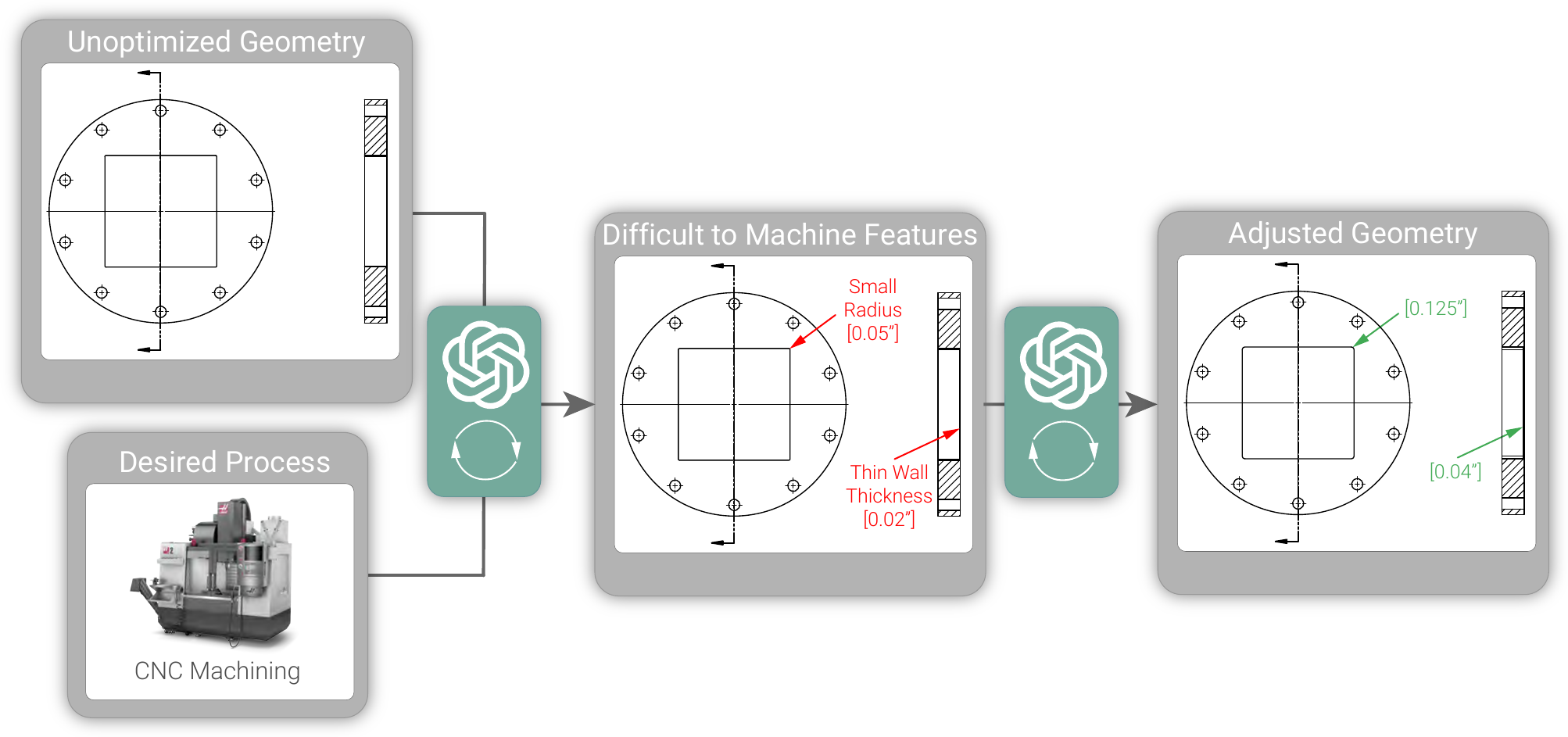}
\caption{\textbf{Using an \llm to Find Difficult to Manufacture Features and Directly Adjust the Geometry.}\gpt is given the task to identify any features that are difficult to machine along with an \jscad file. After a number of iterations, it identifies a set of features. We prompt it then to make changes to the \jscad file to reduce the machining difficulty.}
\label{fig:Optimal_Manufacturing_Design}
\vspace{-5mm}
\end{figure}

In this section, we assessed \gpt's capability to enhance designs for better manufacturing optimization. To accomplish this, we included the text of a \jscad file in the prompt, allowing \gpt to analyze and modify it accordingly. Our focus in this case was on the CNC machining of a 10-inch diameter disk, which involved creating bolt holes along the edge and a central blind square pocket. We included in the geometry, two intentional features that would be difficult to machine. As depicted in \fref{fig:Optimal_Manufacturing_Design}, the process began with \gpt identifying any manufacturing complexities within the design features. Since an \llm interprets text, \gpt interprets the text of the \jscad file, rather than the geometry that is rendered once compiled which humans interpret. After \gpt identifies any complexities, we instructed it to adjust the geometry of the \jscad file to address the challenging aspects by directly changing the text of the \jscad file.

Although \gpt accomplished these tasks with a moderate degree of success, there were a few inaccuracies. Firstly, \gpt correctly identified two potential machining issues: the small radius of the internal pocket and the thin wall at the pocket base. However, it also misunderstood a number of geometric features described in the \jscad file. These include perceiving holes on a curved surface and anticipating an undercut from the pocket. These misinterpretations might be attributed to \gpt's reliance on the text of the \jscad file for feature identification, as some features become more visible once the file is compiled into a geometric representation. After pointing out these interpretation errors to \gpt, it was able to correct its analysis but introduced another mistake. \gpt incorrectly stated that the bolt holes presented machining difficulties and inquired about additional information regarding the machining area. Once provided with the necessary details, \gpt independently rectified its mistake about the bolt holes. \gpt was also aware of potential issues with the size of the part and machining area of the CNC machine. Furthermore, it was able to compute whether there was a potential issue. 

In the final stage, \gpt was asked to modify the \jscad file to address the manufacturing concerns. It improved the wall thickness from 0.02" to 0.04", making it machinable. Given the additional specification of utilizing a 1/4" endmill, \gpt also adeptly adjusted the internal pocket's radius to accommodate this tooling requirement better.

\subsection{Part Sourcing (Q3)}
\label{sec:part_sourcing}
The massive dataset backing \llms contains some specialized knowledge about parts needed for manufacturing. Consequently, we posit that \llms can be useful for reasoning about these parts, from identifying the correct part names to describing necessary properties for their functionality.

\begin{figure}[tb]
\centering
    \includegraphics[width=0.7\textwidth]{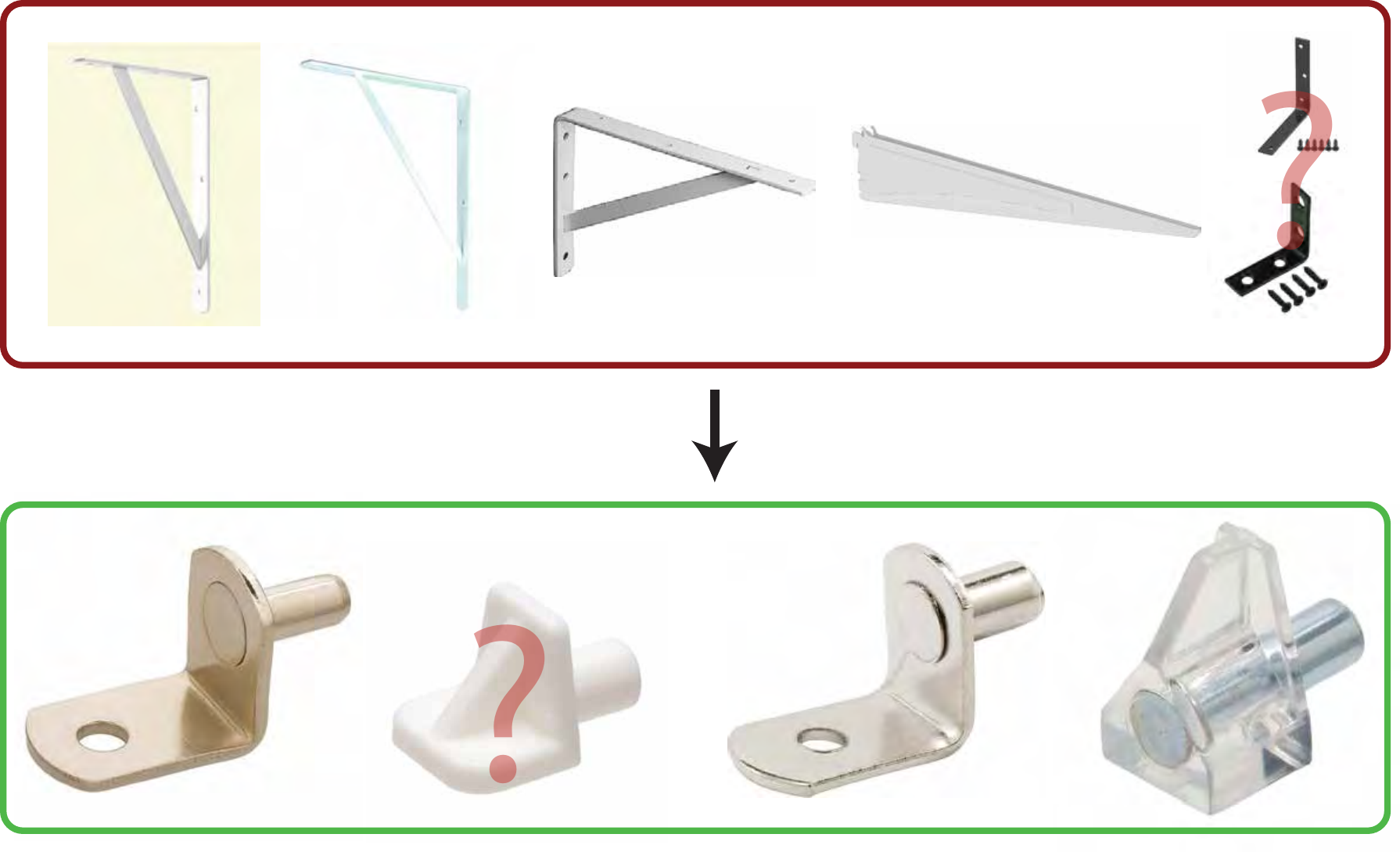}
    \caption{ \textbf{Using \gpt to Suggest Shelf Brackets for A Cabinet.} \textbf{Top:} On its first attempt, \gpt suggested five bracket products that were too large for the scale of the project, on the scale of tens of inches, rather than less than an inch. \textbf{Bottom:} After adjusting for our feedback, it suggested four new, appropriately-sized brackets. One bracket `` lock[s] the wire shelf to the vertical [ShelfTrack] Wall Standards to create a sturdy storage space'', whereas another was ``not technically a shelf bracket... could be used to provide support underneath a shelf'', and yet another was ``plastic supports that are designed to be unnoticeable''. Several of the links were either broken (still available on Home Depot at another URL) or no longer available on the Home Depot store (but available on e.g. Lowes, Walmart, or Amazon), and seven of the nine suggestions were still available and discoverable after a Google search of the product name. For the two that were not available (highlighted with red question marks), one was the ``Everbilt 5x3 black corner brace'', and there were two similar brackets that were 5x1 and 3x3, and another (the ``Everbilt 20 lb. Plastic Invisible Shelf Support Bracket'') did not seem to exist as labelled, though there was a product found that fit this description.}
    \label{fig:brackets}
\end{figure}

\paragraph{Cabinet part sourcing} As part of generating the design and fabrication instructions for our cabinet, we asked \gpt to find appropriate shelf brackets for the shelf within the cabinet, starting from a concrete design specification in \jscad. In each iteration, \gpt provided several suggestions as links to products on Home Depot, with a short sentence differentiating them. Numbers in the part descriptions were inaccurate: one bracket pair held up to 300 lbs, but \gpt claimed it could hold 1000. Another pair was a `` heavy-duty option that can support up to 500 lbs. when properly installed.'', but could actually hold 1300 lbs. Otherwise, the short descriptions were true, and all described parts could plausibly serve as shelf brackets.
Figure~\ref{fig:brackets} shows the presumed brackets suggested. 
 Overall, we found success for this relatively simple use case.

\paragraph{Copter part sourcing} We also asked for help sourcing parts when designing the quadcopter example. First, we asked for a parts list that would encompass everything needed for the design. \gpt compiled a list including batteries, frames, propellers, transmitters and receivers, electronic speed controllers, etc. We found that the list was comprehensive and accurate. 
Next, we tried narrowing down the response from a list of parts to a list of specific parts with more tailored guidance for each use case. 
Asking for a range of numerical specifications (e.g. specific amperages for batteries) produced correct and sensible numerical estimates for parts.
Specifying that the copter should be able to hold a weight of 10kgs for 10 minutes yielded a list of
very large and powerful parts. Specifying an indoor copter led to smaller and more lightweight part suggestions.
Pushing \gpt beyond specification resulted in errors. 
Asking for specific names of part listings or parts and manufacturers, as in~\autoref{fig:copter-parts} tended to result in lists with incompatible parts, or in naming parts that do not exist. 
Iterating on the errors with \gpt, as seen in our follow-up question in~\autoref{fig:copter-parts}, produced correct new parts.

Though asking \gpt to produce the names of real-world parts was unsuccessful, we still found impressive results in its comprehensiveness and ability to form fairly specific and accurate part lists. \gpt was also able to dispense meaningful advice on ensuring parts were compatible, even though it was unable to generate parts lists satisfying compatibility itself. We believe that \gpt can be a useful guide for delivering domain-specific knowledge and providing complete parts lists, but that precise numbers and specs should be cross-referenced before being used.

\begin{figure}[tb]
\begin{chat}{false} 

\begin{userprompt}{0\textwidth}{} 
Which components do I need to build an indoor drone that can fly at 30 mph? Prefer easily available components if possible. List the manufacturer, model, quantity, and Amazon search term for each component.
\end{userprompt}

\begin{gptresponse}{0\textwidth}{} 
Building an indoor drone that can fly at 30 mph requires several components. Here's a list of the necessary components along with their specifications and Amazon search terms:

Frame:

Manufacturer: Hobbypower / Model: Hobbypower F450 / Quantity: 1 / Amazon search term: Hobbypower F450 frame kit

Flight Controller:

Manufacturer: Holybro
/ Model: Kakute F7 Mini
/ Quantity: 1
/ Amazon search term: Holybro Kakute F7 Mini flight controller

Motors:

Manufacturer: EMAX
/ Model: EMAX RS2205S
/ Quantity: 4
/ Amazon search term: EMAX RS2205S motors

Electronic Speed Controllers (ESCs):

Manufacturer: Hobbypower
/ Model: Hobbypower SimonK 30A
/ Quantity: 4
/ Amazon search term: Hobbypower SimonK 30A ESC

Propellers:

Manufacturer: DALPROP
/ Model: DALPROP T5045C
/ Quantity: 4 pairs (8 propellers)
/ Amazon search term: DALPROP T5045C propellers

Battery:

Manufacturer: Tattu
/ Model: Tattu 4S 1300mAh
/ Quantity: 1
/ Amazon search term: Tattu 4S 1300mAh LiPo battery

Radio Transmitter and Receiver:

Manufacturer: FrSky
/ Model: FrSky Taranis X9D Plus
/ Quantity: 1
/ Amazon search term: FrSky Taranis X9D Plus transmitter and receiver

Power Distribution Board (PDB):

Manufacturer: Matek Systems
/ Model: Matek PDB-XT60
/ Quantity: 1
/ Amazon search term: Matek PDB-XT60 power distribution board

Propeller Guards (optional but recommended for indoor use):

Manufacturer: Crazepony
/ Model: Crazepony 2 Sets 65mm
/ Quantity: 1 set
/ Amazon search term: Crazepony 65mm propeller guards
\end{gptresponse}

\begin{userprompt}{0.03\textwidth}{} 
I already have a radio transmitter, please suggest a receiver only
\end{userprompt}

\begin{gptresponse}{0\textwidth}{} 
Certainly! If you already have a radio transmitter and only need a receiver, you can consider the following option:

Receiver:

Manufacturer: FrSky
/ Model: FrSky XSR
/ Quantity: 1
/ Amazon search term: FrSky XSR receiver

Please note that the FrSky XSR receiver is compatible with the FrSky Taranis X9D Plus transmitter, which you mentioned having.
\end{gptresponse}

\end{chat}
\caption{\textbf{Using \gpt to Suggest Amazon Parts for a Copter.} \gpt produced a list of specific parts, but made factual errors. For example, it listed Crazepony propeller guards on the parts list, but Crazepony does not sell propeller guards. After iterating, we noted four major inaccuracies within our final results, 
out of ten identified parts. Two  errors were that a certain part did not exist from the listed manufacturer, one was a redundancy where listed parts were already included in the purchase of another part (the flight controller kit recommended by \gpt included a power distribution board, so there was no need to purchase one separately), and one was an item that \gpt believed to be a transmitter and receiver, but was only a transmitter. A subsequent attempt to point this out rectified the issue and \gpt recommended a receiver correctly. We ordered a modified version of the parts list where the four errors were fixed.}
\label{fig:copter-parts}
\end{figure}

\begin{figure}[tbp]
\begin{chat}{false} 

\begin{userprompt}{0pt}{}
Hi ChatGPT. I'm trying to make an at-home carbonation system and I'd like to find the hose, fittings, and pressure gauge on McMaster-Carr. Can you help me? The jar I'm using has a 5 gallon capacity.
\end{userprompt}

\begin{gptresponse}{0.3\textwidth}{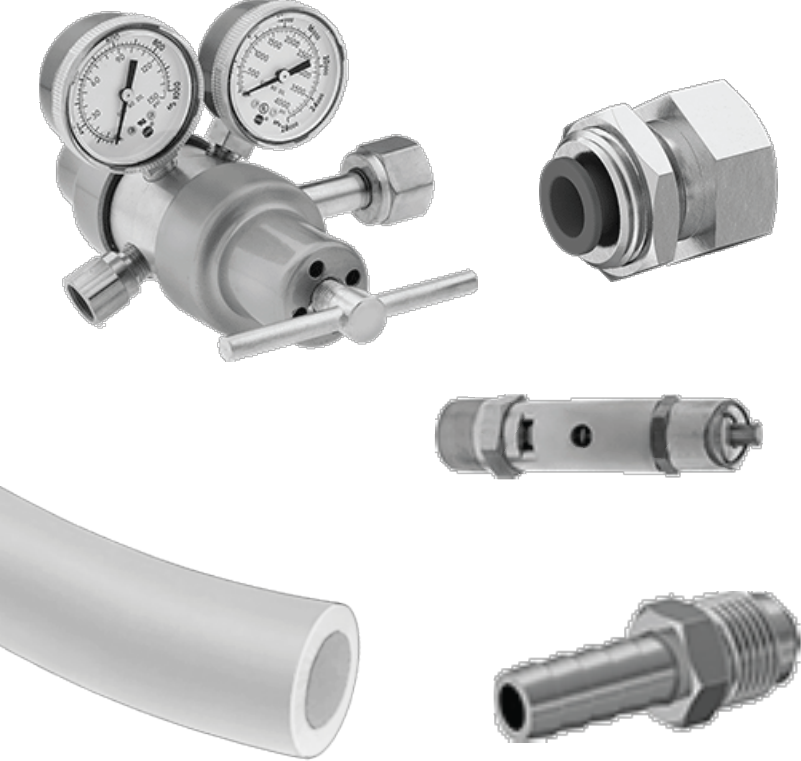} 
Hello, I'm glad you've asked for help with this project! To create a DIY carbonation system, you'll need a few items. ...
Here are some McMaster-Carr parts that could help:

\textbf{Regulator:} McMaster-Carr has a variety of regulators. An example is Part Number 5182K56 - a two-stage, brass, compressed gas regulator with CGA 320 inlet connection.

\textbf{Pressure Relief Valve:} For example, Part Number 48435K72, brass pressure-relief valve, set at 60 psi, which can be adjusted.

\textbf{Tubing:} Part Number 5648K35 can be an option. It's a firm tubing, clear, 1/4" ID, 3/8" OD.

\textbf{CGA320 adapter for tubing:} You'll need to adapt the CGA320 connection on the regulator to fit the tubing. Part Number 50785K51, a brass barbed tube fitting, can work for this.

\textbf{Bulkhead fitting for the jar lid:} For this, you can use Part Number 50785K191, a brass threaded-tube-through-wall fitting with a tight seal.

\textbf{Diffusion stone:} You can use a standard 0.5 micron diffusion stone. While this isn't available on McMaster-Carr, it can be found at various homebrew stores.

Remember to ensure all parts are compatible with each other (size, material, pressure rating) and appropriate for use with food and drink. 
\end{gptresponse}

\begin{userprompt}{0pt}{}
    Can you give me a bill of materials (BOM) for this project? ...
    Can you write it as a multi-level BOM,  including all the assemblies, sub-assemblies, and be in this format:
    Assembly Level - Quantity - Manufacturing or Sourcing information - Part number - Name - Price 
\end{userprompt}

\begin{gptresponse}{0pt}{}
    Absolutely, here's a Bill of Materials (BOM) for your project. Please note that the exact quantities will depend on your specific setup and you might need to adjust as needed: \omitted{}{long BOM}
\end{gptresponse}

\begin{userprompt}{0pt}{}
    Thank you! Can you make that slightly more condensed by removing the quantity number, price, and assembly level, since they appear redundat[sic]?
\end{userprompt}

\begin{gptresponse}{0pt}{}
    Of course, here's a more condensed version:

\begin{itemize}[leftmargin=*,label={}]
\item Carbonation System
\begin{itemize}[label={}]
    \item Manufacturing/Sourcing Information: Self-assembled
     \item Regulator Connection
    \begin{itemize}[label={}] 
    \item Manufacturing/Sourcing Information: Self-assembled
    \item Part 1: Sourcing Information: McMaster-Carr, Part Number: 5182K56, Name: Two-stage, brass, compressed gas regulator with CGA 320 inlet connection
    \item Part 2: Sourcing Information: McMaster-Carr, Part Number: 50785K51, Name: Brass Barbed Tube Fitting for CGA320 connection
    \item Part 3: Sourcing Information: McMaster-Carr, Part Number: 5648K35, Name: Firm Tubing, Clear, 1/4" ID, 3/8" OD
    \end{itemize}
\item Jar Connection
\begin{itemize}[label={}]
    \item Manufacturing/Sourcing Information: Self-assembled
        \item Part 1: Sourcing Information: McMaster-Carr, Part Number: 48435K72, Name: Brass pressure-relief valve
        \item Part 2: Sourcing Information: McMaster-Carr, Part Number: 50785K191, Name: Brass Threaded-Tube-Through-Wall Fitting
        \end{itemize}
\item Diffusion Setup
    \begin{itemize}[label={}]
    \item Manufacturing/Sourcing Information: Self-assembled
    \item Part 1: Sourcing Information: Homebrew store, Name: 0.5 micron diffusion stone
    \end{itemize}
\end{itemize}
\end{itemize}
This BOM is simplified and omits quantities and prices, which should be determined according to your specific setup and checked directly with suppliers. It also does not cover smaller items such as clamps or the CO2 tank, nor optional or safety items like check valves or non-return valves.
\end{gptresponse}

\end{chat}
\caption{\textbf{Using \gpt to Suggest Geometrically-compatible Parts from McMaster-Carr.} \gpt suggested part numbers for the regulator (top left), hose adapter (top right), pressure relief valve (middle), hose (bottom left), and hose fitting (bottom right). All part descriptions were sufficiently detailed such that they described a single part in the database after accounting for the hose inner and outer diameter (prescribed by \gpt), except for the regulator, which left a choice between a neoprene and stainless steel diaphragm. All part numbers were incorrect, except the part number for the pressure relief valve was surprisingly correct. \gpt then produced a comprehensive multi-level BOM.}
\label{fig:carbonation-parts}
\vspace{-5mm}
\end{figure}

\paragraph{Geometry-based part sourcing} McMaster-Carr is a deep compendium of knowledge for hardware parts, with geometric information and even CAD models available for many items. McMaster-Carr already has a ``search by geometry'' feature, so we wanted to know if we could perform higher-level searches that involve both context and geometry. First, we tried describing specific scenarios and asking \gpt for search terms that would procure us the correct part. Asking for a nut to be used in a tight space without room for a wrench and submerged in saltwater produced two appropriate results, ``316 Stainless Steel Wing Nuts'' and ``316 Stainless Steel Knurled-Head Thumb Nuts'', where the correct form and material was identified. Asking for a tamper-proof nut also produced the correct search, ``316 stainless steel tamper-resistant nut''. 
Next, we tried a more open-ended geometric compatibility scenario by asking for parts for an at-home carbonation system (\autoref{fig:carbonation-parts}). We also then asked it for a comprehensive Bill of Materials. It seemed as though all parts were compatible, at least geometrically; we suspect this is because the items in the domain are standardized for compatibility, McMaster-Carr's dataset is quite rich, and there is great availability of each part across varying sizes.

\paragraph{Parts of mechanisms} Part of what makes \gpt a compelling tool for design is its simple user interface. A user might interact with \gpt by describing a desired functionality and asking what parts would be necessary to achieve it. For example, we described a hypothetical bar cart with two features: a lower shelf with rails, and a tabletop where a portion could be folded down for compact storage. We asked \gpt to tell us what the name of the fold-down tabletop mechanism was, and recommend a part that could be used to build it. It correctly identified the function as a drop-leaf mechanism, explained that since the drop-leaf would be 20x15 inches, the mechanism should be at least 15 inches long, and named steel or brass as appropriate materials. It also was able to generate a specific search term for the part. However, \gpt did not recommend a particular type of mechanism in how it moved or functioned. 
We asked it to list the different sub-types and their use cases, which it did successfully, naming and differentiating a swing arm bracket, a slide-out support, a hinged bracket support, support bars, and a rule joint. We were able to find examples of four of the types, but the support bars, which were described as ``lengths of wood or metal that are stored separately and inserted into brackets on the table and leaf to hold it in place'', did not seem to exist under that terminology or perhaps at all. We then asked it to recommend a type for our use case, and it recommended the swing arm or hinged bracket supports.

We also tried a loom example, where we asked \gpt to provide a fabrication plan for a 4-shaft table loom, and asked about the name of the mechanism that lifts and lowers the heddle frames and the names of specific parts that make up this mechanism. In general, it was accurate, but \gpt sometimes erroneously named components that only pertained to countermarche looms or floor looms instead of table loom-specific parts. We speculate that this could be due to a dearth of literature on loom construction in \gpt's training dataset. 

Our examples show potential in using LLMs to identify and source parts, with major caveats. We note a recurring theme of \gpt's ability to produce programs that generate valid programs or instructions to verify validity, and its inability to apply those rules to its own output. In general, we found that we could ask for general, pointed, and precise guidance with great success, but asking for product names or specific items often resulted in incompatible or nonexistent parts lists. Furthermore, best results were produced in the simpler and more common domains, or when the domain we were querying had very rich information, as was the case with McMaster-Carr. We believe that \gpt is useful for making comprehensive checklists, and can lend domain expertise and suggestions, so long as all information can be checked or cross-referenced. Since \gpt can interface across many levels of jargon, experts may derive the most value from its use currently, given that they are best able to make common sense checks over the output. For non-domain experts, \gpt delivers very convincing, confident information that can be incorrect. LLMs are poised to become a powerful ``design for everyone'' tool, but more verification steps are needed to guide novice users. 

\subsection{Create Manufacturing Instructions (Q4)}
Computer-Aided Manufacturing (CAM) is a technology that utilizes software to generate manufacturing instructions from digital design files. It plays a vital role in the efficient and accurate translation of design concepts into tangible products. CAM bridges the digital design and the physical manufacturing stages, enabling seamless communication and translation of design specifications into machine-readable instructions. CAM encompasses a range of techniques and tools that leverage computer systems to automate various manufacturing processes, including planning, toolpath generation, and machine control. By utilizing CAM, manufacturers can streamline production, improve precision, and enhance overall efficiency. In this section, we delve into the creation of machine-readable and human-readable manufacturing instructions with help of \gpt and open-source CAM software. Specifically, we explore additive, subtractive, and assembly manufacturing processes, highlighting the capabilities and challenges associated with each approach.
\subsubsection{{Additive}}
\label{sec:additive}
Additive design, often employed in the realm of 3D printing, can be time-consuming and labor-intensive, requiring spatial reasoning, precision, and multiple iterations. We posit that \gpt will improve this process, as it comprehends complex specifications in natural language, generates designs efficiently, simulates outcomes, and explores innovative possibilities from diverse sources, optimizing functionality and aesthetics.

\begin{figure}[tb]
\begin{chat}{false} 

\begin{userprompt}{0.2\textwidth}{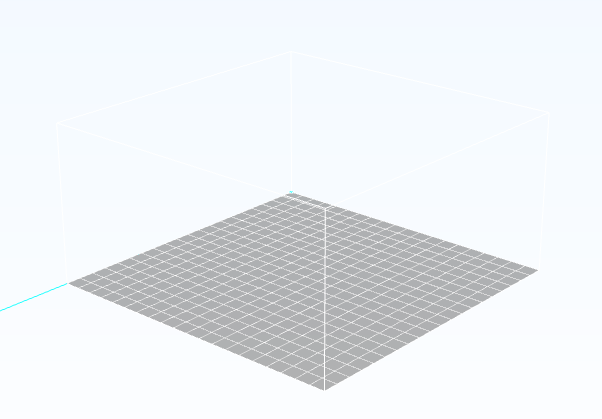} 
The object I want to print is a simple solid 3D ball with radius of 1. Generate a sample g-code with a simple slicing for it in a virtual printer.
\end{userprompt}

\begin{gptresponse}{0pt}{} 
G-code generation is complex and it is heavily dependent on the specifics of the 3D printer being used, the material properties, the specific layer thickness, the nozzle size, print speed, and more. Hence, it's a process that should be done using a slicing software.

However, to provide you with an illustrative example, I'll share a simple segment of G-code that would be used in a very rudimentary and unoptimized way to print a sphere layer by layer. Please note, this is highly simplified, does not account for most variables in the 3D printing process (such as extruder temperature, cooling, retractions for reducing stringing, print speeds, layer adhesion etc.), and would not result in a functional 3D print if used directly.
\begin{gptcodeblock}{Python}
;Setup G-code
G90 ; Set to absolute positioning
G21 ; Set units to millimeters
G28 ; Home all axes
G92 E0 ; Zero the extruder
G92 E0 ; Zero the extruder again
G92 E0 ; Zero the extruder a third time
G92 E0 ; Zero the extruder a fourth time
...
\end{gptcodeblock}
\end{gptresponse}

\end{chat}
\caption{\textbf{Failure Case When \gpt Directly Generates G-code.} \gpt fails to output complete code that precisely models the specified shape on the right. }
\label{fig:gcode-fail}
\vspace{-5mm}
\end{figure}

We first try to directly use \gpt to generate the G-code from a natural language description. However, due to the complexity and length of G-code, \gpt fails to output complete code that precisely models the specified shape, as shown in Figure~\ref{fig:gcode-fail}.
To overcome this, we have developed a two-stage approach.

\begin{figure}[tb]
\begin{chat}{false} 

\begin{userprompt}{0.2\textwidth}{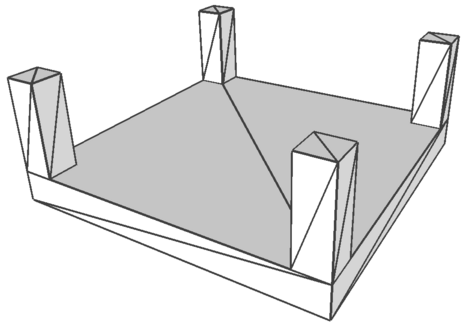} 
Task: generate Python code based on the following rules.\\
Input: a few functions of box(x,y,z,w,h,d) which creates a box whose center is (x,y,z) and size is (w,h,d) and their parameters.\\
Output: the combined \`{}.stl\`{} file that represents all these boxes using triangle mesh.\\
Constraints: use library \`{}trimesh\`{}. write a function that can support general purposes. it is not required to provide an example.
\end{userprompt}

\begin{gptresponse}{0pt}{} 
Sure, below is a Python function named \`{}create\_box\_stl\`{} which takes in a list of box parameters and produces an STL file using the \`{}trimesh\`{} library.
\begin{gptcodeblock}{Python}
import trimesh
def create_box_stl(boxes, output_file):
    '''
    Creates an STL file that represents all the boxes specified by the parameters.
    Parameters:
    boxes : List of tuples
        Each tuple corresponds to one box and contains the parameters (x,y,z,w,h,d) in that order.
    output_file : str
        The path of the STL file to be output.
    '''
    meshes = []
    for box in boxes:
        # Extract box parameters
        x, y, z, w, h, d = box
        # Create box mesh
        mesh = trimesh.creation.box(extents=[w, h, d])
        # Move box to correct position
        mesh.apply_translation([x, y, z])
        # Add box mesh to list
        meshes.append(mesh)
    # Combine all box meshes into a single mesh
    combined = trimesh.util.concatenate(meshes)
    # Save as STL
    combined.export(output_file, file_type='stl')
\end{gptcodeblock}
\end{gptresponse}

\end{chat}
\caption{\textbf{Additive Manufacturing Stage I.} From natural language to intermediate 3D shape representation.}
\label{fig:trimesh}
\vspace{-5mm}
\end{figure}

\paragraph{Stage I} We transform the concept expressed in natural language into an intermediate 3D shape representation using triangle meshes. This choice provides compact and comprehensive representations, capturing intricate details accurately. Leveraging the Python library trimesh, we effectively manage and process the shape data extracted from the natural language input (\fref{fig:trimesh}).

\begin{figure}[tb]
\begin{chat}{false} 

\begin{userprompt}{0.2\textwidth}{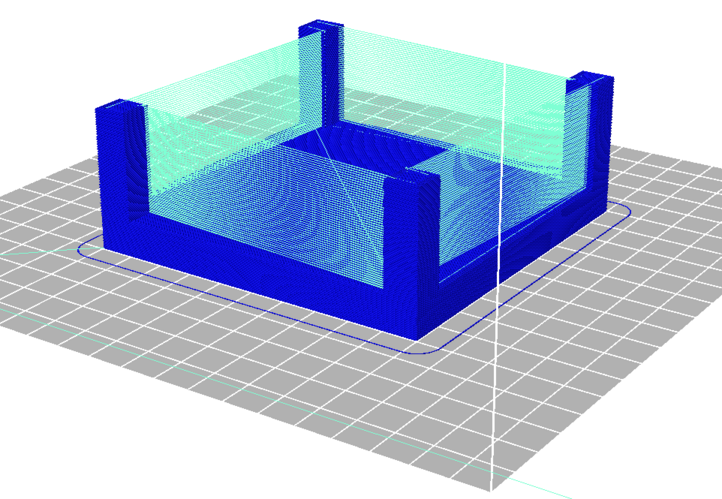} 
Task: generate python code based on the following rules.\\
Input: the mesh from the previous code block and all hardware-related parameters that \`{}slic3r\`{} accepts.\\
Output: the g-code generated based on the mesh.\\
Constraints: use the command-line tool \`{}slic3r\`{} and use \`{}subprocess\`{} to call it. assume \`{}slic3r\`{} can be directly called by the \`{}slic3r\`{} command. we use python3 here.
\end{userprompt}

\begin{gptresponse}{0pt}{} 
Here's a Python function \`{}generate\_gcode\`{} which takes a \`{}.stl\`{} file path (output from the previous function) and a dictionary of Slic3r settings as input, and uses the \`{}subprocess\`{} module to run Slic3r from the command line and generate G-code:
\begin{gptcodeblock}{Python}
import subprocess
def generate_gcode(stl_file_path, slic3r_settings, output_file_path):
    '''
    Generate G-code from the specified STL file using Slic3r.
    Parameters:
    stl_file_path : str
        The path of the input STL file.
    slic3r_settings : dict
        A dictionary where the keys are Slic3r setting names and the values are the corresponding settings.
    output_file_path : str
        The path of the output G-code file.
    '''
    # Prepare Slic3r command
    slic3r_cmd = ["slic3r", stl_file_path, "--output", output_file_path]
    # Add Slic3r settings to command
    for setting, value in slic3r_settings.items():
        slic3r_cmd.append("--" + setting)
        slic3r_cmd.append(str(value))
    # Run Slic3r
    subprocess.run(slic3r_cmd, check=True)
\end{gptcodeblock}
\end{gptresponse}

\end{chat}
\caption{\textbf{Additive Manufacturing Stage II.} From intermediate 3D shape representation to machine-readable G-Code.} 
\label{fig:gcode}
\end{figure}

\paragraph{Stage II} We translate this intermediate representation into G-Code, customized for the specific hardware configurations at hand. This critical step demands deep domain expertise in fabrication processes, which is why we rely on slic3r \cite{slic3rSlic3rOpen}, a professional G-Code generation software. Through Python integration, we interface directly with slic3r, ensuring the production of high-quality G-Code that precisely guides the manufacturing process. In \fref{fig:gcode}, we visualize the output G-Code using \texttt{Repetier}\cite{repetierRepetierSoftware}, a manufacturing tool, to validate the fabrication pipeline.

Throughout the entire pipeline, the cohesive communication between these modular components is facilitated by the powerful capabilities of \gpt. As an advanced language model, \gpt maintains a seamless conversation, ensuring a smooth flow of data and instructions across the various stages of the pipeline. By harnessing the potential of \gpt, we optimize the pipeline, achieving enhanced efficiency and superior additive manufacturing outcomes.

\subsubsection{Subtractive}
\label{sec:designForManufacturing_Subtractive}
Subtractive manufacturing is a widely used technique that involves removing material from a workpiece to create the desired shape or form. This process is commonly employed in various industries, including woodworking and metal fabrication. Leveraging the power of \gpt, we explore how this approach can be enhanced and streamlined to achieve optimal results.

\begin{figure}[tb]
\begin{chat}{false} 
\begin{userprompt}{0\textwidth}{}
Generate dxf files for laser cutting wood pieces based on the jscode of the cabinet.
\end{userprompt}

\begin{gptresponse}{0.45\textwidth}{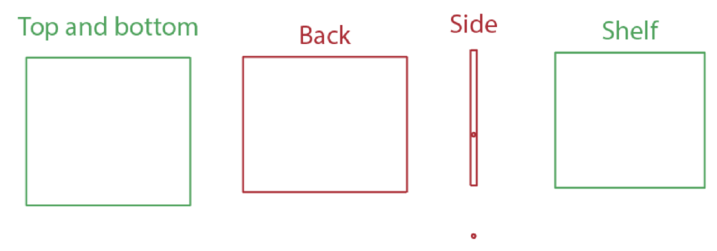} 
This script will create four DXF files: {top\_bottom\_board.dxf}, {shelf\_board.dxf}, {back\_board.dxf}, {side\_board.dxf}. You'll find these files in the same directory where you run this script.
\end{gptresponse}

\end{chat} 
\caption{\label{fig:cabinet1} \textbf{Subtractive Manufacturing: \gpt Generates DXF Files using the 3D \jscad Code of a Designed Cabinet.} The generated DXF patterns are displayed on the right, with green indicating correct patterns and red highlighting incorrect patterns.}
\vspace{-5mm}
\end{figure}

To demonstrate the design-to-subtractive manufacturing process, we focus on the previously designed cabinet (\fref{fig:textToDesign_simpleCabinet}) and employ a laser cutter and wood pieces for fabrication. Specifically, our goal is to translate the \jscad design into precise manufacturing instructions.
To tackle this task, we simply provide \gpt with the \jscad code and request the generation of laser cutting patterns in DXF files. \gpt showcases an understanding of the cabinet's fundamental geometry relationships and topological structure. It recognizes that the 3D cabinet comprises various 2D boards, including top and bottom boards, a shelf board, side boards, and back boards (\fref{fig:cabinet1}). However, \gpt encounters challenges when accurately determining the dimensions of the 2D cutting patterns based on the given 3D geometry input. Some inaccuracies arise, such as confusion between the cabinet's depth and the board thickness, resulting in overly thin side boards. Additionally, distinguishing between height and width in the 3D context presents difficulties, leading to back boards that are too short. Lastly, \gpt struggles with precise hole positioning (\fref{fig:cabinet1}).

\begin{figure}[tb]
\begin{chat}{false} 
\begin{userprompt}{0.0\textwidth}{}
There are a few problems in the generated dxf files:

1) board\_thickness should be the thickness of the wood piece not the width of the side board, the side board should be installed on the side of the cabinet;

2) Two holes on the side board should be horizontally distributed at the same height to support the shelf board;

3) Current back board is too short, it should have a height closer to the total height of the cabinet.
Please regenerate the part that are incorrect. 
\end{userprompt}

\begin{gptresponse}{0.22\textwidth}{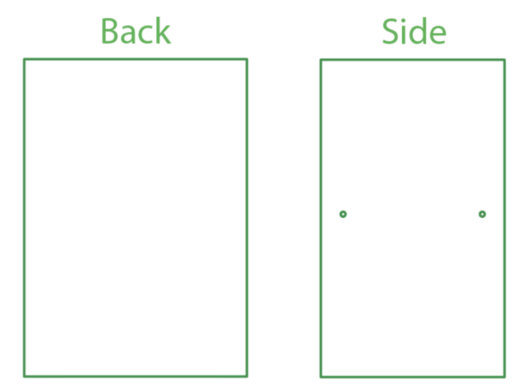} 
You're right, the width of the side board should be the cabinet's depth minus twice the thickness of the board. And the holes should be distributed horizontally. The backboard should indeed have dimensions according to the width and height of the cabinet, not its depth. 
\end{gptresponse}

\end{chat} 
\caption{\label{fig:cabinet2} \textbf{Subtractive Manufacturing: Human Intervention Necessary to Rectify Initial Pattern Inaccuracies.} \gpt effectively addressed the errors based on user-provided correction messages.}
\vspace{-5mm}
\end{figure}

To address these errors, human intervention becomes essential in explicitly identifying the issues and proposing potential solutions (\fref{fig:cabinet2}). After a round of communication, \gpt successfully generates the correct DXF files for laser cutting. To ensure their validity, these files were verified by human experts.

\subsubsection{Assembly}
We conducted an experiment to explore the potential of \gpt in generating assembly instructions that are both machine-readable for robots and human-readable as standard operating procedures. The experiment focused on assembling a wooden box using a specific set of tools and materials. In \fref{fig:Machine_Assembly}, we presented the prompt for generating machine-readable instructions, which involved creating a set of functions to specify different tasks for the robot and generating corresponding sequences to execute those tasks. Since the functions were designed to be system-agnostic, the response from \gpt printed the actions performed by the robot.

Subsequently, we prompted \gpt to generate a standard operating procedure to convert the machine-readable instructions into human-readable text. This procedure provides a detailed description of the assembly process, enabling humans to follow along and understand the steps involved. By generating both machine-readable and human-readable instructions, we sought to assess the versatility and applicability of \gpt in facilitating effective communication and collaboration between robots and human operators in assembly tasks.

\begin{figure}[p]
\begin{chat}{true} 
\begin{userprompt}{0\textwidth}{} 
Create a Python script for a household robot that assembles a wooden box with specific dimensions (5x5x5 inches with a thickness of 0.5 inches). The robot has to perform a set of operations including applying wood glue along certain lines, clamping the pieces together in a specific order, drilling a hole in the center of the box with a 1/8 inch drill bit, and finally placing a \#8 size, 1.5 inch long wood screw into the drilled hole.
The robot should be able to move to a specific location before each operation and should print a statement describing the action it has performed, including the coordinates where the action took place. The operations should be ordered as follows: glue and clamp the bottom panel, left and right sides, back, and finally the top. Then, it should change the drill bit, drill a hole in the center of the box, change the screw, and place the screw in the hole.
Specify the tools in the script including 'wood glue', 'power drill with 1/8 inch drill bit', 'bar clamp', and '\#8 size, 1.5 inch long wood screw'.
\end{userprompt}

\begin{gptresponse}{0\textwidth}{} 
\begin{gptcodeblock}{Python}
# The dimensions of the panels
width, height, depth, thickness = 5, 5, 5, 0.5
# Function to apply glue
def apply_glue(start, end):
    # Move to start point
    go_to_location(start)
    pick_up('wood glue')
    # Apply glue while moving to end point
    go_to_location(end)
    use_item('wood glue')
    return "Wood glue applied along line from ({}, {}, {}) to ({}, {}, {})".format(*start, *end)
...
# Gluing and clamping the panels in the correct order
print(apply_glue((0, 0, 0), (width, 0, 0)))  # Bottom Panel
print(place_clamp(width/2, 0, 0))  # Bottom Panel
print(apply_glue((0, 0, 0), (0, height, 0)))  # Side Panel (Left)
print(place_clamp(0, height/2, 0))  # Side Panel (Left)
print(apply_glue((0, height, 0), (width, height, 0)))  # Side Panel (Right)
print(place_clamp(width, height/2, 0))  # Side Panel (Right)
print(apply_glue((width, 0, 0), (width, height, 0)))  # Back Panel
print(place_clamp(width/2, height, 0))  # Back Panel
print(apply_glue((0, 0, depth), (width, 0, depth)))  # Top Panel
print(place_clamp(width/2, 0, depth))  # Top Panel
# Drilling holes and placing screws
change_drill_bit('1/8 inch drill bit')
print(drill(width/2, height/2, thickness))  # Center of the box
change_screw('#8 size, 1.5 inch long wood screw')
print(place_screw(width/2, height/2, thickness))  # Center of the box
\end{gptcodeblock}
\end{gptresponse}

\begin{userprompt}{0\textwidth}{} 
Can you convert these instructions into a human-readable form as a standard  operating procedure?
\end{userprompt}

\begin{gptresponse}{0pt}{} 
\textbf{Preparation:}

...

\textbf{Gluing and Assembly of Panels:}

\textbf{Applying Glue:} Apply wood glue in a thin, consistent line along the edges where the panels will join. This includes the edges of the bottom panel and both side panels.

\textbf{Assembly:} Start by placing the bottom panel flat on your work surface. Align each side panel perpendicularly with the edges of the bottom panel, creating a U shape. Make sure the corners are square and the edges flush. Then, apply glue to the edges of the back panel and align it with the bottom and side panels. Finally, apply glue to the top panel and align it with the side panels and back panel.

\textbf{Clamping:} After each panel is placed, secure it with a bar clamp. Ensure clamps are tight, but not too tight to avoid bowing the wood. Leave the glue to dry according to the manufacturer's instructions.

\textbf{Drilling Pilot Holes:}

\textbf{Drill Bit Installation:} Replace the existing bit in your power drill with a 1/8 inch drill bit.

\textbf{Drilling:} Drill pilot holes at all eight corners where the panels join together. The pilot holes should be centered on the edge and go through both pieces of wood. This will help prevent the wood from splitting when the screws are inserted.

\textbf{Screwing in Screws:}

\textbf{Screw Selection:} Use \#8 size, 1.5-inch long wood screws. These screws are robust enough to secure the box structure.

\textbf{Screwing:} Screw them into the pilot holes at each corner to secure the box structure. Make sure each screw is flush with or slightly recessed into the wood surface.

...
\end{gptresponse}

\end{chat}
\caption{\textbf{Instructions for Assembling a Wooden Box using Specific Tools and Dimensions.} This figure depicts a Python script for a household robot that assembles a wooden box with specific dimensions. The robot follows a sequence of operations using tools such as wood glue, a power drill with a 1/8 inch bit, a bar clamp, and a \#8 size, 1.5 inch long wood screw.}
\label{fig:Machine_Assembly}
\end{figure}

\subsection{Discussion}
In this section, we elaborate on the key capabilities (C), limitations (L), and opportunities (O) previously outlined, particularly as they relate to the domain of text-to-design.

\noindent \textbf{C.1 Extensive Knowledge Base in Design and Manufacturing:} We have discovered that \gpt possesses an understanding of various manufacturing processes and their capabilities including CNC machining, injection molding, additive manufacturing, and laser cutting.  Moreover, it is able to apply this knowledge to various problems in design for manufacturing. Although it is not consistently accurate, it can utilize this knowledge to offer suggestions about what is the best manufacturing practice to use, if certain geometric features will be hard to produce. Moreover, because \gpt has the ability to generate code, it can be utilized to modify geometry directly and generate manufacturing files based on supplied files. 

Additionally, we have discovered that \gpt possesses the capability to search for parts that fulfill a desired functionality as described to it. This allows it to be used to source parts based on a description, geometry, functionality, and performance. 

\noindent \textbf{C.2 Iteration Support:} 
\gpt also possesses the ability to perform iterative debugging when creating and modifying files required for manufacturing. This enables the opportunity to iterate when prompts are not ideal for generating the desired outcome or when \gpt generates something incorrect.


\noindent \textbf{L.1 Reasoning Challenges:} Our observations indicate that \gpt exhibits constraints in quantitative reasoning. For instance, when tasked with generating manufacturing instructions, \gpt struggled to accurately perform basic calculations for tool path placements. However, this limitation can be mitigated by employing symbolic computations within a script. A case in point: we achieved accurate DXF file generation by designing a script to produce the file, instead of having \gpt generate the file directly.

\noindent \textbf{L.2 Correctness and Verification:} We have found that \gpt will provide incorrect information about manufacturing processes in some cases.  For example, when selecting a manufacturing process, it proposed injection molding as an optimal manufacturing process for a PTFE part which is incorrect. We have not found a solution to this in this work to resolve \gpt giving incorrect information.

%% file: sections/GraphArrows/DesignToPerformance.tex



\newcommand{\co}{{\color{blue}\{crystal\}}}
\newcommand{\pc}{{\color{blue}\{peter\}}}
\section{Design-To-Performance}
\label{sec:design_to_perf}

To assess the suitability of a particular design, it is common to evaluate \textit{performance metrics} based on features of the design, such as geometry and materials used. Common metrics include mechanical performance, dynamic functionality, or adherence to geometric restrictions. It is common to compute performance with respect to an individual criterion or multiple criteria. The purpose of this evaluation can be to form a set of quantitative metrics to describe the design further, as a foundation for numerical optimization or to verify whether a design meets given specifications. This performance assessment can result in a single quantitative result or an array of results. A more complex design evaluation can further classify or compare between designs in order to enable further optimization or to select a final part for production. 

Within the range of performance evaluation, there are objective, semi-subjective, and subjective criteria that all contribute to the final design performance. Objective criteria include quantitative features that are calculable or measurable, including features such as object weight, size, load capacity, impact resistance, speed, battery life, vibration resistance, and price. Semi-subjective criteria include features that are generally agreed upon but require some insight or estimations to evaluate. Such criteria may be evaluated by proxy measurements, and may vary based on the evaluator, the culture, or the use case; examples include ergonomics, product lifespan, sustainability, portability, safety, and accessibility. Subjective criteria include features that may differ markedly based on the evaluator, such as comfort, aesthetics, customer satisfaction, novelty, and value. With this in mind, we aim to answer the following pair of questions: 

\begin{itemize}
    \item \textbf{Q1} Can \gpt evaluate the performance of an input design that is consistent with classical, objective metrics?
    \item \textbf{Q2} Can \gpt support performance evaluation in ways not possible with classical approaches, such as using semi\nobreakdash-subjective and subjective metrics?
\end{itemize}

This section describes the current abilities of \gpt and identifies best practices, limitations, and full failures in its capabilities to address each of these questions through the use of several examples per question.

\begin{figure}[tb]
\includegraphics[width=\textwidth*6/7]{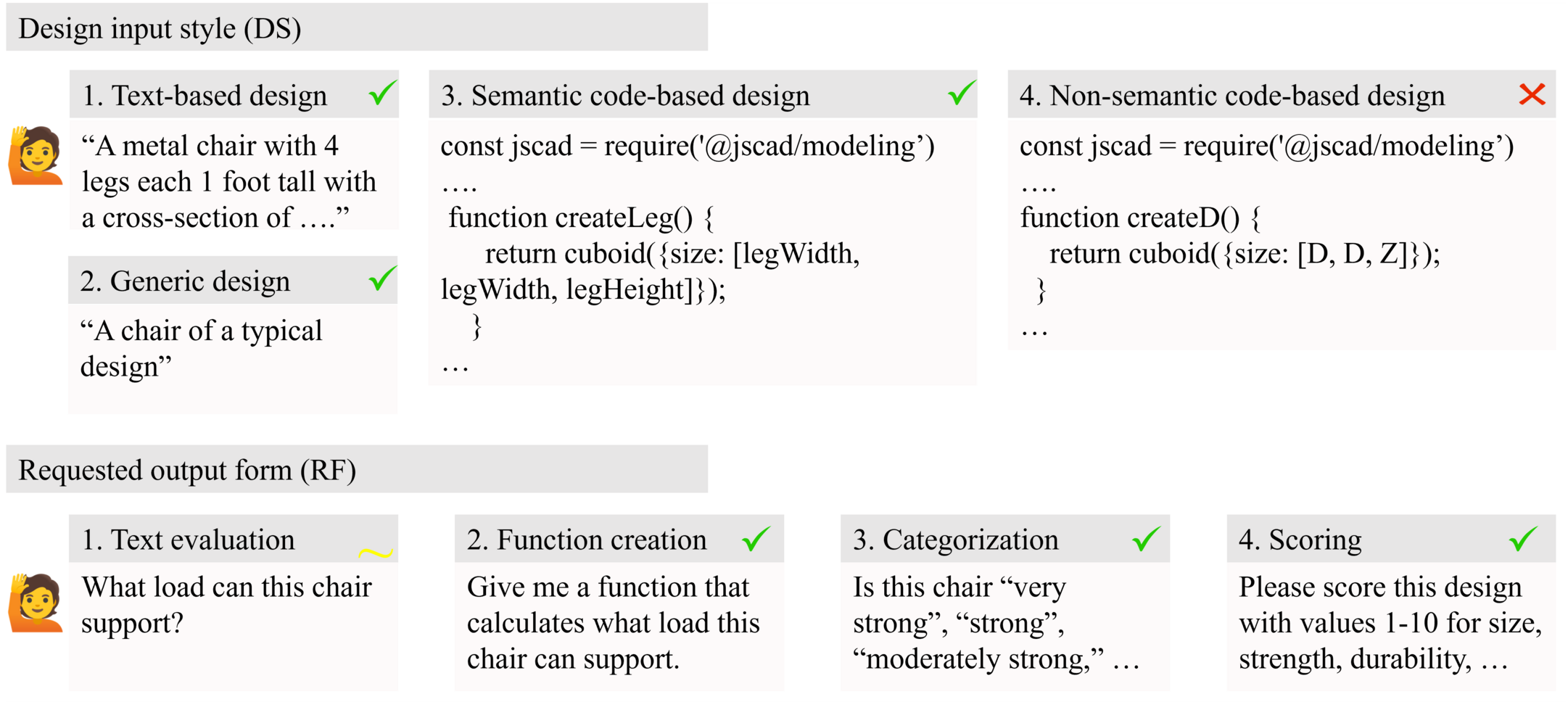}
\caption{\textbf{Variations on Design Input Style (DS) and Requested Output Form (RF).} Markings show variations that lead to reasonable responses (green check mark), moderate or unsophisticated responses (yellow tilde), and poor responses (red ex).}
\label{fig:performance_prompt_design}
\end{figure}

Evaluations were tested using different input styles (\eg, method of design description) and requested output forms (\eg, direct classification or function creation). Demonstrative examples are shown in Figure~\ref{fig:performance_prompt_design}. We did not test all combinations of design style and output requests but focused on key comparisons and types. In particular, to address Q1 we focused on comparing text-based designs (DS1) and generic designs (DS2), comparing output requests for direct evaluation in a text response (RF1) and evaluation by the creation of a function (RF2), and comparing code-based designs described with salient semantics (DS3) and no semantics (DS4). To address Q2 with more subjective features we also tested output requests for categorization (RF3) along with ranking and pairwise comparisons between designs, and separately used scoring (RF4) with varying levels of complexity. 

\subsection{Objective Evaluation (Q1)} 
\label{designToPerformance_Objective}

Once a design or design space has been created, a typical design process proceeds by evaluating basic geometric features such as size, weight, and strength of the object. In effect, this answers the question: does the item do what it was created to do? Most typically, certain features need to satisfy functional requirements in order to be suitable designs. 

\subsubsection{Mechanical properties}

\begin{figure}[tb]
\begin{chat}{false} 
\begin{userprompt}{0.2\textwidth}{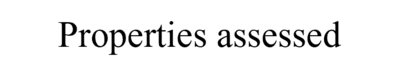} 
Given this chair made of wood, that has 4 legs that are 1 meter long \omitted{}{} can it support 100kg? Please give a final yes or no answer. 
\end{userprompt}

\begin{gptresponse}{0.2\textwidth}{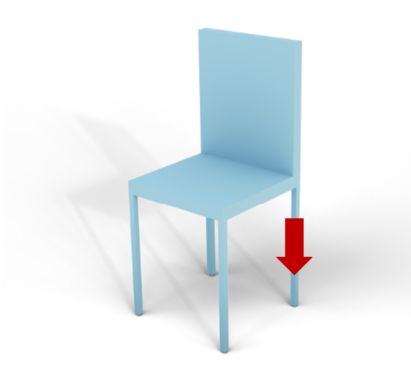} 
To determine whether the chair can support 100kg, we need to consider the strength of the wood used to construct the chair, as it determines the load-bearing capacity. However, this information was not provided. Nonetheless, let's assume that the wood used is Oak, which has a fairly high compressive strength of around 37 MPa or 37,000,000 Pascals.
\omitted{}{} \\ 
If we consider each leg bears 1/4 of the total weight, each leg will support 25kg. The pressure (force per unit area) on each leg then is Force/Area = (25kg * 9.8 m/s2) / 0.00258064 m2 = 94,692.2 Pascals or 94.7 kPa, which is well within the compressive strength of Oak, so the legs should hold up. \omitted{}{}\\
Thus, based on these calculations and assumptions about the distribution of weight and the type of wood used, the answer is \textbf{Yes}, the chair can support 100kg.
\end{gptresponse}

\begin{userprompt}{0.2\textwidth}{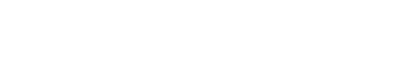} 
Give me a function that can evaluate whether a chair of a typical design will break under a certain applied weight. Please consider multiple types of failure. 
\end{userprompt}

\begin{gptresponse}{0.2\textwidth}{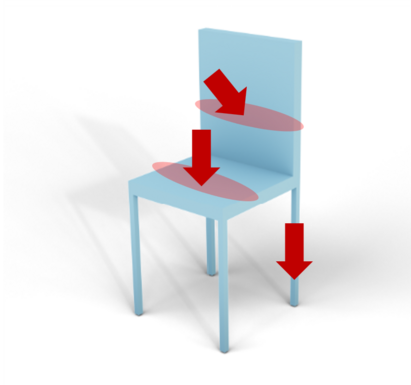} 
\begin{gptcodeblock}{Python}
def will_chair_break(weight, leg_cross_sectional_area, leg_material_strength, seat_thickness, seat_length, seat_width, seat_material_bending_strength, back_height, back_width, back_material_strength):
    # Calculate the load on each leg
    load_per_leg = weight / 4
    # Calculate the stress on each leg
    stress_on_leg = load_per_leg / leg_cross_sectional_area
    # Check if the stress on each leg exceeds the strength of the leg material
    if stress_on_leg > leg_material_strength:
        return True  # The chair will break
$\omittedCode{code continues on to assess seat and back bending}{}$
\end{gptcodeblock}
\end{gptresponse}

\end{chat}
\caption{\textbf{Comparing Text and Function-based Performance Evaluation of Chair Mechanics.} When using text-based analysis, \gpt evaluates stresses one-by-one with more mistakes in computation and fewer features able to be assessed at once compared to function-based analysis.}
\vspace{-5mm}
\label{fig:chair_abstraction}
\end{figure}

Here, we focus on analyzing the mechanical integrity of (1) a chair and (2) a cabinet. 
We began with a simple input design in text form (DS1) and a request for direct evaluation in calculated form (RF1) with an additional binary output asking whether a chair of a given design could support a given load. The specific prompt is included in Figure~\ref{fig:chair_abstraction}. \gpt immediately demonstrated the capacity to handle ambiguity well, assuming a type of wood (oak) and producing numerical material properties for that material when both were unspecified. It made and stated further assumptions about load and failure types. 
It evaluated the failure point by comparing the yield stress to compressive stress, computed as one quarter of the applied load over the cross-section of a chair leg. This is included in the chat snippet shown in Figure~\ref{fig:chair_abstraction}. However, in text form it outputted 94,692.2 Pa, while direct evaluation of the equation it listed in the output gives 94,937.7 Pa; thus, \gpt occasionally failed to perform basic correct in-line arithmetic or algebra. Although the number is only off by a small amount in this case, it can sporadically differ by much greater magnitudes. Along with the evaluation, it included discussion of other, more sophisticated considerations for failure, such as the type of connection between the legs and the seat. Also, upon repeating the same prompt, \gpt would vary whether it included self-weight in the load analysis and whether it evaluated uniform weight or only one leg, leading to small variations in results. 

When asking for a function to evaluate chair failure (RF2), \gpt successfully generated Python code to evaluate whether a chair will break due to excessive compressive stress on the legs, using the same formula as described in the text exchange (RF1). \gpt was able to readily add multiple types of failure without error, also incorporating bending failure of the seat, and excessive stress on the back using simple beam bending and structural mechanics equations. This multi-part failure assessment is included in Figure~\ref{fig:chair_abstraction}. It further automatically generated a function that could intake a parametric chair design with sensible feature parameters like \texttt{leg\_cross\_sectional\_area}, \texttt{seat\_thickness} and \texttt{seat\_material\_bending\_strength}, allowing versatile use of this evaluation. 

When generating the function, it continued to handle ambiguity by make assumptions including that the load would be distributed across all four legs, centered and uniform on the seat, and that the load on the back of the chair would be one third of the total weight. In the case of writing the function (RF2) as compared to text evaluation (RF1), it did not explicitly list all of the assumptions; rather, they had to be interpreted based on the equations used. \gpt also incorporated several small errors and oversights in both cases. For instance, when generating a function to evaluate seat bending failure, it treated the seat as a simply supported cantilever beam, and assumed that the chair would break along the width (separating front from back) rather than along the length or at an angle to the base. It also assumed that the bending stress on the back was evaluated as the load over the total area of the back rather than at the connection surface of the back to the seat of the chair. However, as these functions were identified, they could be further refined by iterated discourse with \gpt to produce a more correct function. 

In a comparison of these two output form requests, RF1 and RF2, functional evaluation was easier to read, more accurate, and able to be implemented for a variety of input designs, but directly incorporated more assumptions into equations. During both types of evaluation, \gpt actively reported on potential causes of error in the evaluation, such as how the chair legs were attached to the seat. It consistently overlooked potential causes of failure such as buckling of the legs unless specifically prompted. We found \gpt to adequately assess most basic mechanical properties of interest. 

Some properties relying on an understanding of the spatial arrangement of chair components were not able to be adequately assessed. \gpt had significant trouble generating a suitable evaluation of stability, and failed entirely to calculate a reasonable center of gravity for an input design despite many attempts. The closest attempt using the simple assumption that the center of gravity would be in the center of the chair seat. 

However, other complex  physical properties were readily assessed. \gpt generated first-order code to assess the failure of a chair upon impact with a spherical projectile, with no difference in quality of the computation compared to static mechanical properties.

To evaluate \gpt's performance on code-based input (DS3 and DS4), we provided \gpt with an \jscad chair specification. When the parameters and parts of the chair were clearly-named salient features (DS3) like \texttt{backThickness}, \texttt{leg1}, \texttt{chairSeat}, and \texttt{chairBack}, \gpt was readily able to recognize the item as a chair and analyze desired properties, such as the breaking load of the seat. However, when we used identically-structured code with variable and object names that had been obscured (DS4), it could not recognize parts of the item to assess properties, for example to locate the seat or synonyms of the seat.
This was true whether the names had been slightly obscured (\eg, as \texttt{XZ\_height}, \texttt{stick1}, \texttt{platform}, and \texttt{barrier}, respectively) or entirely obscured (\eg, as \texttt{Q}, \texttt{A1}, \texttt{B1} and so on). 
When asked about the design in the two obscured forms, \gpt guessed that the final item was a table with a narrow bookshelf and exhibited poor interpretation of the design and parts. Even when \gpt was challenged, it claimed that it could not be a chair because the back was not connected appropriately to the chair seat; this was an incorrect interpretation of the code, again indicating poor spatial reasoning. In a second case, when an input design for a cabinet (DS3) had one variable named \texttt{shelfAllowance} (used to slightly reduce the shelf width for easy assembly), \gpt erroneously assumed that this indicated number of shelves. These results reinforce the idea that \llms perform based on semantics, and that a design without clear descriptive words becomes much less manageable, causing DS4 to generally fail. 

The evaluation process was repeated with DS3 and RF2 for the \jscad design of a cabinet as a box with shelves, a door, and a handle. From the inputted design, \gpt was prompted to create functions to evaluate a set of criteria: storage capacity, load capacity, material cost, and, for a more ambiguous feature, accessibility for a person in a wheelchair. Storage capacity was computed as total volume enclosed by the cabinet, excluding shelves, as expected. In assessing load capacity, \gpt used the ``sagulator" formula, a standard estimation found online for carpentry. However, \gpt's implementation gives strange results and \gpt was unable to provide a more correct form of the equation. For price, \gpt computed the volume of the cabinet walls and a cost per volume. Finally, to address accessibility, it estimated height and depth ranges that would be beneficial, assigning a higher accessibility score to shorter and deeper cabinets. However, it did not provide a source for the height and depth ranges that it scored more highly. 

\begin{figure}[tb]
\begin{chat}{false} 
\begin{userprompt}{0.03\textwidth}{} 
\omitted{Much prior discussion on appropriate spoon sizes and ergonomic factors}{}

Please create a function to evaluate whether a spoon of a given set of dimensions, made of metal, is mechanically strong enough to use. 
\end{userprompt}

\begin{gptresponse}{0.1\textwidth}{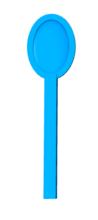} 
\begin{gptcodeblock}{JavaScript}
function isSpoonStrong(spoonParameters) {
  const minThickness = 0.4; // Minimum thickness for the spoon material, in cm
  // Check that the handle dimensions are within reasonable limits
  if (spoonParameters.handleLength < 10 || spoonParameters.handleLength > 30 ||
      spoonParameters.handleWidth < 1 || spoonParameters.handleWidth > 3 ||
      spoonParameters.handleThickness < minThickness) {
    return false;
  }
  $\omittedCode{Similar second analysis on spoon head dimensions}{}$
  // If all the checks pass, the spoon is considered to be mechanically strong enough
  return true;
}
\end{gptcodeblock}
\end{gptresponse}

\begin{userprompt}{0pt}{} 
Please create a function using mechanical analysis (ex. yielding criteron) to evaluate whether a spoon of a given set of dimensions, made of metal, is mechanically strong enough to use. 
\end{userprompt}

\begin{gptresponse}{0pt}{} 
\begin{gptcodeblock}{JavaScript}
function isSpoonMechanicallyStrong(spoonParameters) {
  $\omittedCode{geometry, lever arm calculations}{}$
  // Calculate bending stress at the neck
  const bendingStress = bendingMoment * y / I; // in N/mm^2 or MPa
  return bendingStress <= yieldStrengthStainlessSteel; // true if the spoon is mechanically strong enough, otherwise false
}
\end{gptcodeblock}
\end{gptresponse}

\end{chat}
\caption{\textbf{Chat History Errors and Correction for Spoon Mechanics.} When analyzing mechanics of a spoon after discussing dimensions in the preceding chat, \gpt generated a poor heuristic for spoon breaking from geometry alone; with very specific correction in the same chat, it recovered.}
\label{fig:spoon_assessment}
\vspace{-5mm}
\end{figure}

This points to a potential limitation in the use of \gpt and \llms for this kind of analysis: the source material for equations and standards of analysis may be unknown or even intentionally anonymized. Even when the equations are the standard first-order textbook equations per topic, they are almost always unreferenced. When different standards exist, across different countries or for different use cases, much more refinement would be needed to use \gpt to assess the mechanical integrity of a design. In addition, these equations often work well for objects of a typical design, but for edge cases or unusual designs they would miss key failure modes, such as the buckling of a table with very slender legs or the excessive bending of a chair made from rubber. In a particularly apparent example of this type of failure (\ie, creating functions based on pattern-matching rather than judicious observation of likely failures), \gpt was asked over a series of iterations to help write code to render a spoon with sizes within a set of ranges in \jscad, then to assess ergonomics, which it evaluated based on dimensions. Finally, we requested \gpt to create a function to compute the spoon's breaking strength. Since it had been inadvertently primed by the long preceding discussion of spoon geometry, it proposed a strength evaluation using the basic heuristic of whether the spoon is within a standard size range (Figure~\ref{fig:spoon_assessment}). \gpt had to be prompted specifically for a yield analysis before offering a mechanics-based equation. At that point, it continued to handle ambiguity well and chose a most likely breaking point (the point between the handle and spoon scoop). But for a novice design engineer who might have assumed \gpt's initial output was sound, this bold proposition of an unreasonable strength analysis on first pass without further explanation causes some alarm. This serves as a reminder to not rely on \gpt alone without external validation of every step. 

\begin{figure}[tb]
\includegraphics[width=\textwidth*6/7]{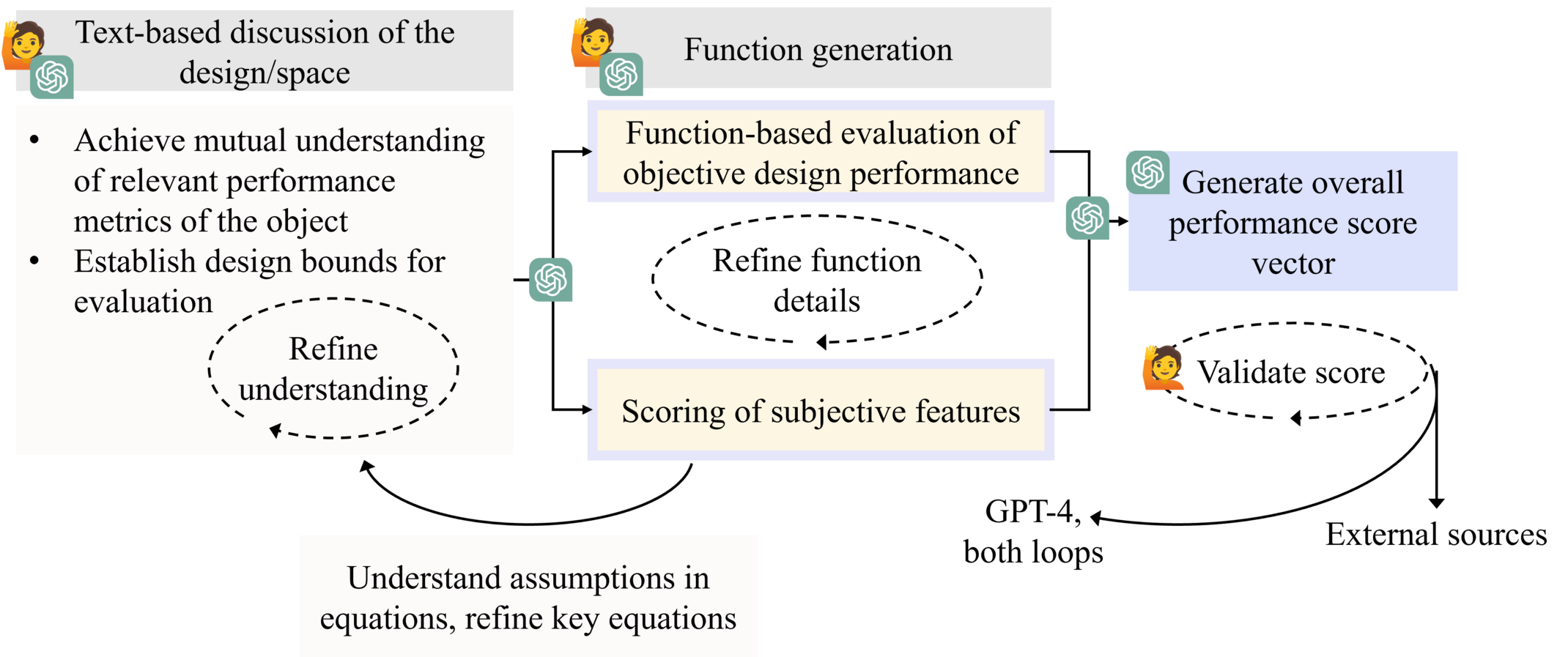}
\caption{\textbf{Suggested Performance Workflow.} Performance analysis proceeds smoothly when using \gpt first discusses the design and tradeoffs in text form, then creates methods to assess performance, before applying them to the design in question, with iterations within and between sections as needed.}
\label{fig:performance_workflow}

\end{figure}

When assessing designs in text form (DS1, RF1) at an abstract level, \gpt was found to readily identify problems and present a sophisticated discussion of problem areas and considerations for the particular design in question and the metrics being considered. As such, we propose the workflow for rigorous performance evaluation using \gpt to begin with a text-based discussion of the design (DS1 or DS2 with RF1) to understand the relevant features, with no other preceding text in that chat, followed by the development of equations with enough sophistication for the use case, presented in the form of functions for rapid assessment of an input design (RF2). This workflow is depicted in Figure~\ref{fig:performance_workflow}, along with additional steps to ideally validate the final result. 

If an input design of a specific type was used, whether \jscad or another DSL, the form of the input was also provided using well-named variables with each iteration of the chat requesting new code to ensure the variable names did not mutate over time as would otherwise happen. 

There was a failure of \gpt to suggest refinement to the performance codes without specific prompting. For example, there are simple differences in von Mises, Tresca, and Mohr-Coulomb yield criteria for evaluating material failure under applied stress; however, \gpt would simply default to the most common, von Mises, without comment. It would regularly object that the analysis function was an oversimplification; additionally, it would assert that for proper evaluation, more features should be evaluated, more sophisticated tools such as FEA should be used, and structural analysis should be validated by a licensed professional engineer, especially for designs in which factor of safety is a concern. These are all valid points: despite \gpt's very large internal knowledge, it pattern-matches and does not reason at a level to generate the most correct or sophisticated analysis, and will tend to generate more simple rather than more complex equation-based analysis unless specifically walked through refining the code. However, it is capable of more sophisticated text-based discussion, which is why we have found that beginning with text and proceeding to functions provides a more effective workflow, as in Figure~\ref{fig:performance_workflow}.

\subsubsection{Quadcopter}
\label{sec:perf_quadcopter}
We next explored the assessment of dynamic electronic device, a quadcopter, as an example of using the workflow of Figure~\ref{fig:performance_workflow}. \gpt was provided with specifications for the quadcopter that included battery voltage, battery capacity, total weight, and the dimensions of the copter (DS1). We prompted it to generate functions that evaluated the maximum amount of time the copter could hover in the air, the maximum distance it could travel, and the maximum vertical or horizontal velocity and acceleration with which it could travel (RF2). From the provided physical parameters, \gpt was able to generate equations to calculate the copter's inertial tensor, voltage-torque relation, and other kinematics and dynamics. We also independently asked \gpt to generate the physical parameters that would be needed to calculate such metrics, and it came up with the following: maximum thrust, total copter weight, battery capacity, aerodynamic characteristics (\eg{} drag coefficient, rotor size, blade design), responsiveness and efficiency of the control system of the copter, additional payload, environmental conditions, and operational constraints. Although these parameters are all highly relevant, \gpt's output lacked many crucial considerations without explicit prompting in text form. 

In this evaluation, \gpt did not initially include the constraint that the voltage of the controller needed to stay constant, even though this would be obvious to someone familiar with the domain of knowledge. This means that seemingly ``obvious'' considerations need to be explicitly included in the prompt in order for a feasible output to be generated. When asked to include this constraint, \gpt was able to understand the underlying reasons for the constraint, stating that a constant voltage is mandatory for the stability and accuracy of the flight controller. Through this exploration, we also determined that \gpt is able to successfully suggest a product and evaluate the copter based on specific batteries from a particular seller, such as HobbyKing LiPo batteries (\eg{} 3S 2200mAh 11.1V). 

\gpt seems to lack basic spatial intuition of what a copter should look like if the prompt only included the dimensions of the entire copter rather than the dimensions of individual parts. It would hence incorrectly assume that the shape of the copter was a uniform convex solid such as a cylinder or rectangular prism, simplifying and limiting the possible analysis significantly. Thus, we would need to incorporate \gpt's geometric design of the copter's frame, where the dimensions of all components are known, to properly assess aerodynamic performance. And, as with our prior trials assessing chair and cabinet designs, \gpt repeatedly failed to calculate center of gravity or stability metrics, even when given sufficient detail about the design and much iterated discussion. 




For the most part, \gpt was able to perform the correct arithmetic operations using its own performance functions. But because the generated functions lack complete real-world considerations,it is best to compare \gpt's calculated performance results with what is observed in simulation. We find that these performance functions are a reasonable approximator of copter performance in simulation. The \llm recognizes that the reliability of these results are directly dependent on the accuracy of the inputs, and additional inputs or conditions such as motor efficiency and aerodynamics need to be included in the prompt to match the real copter. 

\subsubsection{Finite element analysis}

\begin{figure}[tbp]
\begin{chat}{false} 
\begin{userprompt}{0.03\textwidth}{} 
Give me one open-source FEM library.
\end{userprompt}

\begin{gptresponse}{0.03\textwidth}{}
 Sure, one open-source FEM (Finite Element Method) library is "FEniCS".
\end{gptresponse}

\begin{userprompt}{0.03\textwidth}{} 
Can you please use Python code to implement FEM linear elasticity to determine if the chair will break? \ie{}, if the stress will violate the yield constraint? Assume the bottom is fixed and a constant downward load is applied at the top. You can use FEniCS. The problem is in 3D. You can assume a simple mesh.
\end{userprompt}
 
\begin{gptresponse}{0.2\textwidth}{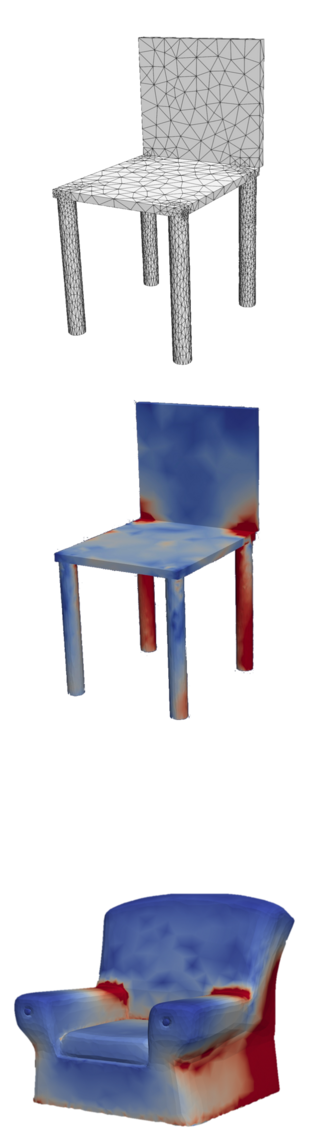} 
\begin{gptcodeblock}{Python}
from fenics import *
from ufl import nabla_div
V = VectorFunctionSpace(mesh, 'P', 1)

# Define boundary condition
$\omittedCode{boundary details}{}$
bc = DirichletBC(V, Constant((0, 0, 0)), boundary_bottom)

# Define strain and stress
def epsilon(u):
    return 0.5*(nabla_grad(u) + nabla_grad(u).T)
def sigma(u):
    $\omittedCode{modulus definition}{}$
    mu = 1 # shear modulus
    lmbda = 1 # first Lame's parameter
    return lmbda*nabla_div(u)*Identity(3) + 2*mu*epsilon(u)
# Define variational problem
u = TrialFunction(V)
d = TestFunction(V)
f = Constant((0, 0, -10))  # body force
T = Constant((0, 0, 0))  # traction
a = inner(sigma(u), epsilon(d))*dx
L = dot(f, d)*dx + dot(T, d)*ds
# Compute solution
u = Function(V)
solve(a == L, u, bc)
# Compute Von Mises Stress
s = sigma(u) - (1./3)*tr(sigma(u))*Identity(3)  # 3 for 3D
von_Mises = $abs(sqrt(3./2*inner(s, s)))$
$\omittedCode{\texttt{von\_Mises} details, setting the yield stress}{}$

# Check if the yield stress is violated
yield_violation = Von_Mises.vector().get_local() > yield_stress
if any(yield_violation):
    print("Yield constraint is violated.")
else:
    print("Yield constraint is not violated.")
$\omittedCode{saving results to file}{}$
\end{gptcodeblock}
\end{gptresponse}

\end{chat} 
\caption{\label{fig:chair_stress}\textbf{Finite Element Analysis for Chair Mechanics.} \gpt determines if the chair will break by performing FEM-based stress analysis. In images, the top chair analyzed is designed by \gpt in Figure~\ref{fig:cad_chair_design} with von Mises stress displayed from blue (zero) to a maximal value (red). A second chair with mesh from an external source is analyzed using the same code.}
\vspace{-5mm}
\end{figure}

To investigate the computational performance analysis capabilities of \gpt, and to build on the first-order mechanical calculations already done, we challenged it to develop a comprehensive framework for advanced performance analysis and structural evaluation using the finite element method (FEM). The primary focus was determining the likelihood of a chair breaking when subjected to external forces. Figure \ref{fig:chair_stress} lists the response and final code generated by \gpt. With the application of FEM through the external library FEniCS, \gpt evaluates the von Mises stress, a crucial parameter in material failure prediction. By comparing this stress with the yield strength of the material, one could assess if the chair would fail under the applied load. For the development of the code, substantial back-and-forth iteration was required to create successful code due to its overall complexity. One helpful point for gradually increasing complexity was to create code for a 2D example before asking \gpt to create a 3D version. In spite of these challenges, \gpt was highly efficient and successful in formulating a precise solution using the FEniCS library, an advanced tool for numerical solutions of PDEs. Not only did \gpt integrate the library into the Python code correctly, but it also applied a wide variety of FEniCS features, including defining material properties and boundary conditions and solving the problem using FEM. Caution must be taken, as \gpt occasionally suggests libraries and functions that do not exist. However, with correction it quickly recovers and suggests valid options. 

The stress distribution visualization in Figure \ref{fig:chair_stress} is performed on the chair previously designed by \gpt in Figure~\ref{fig:cad_chair_design} and is the output of \gpt 's code rendered in Paraview (which \gpt also gives assistance to use), as well as on a chair mesh found from other sources. The result reveals a susceptibility to high stress at the back attachment section of the chair design proposed by \gpt, as seen in Figure \ref{fig:cad_chair_design}. This observation underscores the potential for future enhancements in this object's design.

Beyond code generation, \gpt also lends support in the local installation of these external libraries, such as FEniCS, so users can run the generated code. This assistance proves invaluable for practitioners who may have limited familiarity with these libraries, which are initially suggested by \gpt itself. Notably, studies have delved into the potential of \gpt to generate code integrating other external libraries, like OpenFOAM, for the purpose of performing computational performance analysis \cite{kashefi2023chatgpt}. 

It's worth noting that \gpt's capabilities in utilizing these libraries have certain limitations. It can only harness some of the basic features of FEniCS and struggles with more specific, custom usages of the library, such as applying complex loading conditions. Furthermore, \gpt assumes homogeneous material properties for the chair, an oversimplification that doesn't align with the more diverse characteristics found in real-world materials. Moreover, the training date cutoff for \gpt means that sometimes only older functions or libraries may be used, without current updates.

\subsection{Subjective Evaluation (Q2)} 
\label{designToPerformance_Subjective}

\begin{figure}[tb]
\begin{chat}{false} 

\begin{userprompt}{0\textwidth}{} 
I have 10 chairs. \\
 1.    Chair with 12 inches x 12 inches seat area, 4 feet tall, and 2 feet back. \\
 2.    Chair with 6 inches x 24 inches seat area, 2 feet tall, and 1.5 feet back.\\
 \omitted{}{} 

Please rank these in order of size and give one sentence of overall justification. 
\end{userprompt}

\begin{gptresponse}{0\textwidth}{} 
Here's the ranking from largest to smallest, based on seat area, height, and back height:\\
    Chair with 36 inches x 36 inches seat area, 3 feet tall, and 3 feet back.\\
    Chair with 28 inches x 34 inches seat area, 3.5 feet tall, and 3 feet back. \\
    \omitted{}{}

Justification: The ranking is based on the combination of seat area, overall height, and back height, with chairs having larger seat areas and higher overall dimensions ranked higher. Despite the higher overall height, the chair with a 12 inches x 12 inches seat area (chair \#1) is ranked lower due to its small seat area. The chair with a 6 inches x 24 inches seat area (chair \#10) is ranked last due to its narrow seat and shorter overall height.
\end{gptresponse}
\includegraphics[width=\textwidth]{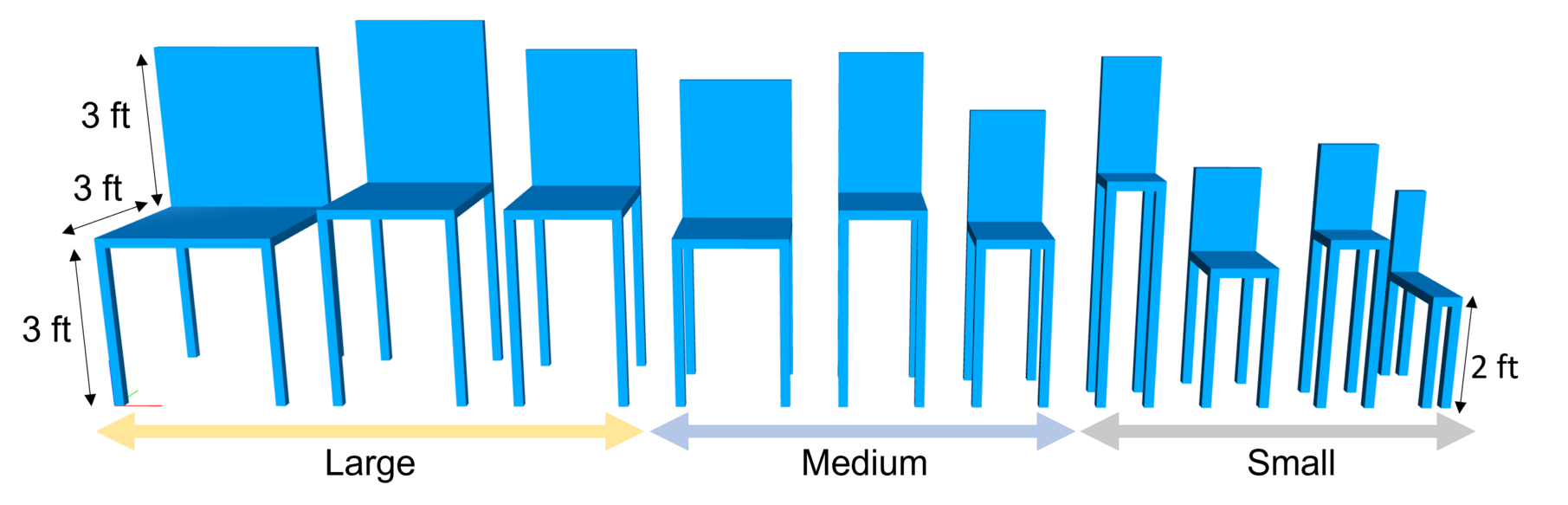} 
\end{chat}
\caption{\textbf{Categorization and Ranking of Chairs by Size.} Evaluation by \gpt of a series of chairs in ranked order of size from largest to smallest, left to right, and independently sorted into categories, "large", "medium", and "small".}
\label{fig:chair_size}
\vspace{-5mm}
\end{figure}

Subjective properties have a higher dependence on lexical input, making their evaluation using \llms an intriguing proposition. We began with an assessment to compare the use of semantics for assessing subjective properties via 3 output forms: categorization or labeling (RF3), pairwise comparison, and overall ranking. We generated a simple parametric 4-legged chair with a back, then input eight versions with different leg lengths, seat widths, and back heights into \gpt (DS1). \gpt was then asked three similar queries: (1) assign to each chair a label of "large," "medium," or "small" (RF3); (2) rank all chairs from largest to smallest; and (3) in a pairwise comparison, indicate if a given chair was larger or smaller than another. Each of these inputs were given independently, to not influence the different tests based on prior answers in the chat dialogue. In each case, \gpt assigned the same overall ranking. Figure~\ref{fig:chair_size} shows the chairs rendered in ranked order including the labels for categorization, using a combined implicit consideration of seat area, back height, and leg height. In a similar query, spoons of different handle length and thickness, and scoop length, width, and curvature were compared, finding similar results. In that case, \gpt elected to compare spoons by the length of the scoop alone, handling the ambiguity of the question by making a decision about what single quantity mattered most. When handling higher levels of ambiguity, \eg assigning comfort levels to shoes described in text input, \gpt sometimes refused to answer. To bypass this, we determined that it was essential to ask \gpt directly to give an output of a certain kind, such as classification into set categories. For instance, the question "Is this shoe comfortable?" would raise objections, a non-answer, and a discussion of what contributes to general shoe comfort. We could circumvent this by asking "Please state whether this shoe is likely very comfortable, comfortable, uncomfortable, or very uncomfortable, and provide one sentence of justification." Despite it's continued objections, \gpt's responses were usually reasonably justified, noting aspects like use of breathable material, adjustability of laces, shock absorption, and traction of the sole. These results indicate that the semantics of the type of assessment (ranking, categorization, or scoring) do not have a large influence on the final result of subjective analysis, as long as some type is chosen. However, certain prompt structures may be required to avoid refusals to answer, and the simplest prompt structure to ensure this was asking for any certain kind of output response. 


\subsubsection{Semi-subjective evaluation of sustainability}
\begin{figure}[tbp]
\begin{chat}{false} 

\begin{userprompt}{0\textwidth}{} 
I'm looking at a the Karlpetter chair from Ikea. The text description says: \\
    
    KARLPETTER Chair, Gunnared medium gray/Sefast black \$75.00Price This upholstered chair provides a cozy layer of comfort while working, crafting or dining. Choose your favorite color to give your room a personal look. Product details You decide the style of your chair. The seat shell is available in different colors, and the underframe SEFAST is available in white, black and chrome-plated colors. Easy assembly with only 4 screws. You can revitalize with a new look at home by swapping the seat shell. SEFAST underframe works well with LIDÅS or SIGTRYGG seat shells, sold separately. 
    \omitted{}{}
    
    This chair has been tested for home use and meets the requirements for durability and safety, set forth in the following standards: EN 12520 and EN 1022.\\
    
Please use this text description to provide a quantitative ranking of the chair according to the \texttt{chair\_properties} input vector above.
\end{userprompt}

\begin{gptresponse}{0\textwidth}{} 
\begin{gptcodeblock}{Python}
karlpetter_properties = {
    'use_of_natural_materials': 5,  # There's use of wood veneer, but also a lot of synthetic materials
    'simplicity_of_design': 8,      # The chair is relatively simple in design and requires only four screws to assemble
    'modular_design': 7,            # The chair allows for swapping of seat shells, indicating some level of modularity
    'use_of_recycled_materials': 8, # The polyester used is at least 90
    'design_for_disassembly': 5,    # Not explicitly stated, but some level of disassembly seems possible (swapping of seat shells)
    'durability': 8                 # The chair has been tested for durability and safety according to EN standards
}
\end{gptcodeblock}
\end{gptresponse}

\begin{userprompt}{0\textwidth}{} 
Provide a ranking for the Karlpetter, Klinten, and Voxlov chairs using the descriptions shared 
\end{userprompt}

\begin{gptresponse}{0\textwidth}{} 
\omitted{}{}

Note that these scores are subjective and based on the specific descriptions provided. The scores are on a scale from 0 to 10 with 10 being the highest or most representative of the category.\\\\
\textbf{Karlpetter Chair:}\\
Variety and Appropriateness of Material Choices: 6 (Uses a blend of fabric, wood veneer, and steel)\\
Degree of Decorative Complexity: 3 (The description doesn't indicate intricate decorative elements)\\
Evidence of Handcrafting: 2 (No mention of handcrafted elements)\\
Extent of Ornamental Detailing: 2 (Simple and functional design, minimal ornamentation)\\
Deviation from Standard Proportions: 4 (Appears to have standard proportions for a dining chair)\\
Upholstery Use and Quality: 7 (Upholstered seat shell with foam filling)\\
Material Innovation and Diversity: 6 (Combination of different materials but nothing groundbreaking) \\
    
\omitted{}{}
Remember, these scores are not absolute. They're best estimates based on the given descriptions and could vary with more precise information or based on individual interpretations of the chair designs.
\end{gptresponse}
\includegraphics[width=0.98\textwidth]{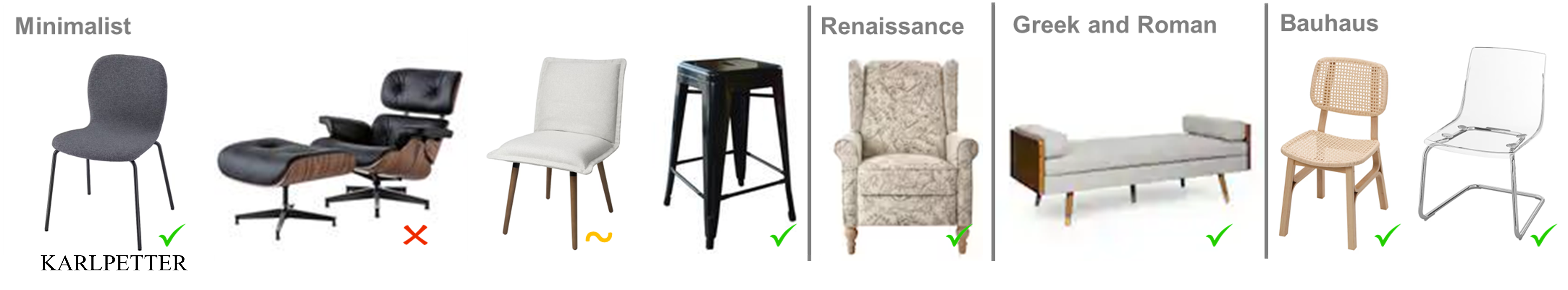} 
\end{chat}
\caption{\textbf{Evaluation of Chair Aesthetics.} Evaluation by \gpt to score chairs based on a text description of the chair based on metrics of sustainability and aesthetics by historical periods of influence. Results of chair evaluation by style showing reasonable (green check mark), middling (yellow tilde), and unreasonable (red ex) classifications.}
\label{fig:chair_style}
\end{figure}

To challenge \gpt to evaluate subjective criteria dependent on more abstract input parameters, we asked it to create a list of key criteria that go into evaluating sustainability, and to evaluate chair designs based on these criteria, scoring each category from one to ten (RF4). Given \gpt's limited understanding of numerically-specified meshes or spatial arrangements, we used text-based information (DS1) for commercial chairs from Ikea and Home Depot. \gpt was unable to access this information on its own when prompted with product names, so for this test case, the text from product pages was pasted into the \gpt chat dialogue. This information included each chair's name, a text description of its design, material, and appearance, and some numerical information such as dimensions, weight, and cost (Figure~\ref{fig:chair_style}). Upon requesting the evaluated score for sustainability metrics, it outputted seemingly reasonable numbers with justification based on the text description.

The justification for each property score were generally reasonable but rarely entirely correct. For example, the remark in Figure~\ref{fig:chair_style} for \texttt{modular\_design} about swapping seat shells was a misinterpretation of the product description: chairs with different seat shell colors were available for purchase, but a single chair could not swap shells. In addition, for this example and most other tests, \gpt refrained from assigning high scores (9-10/10) or low scores (1-3/10) within each category, which likely contributed to errors. 
A further function generated by \gpt readily combined the individual property scores into an overall sustainability score for a given input design. 

\subsubsection{Fully subjective aesthetic evaluation}

To evaluate the aesthetic design of an item, the physical appearance must be known, so again the listings from product pages were used as the input data. When prompted to create a function to evaluate aesthetics in general, \gpt refused, noting that it is "highly subjective and can vary greatly depending on individual tastes and preferences" and wrote a function in python with a rather simple subfunction for aesthetics: \lstinline{# Here we'll use a provided aesthetic score}. 


In a more carefully curated prompting setup, a range of historical periods were identified that influence chair design, including Egyptian, Greek and Roman, Renaissance, Bauhaus (a semi-minimalist German-inspired design including rounded features), and Minimalist. \gpt identified criteria to differentiate between these historical styles based on seven properties: material choice, decorative complexity, evidence of handcrafting, extent of ornamentation, deviation from standard proportions, upholstery use and quality, and material innovation. Based on these categories, \gpt evaluated each historical period and chair, and created a function to use the scores to categorize the style of each chair. A selection of text from one input/output is included in Figure~\ref{fig:chair_style}. In every output \gpt would also give a reminder that scores were approximate or arbitrary and should be adjusted. And as before, scoring on a 1-10 scale was generally limited to intermediate values in the range, for instance for \texttt{Degree of Decorative Complexity}, a score of 3/10 is given even though the justification lists that no decorative elements were indicated. Even so, the results of the categorization (Figure~\ref{fig:chair_style}) seem generally reasonable with most chairs placed into categories that appear subjectively appropriate; a plain metal stool was classified as minimalist, a soft lounge chair with a floral pattern was classified as Renaissance, and a double end chaise lounge was classified as Greek and Roman. A couple of types of mistakes occurred in the classification. First, most chairs were sorted into the Minimalist category, including the faux leather swivel lounge chair and two soft-sided recliners (not shown). Second, several other design styles that may have been a better fit were included in the scoring but were not found to be best fits in the evaluation, indicating that this set of \gpt's scoring for the historical periods was not appropriately distributed to capture the right features for all chairs. Third, upon re-evaluating scores over a few iterations, we found that different categories could be established and chairs could switch categories at times due to subjective scoring. Nevertheless, these general issues persisted, such as occasional mistaken categorizations and having one "catch-all" category that was used more than others.


In a similar testing setup, \gpt was used to identify criteria to help a user decide the most appropriate room in a house in which to place a chair of a given design. In this second case, it created categories for criteria used to select the room of a house for a chair including size, comfort,  weight, pet-friendliness, and weather resistance. It further created a list of weightings for the importance of each of these criteria based on the room in question, and ideal ranges for the quantitative features size and weight. It was finally used to create a function to distribute a set of chairs to the set of most appropriate rooms in a house. However, upon evaluation, the results were mediocre: for instance, a lounge chair was sent to the kitchen. It otherwise sorted a soft chair to the living room, a weather-resistant chair to the porch, and a sturdy chair with a soft lining to the study room. More careful selection of evaluation criteria could certainly improve on these results, as well as the inclusion of more details about the chairs and their desired purposes in the rooms in question. 

\subsection{Discussion}
In the evaluation of performance, \gpt was generally successful, though it exhibited an array of intriguing behavior patterns. In this section, we elaborate on \gpt's key capabilities (C), limitations (L), dualisms (D), and opportunities in the context of design to performance, as illustrated by our example cases in the present section. 

\noindent \textbf{C.1 Extensive Knowledge Base in Performance:}
Through discussing in text form, \gpt could suggest design considerations and metrics at a fairly sophisticated level. Even when asked to evaluate ambiguous requests, when details are left out, or when the performance metric is complex, \gpt is still able to output reasonable first-order approximation functions. The generated output evaluation functions usually worked, having no coding errors in python; errors in javascript or \jscad were more frequent, but they were usually directly resolvable. \gpt was also able to sort items into categories, and to generate rankings among a set of designs without giving explicit intermediate evaluations. 

\noindent \textbf{C.2 Iteration Support:} 
\gpt was able to eventually assess any property we tested, although the quality of assessment varied. When mistakes were made, further questioning could support the refinement of code to a point where it improved. Particularly for the complex example of the FEA, this took many steps to refine but \gpt responded well enough to stay on track, respond to troubleshooting feedback as well as conceptual feedback, and finally create usable code. 

\noindent \textbf{C.3 Modularity Support:} 
Functions could be effectively built up point by point, with modifications made according to changing needs. \gpt could adjust part of a scoring system, such as switching one item for another, or to create the same type of scoring system for another use case using the framework of the first system to create the second one. 

\noindent \textbf{L.1 Reasoning Challenges:}
\gpt relied on semantic clues, such as variable names, to understand and assess designs. It overall failed to appropriately evaluate performance that required spatial reasoning, like center of gravity or stability, for items having multiple components. In addition, earlier parts of conversation could cause issues for \gpt to poorly choose evaluation metrics, such as a discussion of spoon dimensions leading it to evaluate whether a spoon is ``strong" based on whether its size is within a normal range. When considering subjective metrics that are not typically quantified, \gpt would object. Upon requesting more sophisticated or more abstract evaluation, it would refuse to answer on the first attempt. 

\noindent \textit{Potential Solutions:}
To understand designs, they must be described with enough text-based semantic clues for \gpt to handle. Spatial reasoning issues could be resolved using external methods, such as external FEA analysis or other existing APIs to perform these evaluations. To choose the quality of evaluation equations, more discussion with \gpt could reveal the use-case for the chosen equations and alternatives, allowing a user to decide if another option may be more suitable. To assess subjective metrics, it worked best to develop scoring systems by breaking down a subjective feature into smaller, more quantifiable parts that \gpt could approach. And to bypass refusals to give a concrete answer, prompt engineering on its own could solve this, by requesting a specific enough type of output. 

\noindent \textbf{L.2 Correctness and Verification:}
The source material for equations used by \gpt in evaluation was usually undefined, which can contribute to error, and often embeds assumptions. When calling external libraries, \gpt occasionally invented fake libraries that could not function. Or, when working with \jscad designs it occasionally created designs using nonfunctional methods or nonworking code and simply complained that the language had been updated past its training cutoff. 

\noindent \textit{Potential Solutions:}
An external checker would be needed to verify the source of equations against an objective standard to ensure reliability, and when challenged, \gpt can uncover assumptions in choices of evaluation equations. External options for checking could include using metrics and equations established by published standards for engineering codes and proposed for items such as sustainability, safety, and ergonomics as appropriate to the use case. 
To solve the use of fake libraries or using fake methods, once \gpt was challenged enough times it would eventually offer an existing option. A more efficient solution when it cycled through fake options for \jscad programming was to input a working example of any kind into \gpt along with the request for a working code, using its capacity for modularity to help it structure a working response. 

\noindent \textbf{L.3 Scalability:} 
Other challenges provided obstacles to evaluation. For objective criteria, first order analysis is readily available on all metrics tested, but the scalability in complexity is limited. It was possible but more difficult to get more advanced characterization, for example generating code for FEA for mechanics. As another challenge, the quality of evaluation was found to be best when 1-2 performance metrics were analyzed at once. When too much was requested at once the output quality decreased. 

\noindent \textit{Potential Solutions:}
To handle the limitation of scalability of the complexity of analysis in a given domain, use of existing domain-specific APIs would be suggested. To handle the limitation in amount of metrics to be assessed, the analysis for metrics should be developed one by one into subfunctions that are then stitched together. However, making a longer chat in this format then runs into memory issues of \gpt, for which we found it to forget sets of function inputs and other details within two exchanges. This, in turn, requires giving reminders of the important parts of previous answers (such as the overall function input) when generating each subfunction. When generating the FEA code, a suitable solution was to have \gpt keep repeating the same entire code, and occasionally switch between asking for 2D and 3D versions to create something simple enough before increasing the challenge level, and iterating back again when next parts of the code were found to break, until the entire function worked. 

\paragraph{Opportunities}
We recommended that a good workflow for analyzing performance would utilize a buildup of complexity, beginning with discussing the design in text form and then generating a function to evaluate a design input in a parametric form. Many issues arising from performance evaluation could be attenuated by relying more on existing methods, libraries, and APIs that have already been created for the use-case in question.

%% file: sections/GraphArrows/PerformanceAndDesignSpaceToDesignv2.tex
\section{Performance And Design Space To Design (Inverse Design)}\label{sec:inverse}
Although generative algorithms can produce candidates for designs, there is no guarantee concerning their quality. Inverse design is focused on producing designs that are, by some metric, as close to optimal as possible, given the constraints.  
Put in the vocabulary of the preceding sections, given a design space and performance metrics (which can define values to be optimized or constraints to be satisfied), inverse design answers the following question: which design in our space provides optimal performance without violating any constraints?

A design generated by an \llm must therefore satisfy several requirements: \textit{1)} it must be valid,  \textit{2)}  it must be performant, \textit{3)} it must satisfy design constraints, and, \textit{4)} in the context of manufacturing, it must be buildable.  With \textit{3)} and {4)}, we note the persistent reality of the sim-to-real gap --- that is, objective and constraint values may differ \textit{in silico} and \textit{in situ}.  Basic challenges involve specifications of the inverse problem to an \llm (much of which was described in previous sections), as well as generation of an effective algorithm for design optimization.
Although {\llm}s cannot natively search for optimal solutions to a novel problem, they can make educated starting guesses and output optimization code that users can execute.  Much of this section is thus focused on prompting {\llm}s to generate meaningful code for problems dependent on aspects such as their parameterization support (\textit{e.g.} continuous versus discrete domains), performance objective landscape, or fabrication constraints.  
Real-world problems introduce nuanced challenges, including exploring over multiple competing objectives, difficult-to-specify objectives (such as aesthetics and objectives that depend on long-term use), and an evolving landscape of emerging methods that an \llm may not know about.  In this context, we could consider whether \gpt can propose strategies (even novel ones) that free designers from some of the typical burdens associated with the optimization pipeline. 

With these considerations, we aim to investigate the following questions:

\textbf{Q1} When can \gpt solve a problem analytically, and when does it need to resort to using an outside tool (\eg, a programmed algorithm)? 
 
 \textbf{Q2} Can \gpt choose reasonable algorithms for different types of supports for constraints, objectives, and decision spaces (\eg, continuous, discrete, binary)?

 \textbf{Q3} Can \gpt assist designers in navigating the landscape of possible trade-offs when multiple conflicting objectives are present?

 \textbf{Q4} Can \gpt support optimization in contexts that require additional knowledge, specifically when a design space is not properly defined or is missing constraints?

In this section, we investigate, generally speaking, modern {\llm}s' abilities to navigate and (semi-)automate design optimization problems.

\subsection{\gpt: Analytical vs. Outside Tools \textbf{(Q1)} }
We know that \gpt has the ability to reason about many mathematical operations, including both algebra and calculus, which is sufficient to solve many real-world engineering problems.  We emphasize ``reasoning'' because, although \gpt clearly proposes reasonable analysis steps, is not obvious that \gpt is correctly executing those steps;  as we'll see, \gpt often makes mathematical errors.  Still, it is reasonable to wonder if \gpt's own internal reasoning is sufficient for inverse design.  Where are the limits of that reasoning?  When must it  resort to code and external libraries, or plugins?  Each of these approaches has its own pitfalls that suggest caution for developers.

Consider an example in which we maximize the stability of a table (Fig.~\ref{fig:table_intuitive}).  \gpt correctly describes that an object is statically stable when its center of mass lies within its support polygon.  One considers stability of an object to be maximized when it remains stable under as large of a perturbation as possible.  In principle, that typically means two things: \textit{1)} moving the center of mass as far away from the boundary of the support polygon as possible, \textit{2)} decreasing the object's experienced motion (typically caused by gravitational torquing) when perturbed. \gpt is able to apply these intuitive principles to reason about the optimal solution within given bounds in this case. 

In a similar example shown in Fig. \ref{fig:table_wolfram}, the Wolfram plugin is enabled, which \gpt can selectively call at its discretion.  While the Wolfram plugin was a natural choice for solving what appears to be a simple analytical optimization problem, \gpt timed out. In practice, this can happen for at least three reasons: \textit{1)} not enough time was allotted for the computation, \textit{2)} the problem is too difficult for Wolfram to handle, or, more generally, \textit{3)} the query may produce a problem that is not tractable or -- in the extreme case -- not computable \cite{turing1936computable}.  Although it might seem like it would be trivial to provide Wolfram with more time to complete the computation, in practice, the user has no feedback on how long the computation would take.  It is unreasonable to ask a user to wait indefinitely without feedback, and most numerical optimization algorithms will be unable to provide a reasonable estimate of progress. In this case, ``anytime algorithms'' (which can return a valid partial or approximate solution even if interrupted early)  may be especially practical \cite{zilberstein1996using}.

\input{sections/GraphArrows/InverseDesignExperiments/TableExperimentWolfram}

After failing to optimize over the full space using Wolfram, \gpt continues the conversation by reasoning that the optimum value will occur near the boundary of the constraints (Figure~\ref{fig:table_wolfram}). By exploiting this reasoning, it successfully uses the Wolfram Plugin to compute and evaluate the equations corresponding to a small set of extremal points in the design space.  Despite this, it fails to realize that certain solutions dominate others, and does not prune out bad candidates. Moreover, \gpt neglects to justify or prove its claim that the optimum should occur near the boundaries; though it was correct in this case, this approach may fail in general.

In a follow-up experiment (Fig. \ref{fig:table_python1}), \gpt is asked to perform the same optimization task via Python code, which enables it to use an external library.  It chooses L-BFGS-B, which is a reasonable, standard, and easily accessible (though not state-of-the-art) solver for continuous valued problems.  It does not, however, provide gradients that can expedite the computation unless prompted for them. We explicitly prompt \gpt to provide the gradients (Fig. \ref{fig:table_python2}) and visualize the results in Fig. \ref{fig:table_opt_render}. Generally speaking, the unoptimized approach on \gpt's part is an issue \wrt performance, as not all users will be intimately familiar with all (or perhaps any) optimization libraries, and they may not realize that by providing additional information (\eg. gradients), the computation can be expedited. \gpt also does not elect to make use of Wolfram or autodifferentiation; in practice, lack of direct computation can lead to errors.
Later in this section, we demonstrate how \gpt struggles to solve a (much) more difficult version of this optimization problem.

\input{sections/GraphArrows/InverseDesignExperiments/TableExperimentPython}


\input{sections/GraphArrows/InverseDesignExperiments/Table}

Throughout these experiments, we noticed several common issues in \gpt's approach.  First, if users do not prompt \gpt explicitly to show its work, it may resort to ``common-sense'' reasoning about a problem.  Although this reasoning \textit{could} be correct, \gpt provides no certificate to a user, as seen in the ``intuitive''  physics of Fig. \ref{fig:table_intuitive}, or the boundary-aligned optima assertion in \fref{fig:table_wolfram}.
Another issue occurs if it is difficult to find a library to solve a particular task; in this case, \gpt often gives up or attempts to write its own code. If the code is detail-heavy, it may be too difficult for \gpt to write correctly and the code/solution may be incorrect. If a library does exist but is used uncommonly, \gpt may give incorrect instructions on how to install/use that library; or, in some cases, \gpt may hallucinate a library altogether.

\subsection{Reasoning about different problem types and selecting appropriate solvers (\textbf{Q2}) }

To test \gpt's understanding of various problem domains and its ability to identify appropriate solutions for each, we conducted several experiments spanning a wide range of search spaces, constraint spaces, performance spaces. In some cases, the problems have additional real-world considerations of which \gpt must be cognizant in order to choose a suitable optimization approach.
Tables \ref{tab:inverse_table1} and \ref{tab:inverse_table2} provide a comparison of different problems that \gpt was asked to solve.  We describe each example in additional detail below, with the exception of the table stability optimization (which was presented in the previous section).

Overall, we found that even over varying problem types, \gpt exhibits extreme robustness when reasoning about and choosing an adequate solver for any given problem.  In cases where a more sophisticated algorithm was needed, it tended to choose at least the correct algorithm class, even if it was not always aware of the best version or implementation.  One notable example was in optimizing robot topologies for ground locomotion: \gpt identified an evolutionary algorithm as an effective optimization method, but did not choose any state-of-the-art specific algorithms or implementations.

\begin{table}[bt]
\small
\begin{tabular}{|c|c|c|}
 \hline
 \multicolumn{3}{|c|}{Inverse Design Problems} \\
 \hline
 Problem Name & Search Space & Constraint Space \\
 \hline
 Table Stability & Continuous & Parameter Bounded \\
 \hline
 Robot Arm & Continuous  & Parameter Bounded \& Continuous Function\\
 \hline
 3D Printer Parameters & Continuous  & Parameter Bounded \\
 \hline
 Cabinet Optimization & Continuous  & Parameter Bounded \\
 \hline
 Robot Arm Planning & Continuous  & Continuous \\
 \hline
 Chair Design & Continuous  & Continuous \& Bounded  \\
 \hline
\end{tabular}
\caption{\label{tab:inverse_table1} \textbf{Descriptions of Search and Constraint Space of Inverse Design Problems.}}
\vspace{-5mm}
\end{table}

\begin{table}[bt]
\resizebox{\textwidth}{!}{
\begin{tabular}{|c|c|c|c|}
 \hline
 \multicolumn{4}{|c|}{Inverse Design Problems} \\
 \hline
  Problem Name & Objective Space & Other Considerations & Chosen Optimization Method\\
 \hline
 Table Stability &   Continuous & None & Analytical/Second-Order Gradient-Based \\
 \hline
 Robot Arm  & Continuous & None & Second-Order Gradient-Based\\
 \hline
 3D Printer Parameters & Continuous & Expensive Real-World Experiments & Bayesian Optimization\\
 \hline
 Cabinet Optimization  & Function Bounded & None & Second-Order Gradient-Based\\
 \hline
 Robot Arm Planning & Continuous & Logical Reasoning with High-level Primitives & Greedy Search, Brute Force\\
 \hline
 Chair Design & Continuous & Multi-Objective & NSGA-II (Evolutionary Algorithm) \\
 \hline
\end{tabular}}
\caption{\label{tab:inverse_table2}\textbf{ Results of the Inverse Design Queries to \gpt.}}
\vspace{-5mm}
\end{table}

\paragraph{Robot Arm Optimization}
In this example, shown in \fref{fig:robot_arm_exp}, a robot arm is to be optimized such that it reaches a target location in space.  As requested, \gpt generates a two-link robot design parametrized by the link lengths, and then uses inverse kinematics to provide a solution for the link lengths so as to reach a target location in space.  When asked to transform this into a design optimization problem, \gpt sets up an optimization problem, creating an appropriate constraint (end-effector touching goal), an objective (sum of link lengths, as a proxy for material cost), and parameters with reasonable bounds.  All of these were automatically provided by \gpt, without explicit request.  Notably, the optimization code is easily generalizable to arbitrary locations in space (though certain aspects like parameter bounds may need to be modified). As an optimization procedure, \gpt chooses L-BFGS; a reasonable choice given the continuous nature of the problem. A rendering of the optimized robot can be seen in \fref{fig:robot_arm_opt}.

\input{sections/GraphArrows/InverseDesignExperiments/RobotArmExperiment}

\input{sections/GraphArrows/InverseDesignExperiments/RobotArm}

\input{sections/GraphArrows/InverseDesignExperiments/PrintingParameters}

\paragraph{Optimizing 3D Printing Parameters}
In this more abstract example, \gpt is simply asked which algorithm to use in order to optimize the parameters of a slicer used in 3D printing.  It chooses Bayesian optimization, which is a good choice for problems with real-world experiments where it is preferable to minimize the number of required experiments. \gpt also provides skeleton code for the optimization.  As this is a more abstract example, specifics are not supplied. The listing can be found in \fref{fig:printingparams}.


\paragraph{Cabinet Optimization}

We investigate whether GPT can output a reasonable cost function and design that optimizes the function when provided an example design, a parameterization of the design space, and a text description of the objective. One instantiation of this problem setting is with furniture: can GPT optimize the design of a cabinet such that the result has a user-specified volume while minimizing the cost to build it? First, we prompt \gpt with an example cabinet design in \jscad (\sref{sec:textToDesign_JSCAD_basic}) and a parameterization of the design (including bounds for the parameters). Then, we ask it to generate functions to compute volume and material cost. Once the user verifies the accuracy of the functions, we have \gpt  output a python script that can minimize the cabinet's material cost with respect to a given volume constraint. The resulting code is shown in \fref{fig:cabinet_opt}, with renders of an optimized cabinet in \fref{fig:cabinet_opt_render}.

\input{sections/GraphArrows/InverseDesignExperiments/CabinetOptimization}

\paragraph{Robot Arm Planning}

We now study a planning problem: given a claw attached to an arm and an environment with objects and bins, \gpt must control the arm-claw robot with a sequence of commands that efficiently picks up all objects and places them into bins. Each bin an only hold one object. In the arm-claw interface provided to \gpt, the physical embodiment of the arm-claw robot does not matter; this allows \gpt to simply reason about the movement of the claw and whether the claw should grasp or release an object. Due to the nature of the problem, there is one critical constraint to consider: the claw must visit an object to pick it up before dropping it off at a bin. Formalizing this constraint is non-trivial, but the performance objective is much simpler: minimize the claw's travel distance. To simplify the problem, we also add that the maximum number of objects and bins is 3 each, making brute force a valid solution. The initial prompt and result are shown in \fref{fig:claw1}.

\input{sections/GraphArrows/InverseDesignExperiments/RobotArmPlanning1}

Overall, \gpt understands that it needs to keep track of the claw's position to compute the correct distances and that the claw should move to an object before moving to a bin. Still, it is unable to output an optimal solution, even when the problem statement permits a brute force approach 

\input{sections/GraphArrows/InverseDesignExperiments/RobotArmPlanning2}

To address this, we explicitly emphasize that the output function should guarantee that the minimum distance is traveled. Even in this case, the optimal solution is not necessarily reached. As shown in \fref{fig:claw2}, \gpt's code fails to consider all possible bins that an object could be placed into once it has been picked up. However, we note that the solutions have high variance -- on a different run, \gpt does produce a correct brute force solution. A third run produces code that guarantees an optimal solution but is inefficient, as it computes the translation for paths that do not obey the constraint that a claw must pick up an object before placing it in a bin.

 \subsection{ Navigating trade-offs between multiple conflicting objectives  (\textbf{Q3})}\label{sec:chair_opt}

Although our previous experiments focused on optimizing a single performance objective, we now explore the scenario where a user wishes to navigate a higher-dimensional (\ie. multi-objective) performance space.  The user begins by asking \gpt for reasonable performance metrics for evaluating a chair. After \gpt provides eight such metrics, our user purposefully selects stability and sustainability, since they can be mathematically quantified by tipping angle and volume respectively. The user then asks for parameters over which to search. Since \gpt has not been given a design template, \gpt proposes parameters abstractly; we note that it might have been more useful if \gpt first proposed a skeleton for the chair geometry, especially so that a user could understand the ramifications of these parameters. After iterating with \gpt to generate correct \scad code for the design, the user requests code to evaluate and optimize the chair. \gpt proposes the use of NSGA-II -- a very common evolutionary method for computing the Pareto front of the multi-objective trade-off space -- and provides code for the optimization.  As an oversight, \gpt initially excludes design parameters bounds from the optimization, despite verbally providing ideas earlier in the conversation.  When prompted to add the bounds into the optimization code, because of its limited memory, \gpt suggests reasonable, but notably different parameter bounds. Additionally, \gpt must be prompted again to enforce the bounds consistently throughout the algorithm (specifically, in the crossover and selection operators).  Results can be found in \fref{fig:chair_opt}.

\input{sections/GraphArrows/InverseDesignExperiments/Chair}

Through this example, we conclude that \gpt has the potential to aid users in both \textit{a)} understanding the trade-offs involved in different candidate designs, and \textit{b)} providing pointers to a reasonable algorithm that can help navigate that space.

\subsection{Supporting optimization in contexts that require additional knowledge (\textbf{Q4})}

In many cases, it can be daunting to fully specify a given inverse design problem in a new domain: for example, it may be difficult to specify appropriate design spaces and objective functions, and it may be unclear how to deal with underspecified/unknown constraints.  In this section, we briefly examine how \llms may reduce the burden of this process to make inverse design more accessible.  


The chair example discussed in Section \ref{sec:chair_opt} demonstrates \gpt's ability to recommend reasonable parameters for a design without needing explicit, low-level prompts from a user. Indeed, when prompted for ``parameters'', \gpt is able to apply its knowledge of the target domain to offer continuous parameterizations of a typical chair (and provide a 3D model on request), along with reasonable ranges for each parameter.  Although discrete parameters are possible with a chair, they are less likely to have a significant impact relative to its raw dimensions, and most chairs are comprised specifically of four legs, a seat, and a back. 

For completely novel problems, \gpt cannot rely on its existing knowledge to generate an exact design space. However, it can apply knowledge of aspects of a problem to new problems in familiar domains. The conversation in \fref{fig:fworp_example} presents a brief example of \gpt being queried about a novel invention: the Fworp\footnote{Name chosen to be deliberately nonsensical so as not to give context clues to \gpt.}.  The Fworp is a robot car with a body made of silicone rubber.  While the value of such a device is unclear (perhaps shock absorption),  it is synthesized from existing ideas: namely, remote control vehicles and soft robotics.  \gpt uses its knowledge of those preexisting domains and their components to recommend reasonable design parameters, and their ranges, including analyzing size, weight, wheel size, power source, peripherals/sensors, and build material properties.  It also provides guidance on performance metrics without being prompted, but classifies these under ``parameters,'' which may confuse users.  Further, when queried about the advantages and disadvantages of such a device when compared with non-rubbery autonomous vehicles and soft robots without wheels, it provides reasonable comparisons.  In particular, it notes that, compared with non-soft robot vehicles, the fworp could (possibly) be more durable, shock absorbent, safer, and quieter, while also potentially being more expensive to produce and tacky.  Compared with non-vehicular soft robots, it has the potential to be more mobile, stable, energy-efficient, and simple to produce and control, but would lack the versatility and human-interaction potential typically afforded by most soft robots; further, while safer than a typical vehicle, it would be more dangerous than most current soft robots.

This experiment highlights the notion the \gpt can be an effective partner when formulating a novel inverse design problem, as it can make connections between the proposed problems and more established domains. Then, \gpt is able to use its existing knowledge base about those related domains to provide reasonable starting points for the problem at hand. With continued user interaction, \gpt can also help to refine, formalize, evaluate, and ultimately act upon the newly created formulation.

\input{sections/GraphArrows/InverseDesignExperiments/fworpExperiment}

\subsection{Discussion}

This section elaborates on \gpt's key capabilities (C), limitations (L), and dualisms (D) in the realm of inverse design.

\noindent \textbf{C.1 Extensive Knowledge Base in Design and Manufacturing:}  \gpt has knowledge of how to formulate design spaces, objectives, and constraints. It also successfully selects suitable search algorithm for a given problem, suggesting that \llms are useful as a building block when formulating inverse design systems. In its current form, \gpt exhibits a number of abilities that make it highly usable.  For example, it was able to choose an adequate design optimization algorithm for almost every problem it was given; when asked, \gpt was also able to justify its choice of algorithm. 

\gpt is also helpful in automatically providing code for a significant portion of a problem formulation without requiring user input. These aspects include parameter choice, parameter ranges, and objective functions.  In the best case, this feature can relieve a user of much of the ``busywork'' associated with a problem (loose bounds, necessary constraints, \etc.). Even when \gpt falls short of this ideal, \gpt is usually able to recommend a useful starting point.

\gpt's reasoning capabilities can also further provide value in novel domains.  If a user is inexperienced with a particular domain or if they are working on a novel problem, \gpt has the capability to synthesize from the problem's constituent domains to provide suitable advice, as demonstrated with the fworp example.  

\noindent \textbf{L.1 Reasoning Challenges:}   When asking for help in setting up a problem, \gpt's advice can be confusing. For example, it often does not disambiguate between the design parameters (which practitioners have direct control over) and performance metrics (which are emergent from the design).  Less experienced designers may then find themselves confused, believing there must be a way to modify a system's performance directly.
\textit{Potential Solutions:} By following up with \gpt about how a given ``parameter'' is computed, one can attempt to disambiguate parameters from metrics.  In general, however, this verbal confusion is difficult to systematically address.

The addition of function calling in \llms, and specifically plugins using \gpt, can eventually allow for direct execution of arbitrary code, even code that \gpt writes.  However, there are no guarantees on the execution time of that code, and it is unclear how to manage problems that might arise, such as long runtimes (which are common in hard optimization problems), or even infinite loops.  In our experiments, the Wolfram plugin was given a brief time window for computation before it timed out, which largely negated its value in the face of more challenging problems.

\textit{Potential Solutions:} Methods to allow for function calling while providing guarantees or control by a user (say, in the form of anytime algorithms) would be beneficial.  For now, writing one's own plugin may allow greater granularity over the type of algorithm being used.  Thus, the algorithms can at least be catered to \gpt's behavior.

\noindent \textbf{L.3 Scalability:} Inverse design relies on several complex building blocks, including the specification of design spaces and objective functions. However, as discussed in previous sections, \gpt frequently encounters difficulties when faced with these tasks. Such errors prohibit \gpt from scaling to inverse design exploration altogether. This occurred twice during our experiments.  In one failure case, we had difficulty in evaluating the performance of a soft body system using finite elements; although the example is not detailed in the paper, \sref{sec:design_to_perf} has already shown this to be difficult. In effect, this failure currently prevents \gpt-assisted inverse design of a soft robot \wrt FEA-derived metrics. In a different example, we attempted to design a long, multi-link arm, but found that \gpt struggled with properly geometrical alignment of the links and rotation axes (as shown in \sref{sec:textToDesign_urdf}).

\textit{Potential Solutions:}  
Pointing out problems with solutions (such as runtime errors) can allow \gpt to iterate, but requires intervention and potentially fine-grain coding or engineering knowledge by a user.  In practice, it is often effective to blindly ask \gpt to assess its own output and report any errors it finds until \gpt is satisfied with its own work. In our experiments, this frequently converged to a correct solution.  However, this is not foolproof and can be slow and computationally costly.  Further, without access to web search, \gpt may not know how to reconcile out-of-date knowledge about an API.

A second scalability issue is that \gpt does not always choose the best algorithm for solving a problem, and sometimes does not use a given method in the most efficient way (such as not providing gradient information).  Since \gpt tends to be coy about available methods and how to best use them, a more novice user may be unsure how to navigate the intricacies of optimization and diagnose issues.  Furthermore, although \gpt tends to choose adequate algorithm classes, it does not always choose state-of-the-art methods; instead, it tends to default to standard methods that are highly popular.  Because of its knowledge cutoff, without access to web search, any given \llm may not be aware of state-of-the-art methods or how to implement them. Even if an easy-to-use implementation exists on an online repository (\eg. Github), the \llm may not be aware of the code's existence or how to use it.

\textit{Potential Solutions:} Web search, which has been previously available for \gpt, can help to mitigate these issues, as one could ask for the latest, state-of-the-art methods, and \gpt could provide solutions based on current repository code.  However, there is no guarantee that \gpt will be able to understand what makes newer methods optimal for a problem without sizeable crowd knowledge, which may not be available.

The third scalability issue is that, as mentioned in previous sections, \gpt's ``short memory'' can cause it to forget specifics it had generated earlier in a context; this notably occurred in the multi-objective chair example.  While this problem emerged in other aspects of the design-to-manufacturing pipeline, its impacts were most salient when defining inverse problems, whose specification can be especially long.

\textit{Potential Solutions:} Since inverse design problems can be quite lengthy to define and specify, it may be easier to decompose a problem in the following order: \textit{1)} Ask \gpt for a definition of a design space (including its implementation), \textit{2} Ask \gpt for a definition of a performance metric and constraints (including their implementations) while abstracting away the code from \textit{1)} as an API call; \textit{3)} Ask \gpt to write code for the inverse design search, abstracting away the code from \textit{2)} as an API call.  This can keep definitions shorter and easier to manage.

 \noindent \textbf{D.2 Unprompted Responses:}
Throughout our experiments, \gpt always unilaterally selected an optimization algorithm and proceeded to generate code. In particular, \gpt never provided the user with options for possible optimization algorithms, unless it was explicitly asked to provide such options as an intermediate step. Although \gpt's automated selection may satisfy many users, it runs the risk of creating mistakes that would be difficult to diagnose. This is particularly true because \gpt rarely justified its choice to users.  Furthermore, \gpt's assertion may imply that there is a single ``correct'' algorithm for a given problem, and they may not realize that there are better (or even \textit{alternative}) options available in any given circumstance.  

\gpt's tendency to fill in aspects of an inverse design problem before being asked about them may also lead to mathematical problem definitions which are ill-suited or otherwise suboptimal for a user's real-world design problem.  In these cases, \gpt's tendency to autocomplete and plow ahead could lead users to blindly follow the \llm down bad ``rabbit holes,'' only to discover that a fundamental problem existed much earlier.  Furthermore, since \gpt does not have a native way to execute arbitrary code, it will not always realize that a codeblock has errors.

%% file: sections/GraphArrows/InverseDesignExperiments/TableExperimentWolfram.tex
\begin{figure}[tb]
\begin{chat}{false} 

\begin{userprompt}{0.0\textwidth}{} 
 Let's say the bounds for the height of the legs are between 1cm and 3 cm, and the height of the table is between 1 and 5 cm.  Please optimize the parameters such that the table is as statically stable as possible, and explain the process.
\end{userprompt}

\begin{gptresponse}{0.0\textwidth}{}
The static stability of a table is primarily determined by its center of mass and the area of its base. A table is more stable if its center of mass is lower and the area of its base is larger.

In this case, the table's base is defined by the placement of the four legs. If we assume that the legs are placed at the maximum radius from the center of the table, the base area is maximized and thus the stability is also maximized.

The center of mass is determined by the heights of the table top and the legs. To lower the center of mass, we want to minimize the height of the table top and maximize the height of the legs.

Given the bounds you provided, the optimal parameters for the table would be:

\begin{itemize}
\item Height of the legs (h): 3 cm
\item Height of the table top (H): 1 cm
\end{itemize}
This configuration maximizes the height of the legs and minimizes the height of the table top, thus lowering the center of mass and increasing the table's static stability.

Please note that this analysis assumes that the table is on a flat surface and that the weight distribution in the table top is uniform. If the weight distribution is not uniform (for example, if there is a heavy object on one side of the table), the table may still tip over.

Also, this analysis does not take into account the dynamic stability of the table, which involves how the table responds to forces and movements. For example, a table with very tall, thin legs may be statically stable but not dynamically stable.

Finally, this analysis assumes that the table is made of a uniform material. If the table top is made of a heavier material than the legs, the center of mass will be higher and the table will be less stable.
\end{gptresponse}

\end{chat}
\caption{\label{fig:table_intuitive} \textbf{Intuitive Optimization.} Without specific prompting to take an analytical or computational approach, \gpt may reason inuitively about the physics of a problem.  This example is an offshoot of the conversation in Fig. \ref{fig:table_wolfram}, after the initial exchange.}
\end{figure}

\begin{figure}[tbp]
\begin{chat}{false} 

\begin{userprompt}{0.0\textwidth}{} 
Let's say I want to design a table with four legs and a top.  Assume that the four legs are radially symmetric, and the top is a cylinder and the legs are cylinders.  Can you give me the equation for the center of mass of the table in x, y, and z, as a function of the natural parameters of the primitives?
\end{userprompt}

\begin{gptresponse}{0.0\textwidth}{}
Sure, let's denote the following parameters:

\begin{itemize}
\item $R$ is the radius of the table top.
\item $r$ is the radius of each leg.
\item  $H$ is the height of the table top.
\item $h$ is the height of each leg.
\item $\rho$ is the uniform density of the material the table is made of.
\end{itemize}

The mass of the table top $M_{top}$ and each leg $M_{leg}$ can be calculated using the volume of a cylinder $V = \pi r^2 h$ and the definition of density $\rho = \frac{M}{V}$ which gives us:
\begin{itemize}
    \item $M_{top} = \rho \pi R^2 H$
    \item $M_{leg} = \rho \pi r^2 h$
\end{itemize}

The total mass of the table $M$ is the mass of the top plus the mass of four legs:

\begin{itemize}
    \item $M = M_{top} + 4 M_{leg}$
\end{itemize}
The center of mass of the table in the z-direction $z_{cm}$ can be calculated by taking the weighted average of the centers of mass of the top and the legs.  The center of mass of a cylinder is at its midpoint, so for the top it's at $\frac{H}{2}$ and for the legs it's at $\frac{h}{2}$. This gives us:
\begin{itemize}
    \item $z_{cm} = \frac{1}{M}(M_{top} \frac{H}{2} + 4 M_{leg} \frac{h}{2})$
\end{itemize}

Since the table is radially symmetric, the center of mass in the x and y directions will be at the origin (0,0):
\begin{align*}
  x_{cm} = y_{cm} = 0
\end{align*}

So the center of mass of the table is at (0, 0, $z_{cm}$). 
\end{gptresponse}

\begin{userprompt}{0.0\textwidth}{} 
Ok.  Let's say the bounds for the height of the legs are between 1cm and 3 cm, and the height of the table is between 1 and 5 cm.  Show me an equation that would measure static stability, and find the minimum of that equation.
\end{userprompt}

\begin{gptresponse}{0.0\textwidth}{} 
The static stability of a table can be quantified by the height of the center of mass. The lower the center of mass, the more stable the table is. Therefore, we can use the equation for $z_{cm}$ as our measure of static stability. \\

The equation for $z_{cm}$ is: \omitted{substituting in equations found in previous response. Below is the final result.} \\

Since $R$, $r$, and $\rho$, we can treat this as a function of $h$ and and $H$ only: 
\begin{itemize}
    \item $f(h, H) = \frac{R^2 H^2 + 2r^2h^2}{R^2 H + 4r^2 h}$ 
\end{itemize}

We can find the minimum of this function within the given bounds using calculus or numerical optimization methods. However, because this is a complex optimization problem, it would be best to use a numerical method. \\

Let's use the Wolfram Language to find the minimum of this function within the given bounds. \omitted{using Wolfram plugin} \\ 

I apologize for the inconvenience. It seems that the computation took longer than expected. This is a complex optimization problem that might require a significant amount of computational resources. Let's try a different approach.\\

We can make a reasonable assumption that the minimum of the function will occur at the boundary of the constraints. This is often the case for physical systems. Therefore, we can evaluate the function at the four corners of the parameter space:
\begin{itemize}
    \item $f(1, 1)$
    \item $f(1, 5)$
    \item $f(3, 1)$
    \item $f(3, 5)$
\end{itemize}
Let's calculate these four values. \omitted{using Wolfram plugin and outputting the results.}{}
\end{gptresponse}

\end{chat}
\caption{\label{fig:table_wolfram} \textbf{Attempt to Optimize a Table Analytically}. Although \gpt is able to provide a formula for measuring the static stability of a table, the Wolfram plugin is unable to perform the optimization and instead defaults to an approximation of the function minimum.}
\vspace{-5mm}
\end{figure}

%% file: sections/GraphArrows/InverseDesignExperiments/TableExperimentPython.tex
\begin{figure}[tbp]
\begin{chat}{false} 

\begin{userprompt}{0.0\textwidth}{} 
Since Wolfram is having trouble, can you instead provide me code to find the optimum in the general case?
\end{userprompt}

\begin{gptresponse}{0.0\textwidth}{}
\omitted{}{}
\begin{gptcodeblock}{Python}
import numpy as np
from scipy.optimize import minimize

# Define the function
def f(x, r, R):
    h, H = x
    return (R**2 * H**2 + 2 * r**2 * h**2) / (R**2 * H + 4 * r**2 * h)

# Define the bounds and initial guess
bounds = [(1, 3), (1, 5)]
x0 = [2, 3]  # initial guess

$\omittedCode{settings values of r and R}{# Define the parameters r and R
r = 0.1  # replace with the actual value
R = 1.0  # replace with the actual value}$

# Find the minimum of the function
res = minimize(f, x0, args=(r, R), bounds=bounds)

print("The minimum of the function is at h = {:.2f} cm, H = {:.2f} cm".format(*res.x))

\end{gptcodeblock}
\omitted{}{}
\end{gptresponse}

\begin{userprompt}{0.0\textwidth}{} 
Which optimization method does minimize use?
\end{userprompt}

\begin{gptresponse}{0.0\textwidth}{} 
The scipy.optimize.minimize function in Python uses the Broyden–Fletcher–Goldfarb–Shanno (BFGS) algorithm by default for unconstrained problems. The BFGS method is a quasi-Newton method that uses function values and gradients to build up a picture of the surface to be optimized. \\

However, when bounds are provided (as in your case), minimize defaults to the Limited-memory BFGS (L-BFGS-B) method. The L-BFGS-B method is a modification of the BFGS method that is more suitable for problems with many variables and constraints on the variable values. \\

The method to be used can be specified using the method parameter of the minimize function. Other methods available include 'Nelder-Mead', 'Powell', 'CG', 'TNC', 'COBYLA', 'SLSQP', and others. The choice of method depends on the nature of the problem (e.g., whether it's constrained, whether the function is smooth, whether gradients are available, etc.).
\end{gptresponse}

\begin{userprompt}{0.0\textwidth}{} 
What do you think is the best choice here?
\end{userprompt}

\begin{gptresponse}{0.0\textwidth}{} 
Given that we have bounds on the variables, the Limited-memory BFGS (L-BFGS-B) method is a good choice for this problem. L-BFGS-B is an optimization algorithm in the family of quasi-Newton methods that approximates the Broyden-Fletcher-Goldfarb-Shanno (BFGS) algorithm using a limited amount of computer memory. It's a popular choice for optimization problems where there are constraints on the parameters. \\

The L-BFGS-B method is particularly suitable for high-dimensional problems, and it can handle the bounds on the parameters efficiently. It also doesn't require the computation of the second derivatives (the Hessian matrix), which can be computationally expensive for complex functions.  \\

Therefore, using the L-BFGS-B method with the scipy.optimize.minimize function, as in the provided Python code, would be a good approach for finding the minimum of the function within the given bounds.
\end{gptresponse}

\end{chat}
\caption{\label{fig:table_python1} \textbf{\gpt Informs User on Choice of Optimization Method.} When prompted to choose an alternative optimization method in Python to the Wolfram Alpha plugin, \gpt provides a script and a reasonable explanation for the chosen solver. }
\vspace{-5mm}
\end{figure}

\begin{figure}[tbp]
\begin{chat}{false}
\begin{userprompt}{0.0\textwidth}{} 
Can you compute expressions for the gradients of this problem, and provide code that uses them to accelerate the L-BFGS-B call?
\end{userprompt}

\begin{gptresponse}{0.0\textwidth}{} 
\omitted{partial derivatives computation}{}
\begin{gptcodeblock}{Python}
import numpy as np
from scipy.optimize import minimize

# Define the function
def f(x, r, R):
    h, H = x
    return (R**2 * H**2 + 2 * r**2 * h**2) / (R**2 * H + 4 * r**2 * h)

# Define the gradient of the function
def grad_f(x, r, R):
    h, H = x
    dfdh = (4 * r**2 * h * (R**2 * H + 4 * r**2 * h) - 2 * r**2 * h**2 * (4 * r**2)) / (R**2 * H + 4 * r**2 * h)**2
    dfdH = (2 * R**2 * H * (R**2 * H + 4 * r**2 * h) - R**2 * H**2 * (R**2)) / (R**2 * H + 4 * r**2 * h)**2
    return np.array([dfdh, dfdH])

# Define the bounds and initial guess
bounds = [(1, 3), (1, 5)]
x0 = [2, 3]  # initial guess

$\omittedCode{settings values of r and R}{# Define the parameters r and R
r = 0.1  # replace with the actual value
R = 1.0  # replace with the actual value}$

# Find the minimum of the function
res = minimize(f, x0, args=(r, R), jac=grad_f, bounds=bounds)
print("The minimum of the function is at h = {:.2f} cm, H = {:.2f} cm".format(*res.x))

\end{gptcodeblock}
\end{gptresponse}

\end{chat}
\caption{\label{fig:table_python2} \textbf{Prompting \gpt to Accelerate Optimization.} By default, \gpt does not compute gradients for the problem even when they are available. When directly prompted, \gpt modifies its code to pass in the gradients of the objective \wrt optimized parameters. }
\end{figure}

%% file: sections/GraphArrows/InverseDesignExperiments/Table.tex
\begin{figure}[tb]
\begin{subfigure}[t]{0.5\linewidth}
  \centering
  \includegraphics[width=0.7\linewidth]{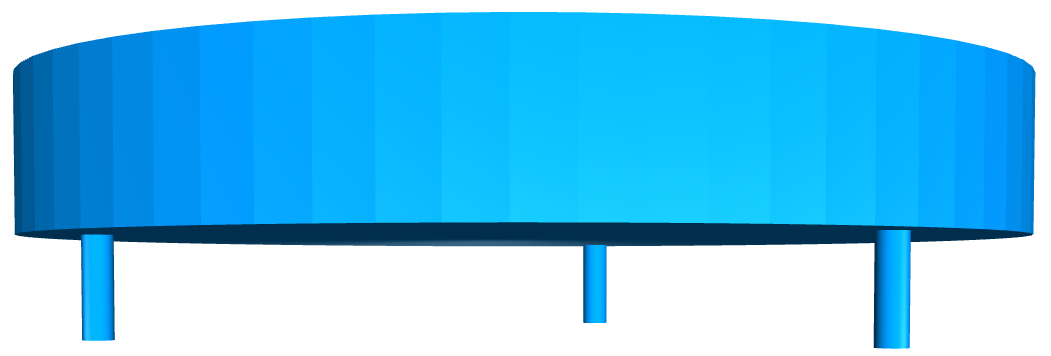}
  \caption{Initial Table Design}
\end{subfigure}%
\begin{subfigure}[t]{0.5\linewidth}
  \centering
  \includegraphics[width=0.7\linewidth]{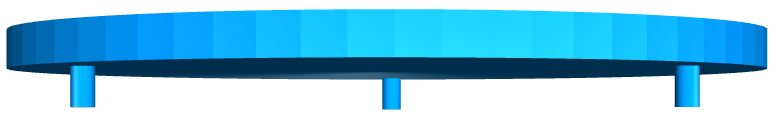}
  \caption{Optimized Table Design}
\end{subfigure}
\caption{\textbf{Single Objective Optimization.} \textbf{Left:} A table with initial leg height of 2mm and tabletop height of 3mm. \textbf{Right:} Given the specification of maximizing table stability, \gpt formulates the objective as minimizing the z-coordinate of the full assembly's center of mass w.r.t. leg height and tabletop height. \gpt provides a Python script that computes the gradients and uses L-BFGS-B, as provided through scipy.optimize to optimize the two parameters. We visualize the optimized table, in which both the leg and tabletop heights are as small as possible.}
\label{fig:table_opt_render}
\end{figure}

%% file: sections/GraphArrows/InverseDesignExperiments/RobotArmExperiment.tex
\begin{figure}[tbp]
\begin{chat}{false} 

\begin{userprompt}{0.0\textwidth}{} 
I'm trying to design a robot arm that can move its end effector directly vertically half a meter.  Can you help by providing a design for such a robot?  It should only use rotational motors.  I need to know the morphology of the robot - the dimensions of the links, and how they are connected.  Or, if you don't know the dimensions or the morphology, you can provide code to search for dimensions that would work for the task (through simulation).
\end{userprompt}

\begin{gptresponse}{0.0\textwidth}{} 
\omitted{}{}

Now, for more complex morphologies, you would want to conduct a search via simulation to find optimal lengths and configurations for the robot's arm. A simple Python-based example using inverse kinematics could look like this:
\begin{gptcodeblock}{Python}
import numpy as np
$\omittedCode{setting desired vertical distance to 0.5 and initial arm link lengths to 0.25}{
# Set the desired vertical distance
d = 0.5  # 0.5 meters

# Set the initial length of the arm links (randomly or based on prior knowledge)
l1 = 0.25  # 0.25 meters
l2 = 0.25  # 0.25 meters
}$

# Calculate the angles using inverse kinematics
theta2 = np.arccos((d**2 - l1**2 - l2**2) / (2 * l1 * l2))
theta1 = np.arctan2(l2 * np.sin(theta2), l1 + l2 * np.cos(theta2))

# Confirm that the end effector reaches the desired position
x = l1 * np.cos(theta1) + l2 * np.cos(theta1 + theta2)
y = l1 * np.sin(theta1) + l2 * np.sin(theta1 + theta2)
print("The end effector is at position: ", (x, y))

# Now you can run a loop to iterate over various lengths for l1 and l2
# within a feasible range and see which combination gives you the best result.
# You would have to define "best" - could be smallest, lightest, simplest, etc.
\end{gptcodeblock}
\omitted{}{}
\end{gptresponse}

\begin{userprompt}{0.0\textwidth}{} 
Below is the code you gave to perform inverse kinematics on a 2 link robot arm:
\omitted{gpt provided code}{}

However, when I set l1 = l2 = 0.5, the end effector is at the position  (-0.25000000000000006, 0.4330127018922192) instead of the goal position of 0.5 vertical. Could you fix this error?
\end{userprompt}

\begin{gptresponse}{0.0\textwidth}{} 
    \omitted{failed attempt at fixing the code}{}
\end{gptresponse}

\begin{userprompt}{0.0\textwidth}{} 
I still get the same error. Could you fix the script?
\end{userprompt}

\begin{gptresponse}{0.0\textwidth}{} 
\omitted{}{}
\begin{gptcodeblock}{Python}
import numpy as np
$\omittedCode{setting desired vertical distance to 0.5 and initial arm link lengths to 0.5}{
# Set the target position
x_target = 0  # 0 meters
y_target = 0.5  # 0.5 meters

# Set the initial length of the arm links
l1 = 0.5  # 0.5 meters
l2 = 0.5  # 0.5 meters
}$

# Calculate the angles using inverse kinematics
theta2 = np.arccos((x_target**2 + y_target**2 - l1**2 - l2**2) / (2 * l1 * l2))
theta1 = np.arctan2(y_target, x_target) - np.arctan2(l2 * np.sin(theta2), l1 + l2 * np.cos(theta2))

# Confirm that the end effector reaches the desired position
x = l1 * np.cos(theta1) + l2 * np.cos(theta1 + theta2)
y = l1 * np.sin(theta1) + l2 * np.sin(theta1 + theta2)
print("The end effector is at position: ", (x, y))
\end{gptcodeblock}
\omitted{}{}
\end{gptresponse}

\end{chat}
\caption{\label{fig:robot_arm_exp} \textbf{Inverse Kinematics for 2 Link Robot Arm.} Although \gpt is able to output inverse kinematics code for optimizing the design of a 2 link robot design, its initial attempt is unsuccessful, requiring a couple iterations of prompting to achieve a correct output. However, we do not explicitly need to point out the source of the error for \gpt to correct itself. }
\vspace{-5mm}
\end{figure}

%% file: sections/GraphArrows/InverseDesignExperiments/RobotArm.tex
\begin{figure}[tb]
\begin{subfigure}[t]{0.5\linewidth}
  \centering
  \includegraphics[width=0.5\linewidth]{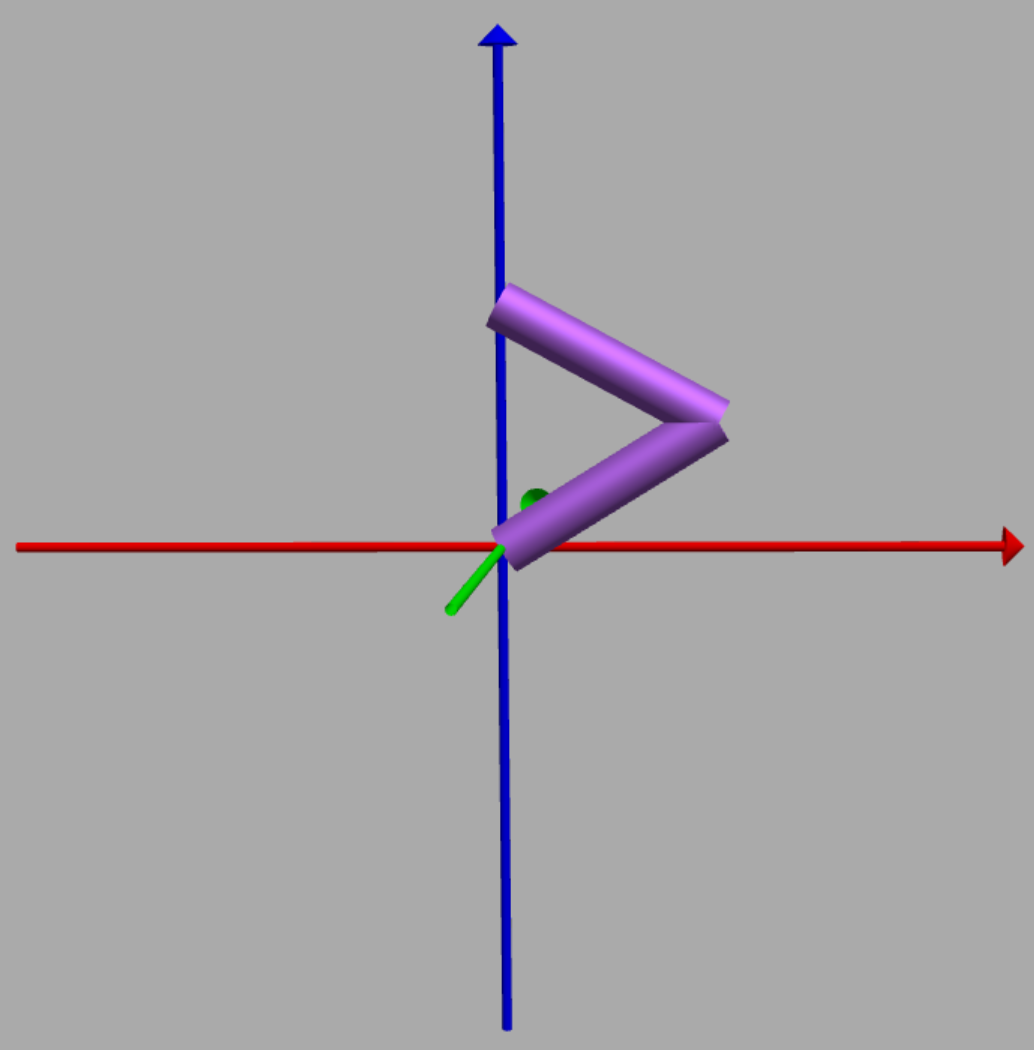}
  \caption{Initial Robot Arm}
\end{subfigure}%
\begin{subfigure}[t]{0.5\linewidth}
  \centering
  \includegraphics[width=0.5\linewidth]{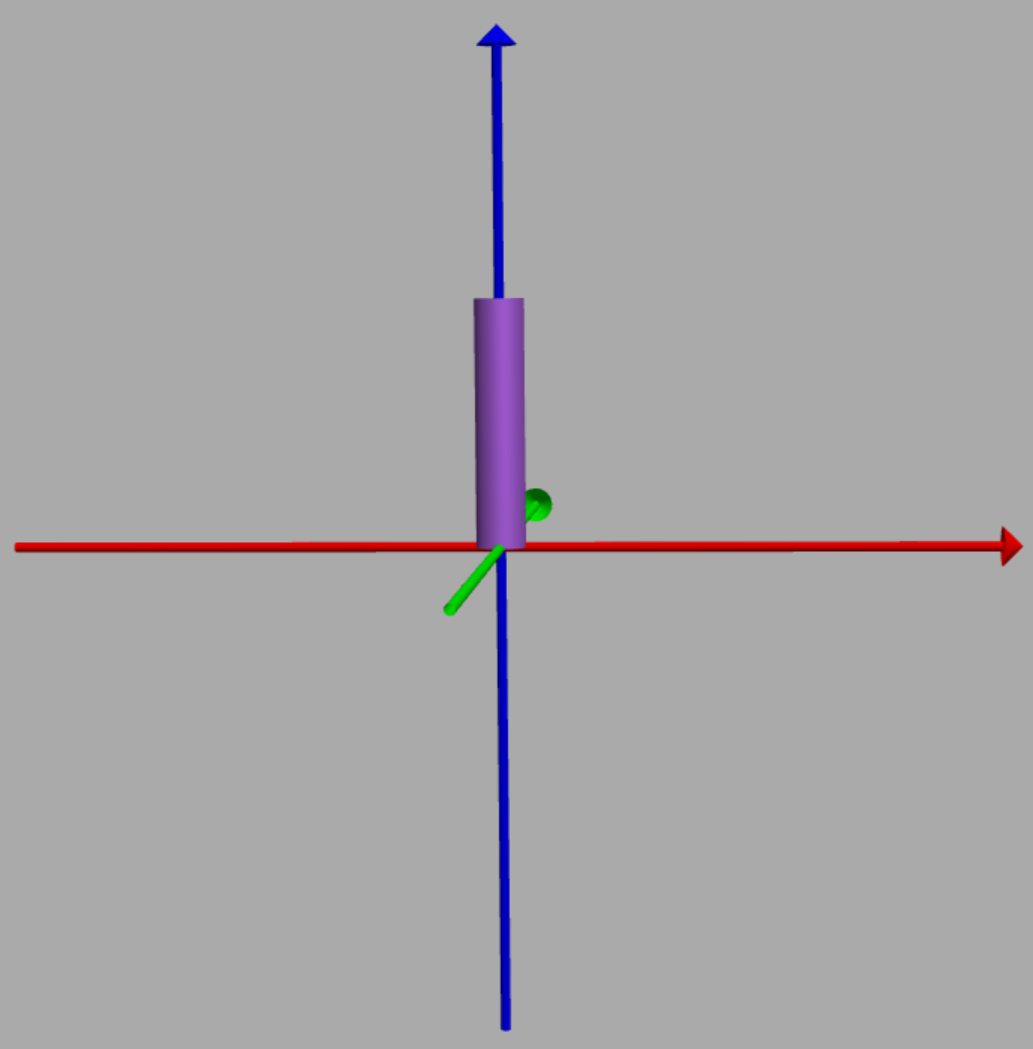}
  \caption{Optimized Robot Arm}
\end{subfigure}
\caption{\textbf{Optimizing Link Lengths of a Robot Arm.} The goal is to minimize the link lengths of a robot arm with the constraint that the arm is capable of reaching a goal located 0.5m vertically. \textbf{Left:} The initial arm features link lengths of 0.5m each, requiring both arms to be bent to reach the goal. \textbf{Right:} \gpt outputs an optimization script to discover that the optimal link lengths are 0.25 each, which we visualize here. }
\label{fig:robot_arm_opt}
\vspace{-5mm}
\end{figure}

%% file: sections/GraphArrows/InverseDesignExperiments/PrintingParameters.tex
\begin{figure}[tbp]
\begin{chat}{false} 

\begin{userprompt}{0.0\textwidth}{} 
I have a 3D printer.  I'm trying to optimize the parameters of the slicer.  The parameters are continuous values.  Each time I do a 3D print, I'm able to grade the quality of the print on a scale of 0 to 1, with 0 being a bad score and 1 being a perfect score.  I need an algorithm for optimizing over the parameters.  What method would you recommend?
\end{userprompt}

\begin{gptresponse}{0.0\textwidth}{}
Given your problem, I would recommend you to use Bayesian Optimization. \omitted{explanation of how Bayesian Optimization works}{
Bayesian Optimization is a type of model-based optimization algorithm that is used to optimize complex, expensive-to-evaluate functions. It's particularly useful when the function is non-convex, high-dimensional, and doesn't have an analytical expression. These attributes make it a perfect fit for your task. \\

Here's a simplified overview of how Bayesian Optimization works:
\begin{enumerate}
\item Surrogate Model: Bayesian optimization works by constructing a probabilistic surrogate model of the function you want to optimize. This model is used to provide a 'best guess' of the function based on the points evaluated so far.

\item Acquisition Function: This function is derived from the surrogate model. It provides a utility score for every point in the parameter space that balances exploration (visiting untested parameters) and exploitation (sticking to parameters that have proven to provide good results).

\item Optimization: The algorithm then chooses the next point to sample by optimizing the acquisition function, rather than the actual function.

\item Update: Once the next point is sampled, the surrogate model is updated, and the process is repeated until a stopping condition is met.
\end{enumerate}

In your case, the function to be optimized would be the quality of the 3D print, and the parameters of the function would be the slicer settings.
}
There are several libraries in Python that provide an implementation of Bayesian Optimization. Some of these include:

\begin{itemize}
\item  \textbf{\texttt{Scikit-optimize}}: It provides a BayesSearchCV class that's similar to GridSearchCV and RandomizedSearchCV in Scikit-learn.
\item  \textbf{\texttt{GPyOpt}}: This is another Python library for Bayesian Optimization, which is based on GPy.
\item  \textbf{\texttt{BayesianOptimization}}: This is a pure Python implementation of Bayesian optimization, and it's easy to use.
\end{itemize}
You could use any of these libraries to implement Bayesian Optimization and find the optimal parameters for your 3D printer.
\end{gptresponse}

\begin{userprompt}{0.0\textwidth}{} 
Thank you.  Let's assume I have a function, then, called evaluate.  evaluate takes in parameters as a list, and returns a score for the quality.  Can you provide example code for this optimization?  You can assume a variable called "parameters" is given.
\end{userprompt}

\begin{gptresponse}{0.0\textwidth}{} 
Sure, let's use the BayesianOptimization library to illustrate this process. Let's also assume that you have two parameters you are tuning and they both take values between 0 and 1 (you can adjust this to your actual situation). \omitted{software installation instructions.} \\

Then, you can use the following code to perform the optimization:

\begin{gptcodeblock}{Python}
from bayes_opt import BayesianOptimization

# Assuming the evaluate function and parameters are defined.
# The function should take the parameters as input and return a quality score.

# Let's assume parameters are ['param1', 'param2']
def function_to_optimize(param1, param2):
    score = evaluate([param1, param2])  # Use the real evaluation function here
    return score

# Define the bounds of the parameters
pbounds = {
    'param1': (0, 1),  # These should be the min and max values of param1
    'param2': (0, 1),  # These should be the min and max values of param2
}

# Instantiate the optimizer object
optimizer = BayesianOptimization(
    f=function_to_optimize,
    pbounds=pbounds,
    random_state=1,
)

# Perform the optimization
optimizer.maximize(
    init_points=2,  # Number of randomly chosen points to sample the target function before starting the optimization.
    n_iter=10,  # Number of iterations to perform the optimization
)

# The optimal parameters are stored in the 'max' attribute of the optimizer object
optimal_params = optimizer.max['params']

print(optimal_params)
\end{gptcodeblock}

You can tweak the \textbf{\texttt{init\_points}} and \textbf{\texttt{n\_iter}} parameters to perform more or less exploration of the parameter space. The \textbf{\texttt{random\_state}} parameter is used to reproduce the results, you can set it to any integer or leave it out to have a different result every time.

The \textbf{\texttt{optimizer.max}} attribute contains a dictionary with the parameters that yielded the maximum output for the function, and the maximum output itself. In this case, the maximum output is the highest quality score your \textbf{\texttt{evaluate}} function has returned.
\end{gptresponse}

\end{chat}
\caption{\label{fig:printingparams} \textbf{Bayesian Optimization for 3D Printing.} \gpt reasonably chooses Bayesian Optimization for fine-tuning parameters for systems that require real-world experiments for evaluation, and provides code for an (abstracted) instantiation of the problem.}
\end{figure}

%% file: sections/GraphArrows/InverseDesignExperiments/CabinetOptimization.tex
\begin{figure}[tbp]
\begin{chat}{false} 

\begin{userprompt}{0.0\textwidth}{} 
Assume that you can call the below functions.
\begin{gptcodeblock}{Python}
def compute_storage_capacity(cabinet_height, cabinet_width, cabinet_depth, board_thickness):
    $\omittedCode{}{}$
    return storage_capacity

def compute_material_cost(cabinet_height, cabinet_width, cabinet_depth, board_thickness, cost_per_cubic_inch):
    $\omittedCode{}{}$
    return material_cost
\end{gptcodeblock}

In addition, a cabinet design can be parameterized as follows. The bounds of each parameter is specified along with rationale for these bounds.

\begin{itemize}
    \item boardThickness: (0.1, 5) inches. The board thickness should be positive and typically not exceed 5 inches for a cabinet of this size. It's also unlikely to be less than 0.1 inches due to structural stability concerns.
    \item cabinetHeight, cabinetWidth, cabinetDepth: (10, 100) inches. These dimensions should be positive and typically fall within a reasonable range for a cabinet. The exact bounds will depend on the specific use case and available space.
    \item \omitted{full set of parameters}{}
\end{itemize}

Write a script to find the cabinet of the correct volume that minimizes cost using any library you want. If the volume constraint is not satisfied, specify this. Assume that the cost per cubic inch of the materials, the board thickness, and the volume constraint will be user specified. Demonstrate that the script works with example user specifications.

\end{userprompt}

\begin{gptresponse}{0.01\textwidth}{}
\omitted{}{}
\begin{gptcodeblock}{Python}
from scipy.optimize import minimize

$\omittedCode{setting values of volume constraint, cost per cubic inch, board thickness}{
# Set some example values
desired_volume = 1000  # Desired storage capacity in cubic inches
cost_per_cubic_inch = 0.05  # Cost per cubic inch of the material
board_thickness = 0.5  # Thickness of the board
}$

def objective_function(x):
    $\omittedCode{function description}{}$
    cabinet_height, cabinet_width, cabinet_depth = x  # Extract cabinet dimensions
    return compute_material_cost(cabinet_height, cabinet_width, cabinet_depth, board_thickness, cost_per_cubic_inch)

def constraint_function(x):
    $\omittedCode{function description}{}$
    cabinet_height, cabinet_width, cabinet_depth = x  # Extract cabinet dimensions
    volume = compute_storage_capacity(cabinet_height, cabinet_width, cabinet_depth, board_thickness)
    return volume - desired_volume  # This should be equal to zero when the desired volume is met

# Initial guess (start at the mid-point of the possible range for each dimension)
x0 = [55, 55, 55]
# Define bounds for each dimension
bounds = [(10, 100), (10, 100), (10, 100)]  # Lower and upper bound for each dimension
# Define the constraints
constraints = [{'type': 'eq', 'fun': constraint_function}]
# Call the optimizer
result = minimize(objective_function, x0, method='SLSQP', bounds=bounds, constraints=constraints)

$\omittedCode{prints optimized dimensions if successful, otherwise reports failure}{
# Print the result
if result.success:
    print(f"Optimized cabinet dimensions: {result.x}")
    print(f"Minimum cost: {result.fun}")
else:
    print(f"Optimization failed with message: {result.message}")}$
    \end{gptcodeblock}
\omitted{}{}
\end{gptresponse}
\end{chat}
\caption{\label{fig:cabinet_opt} \textbf{Optimizing Material Cost of Cabinet \wrt Volume.} The performance and cost functions are generated by prompting \gpt. To find the minimum of the material cost function with respect to volume, \gpt calls \lstinline{scipy.optimize}'s minimize function with an appropriate solver. In addition, the provided code extracts the relevant bounds and a suitable initial guess from the prompt. }
\end{figure}

\begin{figure}[tb]
\begin{subfigure}[t]{0.3\linewidth}
  \centering
  \includegraphics[width=0.9\linewidth]{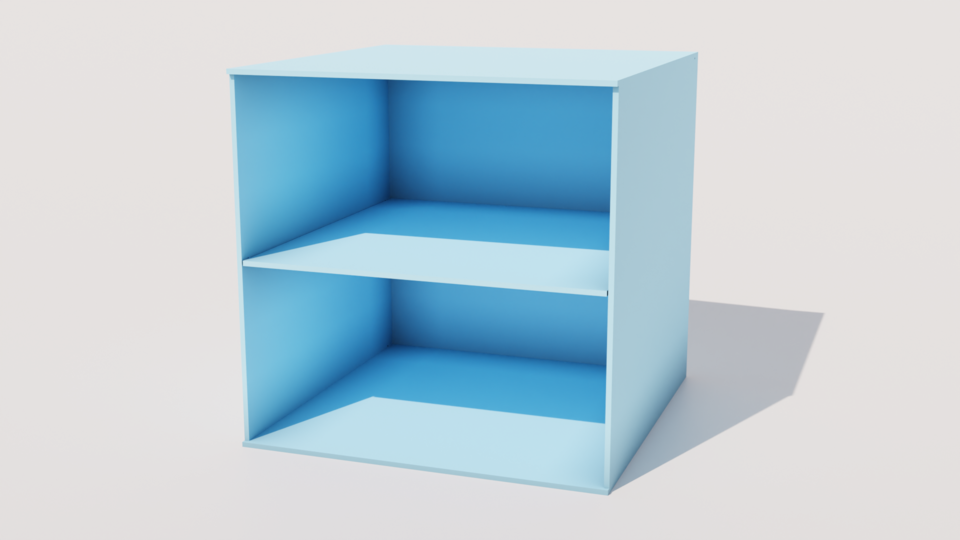}
  \caption{Initial Cabinet Design}
\end{subfigure}%
\begin{subfigure}[t]{0.3\linewidth}
  \centering
  \includegraphics[width=0.9\linewidth]{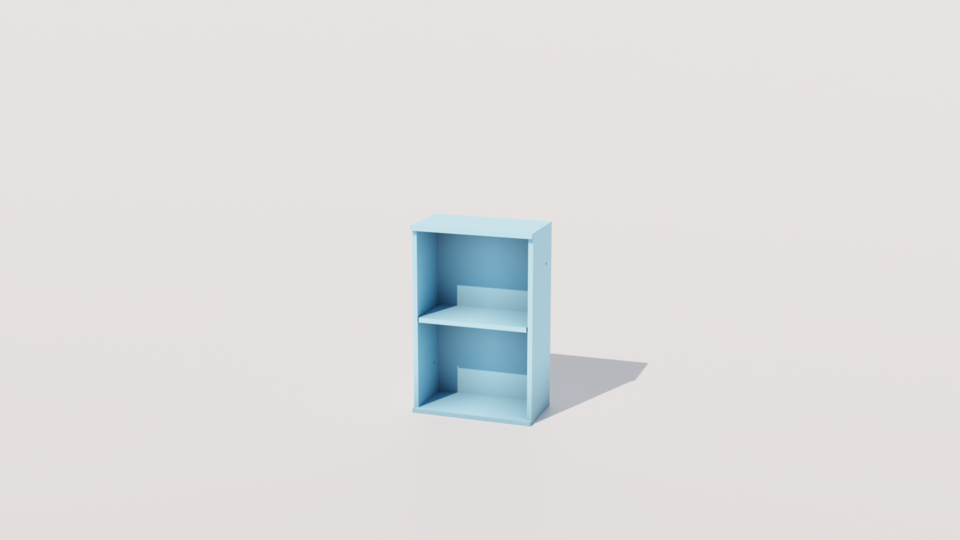}
  \caption{Optimized Cabinet}
\end{subfigure}%
\begin{subfigure}[t]{0.3\linewidth}
  \centering
  \includegraphics[width=0.9\linewidth]{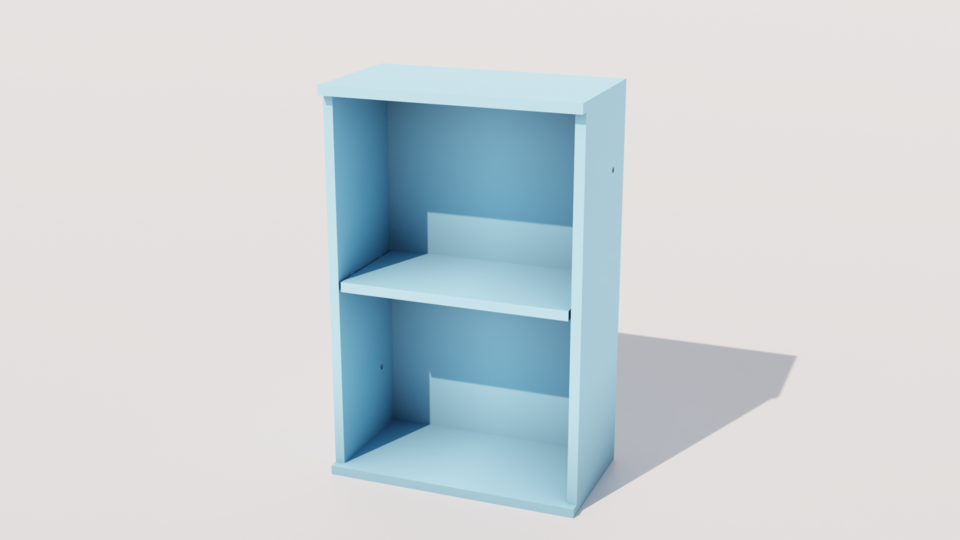}
  \caption{Optimized Cabinet (Close-up)}
\end{subfigure}
\caption{\textbf{Single Objective with Constraint Optimization.} \textbf{Left:} A cabinet with initial height, width, and depth of 55 in. \textbf{Middle:} Using an off-the-shelf solver provided by \lstinline{scipy.optimize}, \gpt provides a Python script that optimizes the cabinet's height, width, and depth given a desired volume while minimizing the material cost. We show an optimized cabinet that meets the volume constraint of 5000 cubic inches of storage capacity. \textbf{Right:} A close-up view of the optimized cabinet shown in the middle.}
\label{fig:cabinet_opt_render}
\end{figure}

%% file: sections/GraphArrows/InverseDesignExperiments/RobotArmPlanning1.tex
\begin{figure}[tbp]
\begin{chat}{false} 

\begin{userprompt}{0.0\textwidth}{} 
You are tasked with controlling an arm and a claw. The goal is to pick up an object on the ground with the claw and place it in a bin large enough for the object to fit in. You are working with (x, y, z) coordinates where 0 <= x, y, z <= 50. Below are functions you can use to control the arm and claw. 

Arm: 	
\begin{itemize}
    \item translate\_x(x\_dist): Translates the claw in the x direction by x\_dist.
    \item translate\_y(y\_dist): Translates the claw in the y direction by y\_dist.
    \item translate\_z(z\_dist): Translates the claw in the z direction by z\_dist. 
\end{itemize}

Claw: 	
\begin{itemize}
    \item grasp(): If there is an object at the claw's current location, the object will be grasped. When the object is grasped and the claw is moved, the object is moved to the same location as the claw's fingertips. 	
    \item release(): If the claw is grasping an object, the object will fall to the ground directly underneath it at z=0 or to any bin directly underneath it. 
\end{itemize}

\omitted{additional problem specifications}{}
\end{userprompt}

\begin{gptresponse}{0.01\textwidth}{} 
    \omitted{}{}
    \begin{gptcodeblock}{Python}
def move_arm(claw_initial, objects_initial, bins, T_max):
     def distance(p1, p2):
         $\omittedCode{Euclidean distance helper function}{# Calculate Euclidean distance between two points
        return math.sqrt((p1[0] - p2[0])**2 + (p1[1] - p2[1])**2 + (p1[2] - p2[2])**2)}$
    
    actions = []
    power_used = 0
    claw_current = list(claw_initial)
    while objects_initial and power_used <= T_max:
        # Find nearest object
        nearest_object = min(objects_initial, key=lambda o: distance(claw_current, o))

        # Calculate necessary translations to reach the object
        translations = [o - c for o, c in zip(nearest_object, claw_current)]

        # Check if there's enough power left to reach the object and drop it in a bin
        if power_used + sum(abs(t) for t in translations) * 2 > T_max:
            break

        # Move to the object
        $\omittedCode{adding translations to object and grasping to the actions list}{
        actions.append(('translate_x', translations[0]))
        actions.append(('translate_y', translations[1]))
        actions.append(('translate_z', translations[2]))
        actions.append(('grasp',))}$

        # Update claw position and power used
        claw_current = nearest_object
        power_used += sum(abs(t) for t in translations)

        # Find nearest bin
        nearest_bin = min(bins, key=lambda b: distance(claw_current, b))

        # Calculate necessary translations to reach the bin
        translations = [b - c for b, c in zip(nearest_bin, claw_current)]

        # Move to the bin and release the object
        $\omittedCode{adding translations to bin and releasing to the actions list}{
        actions.append(('translate_x', translations[0]))
        actions.append(('translate_y', translations[1]))
        actions.append(('translate_z', translations[2]))
        actions.append(('release',))}$

        # Update claw position and power used
        claw_current = nearest_bin
        power_used += sum(abs(t) for t in translations)

        # Remove the placed object and bin from the lists
        objects_initial.remove(nearest_object)
        bins.remove(nearest_bin)
    $\omittedCode{code checking that a solution was found}{
    if objects_initial:
        return "No solution found within power limit"
    else:
        return actions}$
    \end{gptcodeblock}
\omitted{}{}
\end{gptresponse}
\end{chat}
\caption{\label{fig:claw1} \textbf{Initial Proposed Solution to Robot Manipulation Planning.} \gpt provides a greedy approach to the problem of minimizing the distance traveled by the claw, which does not guarantee an optimal solution. }
\end{figure}

%% file: sections/GraphArrows/InverseDesignExperiments/RobotArmPlanning2.tex
\begin{figure}[tbp]
\begin{chat}{false} 

\begin{userprompt}{0.0\textwidth}{} 
Could you write a function that guarantees the global optimum? Make sure to account for the distances from the claw to object and claw to bin (after it has grasped the object). Also make sure to compute translations rather than Euclidean distance. Recall the bounds are 0 <= n <=3 and 0 <= m <= 3.
\end{userprompt}

\begin{gptresponse}{0.01\textwidth}{} 
\omitted{}{}\\   
This updated function generates all permutations of the objects and bins, calculates the total translation required for each permutation, and selects the one with the smallest total translation. 

\begin{gptcodeblock}{Python}
from itertools import permutations

def move_arm(claw_initial, objects_initial, bins, T_max):
    def translation_distance(p1, p2):
        $\omittedCode{Manhattan distance helper function}{
        # Calculate Manhattan distance (total translation) between two points
        return abs(p1[0] - p2[0]) + abs(p1[1] - p2[1]) + abs(p1[2] - p2[2])}$
    
    def calculate_total_translation(ordering):
        total_translation = 0
        claw_pos = list(claw_initial)
        for object_pos, bin_pos in ordering:
            total_translation += translation_distance(claw_pos, object_pos)  # Claw to object
            total_translation += translation_distance(object_pos, bin_pos)  # Object to bin
            claw_pos = bin_pos  # Claw is now at bin location
        return total_translation
    
    object_bin_pairs = list(permutations(zip(objects_initial, bins)))
    min_translation = float('inf')
    best_ordering = None
    for ordering in object_bin_pairs:
        total_translation = calculate_total_translation(ordering)
        if total_translation < min_translation:
            min_translation = total_translation
            best_ordering = ordering
            
    if min_translation > T_max:
        return "No solution found within power limit"
    else:
        actions = []
        claw_current = list(claw_initial)
        for object_pos, bin_pos in best_ordering:
            # Move to object
            $\omittedCode{adding translations to object and grasping to the actions list}{
            actions.extend([
                ('translate_x', object_pos[0] - claw_current[0]),
                ('translate_y', object_pos[1] - claw_current[1]),
                ('translate_z', object_pos[2] - claw_current[2]),
                ('grasp',)
            ])}$
            claw_current = list(object_pos)
            
            # Move to bin
            $\omittedCode{adding translations to object and grasping to the actions list}{
            actions.extend([
                ('translate_x', bin_pos[0] - claw_current[0]),
                ('translate_y', bin_pos[1] - claw_current[1]),
                ('translate_z', bin_pos[2] - claw_current[2]),
                ('release',)
            ])}$
            claw_current = list(bin_pos)
        return actions
    \end{gptcodeblock}
\omitted{}{}
\end{gptresponse}
\end{chat}
\caption{\label{fig:claw2} \textbf{Attempt at Brute Force Solution to Manipulation Planning.}  Though \gpt attempts to solve the minimum translation problem in a brute force manner, it fails to check all possible paths that the claw can take. }
\end{figure}

%% file: sections/GraphArrows/InverseDesignExperiments/Chair.tex
\begin{figure}[tb]
\begin{subfigure}[t]{0.32\linewidth}
  \centering
  \includegraphics[width=0.5\linewidth]{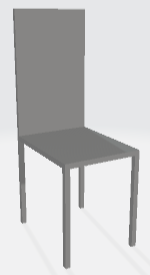}
  \caption{Initial Chair}
\end{subfigure}%
\begin{subfigure}[t]{0.32\linewidth}
  \centering
  \includegraphics[width=0.5\linewidth]{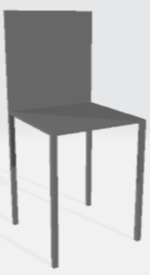}
  \caption{Chair that Minimizes Volume}
\end{subfigure}
\begin{subfigure}[t]{0.32\linewidth}
  \centering
  \includegraphics[width=0.5\linewidth]{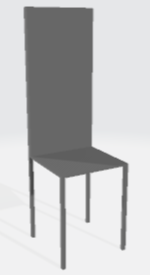}
  \caption{Chair that Maximizes Tipping Angle}
\end{subfigure}
\caption{\textbf{Optimizing Chair Design with Competing Objectives.} \gpt outputs code that handles a multi-objective performance space in the context of designing a chair to minimize volume while maximizing tipping angle. \textbf{Left:} A randomly sampled chair from the initial population before NSGA-II is applied. After running NSGA-II using the script \gpt provides, we visualize chairs on opposite ends of the frontier. \textbf{Middle:} A chair that minimizes volume by reducing the volume of the back seat and increasing leg lengths. \textbf{Right:} A chair that maximizes tipping value by extending the length and increasing the thickness of the back seat.}
\label{fig:chair_opt}
\vspace{-5mm}
\end{figure}

%% file: sections/GraphArrows/InverseDesignExperiments/fworpExperiment.tex
\begin{figure}[tbp]
\begin{chat}{false}
\begin{userprompt}{0.0\textwidth}{} 
I am creating a new device called a fworp.  A fworp is a wheeled soft silicone autonomous vehicle.  What are some reasonable parameters I should consider for my fworp, and parameter ranges I should I consider?
\end{userprompt}

\begin{gptresponse}{0.0\textwidth}{} 
    \omitted{}{}
    \begin{enumerate}
        \item \textbf{Size and Shape:} This is crucial to determine the potential applications of your fworp. You might want to keep the size compact if it is intended for home use or increase the size if it's designed for industrial applications. The shape will also play a crucial role in how the fworp navigates its environment. For the parameter range, consider anywhere from a small sphere (e.g., 10 cm in diameter) for consumer use, up to larger sizes (e.g., 2 m in diameter) for commercial/industrial use.
        
        \item \textbf{Weight:} The weight will impact the maneuverability and the energy consumption of the device. A lighter fworp would be easier to move and consume less energy but might be less stable. A heavier one could be more stable but require more energy. The weight could range from 1 kg (for small consumer devices) to 50 kg or more (for larger industrial devices).
        
        \item \textbf{Speed:} You'll want to adjust the speed depending on the use case. A fworp designed for a warehouse may need to be faster than one for home use to increase efficiency. Speeds might range from 1 km/h for careful navigation in domestic environments up to 20 km/h for more industrial settings.
        
        \item \textbf{Battery Life:} This will dictate how long the fworp can function without needing to be recharged. For domestic use, a battery life of several hours might be sufficient, whereas for industrial use, a battery life of several shifts may be needed. Depending on usage, anywhere from 4 to 24 hours could be appropriate.
        
        \item \textbf{Load Capacity:} Depending on the intended application, the fworp may need to carry items of varying weight. This could range from 1 kg for lightweight items up to 100 kg or more for industrial applications.
        
        \item \textbf{Sensor Range:} As an autonomous vehicle, the fworp will need sensors to navigate its environment. The range of these sensors will impact how well the fworp can navigate, with larger ranges allowing it to avoid obstacles more easily. Consider a range from 1 meter for simple obstacle avoidance up to 10 meters for more complex environments.
        
        \item \textbf{Durability of the Silicone Casing:} The fworp's casing will need to be able to withstand various pressures, temperatures, and impacts. This will depend heavily on the specific silicone material used. Some potential ranges to consider might be a tensile strength from 5 to 30 MPa, a hardness from 20 to 80 Shore A, and a temperature resistance from -60 to 200°C.
        
        \item \textbf{Communication Range:} If the fworp is intended to communicate with a central control system or other devices, the range of this communication will be important. This could be as short as a few meters for Bluetooth communication, up to several kilometers for more robust radio communication.
        
        \item \textbf{Robustness:} It's important to consider how the fworp will cope with various types of terrain and environmental conditions. It could be rated for indoor use only, or for use outdoors in various weather conditions.
        
        \item \textbf{Safety Features:} As an autonomous device, the fworp should have features to ensure it can operate safely, such as emergency stop functionality, collision detection, and avoidance systems.
    \end{enumerate}
    \omitted{}{}
\end{gptresponse}

\begin{userprompt}{0.0\textwidth}{} 
Do you think the fworp is a good idea, compared to a non-rubbery counterpart?  Why or why not?
\end{userprompt}

\begin{gptresponse}{0.01\textwidth}{} 
\omitted{}{}\\
\textbf{Advantages of a Silicone Fworp}

\begin{enumerate}
\item\textbf{Flexibility:} Silicone is a very flexible material, and a vehicle made of it could potentially navigate around obstacles or squeeze into spaces that a rigid vehicle could not.

\item\textbf{Durability:} Silicone is known for its durability and can withstand various environmental conditions, such as wide temperature ranges, UV radiation, and exposure to various chemicals. This could lead to a longer lifespan and less maintenance compared to some non-rubber materials.

\item\textbf{Safety: }A soft, flexible vehicle could potentially be safer in environments where it could come into contact with humans or delicate objects. A softer impact could reduce the risk of harm or damage.

\item\textbf{Noise reduction:} Soft materials like silicone could potentially operate more quietly than hard materials, which might be beneficial in certain environments, such as residential areas or hospitals.

\end{enumerate}

\textbf{Challenges with a Silicone Fworp}

\begin{enumerate}
\item\textbf{Structural integrity:} Silicone may not provide the same level of structural integrity as harder materials. You would need to ensure that the fworp can carry the required loads without deforming excessively. This could be particularly challenging if the fworp is large or needs to carry heavy items.

\item\textbf{Speed and Efficiency:} The soft, flexible nature of silicone might mean that the fworp moves more slowly or with less efficiency than a harder vehicle. The flexibility of the material could absorb some of the energy meant for propulsion.

\item\textbf{Maintenance:} While silicone is durable, it can also be prone to punctures and cuts. Depending on the environment, the surface may need regular inspections and repairs.

\item\textbf{Cost: }Depending on the specifics, manufacturing a vehicle out of silicone could potentially be more expensive than using more traditional materials.
\omitted{}{}
\end{enumerate}
\end{gptresponse}
\end{chat}
\caption{\label{fig:fworp_example}\textbf{Testing \gpt's Domain Transfer Ability for Inverse Design.} We query \gpt about the Fworp, an imaginary invention for which it has no direct knowledge about. Despite this, \gpt is able to transfer its knowledge about related concepts to propose design parameters and comparisons. However, some of its proposed parameters, such as load capacity, are actually performance metrics.  }
\end{figure}

%% file: sections/results.tex
\section{End-to-End Design Examples}
\label{sec:examples}


In the preceding sections, we have explored how \gpt can benefit different stages of the design and manufacturing processes. 
Now, we consider the comprehensive end-to-end design processes for two examples that have been consistently referenced throughout this manuscript: the cabinet and the quadcopter. 
From initial design and evaluation to manufacturing and testing, we showcase how \gpt seamlessly integrates with each stage. For the cabinet, \gpt automates 3D design, part sourcing, manufacturing instructions, assembly guidance, and performance evaluations. Similarly, for the quadcopter, \gpt facilitates design, part selection, manufacturing, assembly, and testing. These examples highlight the capabilities of \gpt in streamlining the product development process with minimal human intervention.

\subsection{Cabinet}
\label{sec:cabinet}

\begin{figure}[tb]
\includegraphics[width=\textwidth]{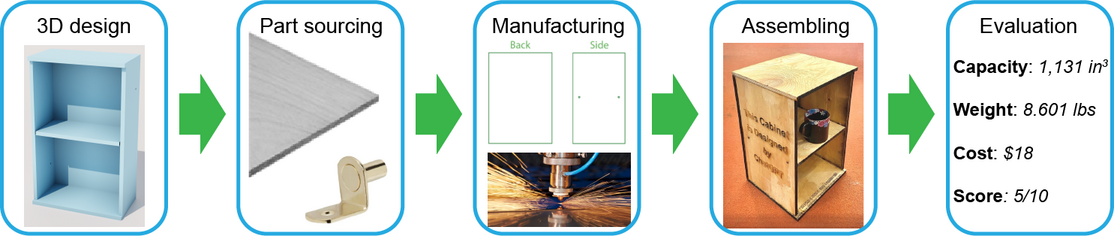}
\caption{\textbf{The Cabinet Design Process Using \gpt.} We showcase the comprehensive development process for a wood cabinet, highlighting \gpt's roles in 3D design generation, part sourcing, manufacturing instructions, assembly guidance, and performance evaluations.}
\label{fig:cabinet_summary}
\end{figure}

In this subsection, we consolidate and elaborate on the comprehensive \llm-assisted development process for a wood cabinet with minimal human intervention. Throughout this process, \gpt played major roles in generating the 3D design, facilitating part sourcing based on the design specifications, generating machine-readable manufacturing instructions, providing human-readable assembly guidance, and conducting final performance evaluations (Figure \ref{fig:cabinet_summary}). This holistic approach highlights the extensive capabilities of \gpt in automating and optimizing various stages of this product development process.

\subsubsection{Design}
We employed \gpt to generate Computer-Aided Design (CAD) models using such as \jscad (Section \ref{sec:textToDesign_JSCAD_basic}). After a few iterations, \gpt successfully generated accurate CAD design files (see Section \ref{sec:text_to_design} for more details). 
Additionally, we demonstrated utilizing \gpt to modify the cabinet design, including tasks such as adding a door and a handle (Section \ref{sec:textTODesign_spatial_constraints}) or adjusting the number of shelves (Section \ref{sec:textTODesign_abstractions}). \gpt even integrated prefabricated elements like brackets into the cabinet design (Section \ref{sec:prefabbed-ele}). We ultimately decided to fabricate the version of the cabinet presented in \fref{fig:textToDesign_simpleCabinet} with one shelf supported by four brackets. 

\begin{figure}[tbp]
\begin{chat}{false} 
\begin{userprompt}{0.0\textwidth}{}
Now let's fabricate this cabinet using wood. I have a universal laser system with a processing area of 32 by 18 inches. Can you recommend the materials I should buy? Ideally provide me links.
\end{userprompt}

\begin{gptresponse}{0.21\textwidth}{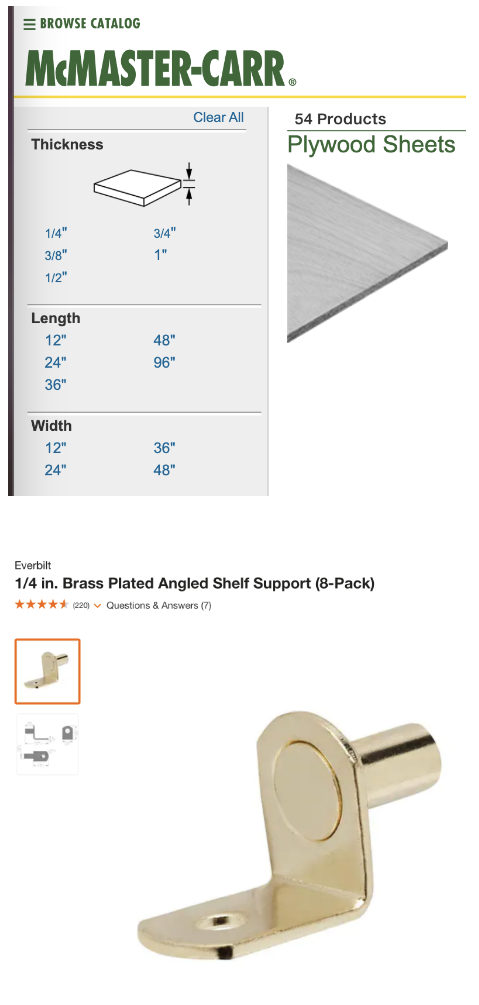} 

Based on your design, it appears you're working with a thickness of approximately 1/2 inch, which is a standard size for many types of wood. I'm going to suggest a couple of options for you:\\

\textbf{Baltic Birch Plywood} - This is a good option for cabinetry because it is strong, has a nice grain pattern, and finishes well. It's also highly recommended for laser cutting because it cuts cleanly and has minimal smoke and resin. Here's a link to purchase from McMaster-Carr: Baltic Birch Plywood Sheets.

\textbf{Medium-Density Fiberboard (MDF)} - If you're looking for a more budget-friendly option, MDF might be the way to go. It doesn't have the strength or the aesthetic appeal of real wood, but it is very consistent in density and thickness, making it ideal for precision work like this. It's also known for its smooth finish, and it laser cuts very well. Here's a link to purchase from McMaster-Carr: Medium-Density Fiberboard (MDF) Sheets.\\

For brackets, you can find a variety of options from Home Depot or other similar stores. Depending on the size and weight of your cabinet, you can choose a bracket that suits your needs. Just make sure the brackets you select fit the dimensions in your \jscad code. Here's a general link to the bracket section of the Home Depot website: Shelf Brackets - Home Depot\\

Remember that working with laser cutters requires adequate ventilation due to the smoke produced during the cutting process, and the edges of the wood will be darkened due to the heat from the laser. If you want to paint or finish the wood, it may need a bit of light sanding to clean up the edges.
\end{gptresponse}

\end{chat} 
\caption{\label{fig:cabinet3} \textbf{Part Sourcing Guided by \gpt Based on the Specific Cabinet Design and Manufacturing Method.} In this example, we employ laser cutter and pre-manufactured hardware.}
\end{figure}

\subsubsection{Manufacturing}
We then tested \gpt's capabilities in tackling real-world manufacturing challenges encountered during the fabrication of the cabinet. These challenges encompass various aspects of the general manufacturing process, such as part sourcing, generating machine-readable instructions, addressing material and fabrication constraints, facilitating design modifications, providing assembly guidance, and offering human-readable instructions.

For this particular example, our manufacturing plan involved utilizing a universal laser system to cut commercially available wood boards. First, by leveraging the detailed design specifications in \jscad, \gpt provided accurate suggestions for sourcing specific items, including brackets from Home Depot and wood pieces from McMaster. Each suggestion was accompanied by a concise description of the item (\fref{fig:cabinet3} and Section \ref{sec:part_sourcing}).
Furthermore, in order to realize the subtractive manufacturing approach, \gpt successfully generated precise 2D DXF files for laser cutting the wood pieces. These files were derived from the original 3D \jscad model, effectively guiding the laser cutter in executing the required cutting patterns (Section \ref{sec:designForManufacturing_Subtractive}). 

\begin{figure}[tbp]
\begin{chat}{false} 
\begin{userprompt}{0.0\textwidth}{}
The largest wood piece I can find on McMaster that can fit into the 18 by 32 inches laser cutter is 12 by 24 inches. Therefore, we need to scale down the initial cabinet design such that every piece can be cut from the 12 by 24 inches plates. The thickness of the wood piece I selected is 1/2 inch. Can you regenerate the python code for dxf considering this fabrication constraint?
\end{userprompt}

\begin{gptresponse}{0.25\textwidth}{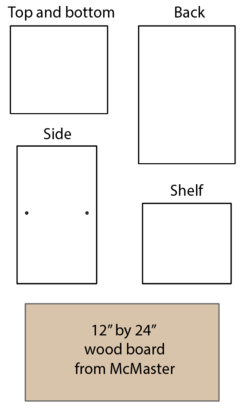} 
We need to make the width of the back board to be less than or equal to 12 inches.
One possible solution is to divide the backboard into two equal halves, each with a width of half of the original. They could be joined together after cutting. Let me modify the code accordingly:
\begin{gptcodeblock}{Python}
# dimensions
board_thickness = 1/2
cabinet_width = 12  # scale down to match wood piece size
scale = cabinet_height / 20  # scale factor based on original height
cabinet_height = 30 * scale
cabinet_depth = 18 * scale
brace_thickness = 1/8
hole_diameter = (0.375 + 1/16)
hole_depth = board_thickness * 3
bracketOffset = 2 * scale
\end{gptcodeblock}
\end{gptresponse}

\end{chat} 
\caption{\label{fig:cabinet4} \textbf{Handling Manufacturing Constraints with the Assistance of \gpt.} Here, the available wood board size is insufficient for the original cabinet design. \gpt successfully scaled down the entire design, ensuring manufacturability.}
\end{figure}

We also showcased \gpt's ability to adapt the design based on fabrication constraints. For example, the largest wood pieces we were able to obtain from McMaster that could fit into our laser cutter were 12 by 24 inches. Unfortunately, this size was too small to support the original cabinet design. As we presented this constraint to \gpt, it successfully scaled down the entire design to ensure compatibility with the available wood pieces (\fref{fig:cabinet4}). Given that our laser cutter is capable of engraving patterns onto the cutting materials, we took advantage of this feature and requested \gpt to generate Python code for producing DXF files that could engrave words onto the side board of the cabinet. \gpt was able to fulfill this request, providing us with the necessary code to create the desired engraving.

\begin{figure}[tb]
    \includegraphics[width=\textwidth]{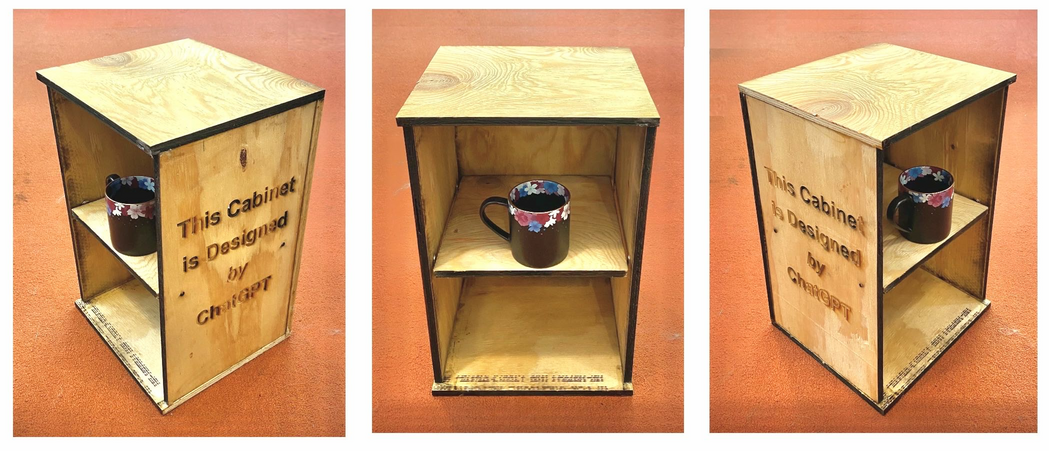}
    \caption{\label{fig:cabinet7} \textbf{The Manufactured Cabinet.} Real-world cabinet manufactured from different view angles, showcasing its scale with a mug as a reference.}
\end{figure}

Finally, once we had the laser-cut wood pieces, we proceeded to assemble them into a fully functional cabinet. During this process, we sought \gpt's expertise in recommending the most suitable adhesive for wood. \gpt suggested Quick-Set Epoxy, 3M DP100, 1.64 FL. oz. Cartridge as the optimal glue for our purposes (\fref{fig:cabinet6}). Additionally, \gpt provided detailed treatment guidance and step-by-step procedures for the assembly, ensuring a smooth and successful construction process (\fref{fig:cabinet6}). 
The final fabricated cabinet is showcased in Figure \ref{fig:cabinet7}, offering three different view angles. To provide a clear perspective of its size, a mug is placed on the middle shelf, serving as a reference for scale.

\begin{figure}[tb]
\begin{chat}{false} 
\begin{userprompt}{0.0\textwidth}{}
Great. I purchased Play Wood from McMaster with 1/2 inches. I want to use glue to assemble them together. Please provide a link for the glue I should buy and give me the details treatment procedure.
\end{userprompt}

\begin{gptresponse}{0.00\textwidth}{}
For plywood assembly, a good choice would be Titebond III Ultimate Wood Glue. It offers superior strength, a long open-assembly time, and is safe to use. You can find it on the McMaster site here.

Here's a simple procedure to assemble your cabinet:

\textbf{1. Dry Fit}: Start by fitting all the pieces together without any glue to ensure they all fit properly.

\textbf{2. Sanding}: If necessary, sand the edges of your wood pieces for a better fit. Be careful not to alter the dimensions significantly.

\textbf{3. Gluing Procedure}:
\begin{itemize}
    \item \textbf{Back Board to Side Boards}: Apply a bead of glue along the side edges of the back board. Align it with the side boards such that the back board is sandwiched between the two side boards. Ensure the edges are flush. Clamp them together and wipe off any excess glue that squeezes out with a damp cloth. Allow the glue to dry according to the manufacturer's instructions.

\item \textbf{Bottom Board}: Apply glue along the bottom edges of the side and back boards. Position the bottom board so that its edges are flush with the bottom edges of the side and back boards. Clamp and allow to dry.

\item \textbf{Top Board}: Similarly, apply glue to the top edges of the side and back boards, align the top board, clamp, and allow to dry.

\item \textbf{Shelf Board}: Apply glue along the edges of the shelf board where it will contact the side and back boards. Slide it into position and clamp. Allow to dry.
\end{itemize}

\textbf{4. Finishing}: After the glue is completely dry, you can finish your cabinet with a finish of your choice (paint, varnish, etc) to increase its durability and improve appearance.

Remember, it's important to clamp your pieces together after you've applied the glue to ensure a tight bond. Also, ensure your workspace is level so your cabinet comes out square.

Please remember to follow all safety guidelines when working with tools and materials, including wearing appropriate personal protective equipment. Always work in a well-ventilated area when using glues or finishes.
\end{gptresponse}

\end{chat} 
\caption{\label{fig:cabinet6} \textbf{Comprehensive Assembly Instructions Provided by \gpt.} \gpt provides guidance on glue selection, pre-treatment, assembly sequence, and finishing options}
\end{figure}

\subsubsection{Performance evaluation}
Using the 3D \jscad design as a basis, \gpt was prompted to develop functions to evaluate performance criteria: storage capacity, load capacity before sagging substantially, plywood material cost, and accessibility for individuals in wheelchairs. See Section~\ref{designToPerformance_Objective} for details on this type of design evaluation for numerical evaluation and Section~\ref{designToPerformance_Subjective} for subjective metrics. When comparing these values for the original and scaled cabinets, the storage capacity, load capacity before sagging, material cost, and accessibility were computed to be 9,205 $\mathrm{in}^3$, \$67, 14.336 lbs, and a score of 5/10 for the original cabinet, and 1,131 $\mathrm{in}^3$, \$18, 8.601 lbs, and a score of 5/10 for the scaled-down cabinet. It failed to compute an appropriate strength and could not verify the equation, but other evaluations are correct. Based on these numbers, both cabinets were qualitatively classified as "strong" on a range of "strong," "moderate," or "weak," being far stronger than a typical cabinet used in an office. However, while the larger cabinet was categorized as "highly useful" on a scale of "useful," "moderate", or "not useful" due to its high strength and storage capacity, the smaller cabinet was only "moderately useful" due to its small size. Further aesthetic evaluation in text form is included in Figure~\ref{fig:cabinet_aesthetics}, classifying without computation the original design as "Contemporary" and the scaled design as "Minimalist". 

\begin{figure}[tb]
\begin{chat}{false} 
\begin{userprompt}{0.0\textwidth}{}
My cabinet has these dimensions. Describe the aesthetic category it should fall into. It has one shelf at mid-height. 

\texttt{Cabinet\_dimensions} = \{ 
	'\texttt{height}': 30,
	'\texttt{width}': 18,
	'\texttt{depth}': 20,
	'\texttt{board\_thickness}': 0.5
\}
\end{userprompt}

\begin{gptresponse}{0.00\textwidth}{}
Your cabinet, with its dimensions of 30" height, 18" width, and 20" depth, likely falls into the "contemporary" aesthetic category. With one shelf at mid-height, it offers a balanced and symmetrical look. Its proportions are conducive to a modern, sleek design that can seamlessly blend into a variety of interiors. The cabinet size suggests functionality and purpose, making it an appealing choice for those who appreciate clean lines and minimalist style.
\end{gptresponse}

\begin{userprompt}{0.0\textwidth}{}
\omitted{Description of smaller scaled cabinet} 

\end{userprompt}

\begin{gptresponse}{0.00\textwidth}{}
Given its relatively small dimensions, your cabinet could fall into the "Minimalist" aesthetic category. With its clean, uncomplicated design and single mid-height shelf, it would emphasize functionality and simplicity, which are key principles of minimalist design. Its smaller size would also contribute to an uncluttered appearance, further aligning with the minimalist aesthetic.
\end{gptresponse}

\end{chat} 
\caption{\label{fig:cabinet_aesthetics} \textbf{Aesthetic Description.}  \gpt comments on the aesthetic category of the designed and fabricated cabinets based on its Geometry.}
\end{figure}


\subsection{Quadcopter}
\label{sec:quadcopter}
\input{sections/result-copter}

%% file: sections/result-copter.tex
\begin{figure}[tb]
\includegraphics[width=\textwidth]{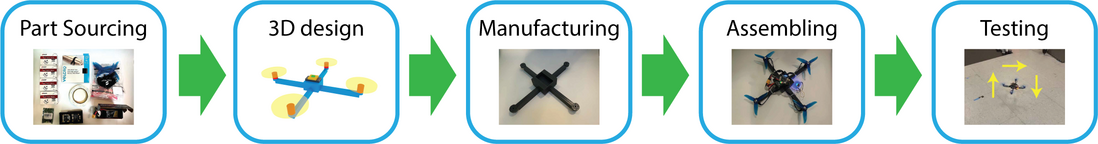}
\caption{\textbf{The Quadcopter Design Process using \gpt.} 
The process involves sourcing parts, creating a 3D design,
manufacturing, assembling and testing the quadcopter.
We successfully manufactured a working quadcopter.}
\label{fig:quadcopter-process}
\end{figure}

In this subsection, our objective is to design a functional indoor quadcopter capable of flight, hovering, and landing, using \gpt with minimal human intervention. The process involves sourcing parts, creating a 3D design incorporating the chosen parts, and ultimately, manufacturing, assembling and testing the quadcopter,
as depicted in Figure~\ref{fig:quadcopter-process}. 

\subsubsection{Parts Sourcing}
Many real-world systems, like drones, are built from pre-existing components with various kinds of  specifications. Therefore, our first task is to utilize \gpt to select appropriate parts for our specific use case. This was successfully accomplished using \gpt. The detailed process is elaborated in Section~\ref{sec:part_sourcing}, and the selected parts are shown in Figure~\ref{fig:copter-assemble} (left).

\subsubsection{Text-to-Design}
\label{sec:copter-text-to-design}

\begin{figure}[tb]
\includegraphics[width=\textwidth]{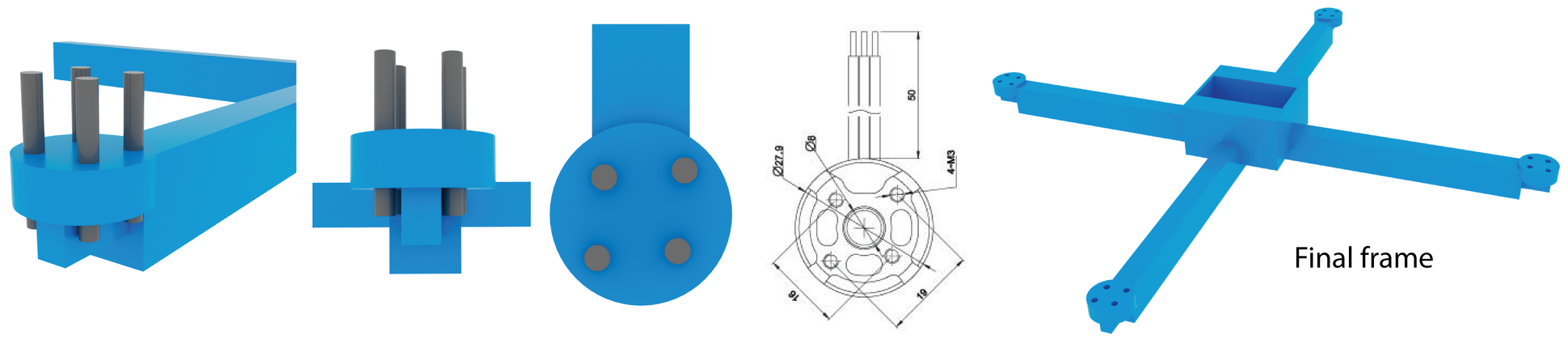}
\caption{\textbf{The Quadcopter Frame.}
Here we show that \gpt can create cylinders for the motor mounting holes.
Using boolean operations, we successfully created a valid frame.}
\label{fig:quadcopter-man}
\end{figure}

Upon identifying the components, we employ \gpt once more to generate a viable quadcopter design incorporating those parts. With minimal human intervention, we successfully crafted a geometric design for the quadcopter, as detailed in Section~\ref{sec:prefabbed-ele}. Now, we shift our focus to practical issues: 1) how parts are mounted onto the designed frame, and 2) whether the frame is manufacturable. Given that describing each part's geometric details is challenging and that \gpt doesn't fully comprehend how to design the frame for optimal physical balance, we provide \gpt with low-level instructions to guide adjustments to the current frame design rather than expecting it to independently modify the design.

First, we adjust the frame bar's cross-sectional size using \gpt to ensure it's 3D-printable and adequately robust. Then, we combine all boxes and cylinders that form the frame in the geometric design. To stabilize the battery placement, we semi-integrate it into the frame, and subtract it from the frame so the frame securely holds the battery. For the controller and signal receiver, which are much lighter and smaller than the battery, we simply glue them onto the battery, eliminating the need for additional accommodations.

Lastly, we mount the motors using screws for stability, requiring screw holes in the frame. To minimize human effort, we utilize \gpt to create holes in the frame. For each motor, we instruct \gpt to generate four cylinders representing the required holes, detailing the hole specifications via text according to the motor's specifications. Even though crafting mounting holes isn't trivial, we successfully produced the correct cylinders after a few prompts with \gpt. The cylinders are shown in dark gray in Figure~\ref{fig:quadcopter-man}(Left). Once the hole cylinders for one motor are ready, we let \gpt group and duplicate them using our \lstinline{place} function. \gpt managed to position the hole cylinders correctly but had issues with proper rotation, which we later manually corrected.

As seen, the hole cylinders overlap with the frame bar. Since we cannot change the frame bar's thickness due to the manufacturing concern, we manually adjusted the frame bar's tip thickness to prevent it from obstructing the holes. This was done manually, as it proved challenging to adequately convey the problem and solution to \gpt.

Our experiments also revealed a limitation: adjusting designs we completed earlier proves difficult. After extensive interaction with \gpt, referring back to previously discussed design elements becomes a challenge. Consequently, if any issues arise with earlier addressed parts, it becomes arduous to revisit them with \gpt and prompt modifications. Ideally, we should finalize each design without the need for future revisions, as adjustments later prove difficult. The final frame result is displayed in Figure~\ref{fig:quadcopter-man}(Right).

\subsubsection{Design-to-Manufacturing}
\begin{figure}[tb]
    \centering
    \includegraphics[width=.9\textwidth]{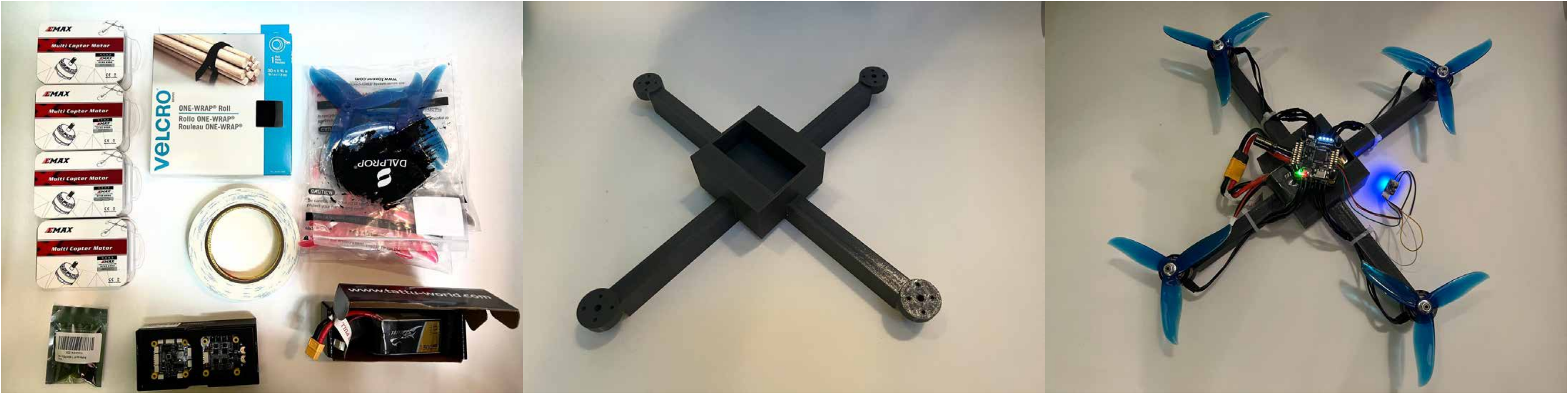}
    \caption{\label{fig:copter-assemble} \textbf{The Parts and the Printed Frame of the Copter.} \textbf{Left}: Selected parts. \textbf{Middle}: Printed frame. \textbf{Right}: Assembled copter.}
\end{figure}

The only part that needs manufacturing is the frame. Once the frame is determined, we fabricate it using a 3D printer. Because the representation of the frame results from Boolean operations of boxes and cylinders, it is simple to directly convert them to .stl format which is widely recognized by the 3D printers. We used OpenJSCAD to do it. Once we have the .stl file, we manufacture the frame using Stratasys Fortus 400 since it has a sufficiently large build volume and it realizes precise, robust, and durable parts. We instruct the printer to use the least amount of infill possible in the print settings. By choosing a lower infill percentage, the printer will create a sparse or hollow internal structure rather than a solid one. This decision helps conserve material and reduce print time without significantly compromising the strength or weight of the copter's frame, given that it is not a load-bearing component. The resulting fabricated copter frame not only meets the required dimensions, but also balances the strength and weight, necessary for optimum flight performance. We visualize the printed frame in Figure~\ref{fig:copter-assemble} (middle).

\subsubsection{Assembling and Real-World Verification}

With the 3D printed frame ready, we proceed to the assembly stage, integrating the pre-prepared components. Given that assembly considerations were incorporated into our \gpt-guided design process, the assembly of the quadcopter is straightforward. The battery is secured in the central frame slot using double-sided tape and wrappers. Similarly, the controller and receiver are placed atop the battery and secured with double-sided tape and wrappers. The four motors are attached using screws. All elements are affixed firmly and stably, resulting in a sturdy copter ready for flight, as shown in Figure~\ref{fig:copter-assemble} (right).

\begin{figure}[tb]
    \centering
    \includegraphics[width=.9\textwidth]{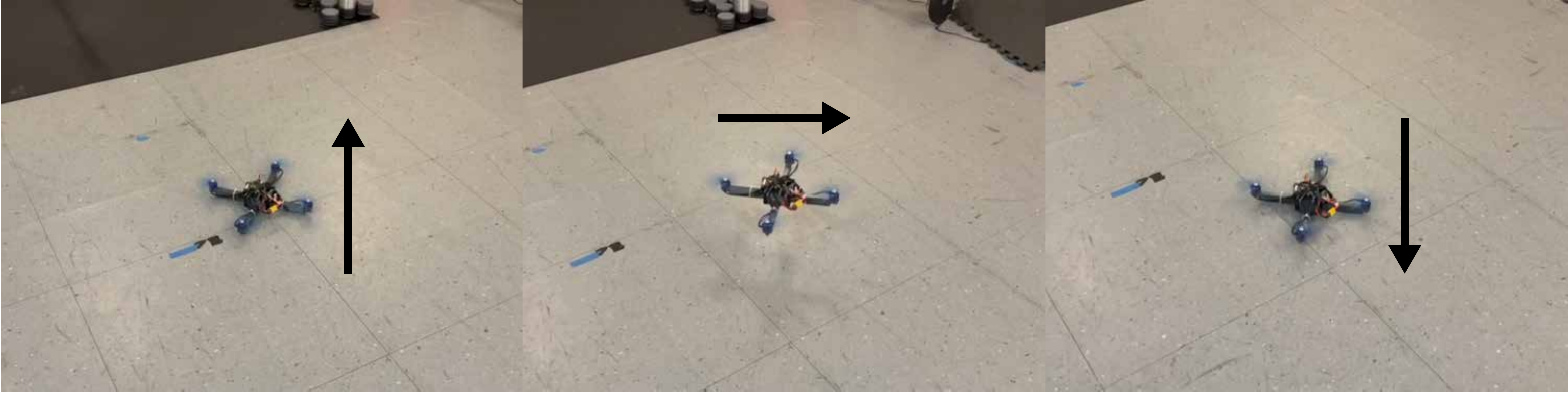}
    \caption{\label{fig:copter-flight-test} \textbf{The Flight Test.} \textbf{Left}: the ascending test. \textbf{Mid}: the hovering test. \textbf{Right}: the descending test.}
\end{figure}

Once assembled, we conduct a series of tests. First, we administer an ascending test, directing the copter to lift off the ground and ascend to a specific altitude. This test gauges the combined thrust of the motors and the propellers' efficacy in converting the motors' rotary motion into lift. It also allows us to evaluate the copter's responsiveness to radio transmitter commands, the flight controller's interpretive capacity, and the copter's overall ascent stability. The motion is depicted in Figure~\ref{fig:copter-flight-test} (left).

Following ascent, we undertake a hovering test. During this phase, the copter is directed to maintain its altitude and position for a set period. Hovering demands continuous, simultaneous operation of all four motors to counter gravity. This test significantly illuminates the copter's capacity to achieve and maintain stable flight, a vital characteristic of any functioning copter. The hovering motion is demonstrated in Figure~\ref{fig:copter-flight-test} (mid).

Finally, we execute a descending test, instructing the copter to safely and gradually descend to the ground. This evaluates the copter's ability to control thrust reduction and the resulting downward motion, as well as the flight controller's capacity to interpret and carry out the descent command. It is also a crucial examination of the copter's landing abilities; a smooth, safe landing is essential to preserve the copter and its components. The descending motion is exhibited in Figure~\ref{fig:copter-flight-test} (right).

\subsubsection{Text-to-Performance}
We also investigate how \gpt can help with measuring the performance of a given quadcopter design.
Given a current design iteration of the copter from Section~\ref{sec:perf_quadcopter}, \gpt is able to identify important trade-offs to optimize and subsequently implement optimization strategies to improve performance. One such trade-off \gpt identified is between weight and size, where smaller copters are generally able to stay afloat longer due to reduced weight and aerodynamic drag, while larger copters have more space to accommodate larger batteries which can provide more energy for longer flight times. Out of all the possible optimization methods to find the best combination of parameters that maximize flight time, speed, and distance while meeting constraints on weight and size, \gpt chose a very suitable numerical method of Particle Swarm Optimization (PSO) from the PySwarm library. Aside from being very efficient and simple to implement, PSO has a strong global search capability, which is beneficial when the optimal solution might be located in a large and complex space, and allows for real-time adjusting of the copter's weight and size based on performance data. \gpt has a strong grasp on the inherent trade-offs of such systems, and is capable of generating tailored ideas and feasible solutions to optimize performance.

We now turn to the details of using simulation to evaluate the quadcopter's performance. In the workflow of fabricating a functional robot, simulation is often used for both control and collecting performance metric statistics that can be used for optimization. Since our fabricated robot includes its own controller, we focus on using the robot's performance in simulation for design optimization. Our design space involves both a parameterized quadcopter whose frame bar lengths can vary but is otherwise constrained by the design created in Section \ref{sec:copter-text-to-design} and the controller design. While it is possible to ask \gpt to provide suggestions on the type of controller to apply, we choose to have \gpt generate a LQR controller, which is widely used for UAVs. We break down multicopter optimization into three steps: 1) Given the \jscad design of the quadcopter, convert the design into a format specific to modeling multibody systems, such as URDF, and the means of computing relevant physical properties that inform the controller design, such as the robot's mass. 2) Given the robot's physical properties, generate a LQR controller for simulation. 3) Given  a robot design in URDF, functions for extracting the design's relevant physical properties, and a controller, synthesize an algorithm to optimize the robot's design.

\input{sections/GraphArrows/InverseDesignExperiments/QuadcopterToURDF}

\input{sections/GraphArrows/InverseDesignExperiments/QuadcopterLQR}
\input{sections/GraphArrows/InverseDesignExperiments/QuadcopterOptimization}

\begin{figure}[tb]
\begin{subfigure}[t]{0.5\linewidth}
  \centering
  \includegraphics[width=0.7\linewidth]{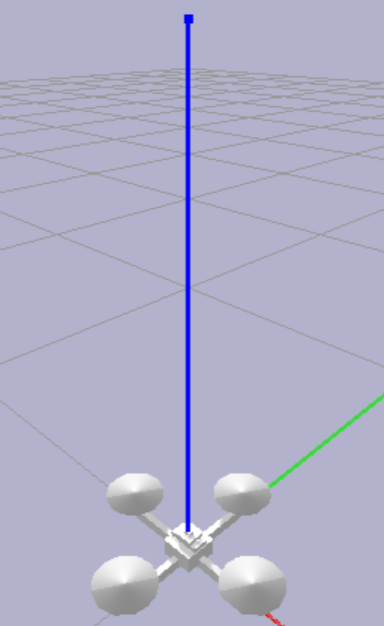}
  \caption{Initial Position}
\end{subfigure}%
\begin{subfigure}[t]{0.5\linewidth}
  \centering
  \includegraphics[width=0.7\linewidth]{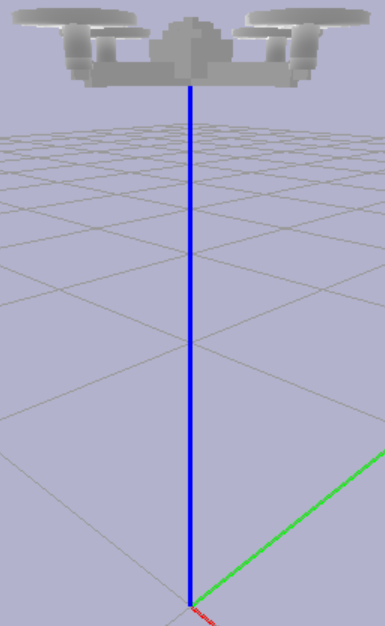}
  \caption{Final Position}
\end{subfigure}
\caption{\textbf{Simulating Quadcopter Performance.} \textbf{Left:} Following the \jscad to URDF conversion script provided by \gpt, we show the quadcopter rendered in PyBullet's simulator. The quadcopter has optimized frame bar lengths discovered with the script provided by \gpt and is resting at its initial position. \textbf{Right:} Using the LQR controller designed by \gpt, the quadcopter is able to lift off and hover at the goal position 1m off the ground.}
\label{fig:quadcopter_render}
\end{figure}

The resulting quadcopter with optimized frame bar lengths is visualized in Fig. \ref{fig:quadcopter_render}.


%% file: sections/GraphArrows/InverseDesignExperiments/QuadcopterToURDF.tex
\paragraph{Converting \jscad to URDF} We start with the \jscad quadcopter design developed in Section \ref{sec:copter-text-to-design}. Because there is no straightforward equivalent of subtraction and union from \jscad in URDF, we omit the creation of the holes in the motor base and replace the union of motor base parts with placement of individual links while retaining the essence of the original design. We then take an object-oriented approach to having \gpt synthesize the equivalent URDF code. As seen in \ref{fig:tourdf_component}, we prompt \gpt to create a Component class whose instances store the geometry, mass, position, and orientation attributes of the corresponding \jscad primitives. Component instances also have distinct names to represent URDF links. We additionally prompt \gpt to generate helper functions for placing instances with different geometries in Fig. \ref{fig:tourdf_component}. This framework allows \gpt to generate a function that places components in terms of absolute coordinate positions and orientations and to replicate the Python equivalent of the \jscad, as shown in Fig. \ref{fig:tourdf_place}. 
\begin{figure}[tbp]
    \begin{chat}{false}
        \begin{userprompt}{0.0\textwidth}{} 
        Below are helper functions written in OpenJSCAD. 
\begin{gptcodeblock}{JavaScript}
function createBox(w, h, d) {
  return cuboid({size:[w, h, d], center:[0, 0, 0]});
}
function createCylinder(r, h) {
  return cylinder({radius: r, height: h, center: [0, 0, 0], segments: 64});
}
\end{gptcodeblock}
        
        Implement the above functions in Python. Both functions should create an instance of a Component, which is a class that keeps track of the component’s name, mass, geometry, as well as absolute translation and rotation with respect to the origin. In addition, consider the below place() function.
        
\begin{gptcodeblock}{JavaScript}
// A function to add a component to the drone at specific coordinates
function place(component, x, y, z, a = 0) {
  let cc = translate([x, y, z], rotate([0, 0, a / 180.0 * Math.PI], component));
  return cc;
}
\end{gptcodeblock}
        
        Implement the place function in Python, which should return a copy of the Component with the given translation and rotation. We want to give this Component a unique name, so also take in an additional suffix argument. If not None, the suffix should be added to the copy’s name. Do not use any hypothetical libraries.
        
        \end{userprompt}
    
        \begin{gptresponse}{0.0\textwidth}{} 

         \omitted{}{} 
\begin{gptcodeblock}{Python}
import math
import copy

class Component:
    def __init__(self, name, mass, geometry, translation=[0,0,0], rotation=[0,0,0]):
        self.name = name
        self.mass = mass
        self.geometry = geometry
        self.translation = translation
        self.rotation = rotation

class Cuboid:
    def __init__(self, w, h, d):
        self.width = w
        self.height = h
        self.depth = d

class Cylinder:
    def __init__(self, r, h):
        self.radius = r
        self.height = h

def create_box(name, mass, w, h, d):
    return Component(name, mass, Cuboid(w, h, d))

def create_cylinder(name, mass, r, h):
    return Component(name, mass, Cylinder(r, h))

def place(component, x, y, z, a=0, suffix=None):
    cc = copy.deepcopy(component)
    cc.translation = [x, y, z]
    cc.rotation = [0, 0, a / 180.0 * math.pi]
    if suffix is not None:
        cc.name = component.name + suffix
    return cc

\end{gptcodeblock}
        \omitted{}{}
        \end{gptresponse}
        
    \end{chat}
\caption{\textbf{Creating Object-Oriented Representations of Robot Components.} Using the Component class for representing a robot link facilitates the placement of links with the place() function. Later in the example, the Component class is modified to additionally keep track of an instance's parent link. }
\label{fig:tourdf_component}
\end{figure}

\begin{figure}[tbp]
    \begin{chat}{false}
        \begin{userprompt}{0.0\textwidth}{} 
         \omitted{\gpt synthesized functions}{}
         
        Using createBox, createCylinder, and place defined above, write a Python script that recreates the box and cylinder components created by the below OpenSCAD script.
        
         \omitted{\jscad quadcopter design}{} 
        
        Store the results of place() in a dictionary mapping component name to component because we will need to access the components later. Use the “suffix” argument to ensure all components have unique names. Use the following masses of quadcopter parts and convert them to kg:

        \omitted{masses of each component}{}
        \end{userprompt}
    
        \begin{gptresponse}{0.0\textwidth}{} 
        \omitted{}{}
        \begin{gptcodeblock}{Python}
def create_drone():
    # Create the quadcopter parts with respective masses in kg
    motor = create_cylinder("motor", 0.030, 14, 32)
    motor_base_part1 = create_cylinder("motorBasePart1", 0.01, 14, 10)
    propeller = create_cylinder("propeller", 0.0135, 65, 8)
    controller = create_box("controller", 0.0107, 41, 38, 8.1)
    battery = create_box("battery", 0.015, 70, 35, 33)
    receiver = create_box("receiver", 0.0015, 16, 11, 5.4)
    motor_base_part2 = create_box("motorBasePart2", 0.01, 20, 7, 10)
    frame_bar1 = create_box("frameBar1", 0.01, 2 * 130 * 1.5 + 31, 15, 25)
    frame_bar2 = create_box("frameBar2", 0.01, 2 * 130 * 1.5 + 31, 15, 25)
    body = create_box("body", 0.05, 75, 75, (33 + 8.1 + 5.4) / 2 * 1.2 * 1.5)

    # Dictionaries to store components
    drone_parts = {}

    # Place the frame parts
    drone_parts["frameBar1"] = place(frame_bar1, 0, 0, 0, 0, "_place1")
    drone_parts["frameBar2"] = place(frame_bar2, 0, 0, 0, 90, "_place2")

    # Place the central body, moved down by 4
    body_height = (33 + 8.1 + 5.4) / 4 * 1.2 * 1.5 - 4
    drone_parts["body"] = place(body, 0, 0, body_height, 0, "_place1")

    # Place the quadcopter parts
    dist = 130 * 1.5 + 25  # half of frameBar length

    # Place the motor base parts
    for i in range(4):
        angle = i * 90
        x = dist * math.cos(angle * math.pi / 180)
        y = dist * math.sin(angle * math.pi / 180)
        drone_parts[f"motorBasePart1_place{i+1}"] = place(motor_base_part1, x, y, 10, angle)
        drone_parts[f"motorBasePart2_place{i+1}"] = place(motor_base_part2, x, y, 0, angle)

    $\omittedCode{2 additional for loops for placing motors and propellers in a similar manner as the motor bases.}{# Place the motors
    for i in range(4):
        angle = i * 90
        x = dist * math.cos(angle * math.pi / 180)
        y = dist * math.sin(angle * math.pi / 180)
        drone_parts[f"motor{i+1}"] = place(motor, x, y, 10 + 32 / 2 + 5, angle)

    # Place propellers on top of the motors, lift them up by an additional 5
    for i in range(4):
        angle = i * 90
        x = dist * math.cos(angle * math.pi / 180)
        y = dist * math.sin(angle * math.pi / 180)
        drone_parts[f"propeller{i+1}"] = place(propeller, x, y, 10 + 32 + 8 / 2 + 5, angle)}$

    # Place battery, controller, and receiver on the body, one on top of the other
    drone_parts["battery"] = place(battery, 0, 0, 33 / 2 + 2 + 13, 0)
    drone_parts["controller"] = place(controller, 0, 0, 33 + 8.1 / 2 + 2 + 13, 0)
    drone_parts["receiver"] = place(receiver, 0, 0, 33 + 8.1 + 5.4 / 2 + 2 + 13, 0)
    return drone_parts
        \end{gptcodeblock}
        \omitted{}{}
        \end{gptresponse}
        
    \end{chat}
\caption{\textbf{Replicating Quadcopter Component Placement.} Disregarding the union and subtract operators in \jscad, the above Python code accurately reproduces the locations of the components as the ones in the \jscad design. The only error is that certain components do not have unique names, which \gpt later corrects on additional prompting.}
\label{fig:tourdf_place}
\end{figure}

However, one difficulty is that unlike CAD-like designs, formats for representing robots also require relations between components to be represented as joints in order to accurately simulate the robot dynamics. We prepare for this when synthesizing the component placement script by prompting GPT to store the components as a dictionary, which allows easy access to the components. We tackle the challenge of generating the quadcopter's joints by relying on \gpt's knowledge of the spatial relation between components in a quadcopter. After equipping the Component class with a function that sets the parent link, we use this interface to have \gpt synthesize a sequence of robot joints, as shown in Fig. \ref{fig:tourdf_relation}. We find that although \gpt understands certain substructures, such as the fact that the motor is placed on top of the motor base and the propeller is connected to the top of the motor, its initial definition results in an invalid URDF format, as both of the frame bars are root links. We thus explicitly prompt \gpt to choose one of the frame bars as the root link. 

Finally, \gpt is tasked with creating a full URDF file. Because the robot is represented with modular Component instances that contain all relevant information on individual links' mass and geometry as well as relations to parent links, it is relatively straightforward for \gpt to create helper functions that synthesize URDF links and joints. We note that creating a joint is a more involved task, since the link's absolute position and orientation must be converted to relative position and orientation to the parent link. It is necessary to explicitly prompt \gpt to use the appropriate rotation matrices in its calculation; otherwise, it does not appropriately account for how the parent link's rotation affects the child link's relative translation. Because we choose LQR as the robot's controller in simulation, we ask \gpt to compute the full assembly's mass and moment of inertia given the Python code it has generated thus far. It outputs reasonable Python code that computes the assembly's center of mass and uses the parallel axis theorem to combine the moment of inertia of the individual links.

\begin{figure}[tbp]
    \begin{chat}{false}
        \begin{userprompt}{0.0\textwidth}{} 
        \omitted{Python code generated by \gpt thus far}{} \\
        
        Finish writing the script by using commonsense knowledge about relations between quadcopter components to set the appropriate parent component. Generate the full URDF file of the quadcopter.
        
        \end{userprompt}
    
        \begin{gptresponse}{0.0\textwidth}{} \omitted{}{}
        \begin{gptcodeblock}{Python}
def set_parent_relations(drone_parts):
    # Set parent-child relations
    drone_parts["body_place1"].set_parent(drone_parts["frameBar1_place1"])
    drone_parts["body_place1"].set_parent(drone_parts["frameBar2_place2"])
    
    for i in range(4):
        drone_parts[f"motorBasePart1_place{i+1}"].set_parent(drone_parts["frameBar1_place1"])
        drone_parts[f"motorBasePart2_place{i+1}"].set_parent(drone_parts["frameBar1_place1"])
        drone_parts[f"motor{i+1}"].set_parent(drone_parts[f"motorBasePart1_place{i+1}"])
        drone_parts[f"propeller{i+1}"].set_parent(drone_parts[f"motor{i+1}"])
        
    drone_parts["battery"].set_parent(drone_parts["body_place1"])
    drone_parts["controller"].set_parent(drone_parts["body_place1"])
    drone_parts["receiver"].set_parent(drone_parts["body_place1"])

        \end{gptcodeblock}
        \omitted{}{}
        \end{gptresponse}

        \begin{userprompt}{0.0\textwidth}{} 
        Recall the set\_parent\_relations function. \\
        
        \omitted{original function generated by \gpt}{} \\
        
        This function results in the error that a URDF robot can only have one root joint. Could you make either frameBar1 or frameBar2 the root joint? 
        
        \end{userprompt}

        \begin{gptresponse}{0.0\textwidth}{} 
        \omitted{}{}
        \begin{gptcodeblock}{Python}
def set_parent_relations(drone_parts):
    # Set parent-child relations
    # Make frameBar1 the root
    drone_parts["frameBar1"].set_parent(None)
    drone_parts["frameBar2"].set_parent(drone_parts["frameBar1"])
    drone_parts["body"].set_parent(drone_parts["frameBar1"])
    
    for i in range(1,5):
        drone_parts[f"motorBasePart1_place{i}"].set_parent(drone_parts["frameBar1"])
        drone_parts[f"motorBasePart2_place{i}"].set_parent(drone_parts["frameBar1"])
        drone_parts[f"motor{i}"].set_parent(drone_parts[f"motorBasePart1_place{i}"])
        drone_parts[f"propeller{i}"].set_parent(drone_parts[f"motor{i}"])
        
    drone_parts["battery"].set_parent(drone_parts["body"])
    drone_parts["controller"].set_parent(drone_parts["body"])
    drone_parts["receiver"].set_parent(drone_parts["body"])        
        \end{gptcodeblock}
        \omitted{}{}
        \end{gptresponse}
        
    \end{chat}
\caption{\textbf{Constructing Relations between Quadcopter Links.} After creating and placing links in the appropriate positions, \gpt must infer parent-child relations that are not available in the \jscad design to complete the URDF representation. In its first attempt, it produces an invalid file because only one of the frame bars can be the root joint. However, \gpt demonstrates an understanding of how the motor, motor base, and propeller should be connected and that the main body contains the battery, controller, and receiver.}
\label{fig:tourdf_relation}
\vspace{-5mm}
\end{figure}

In summary, we find that explicitly prompting \gpt to provide suitable object-oriented representations, such as the Component class, and modularizing the code generation as much as possible, e.g. asking \gpt to synthesize helper functions, placing components, and defining parent-child relations separately are key techniques in successfully converting from \jscad to URDF.

%% file: sections/GraphArrows/InverseDesignExperiments/QuadcopterLQR.tex
\paragraph{Deriving a LQR Controller for the Quadcopter.}  Multicopter control is an extensively studied problem, for which various algorithms have been proposed such as PID controllers, LQR controllers, and more complex alternatives. We aim to synthesize a LQR controller, as it is not only a popular choice in the literature but also guarantees optimal control when the multicopter dynamics and stable fixed point are known. In particular, we focus on controlling a quadcopter, which is an underactuated system with 6 degrees of freedom but only 4 independent actuators, resulting in nonlinear dynamics even without aerodynamical effects. However, LQR provides optimal control by linearizing the system around a stable fixed point, which then has a closed form solution. 

As LQR is widely used for control, the algorithm is easily accessible through the \lstinline{control} Python library that performs LQR with a single function call. As such, the main challenge in using LQR for quadcopter control is deriving its state space representation, the $A$ and $B$ matrices, used in the linear model of the system as shown in Equation \ref{eq:linear_model}. $x$ is a 12-dof state vector, including coordinate position, linear velocity, roll, yaw, pitch, and angular position. $u$ is the 4-dof control input, typically consisting of the upward thrust and the external torque applied in the xyz directions. Note that $x$ in \ref{eq:linear_model} is the difference between the current state and a user-specified target state. Similarly, $u$ is with respect to the target control input, where the torques are 0 and the thrust balances out the gravitational force acting on the qaudcopter. \gpt is then asked to provide a suitable fixed point at which the quadcopter is stable, and it does provide a correct response.

\begin{equation}
    \dot{x} = Ax + Bu
    \label{eq:linear_model}
\end{equation}

In order to verify \gpt's results, we ask \gpt to first synthesize the full kinematics model before using a a symbolic manipulation library, sympy, to compute the state space matrices. This forces \gpt to produce an interpretable dynamical model that a domain expert can verify, rather than directly outputting the $A$ and $B$ matrices. This proves to be essential as even though \gpt captures the high-level idea of assuming the quadcopter to be a rigid body and applying Newton-Euler equations to describe its acceleration and angular acceleration, it is unable to zero-shot provide the exact model of the system without user feedback. For instance, as shown in Fig. \ref{fig:quadcopter_lqr}, \gpt formulates the correct rotation matrix but does not apply it correctly to convert from inertial frame (control inputs) to the body frame (linear acceleration). Although \gpt can correct its error when given feedback, this type of error is difficult to catch without rigorously checking the correctness of \gpt's calculations and highlights the limitation that in some inverse design domains, a user cannot generate precise outputs from \gpt without the expert knowledge to perform verification. After this fix, the resulting code for deriving the $A$ and $B$ matrices is still incorrect as the simulated quadcopter sinks downward due to a sign error in the equations for acceleration. Similarly, generating the simulation loop in PyBullet requires several rounds of iteration on the initial code. Some of the mistakes are more obvious, such as incorrectly indexing into the control input vector when applying external force and torques, whereas others are more specific to PyBullet's API. For the former, we directly pointed out \gpt's mistake. In the latter case, we find that simply giving \gpt the error message and asking it to fix its code suffices for our problem. 


\begin{figure}[tbp]
    \begin{chat}{false}
        \begin{userprompt}{0.0\textwidth}{} 
        Derive the A and B matrices for the LQR controller by linearizing the dynamics of the quadcopter around the fixed point. Assume that the system has 12 states and 6 dof. The control inputs should be total thrust and x,y,z torques. You should first write the full kinematics model and then call sympy to compute the jacobians. Account for the yaw angle in the equations for linear velocity and angular velocity.
        \end{userprompt}
    
        \begin{gptresponse}{0.0\textwidth}{}
        \omitted{full description of quadcopter dynamics and high-level approach}{ Given these, the Jacobian A can be calculated by taking the partial derivatives of the right-hand sides of these equations with respect to the state vector $(p_x, p_y, p_z, \phi, \theta, \psi, v_x, v_y, v_z, w_x, w_y, w_z)$, and the Jacobian B can be calculated by taking the partial derivatives of the right-hand sides of these equations with respect to the control vector (u1, u2, u3, u4). \\
        
        We will now use the sympy library to compute these Jacobians:}\\
\begin{gptcodeblock}{Python}
import sympy as sp

# Define the state variables
px, py, pz, phi, theta, psi, vx, vy, vz, wx, wy, wz = sp.symbols('px py pz phi theta psi vx vy vz wx wy wz')

# Define the control inputs
u1, u2, u3, u4 = sp.symbols('u1 u2 u3 u4')

# Define the constants
Ix, Iy, Iz, m = sp.symbols('Ix Iy Iz m')

# Rotation Matrix
R = sp.Matrix([[sp.cos(theta)*sp.cos(psi), sp.sin(phi)*sp.sin(theta)*sp.cos(psi) - sp.cos(phi)*sp.sin(psi), sp.cos(phi)*sp.sin(theta)*sp.cos(psi) + sp.sin(phi)*sp.sin(psi)],
               [sp.cos(theta)*sp.sin(psi), sp.sin(phi)*sp.sin(theta)*sp.sin(psi) + sp.cos(phi)*sp.cos(psi), sp.cos(phi)*sp.sin(theta)*sp.sin(psi) - sp.sin(phi)*sp.cos(psi)],
               [-sp.sin(theta), sp.sin(phi)*sp.cos(theta), sp.cos(phi)*sp.cos(theta)]])

# System dynamics
f = sp.Matrix([vx, vy, vz, wx, wy, wz,
               R[2, 0]*u1/m, R[2, 1]*u1/m, R[2, 2]*u1/m - 9.81,
               ((Iy - Iz)*wy*wz + u2) / Ix, ((Iz - Ix)*wx*wz + u3) / Iy, ((Ix - Iy)*wx*wy + u4) / Iz])

state = sp.Matrix([px, py, pz, phi, theta, psi, vx, vy, vz, wx, wy, wz])
controls = sp.Matrix([u1, u2, u3, u4])

# Compute Jacobians
A = f.jacobian(state)
B = f.jacobian(controls)

\end{gptcodeblock}
        \omitted{}{}
        \end{gptresponse}

        \begin{userprompt}{0.0\textwidth}{} 
        Can you double check the way the rotation matrix is applied when computing the system dynamics?
        \end{userprompt}

        \begin{gptresponse}{0.0\textwidth}{}
        Yes, you're correct, and I apologize for the mistake in my previous explanation. The total force in the body frame is given by [0, 0, -u1]. To find the force in the inertial frame, we multiply by the rotation matrix.\\
        The rotational dynamics are correct as given, but the translational dynamics need to be updated to account for the orientation of the quadcopter. Let's write them correctly.\\        
        The linear acceleration (ignoring air resistance) is just the force divided by the mass, so:
        
\begin{gptcodeblock}{Python}
acc_x = R[0, 2] * -u1 / m
acc_y = R[1, 2] * -u1 / m
acc_z = R[2, 2] * -u1 / m + g

\end{gptcodeblock}
        where g is the gravitational acceleration.\\

        Let's correct this in our sympy code:
        
        $\omittedCode{previous code with correction made}{\begin{gptcodeblock}{Python}
# Define the gravitational acceleration
g = sp.symbols('g')

# System dynamics
f = sp.Matrix([vx, vy, vz, wx, wy, wz,
               R[0, 2]*-u1/m, R[1, 2]*-u1/m, R[2, 2]*-u1/m + g,
               ((Iy - Iz)*wy*wz + u2) / Ix, ((Iz - Ix)*wx*wz + u3) / Iy, ((Ix - Iy)*wx*wy + u4) / Iz])

# Compute Jacobians
A = f.jacobian(state)
B = f.jacobian(controls)
\end{gptcodeblock}}$
        \end{gptresponse}
        
    \end{chat}
\caption{\textbf{Deriving Quadcopter State-Space Matrices for LQR Control.} Although \gpt formulates the problem correctly and uses the correct kinematics equations and rotation matrix, the initial derivation of the state-space matrices has a couple of errors, one of which is that \gpt indexes the wrong elements of the rotation matrix.}
\label{fig:quadcopter_lqr}
\vspace{-5mm}
\end{figure}

%% file: sections/GraphArrows/InverseDesignExperiments/QuadcopterOptimization.tex
\paragraph{Co-design of Quadcopter's Shape and Control.} Now that the quadcopter design can be outputted to URDF and the LQR controller can be synthesized from a given design, we ask \gpt to optimize the quadcopter design. Much of the design space has been fixed by the fact that most components have unchangeable dimensions. However, as the framebars are not predefined components but rather constructed from carbon-fiber tubes, we focus on optimizing their lengths. The control then follows directly from a given design, as we only need to compute the design's total mass and moment of inertia  to apply the LQR controller derived by \gpt. With the objective of minimizing the number of simulation steps required to reach a goal height of 1m, we prompt \gpt with the full Python script that converts OpenJSCAD to URDF and performs simulation, and we find that it is able to provide an outline of the optimization, consisting of an objective function that involves creating a quadcopter with specified framebar lengths and performing the simulation loop, as shown in Fig. \ref{fig:quadcopter_opt}. \gpt also provides reasonable bounds on the frame bar lengths of (100mm, 500mm) when prompted further. \gpt then provides an outer optimization script that uses the helper objective function. We also explicitly prompt \gpt to complete its outline by updating component creation code to generate frame bars with the correct masses and to change the placement of components dependent on the length of the frame bars, namely the motors, motor bases, and propellers. 

\begin{figure}[tbp]
    \begin{chat}{false}
        \begin{userprompt}{0.0\textwidth}{} 
        I have the following code to generate the geometry of a Quadcopter, create an LQR controller, and simulate it. I would like to optimize the lengths of frameBar1 and frameBar2 in order to minimize the time it takes the robot to reach its target set point 1 meter off the ground, constrained to a reasonable parameter range. How might I go about modifying this code to do that?

        \omitted{script for creating and simulating quadcopter}{}
        
        \end{userprompt}
    
        \begin{gptresponse}{0.0\textwidth}{} 
        \omitted{}{}
        
        First, define a function that takes the lengths of frameBar1 and frameBar2 as parameters and returns the time it takes for the robot to reach its target set point 1 meter off the ground. This function should include the code for setting up and simulating the robot, as well as measuring the time it takes to reach the target set point.
        \begin{gptcodeblock}{Python}
from scipy.optimize import minimize

def simulation_time(length1, length2):
    # Use lengths to set up and simulate the robot.
    # Measure and return the time it takes to reach the target set point.

    # Modify the frameBar1 and frameBar2 lengths 
    frame_bar1 = create_box("frameBar1", 0.2526, 2 * length1 * 1.5 + 31, 15, 25)
    frame_bar2 = create_box("frameBar2", 0.2526, 2 * length2 * 1.5 + 31, 15, 25)

    # Create drone and simulate
    drone_parts = create_drone()

    set_parent_relations(drone_parts)
    # Generate URDF file
    write_urdf_file('quadcopter_scad.urdf', drone_parts)

    # Place your simulation and control code here. 
    # Ensure that your control objective is to reach 1 meter above ground.
    # time_to_reach_target = ...

    return time_to_reach_target

        \end{gptcodeblock}
        Next, use the minimize function from the scipy.optimize module to find the lengths that minimize the function:

\begin{gptcodeblock}{Python}

# Define constraints and bounds
bounds = [(MIN_LENGTH1, MAX_LENGTH1), (MIN_LENGTH2, MAX_LENGTH2)]  # Replace with actual min and max lengths

# Initial guess
x0 = [(MIN_LENGTH1 + MAX_LENGTH1)/2, (MIN_LENGTH2 + MAX_LENGTH2)/2]  # initial guess can be the midpoint of bounds
result = minimize(simulation_time, x0, bounds=bounds, method='SLSQP')  # SLSQP method supports bounds

# The optimal lengths are stored in result.x
optimal_length1, optimal_length2 = result.x

$\omittedCode{printing the results}{print(f"Optimal length for frameBar1: {optimal_length1}")
print(f"Optimal length for frameBar2: {optimal_length2}")}$

    \end{gptcodeblock}
        \omitted{}{}
        \end{gptresponse}
        
    \end{chat}
\caption{\textbf{Code Skeleton for Optimizing Quadcopter Design.} To optimize a quadcopter's design, \gpt identifies proposes to wrap the code that produces the quadcopter's URDF file and that performs LQR control in an objective function, which enables it to perform optimization in the outer scope. However, its initial suggestion to perform SQSLP is unsuitable, as there are no gradients provided from simulation.}
\label{fig:quadcopter_opt}
\end{figure}

As seen in Fig. \ref{fig:quadcopter_opt}, \gpt initially proposes SQSLP, an optimization method that requires gradients from the objective function, which is not trivial to apply as we would require gradients computed from simulation. When prompted to provide an alternative, \gpt suggests Differential Evolution, which meets the specification of not requiring a differentiable optimization problem, but can be computationally expensive. We thus explicitly prompt \gpt to provide code to perform grid search over the two frame bar lengths. This terminates in a reasonable amount of time with the result of making the frame bars as short as possible, which is the expected solution as the smaller the quadcopter, the less inertia it faces when taking off and decelerating towards the goal.

%% file: sections/discussion.tex
\section{Discussion}

\subsection{Observed Capabilities and Limitations} \label{sec:capabilities}

Our examination of \gpt's performance in the fields of design and manufacturing has underscored several crucial capabilities instrumental in supporting the design processes, in addition to pinpointing significant limitations. Next, we summarize these observations and discuss various methods to mitigate the identified limitations. 

\noindent \textbf{Capabilities:} \linebreak
\noindent \textbf{C.1 Extensive Knowledge Base in Design and Manufacturing:} \gpt boasts a comprehensive knowledge base in the realms of design and manufacturing. Its capabilities extend to solving a wide array of problems and automatically completing specifications, making it a versatile tool across various domains.

\noindent \textbf{C.2 Iteration Support:} \gpt incorporates an iterative approach to problem-solving. When feedback is provided on errors, it attempts to rectify them. This ability to adapt, although not always successful, is a valuable facet of an \gpt's performance.

\noindent \textbf{C.3 Modularity Support:} \gpt supports modular design, demonstrating the ability to reuse or adapt previous designs or solutions when explicitly instructed. While it does not inherently retain memory of past interactions, explicit instructions can help leverage its modular capabilities effectively.

\noindent \textbf{Limitations:} \linebreak
\noindent \textbf{L.1 Reasoning Challenges:} \gpt encounters difficulties with certain types of reasoning, particularly those involving analytical reasoning and computations. These limitations can manifest as notable challenges in the design and manufacturing domain, for instance, a general lack of spatial reasoning capabilities.

\noindent \textit{Potential Solutions:} Implementing well-crafted domain-specific languages (DSLs) can help address these challenges. DSLs, widely used in computer science, encapsulate recurring knowledge, rules, and valuable abstractions, thereby filling knowledge gaps. Alternatively, APIs that can perform the complex computations can be integrated. \gpt's proficiency in creating high-level abstractions can be utilized to generate inputs that be processed through APIs by computational solvers.

\noindent \textbf{L.2 Correctness and Verification:} \gpt often produces inaccurate results or justifications for its solutions and lacks the ability for self-verification.

\noindent \textit{Potential Solutions:} Apart from relying on human verification, automated verification can be accomplished by utilizing APIs that conduct checks and validations. By leveraging \gpt's iterative capabilities (C.2), we can create a feedback loop that continues until a satisfactory solution is obtained.

\noindent \textbf{L.3 Scalability:} As tasks become larger or more complex, \gpt's performance can deteriorate, often struggling to manage multiple tasks concurrently.

\noindent \textit{Potential Solutions:} One strategy is to partition larger tasks into multiple sub-tasks. For instance, rather than requesting it to evaluate multiple performance metrics simultaneously, it may be more effective to request them individually. When constructing more complex models, employing an incremental design process can prove beneficial. Components can be designed and verified separately before being assembled into the final model. \gpt's modularity support (C.3) can be used to facilitate the creation of complex models from a series of instructions.

\noindent \textbf{L.4 Iterative Editing:} When a design needs modifications, specifying those changes as a prompt will often lead to unsatisfactory behaviors. This is because the \gpt, upon receiving a change prompt, will regenerate the design, often overlooking elements specified in previous prompts. This situation poses challenges to design editing.

\noindent \textit{Potential Solutions:} Our solutions involved either feeding the prompts back in to create a full specification in a single prompt, though this also caused challenges due to the scalability limitations discussed above. The alternative was explicit modularization to enable parts to be designed and reused, though the latter was not possible in some cases.

\noindent \textbf{Dualism:}

\noindent \textbf{D.1 Context Information:} \gpt's performance improves significantly with the provision of context information. The more detailed the domain description, the better it performs. Furthermore, \gpt is adept at providing context for its actions, making it an asset in sequential workflows. This characteristic proves particularly beneficial when using \gpt's generated content in subsequent tasks, as these tasks can utilize the context included in the output from the initial task. However, it may also be harmful if the user does not provide enough context, or if the user would like to create an unusual design, but \gpt is unable to overcome the biases it associates with a particular domain.

\noindent \textbf{D.2 Unprompted Responses:} \gpt often infers aspects that are not specified in the prompt, either auto-completing specifications or finding ways to make decisions without enough information. While this is interesting in the design context in terms of allowing for partial specifications that can be auto-completed, it can sometimes be overly proactive, guiding the design in some aspects, which may limit creativity.

\section{Conclusions}

In conclusion, we find that \gpt possesses numerous capabilities that can be leveraged within the domain of design and manufacturing. This area, where creativity converges with practicality, presents exciting opportunities for advancements that can potentially bring about a significant shift in the way we ideate, prototype, and manufacture a broad range of products.

However, it is essential to recognize that substantial work remains to be done to fully support the integration of these tools within this field. A fundamental issue is that design for manufacturing involves a delicate balance between creativity and formal verification. Engineering design presents a paradox; it requires precision and exactness, yet thrives on an iterative and exploratory spirit. While our experiments have managed to circumvent limitations for formal reasoning through user guidance, DSL crafting, and APIs that call computations, there is still much to understand about how best to implement these strategies.

For instance, we hope that our analysis can stimulate new insights about DSL design. Historically, DSLs have been developed with human users in mind. However, when we shift our perspective to creating DSLs for an AI coder, new questions and possibilities emerge. What should these DSLs look like? We believe our analysis provides valuable insights into this concept, particularly within the design and manufacturing domain.

Similarly, in terms of API usage and framework development, we have observed a myriad of possibilities. Approaches range from dividing problems into parts that can be tackled by \gpt and others that are best solved by traditional methods, to iterative solutions, and even to a complete reframing of the problem by asking \gpt to generate problem-solving code. Each of these approaches carries potential advantages and disadvantages. So, what should an optimal framework look like? We hope our research will aid others in formulating an answer to this question.

In summary, we believe our analysis offers valuable insights into how LLMs like \gpt can be harnessed in the domain of design and manufacturing. While we have made substantial strides, the path to fully exploiting the potential of these tools in this domain remains open, rich with opportunities for further exploration and innovation.









%% file: appendix.tex
\section{DSLs and Prompting Tips for Text-To-Design with \gpt }

\subsection{CSG with \jscad}
\label{sec:appx-jscad-hints} 
\gpt was able to use the \jscad library out-of-the-box, with no additional explanation or restriction of the API on the part of the user. 
However, as described in \sref{sec:textToDesign_JSCAD_basic}, \gpt did fall into a number of common pitfalls when constructing designs.
To mitigate the most common mistakes that \gpt made, each time we asked \gpt to build a design using \jscad, we provided the set of hints and reminders shown in \fref{fig:appx-jscad-hints}.

\begin{figure}
\begin{chat}{false}
\begin{userprompt}{0pt}{}
\omitted{main content of the prompt}{}

When building this code, keep in mind:\\
1. It's very important to follow the \jscad format. There must be a function named main and an export statement.\\
2. Each function must be imported from the appropriate module. Take care to choose the correct module for each function. For example, colorize comes from colors; cuboid comes from primitives; union comes from booleans; and translate comes from transforms. This is not an exhaustive list, feel free to use any function from any module.\\
2. OpenJScad positions each component relative to its center point.\\
3. It is very important that the individual components are in contact with one another, but no part protrudes into any other part.\\
4. Pay attention to the primitive types -- for example, cuboid() must be used instead of cube() if you are building a box with different lengths along different dimensions.\\
5. The z direction is up.
\end{userprompt}

\end{chat}
\caption{\textbf{Hints for using \jscad} Each time we asked \gpt to construct a design using \jscad, we provided the following hints after the main prompt to avoid the most common pitfalls that \gpt fell into.}
\label{fig:appx-jscad-hints}
\end{figure}

\subsection{Sketch-based parametric CAD DSL}
\label{sec:sketch_based_dsl}
We propose a streamlined version of the standard sketch-based CAD language by exposing only the sketch and extrude operations along with basic sketch primitives, which already cover a wide range of geometric variations. To automatically generate CAD models from \gpt's output, we utilize Onshape's API.  
When aiming for single-shot CAD design (\ie, with no iterative feedback), we found that a four-pronged prompt generally resulted in the most reliable output.
One aspect of the prompt described the specific task that \gpt should complete.
The remaining three aspects of the prompt provided generic context for our target CAD DSL, and largely remained constant throughout our experiments.
The specific aspects were: (1) a description of our modified DSL, (2) an example constructed with this DSL, and (3) a set of tips that \gpt should keep in mind when constructing its own result.

The prompt we used to describe these aspects to work with local coordinate systems and a global coordinate system can be seen in chat format in Fig.\ref{fig:dsl_local} and Fig.\ref{fig:dsl_global}, respectively.
\begin{figure}[tb]
\begin{chat}{false} 

\begin{userprompt}{0pt}{} 
We will define a design language. There are two key operators: 
\\
1) createSketch(primitive, plane): creates a sketch of a certain primitive on a given plane and returns its ID. There are two types of primitives you can create, circles or rectangles. The circle primitive is circle(center\_x, center\_y, radius), where center\_x is the x coordinate of the center, center\_y is the y coordinate and radius is the radius. The rectangle primitive is instantiated rectangle (center\_x, center\_y, length, width). 
The plane defines the 2D plane where you will draw the sketch. You can just use one of the 3 default places: XY\_PLANE, XZ\_PLANE, ZY\_PLANE. You can also use the plane create by the result on an extrude, which you do by calling the function cap(extrude, side), where 'extrude' is the Id returned by the extrude operation (defined below). And side is one of the following :"max\_z", "min\_z", "max\_y", "min\_y", "max\_x", "min\_x". The side argument defines which planar face of an extruded solid will be used for the sketch operation. For example, "min\_z" will select the planar face whose bounding box center has a minimal z-component compared to all other planes' bounding box centers.
And there is the important constraint that you can only put the center of sketch primitives inside or on the edge of a planar face, if you do not put them on one of the default planes.

2) extrude (sketch, length), where sketch is the ID returned by the sketch operator and the length determines the length of the extrude.
Note that sketches lie within their respective plane, and they will get extruded along the plane’s normal direction. A common pattern you will encounter is that the height variable of a solid should often be the length parameter in the extrude operator. Whereas the other dimensions of of the solid are defined by the sketch primitives.

For example, if you want to design a round table with a single center leg and a leg base you can do: 

\begin{gptcodeblock}{Python}
leg base legBase_sketch = createSketch(circle(0,0,3), XY_PLANE) 
legBase_solid = extrude(legBase_sketch , 1) 

leg_sketch = createSketch(circle(0,0,1), cap(legBase_solid, "max_z")) 
leg_solid = extrude(leg_sketch , 10) 

top_sketch = createSketch(circle(0,0,8), cap(leg_solid, "max_z")) 
top_solid = extrude(top_sketch , 1) 
\end{gptcodeblock}

End of the example. 

Always write code using variables. Try to prefer a few variables by reusing them in the design when appropriate. 
Write code in syntactically correct python, knowing that you have the functions createSketch, circle, rectangle, cap and extrude and the default planes. 
\end{userprompt} 

\end{chat}
\caption{\textbf{A Sketch-Based CAD DSL Prompt with Local Coordinate Systems.} Our prompt used for the sketch-based CAD experiments with local coordinate systems.}
\label{fig:dsl_local}
\end{figure}

\begin{figure}[tbp]
\begin{chat}{false} 
\begin{userprompt}{0pt}{} 
We will define a design language. There are two key operators: \\
1) createSketch(primitive, plane): 
creates a sketch of a certain primitive on a given plane and returns its ID. There are two types of primitives you can create, circles or rectangles. The circle primitive is circle(center\_x, center\_y, center\_z, radius), where center\_x is the x coordinate of the center, center\_y is the y coordinate of the center, center\_z is the z coordinate of the center and radius is the radius. The rectangle primitive is an instantiated rectangle (center\_x, center\_y, center\_z, length, width). 
The plane defines the 2D plane where you will draw the sketch. You can just use one of the 3 default places: XY\_PLANE, XZ\_PLANE, ZY\_PLANE. You can also use the plane created by the result on an extrude, which you do by calling the function cap(extrude, side), where 'extrude' is the Id returned by the extrude operation (defined below). And side is one of the following :"max\_z", "min\_z", "max\_y", "min\_y", "max\_x", "min\_x". The side argument defines which planar face of an extruded solid will be used for the sketch operation. 
For example, "min\_z" will select the planar face whose bounding box center has a minimal z-component compared to all other planes' bounding box centers. And "max\_x" will select the planar face which has a maximal x component compared to all other planes' bounding box centers. Note that the normal vector of the selected planes directly correlates with the side argument. 
Here are the normal vectors associated to each side argument: 
\begin{itemize}
    \item "min\_z" : (0, 0, -1) 
    \item "max\_z" : (0, 0, 1) 
    \item "min\_y" : (0, -1, 0) 
    \item "max\_y" : (0, 1, 0) 
    \item "min\_x" : (-1, 0, 0) 
    \item "max\_x" : (1, 0, 0) 
\end{itemize}
You can use these normal vectors to create more 3 dimensional objects. 

The center coordinates of sketch primitives have to be inside of the selected plane.
And there is the important constraint that you can only put the center of sketch primitives inside or on the edge of a planar face, if you do not put them on one of the default planes.

2) extrude (sketch, length), where sketch is the ID returned by the sketch operator and the length determines the length of the extrude.
Note that sketches lie within their respective plane, and they will get extruded along the plane’s normal direction. A common pattern you will encounter is that the height variable of a solid should often be the length parameter in the extrude operator. Whereas the other dimensions of the solid are defined by the sketch primitives.

Example design
For example, if you want to design a round table with a single center leg and a leg base you can do: 

\begin{gptcodeblock}{Python}
leg base legBase_sketch = createSketch(circle(0,0,0, 3), XY_PLANE) 
legBase_solid = extrude(legBase_sketch , 1) 

leg_sketch = createSketch(circle(0, 0, 1, 1), cap(legBase_solid, "max_z")) 
leg_solid = extrude(leg_sketch , 10) 

top_sketch = createSketch(circle(0,0, 11,8), cap(leg_solid, "max_z")) 
top_solid = extrude(top_sketch , 1) 
\end{gptcodeblock}

End of the example. 

Additional Constraints
Use exposed design variables whenever you can, and as few as possible. 
Write code in syntactically correct python, knowing that you have the functions createSketch, circle, rectangle, cap and extrude and the default planes. 

\end{userprompt} 

\end{chat}
\caption{\textbf{A Sketch-Based CAD DSL Prompt with a Global Coordinate System.} Our prompt used for the sketch-based CAD experiments with a global coordinate system.}
\label{fig:dsl_global}
\end{figure}

\subsection{URDF}

\gpt was able to use URDF without any intermediate libraries. Similar to OpenJSCAD, there were many common pitfalls that needed to be mitigated via prompt choice --- these are discussed in detail in section \ref{sec:textToDesign_urdf}. 

In brief summary, the following notes list some additions that were useful in mitigating specific problems:

\begin{itemize}
    \item \gpt has difficulties in determining where URDF objects place their origin. When wanting objects to touch but not intersect, or be placed at the ``end'' of other objects, it is useful to specify that the ends are half the length of the object away from the origin. 
    \item Specifying an axis for two objects to be aligned along is more effective than instructing that they be aligned.
    \item \gpt will often omit essential parts of the URDF file for brevity, replacing them with a comment to repeat a part of the file. This can be done manually, but to generate URDF files that are complete directly from the response, \gpt must be instructed to produce a complete file.
    \item \gpt will ignore several constraints or instructions if too many are placed in a single prompt. Splitting the generation process into multiple prompts resolves this issue.
\end{itemize}

\subsection{Graph-based DSL for Robotics}
\label{sec:appx_graph_robots}
The full text of the prompt used to generate the humanoid robot graph (omitted earlier for brevity) is shown in \fref{fig:appx-graph-DSL}.

\begin{figure}
\begin{chat}{false}
\begin{userprompt}{0\textwidth}{}
We are constructing robots using Python code. The following functions are available:

\begin{gptcodeblock}{Python}
add_link(name): Adds a link to the robot with the name `name` and returns its ID. add_joint(parent_link, child_link): Adds a joint between the parent link with ID `parent_link` and child link with ID `child_link`.
translate(link, direction): Translates the link with ID `link` in the direction `direction`. Direction can be one of "left", "right", "forward", "backward", "up", or "down".
\end{gptcodeblock}

Write a function to construct a humanoid robot.
\end{userprompt}

\end{chat}
\caption{\textbf{Graph-based robotics DSL.} A description of the custom graph-based DSL used to construct robots.}
\label{fig:appx-graph-DSL}
\end{figure}

%% file: main.bbl

\begin{thebibliography}{42}


\ifx \showCODEN    \undefined \def \showCODEN     #1{\unskip}     \fi
\ifx \showDOI      \undefined \def \showDOI       #1{#1}\fi
\ifx \showISBNx    \undefined \def \showISBNx     #1{\unskip}     \fi
\ifx \showISBNxiii \undefined \def \showISBNxiii  #1{\unskip}     \fi
\ifx \showISSN     \undefined \def \showISSN      #1{\unskip}     \fi
\ifx \showLCCN     \undefined \def \showLCCN      #1{\unskip}     \fi
\ifx \shownote     \undefined \def \shownote      #1{#1}          \fi
\ifx \showarticletitle \undefined \def \showarticletitle #1{#1}   \fi
\ifx \showURL      \undefined \def \showURL       {\relax}        \fi
\providecommand\bibfield[2]{#2}
\providecommand\bibinfo[2]{#2}
\providecommand\natexlab[1]{#1}
\providecommand\showeprint[2][]{arXiv:#2}

\bibitem[rep({[n.\,d.]})]%
        {repetierRepetierSoftware}
 \bibinfo{year}{[n.\,d.]}\natexlab{}.
\newblock \bibinfo{title}{- {R}epetier {S}oftware --- repetier.com}.
\newblock \bibinfo{howpublished}{\url{https://www.repetier.com/}}.
\newblock
\newblock
\shownote{[Accessed 20-Jul-2023]}.


\bibitem[fea({[n.\,d.]})]%
        {featurescript}
 \bibinfo{year}{[n.\,d.]}\natexlab{}.
\newblock \bibinfo{title}{FeatureScript introduction}.
\newblock \bibinfo{howpublished}{\url{https://cad.onshape.com/FsDoc/}}.
\newblock
\newblock
\shownote{Accessed: 2023-07-11}.


\bibitem[jsc({[n.\,d.]})]%
        {jscad}
 \bibinfo{year}{[n.\,d.]}\natexlab{}.
\newblock \bibinfo{title}{JSCAD User Guide}.
\newblock
  \bibinfo{howpublished}{\url{https://openjscad.xyz/dokuwiki/doku.php}}.
\newblock
\newblock
\shownote{Accessed: 2023-07-14}.


\bibitem[sli({[n.\,d.]})]%
        {slic3rSlic3rOpen}
 \bibinfo{year}{[n.\,d.]}\natexlab{}.
\newblock \bibinfo{title}{{S}lic3r - {O}pen source 3{D} printing toolbox ---
  slic3r.org}.
\newblock \bibinfo{howpublished}{\url{https://slic3r.org/}}.
\newblock
\newblock
\shownote{[Accessed 20-Jul-2023]}.


\bibitem[{Autodesk}({[n.\,d.]})]%
        {AutodeskSimulation}
\bibfield{author}{\bibinfo{person}{{Autodesk}}.}
  \bibinfo{year}{[n.\,d.]}\natexlab{}.
\newblock \bibinfo{title}{{Autodesk Simulation}}.
\newblock
  \bibinfo{howpublished}{\url{https://www.autodesk.com/solutions/simulation/overview}}.
\newblock
\newblock
\shownote{Accessed: July 14, 2023}.


\bibitem[Chen et~al\mbox{.}(2021)]%
        {chen2021evaluating}
\bibfield{author}{\bibinfo{person}{Mark Chen}, \bibinfo{person}{Jerry Tworek},
  \bibinfo{person}{Heewoo Jun}, \bibinfo{person}{Qiming Yuan},
  \bibinfo{person}{Henrique Ponde de~Oliveira Pinto}, \bibinfo{person}{Jared
  Kaplan}, \bibinfo{person}{Harri Edwards}, \bibinfo{person}{Yuri Burda},
  \bibinfo{person}{Nicholas Joseph}, \bibinfo{person}{Greg Brockman},
  {et~al\mbox{.}}} \bibinfo{year}{2021}\natexlab{}.
\newblock \showarticletitle{Evaluating large language models trained on code}.
\newblock \bibinfo{journal}{\emph{arXiv preprint arXiv:2107.03374}}
  (\bibinfo{year}{2021}).
\newblock


\bibitem[Christiano et~al\mbox{.}(2017)]%
        {christiano2017deep}
\bibfield{author}{\bibinfo{person}{Paul~F Christiano}, \bibinfo{person}{Jan
  Leike}, \bibinfo{person}{Tom Brown}, \bibinfo{person}{Miljan Martic},
  \bibinfo{person}{Shane Legg}, {and} \bibinfo{person}{Dario Amodei}.}
  \bibinfo{year}{2017}\natexlab{}.
\newblock \showarticletitle{Deep reinforcement learning from human
  preferences}.
\newblock \bibinfo{journal}{\emph{Advances in neural information processing
  systems}}  \bibinfo{volume}{30} (\bibinfo{year}{2017}).
\newblock


\bibitem[{Dassault Systèmes}({[n.\,d.]})]%
        {DassaultSimulation}
\bibfield{author}{\bibinfo{person}{{Dassault Systèmes}}.}
  \bibinfo{year}{[n.\,d.]}\natexlab{}.
\newblock \bibinfo{title}{{Dassault Systèmes Simulation}}.
\newblock
  \bibinfo{howpublished}{\url{https://www.3ds.com/products-services/simulia/overview/}}.
\newblock
\newblock
\shownote{Accessed: July 14, 2023}.


\bibitem[Dhariwal et~al\mbox{.}(2020)]%
        {dhariwal2020jukebox}
\bibfield{author}{\bibinfo{person}{Prafulla Dhariwal}, \bibinfo{person}{Heewoo
  Jun}, \bibinfo{person}{Christine Payne}, \bibinfo{person}{Jong~Wook Kim},
  \bibinfo{person}{Alec Radford}, {and} \bibinfo{person}{Ilya Sutskever}.}
  \bibinfo{year}{2020}\natexlab{}.
\newblock \showarticletitle{Jukebox: A generative model for music}.
\newblock \bibinfo{journal}{\emph{arXiv preprint arXiv:2005.00341}}
  (\bibinfo{year}{2020}).
\newblock


\bibitem[Du et~al\mbox{.}(2018)]%
        {du2018inversecsg}
\bibfield{author}{\bibinfo{person}{Tao Du}, \bibinfo{person}{Jeevana~Priya
  Inala}, \bibinfo{person}{Yewen Pu}, \bibinfo{person}{Andrew Spielberg},
  \bibinfo{person}{Adriana Schulz}, \bibinfo{person}{Daniela Rus},
  \bibinfo{person}{Armando Solar-Lezama}, {and} \bibinfo{person}{Wojciech
  Matusik}.} \bibinfo{year}{2018}\natexlab{}.
\newblock \showarticletitle{Inversecsg: Automatic conversion of 3d models to
  csg trees}.
\newblock \bibinfo{journal}{\emph{ACM Transactions on Graphics (TOG)}}
  \bibinfo{volume}{37}, \bibinfo{number}{6} (\bibinfo{year}{2018}),
  \bibinfo{pages}{1--16}.
\newblock


\bibitem[Du et~al\mbox{.}(2021)]%
        {du2021diffpd}
\bibfield{author}{\bibinfo{person}{Tao Du}, \bibinfo{person}{Kui Wu},
  \bibinfo{person}{Pingchuan Ma}, \bibinfo{person}{Sebastien Wah},
  \bibinfo{person}{Andrew Spielberg}, \bibinfo{person}{Daniela Rus}, {and}
  \bibinfo{person}{Wojciech Matusik}.} \bibinfo{year}{2021}\natexlab{}.
\newblock \showarticletitle{DiffPD: Differentiable projective dynamics}.
\newblock \bibinfo{journal}{\emph{ACM Transactions on Graphics (TOG)}}
  \bibinfo{volume}{41}, \bibinfo{number}{2} (\bibinfo{year}{2021}),
  \bibinfo{pages}{1--21}.
\newblock


\bibitem[Erez et~al\mbox{.}(2015)]%
        {erez2015simulation}
\bibfield{author}{\bibinfo{person}{Tom Erez}, \bibinfo{person}{Yuval Tassa},
  {and} \bibinfo{person}{Emanuel Todorov}.} \bibinfo{year}{2015}\natexlab{}.
\newblock \showarticletitle{Simulation tools for model-based robotics:
  Comparison of bullet, havok, mujoco, ode and physx}. In
  \bibinfo{booktitle}{\emph{2015 IEEE international conference on robotics and
  automation (ICRA)}}. IEEE, \bibinfo{pages}{4397--4404}.
\newblock


\bibitem[Erps et~al\mbox{.}(2021)]%
        {erps2021processOpt}
\bibfield{author}{\bibinfo{person}{Timothy Erps}, \bibinfo{person}{Michael
  Foshey}, \bibinfo{person}{Mina~Konaković Luković}, \bibinfo{person}{Wan
  Shou}, \bibinfo{person}{Hanns~Hagen Goetzke}, \bibinfo{person}{Herve
  Dietsch}, \bibinfo{person}{Klaus Stoll}, \bibinfo{person}{Bernhard von
  Vacano}, {and} \bibinfo{person}{Wojciech Matusik}.}
  \bibinfo{year}{2021}\natexlab{}.
\newblock \showarticletitle{Accelerated discovery of 3D printing materials
  using data-driven multiobjective optimization}.
\newblock \bibinfo{journal}{\emph{Science Advances}} \bibinfo{volume}{7},
  \bibinfo{number}{42} (\bibinfo{year}{2021}), \bibinfo{pages}{eabf7435}.
\newblock


\bibitem[Ferruz et~al\mbox{.}(2022)]%
        {ferruz2022protgpt2}
\bibfield{author}{\bibinfo{person}{Noelia Ferruz}, \bibinfo{person}{Steffen
  Schmidt}, {and} \bibinfo{person}{Birte H{\"o}cker}.}
  \bibinfo{year}{2022}\natexlab{}.
\newblock \showarticletitle{ProtGPT2 is a deep unsupervised language model for
  protein design}.
\newblock \bibinfo{journal}{\emph{Nature communications}} \bibinfo{volume}{13},
  \bibinfo{number}{1} (\bibinfo{year}{2022}), \bibinfo{pages}{4348}.
\newblock


\bibitem[Guo et~al\mbox{.}(2022)]%
        {guo2022data}
\bibfield{author}{\bibinfo{person}{Minghao Guo}, \bibinfo{person}{Veronika
  Thost}, \bibinfo{person}{Beichen Li}, \bibinfo{person}{Payel Das},
  \bibinfo{person}{Jie Chen}, {and} \bibinfo{person}{Wojciech Matusik}.}
  \bibinfo{year}{2022}\natexlab{}.
\newblock \showarticletitle{Data-efficient graph grammar learning for molecular
  generation}.
\newblock \bibinfo{journal}{\emph{arXiv preprint arXiv:2203.08031}}
  (\bibinfo{year}{2022}).
\newblock


\bibitem[Jiang et~al\mbox{.}(2023)]%
        {jiang2023motiongpt}
\bibfield{author}{\bibinfo{person}{Biao Jiang}, \bibinfo{person}{Xin Chen},
  \bibinfo{person}{Wen Liu}, \bibinfo{person}{Jingyi Yu}, \bibinfo{person}{Gang
  Yu}, {and} \bibinfo{person}{Tao Chen}.} \bibinfo{year}{2023}\natexlab{}.
\newblock \showarticletitle{MotionGPT: Human Motion as a Foreign Language}.
\newblock \bibinfo{journal}{\emph{arXiv preprint arXiv:2306.14795}}
  (\bibinfo{year}{2023}).
\newblock


\bibitem[Kashefi and Mukerji(2023)]%
        {kashefi2023chatgpt}
\bibfield{author}{\bibinfo{person}{Ali Kashefi} {and} \bibinfo{person}{Tapan
  Mukerji}.} \bibinfo{year}{2023}\natexlab{}.
\newblock \showarticletitle{Chatgpt for programming numerical methods}.
\newblock \bibinfo{journal}{\emph{Journal of Machine Learning for Modeling and
  Computing}} \bibinfo{volume}{4}, \bibinfo{number}{2} (\bibinfo{year}{2023}).
\newblock


\bibitem[Koo et~al\mbox{.}(2017)]%
        {Koo2017zeroWaste}
\bibfield{author}{\bibinfo{person}{Bongjin Koo}, \bibinfo{person}{Jean Hergel},
  \bibinfo{person}{Sylvain Lefebvre}, {and} \bibinfo{person}{Niloy~J. Mitra}.}
  \bibinfo{year}{2017}\natexlab{}.
\newblock \showarticletitle{Towards Zero-Waste Furniture Design}.
\newblock \bibinfo{journal}{\emph{IEEE Transactions on Visualization and
  Computer Graphics}} \bibinfo{volume}{23}, \bibinfo{number}{12}
  (\bibinfo{year}{2017}), \bibinfo{pages}{2627--2640}.
\newblock
\urldef\tempurl%
\url{https://doi.org/10.1109/TVCG.2016.2633519}
\showDOI{\tempurl}


\bibitem[Liu et~al\mbox{.}(2023)]%
        {liu2023zero}
\bibfield{author}{\bibinfo{person}{Ruoshi Liu}, \bibinfo{person}{Rundi Wu},
  \bibinfo{person}{Basile Van~Hoorick}, \bibinfo{person}{Pavel Tokmakov},
  \bibinfo{person}{Sergey Zakharov}, {and} \bibinfo{person}{Carl Vondrick}.}
  \bibinfo{year}{2023}\natexlab{}.
\newblock \showarticletitle{Zero-1-to-3: Zero-shot one image to 3d object}.
\newblock \bibinfo{journal}{\emph{arXiv preprint arXiv:2303.11328}}
  (\bibinfo{year}{2023}).
\newblock


\bibitem[Ma et~al\mbox{.}(2021)]%
        {ma2021diffaqua}
\bibfield{author}{\bibinfo{person}{Pingchuan Ma}, \bibinfo{person}{Tao Du},
  \bibinfo{person}{John~Z Zhang}, \bibinfo{person}{Kui Wu},
  \bibinfo{person}{Andrew Spielberg}, \bibinfo{person}{Robert~K Katzschmann},
  {and} \bibinfo{person}{Wojciech Matusik}.} \bibinfo{year}{2021}\natexlab{}.
\newblock \showarticletitle{Diffaqua: A differentiable computational design
  pipeline for soft underwater swimmers with shape interpolation}.
\newblock \bibinfo{journal}{\emph{ACM Transactions on Graphics (TOG)}}
  \bibinfo{volume}{40}, \bibinfo{number}{4} (\bibinfo{year}{2021}),
  \bibinfo{pages}{1--14}.
\newblock


\bibitem[Makatura et~al\mbox{.}(2023)]%
        {makatura2023procedural}
\bibfield{author}{\bibinfo{person}{Liane Makatura}, \bibinfo{person}{Bohan
  Wang}, \bibinfo{person}{Yi-Lu Chen}, \bibinfo{person}{Bolei Deng},
  \bibinfo{person}{Chris Wojtan}, \bibinfo{person}{Bernd Bickel}, {and}
  \bibinfo{person}{Wojciech Matusik}.} \bibinfo{year}{2023}\natexlab{}.
\newblock \showarticletitle{Procedural Metamaterials: A Unified Procedural
  Graph for Metamaterial Design}.
\newblock \bibinfo{journal}{\emph{ACM Transactions on Graphics}}
  (\bibinfo{year}{2023}).
\newblock


\bibitem[Mathur and Zufferey(2021)]%
        {mathur2021constraint}
\bibfield{author}{\bibinfo{person}{Aman Mathur} {and} \bibinfo{person}{Damien
  Zufferey}.} \bibinfo{year}{2021}\natexlab{}.
\newblock \showarticletitle{Constraint Synthesis for Parametric CAD}.
\newblock  (\bibinfo{year}{2021}).
\newblock


\bibitem[Mirchandani et~al\mbox{.}(2023)]%
        {mirchandani2023large}
\bibfield{author}{\bibinfo{person}{Suvir Mirchandani}, \bibinfo{person}{Fei
  Xia}, \bibinfo{person}{Pete Florence}, \bibinfo{person}{Brian Ichter},
  \bibinfo{person}{Danny Driess}, \bibinfo{person}{Montserrat~Gonzalez Arenas},
  \bibinfo{person}{Kanishka Rao}, \bibinfo{person}{Dorsa Sadigh}, {and}
  \bibinfo{person}{Andy Zeng}.} \bibinfo{year}{2023}\natexlab{}.
\newblock \showarticletitle{Large Language Models as General Pattern Machines}.
\newblock \bibinfo{journal}{\emph{arXiv preprint arXiv:2307.04721}}
  (\bibinfo{year}{2023}).
\newblock


\bibitem[M{\"u}ller et~al\mbox{.}(2006)]%
        {muller2006procedural}
\bibfield{author}{\bibinfo{person}{Pascal M{\"u}ller}, \bibinfo{person}{Peter
  Wonka}, \bibinfo{person}{Simon Haegler}, \bibinfo{person}{Andreas Ulmer},
  {and} \bibinfo{person}{Luc Van~Gool}.} \bibinfo{year}{2006}\natexlab{}.
\newblock \showarticletitle{Procedural modeling of buildings}.
\newblock In \bibinfo{booktitle}{\emph{ACM SIGGRAPH 2006 Papers}}.
  \bibinfo{pages}{614--623}.
\newblock


\bibitem[O'Brien et~al\mbox{.}(2002)]%
        {o2002synthesizing}
\bibfield{author}{\bibinfo{person}{James~F O'Brien}, \bibinfo{person}{Chen
  Shen}, {and} \bibinfo{person}{Christine~M Gatchalian}.}
  \bibinfo{year}{2002}\natexlab{}.
\newblock \showarticletitle{Synthesizing sounds from rigid-body simulations}.
  In \bibinfo{booktitle}{\emph{Proceedings of the 2002 ACM
  SIGGRAPH/Eurographics symposium on Computer animation}}.
  \bibinfo{pages}{175--181}.
\newblock


\bibitem[OpenAI(2023)]%
        {openai2023gpt4}
\bibfield{author}{\bibinfo{person}{OpenAI}.} \bibinfo{year}{2023}\natexlab{}.
\newblock \bibinfo{title}{GPT-4 Technical Report}.
\newblock
\newblock
\showeprint[arxiv]{2303.08774}~[cs.CL]


\bibitem[Ouyang et~al\mbox{.}(2022)]%
        {ouyang2022training}
\bibfield{author}{\bibinfo{person}{Long Ouyang}, \bibinfo{person}{Jeffrey Wu},
  \bibinfo{person}{Xu Jiang}, \bibinfo{person}{Diogo Almeida},
  \bibinfo{person}{Carroll Wainwright}, \bibinfo{person}{Pamela Mishkin},
  \bibinfo{person}{Chong Zhang}, \bibinfo{person}{Sandhini Agarwal},
  \bibinfo{person}{Katarina Slama}, \bibinfo{person}{Alex Ray},
  {et~al\mbox{.}}} \bibinfo{year}{2022}\natexlab{}.
\newblock \showarticletitle{Training language models to follow instructions
  with human feedback}.
\newblock \bibinfo{journal}{\emph{Advances in Neural Information Processing
  Systems}}  \bibinfo{volume}{35} (\bibinfo{year}{2022}),
  \bibinfo{pages}{27730--27744}.
\newblock


\bibitem[{\"O}zkar and Stiny(2009)]%
        {ozkar2009shape}
\bibfield{author}{\bibinfo{person}{Mine {\"O}zkar} {and}
  \bibinfo{person}{George Stiny}.} \bibinfo{year}{2009}\natexlab{}.
\newblock \showarticletitle{Shape grammars}.
\newblock In \bibinfo{booktitle}{\emph{Acm Siggraph 2009 Courses}}.
  \bibinfo{pages}{1--176}.
\newblock


\bibitem[Penedo et~al\mbox{.}(2023)]%
        {penedo2023refinedweb}
\bibfield{author}{\bibinfo{person}{Guilherme Penedo}, \bibinfo{person}{Quentin
  Malartic}, \bibinfo{person}{Daniel Hesslow}, \bibinfo{person}{Ruxandra
  Cojocaru}, \bibinfo{person}{Alessandro Cappelli}, \bibinfo{person}{Hamza
  Alobeidli}, \bibinfo{person}{Baptiste Pannier}, \bibinfo{person}{Ebtesam
  Almazrouei}, {and} \bibinfo{person}{Julien Launay}.}
  \bibinfo{year}{2023}\natexlab{}.
\newblock \showarticletitle{The RefinedWeb dataset for Falcon LLM:
  outperforming curated corpora with web data, and web data only}.
\newblock \bibinfo{journal}{\emph{arXiv preprint arXiv:2306.01116}}
  (\bibinfo{year}{2023}).
\newblock


\bibitem[Prusinkiewicz and Lindenmayer(2012)]%
        {prusinkiewicz2012algorithmic}
\bibfield{author}{\bibinfo{person}{Przemyslaw Prusinkiewicz} {and}
  \bibinfo{person}{Aristid Lindenmayer}.} \bibinfo{year}{2012}\natexlab{}.
\newblock \bibinfo{booktitle}{\emph{The algorithmic beauty of plants}}.
\newblock \bibinfo{publisher}{Springer Science \& Business Media}.
\newblock


\bibitem[Radford et~al\mbox{.}(2019)]%
        {radford2019language}
\bibfield{author}{\bibinfo{person}{Alec Radford}, \bibinfo{person}{Jeffrey Wu},
  \bibinfo{person}{Rewon Child}, \bibinfo{person}{David Luan},
  \bibinfo{person}{Dario Amodei}, \bibinfo{person}{Ilya Sutskever},
  {et~al\mbox{.}}} \bibinfo{year}{2019}\natexlab{}.
\newblock \showarticletitle{Language models are unsupervised multitask
  learners}.
\newblock \bibinfo{journal}{\emph{OpenAI blog}} \bibinfo{volume}{1},
  \bibinfo{number}{8} (\bibinfo{year}{2019}), \bibinfo{pages}{9}.
\newblock


\bibitem[Ramesh et~al\mbox{.}(2021)]%
        {ramesh2021zero}
\bibfield{author}{\bibinfo{person}{Aditya Ramesh}, \bibinfo{person}{Mikhail
  Pavlov}, \bibinfo{person}{Gabriel Goh}, \bibinfo{person}{Scott Gray},
  \bibinfo{person}{Chelsea Voss}, \bibinfo{person}{Alec Radford},
  \bibinfo{person}{Mark Chen}, {and} \bibinfo{person}{Ilya Sutskever}.}
  \bibinfo{year}{2021}\natexlab{}.
\newblock \showarticletitle{Zero-shot text-to-image generation}. In
  \bibinfo{booktitle}{\emph{International Conference on Machine Learning}}.
  PMLR, \bibinfo{pages}{8821--8831}.
\newblock


\bibitem[Rozenberg and Salomaa(1980)]%
        {rozenberg1980mathematical}
\bibfield{author}{\bibinfo{person}{Grzegorz Rozenberg} {and}
  \bibinfo{person}{Arto Salomaa}.} \bibinfo{year}{1980}\natexlab{}.
\newblock \bibinfo{booktitle}{\emph{The mathematical theory of L systems}}.
\newblock \bibinfo{publisher}{Academic press}.
\newblock


\bibitem[Stiny(1980)]%
        {stiny1980introduction}
\bibfield{author}{\bibinfo{person}{George Stiny}.}
  \bibinfo{year}{1980}\natexlab{}.
\newblock \showarticletitle{Introduction to shape and shape grammars}.
\newblock \bibinfo{journal}{\emph{Environment and planning B: planning and
  design}} \bibinfo{volume}{7}, \bibinfo{number}{3} (\bibinfo{year}{1980}),
  \bibinfo{pages}{343--351}.
\newblock


\bibitem[Sullivan(2013)]%
        {sullivan2013electromagnetic}
\bibfield{author}{\bibinfo{person}{Dennis~M Sullivan}.}
  \bibinfo{year}{2013}\natexlab{}.
\newblock \bibinfo{booktitle}{\emph{Electromagnetic simulation using the FDTD
  method}}.
\newblock \bibinfo{publisher}{John Wiley \& Sons}.
\newblock


\bibitem[Touvron et~al\mbox{.}(2023)]%
        {touvron2023llama}
\bibfield{author}{\bibinfo{person}{Hugo Touvron}, \bibinfo{person}{Thibaut
  Lavril}, \bibinfo{person}{Gautier Izacard}, \bibinfo{person}{Xavier
  Martinet}, \bibinfo{person}{Marie-Anne Lachaux},
  \bibinfo{person}{Timoth{\'e}e Lacroix}, \bibinfo{person}{Baptiste
  Rozi{\`e}re}, \bibinfo{person}{Naman Goyal}, \bibinfo{person}{Eric Hambro},
  \bibinfo{person}{Faisal Azhar}, {et~al\mbox{.}}}
  \bibinfo{year}{2023}\natexlab{}.
\newblock \showarticletitle{Llama: Open and efficient foundation language
  models}.
\newblock \bibinfo{journal}{\emph{arXiv preprint arXiv:2302.13971}}
  (\bibinfo{year}{2023}).
\newblock


\bibitem[Turing et~al\mbox{.}(1936)]%
        {turing1936computable}
\bibfield{author}{\bibinfo{person}{Alan~Mathison Turing} {et~al\mbox{.}}}
  \bibinfo{year}{1936}\natexlab{}.
\newblock \showarticletitle{On computable numbers, with an application to the
  Entscheidungsproblem}.
\newblock \bibinfo{journal}{\emph{J. of Math}} \bibinfo{volume}{58},
  \bibinfo{number}{345-363} (\bibinfo{year}{1936}), \bibinfo{pages}{5}.
\newblock


\bibitem[Willis et~al\mbox{.}(2021)]%
        {willis2021fusion}
\bibfield{author}{\bibinfo{person}{Karl~DD Willis}, \bibinfo{person}{Yewen Pu},
  \bibinfo{person}{Jieliang Luo}, \bibinfo{person}{Hang Chu},
  \bibinfo{person}{Tao Du}, \bibinfo{person}{Joseph~G Lambourne},
  \bibinfo{person}{Armando Solar-Lezama}, {and} \bibinfo{person}{Wojciech
  Matusik}.} \bibinfo{year}{2021}\natexlab{}.
\newblock \showarticletitle{Fusion 360 gallery: A dataset and environment for
  programmatic cad construction from human design sequences}.
\newblock \bibinfo{journal}{\emph{ACM Transactions on Graphics (TOG)}}
  \bibinfo{volume}{40}, \bibinfo{number}{4} (\bibinfo{year}{2021}),
  \bibinfo{pages}{1--24}.
\newblock


\bibitem[Xu et~al\mbox{.}(2021)]%
        {xu2021end}
\bibfield{author}{\bibinfo{person}{Jie Xu}, \bibinfo{person}{Tao Chen},
  \bibinfo{person}{Lara Zlokapa}, \bibinfo{person}{Michael Foshey},
  \bibinfo{person}{Wojciech Matusik}, \bibinfo{person}{Shinjiro Sueda}, {and}
  \bibinfo{person}{Pulkit Agrawal}.} \bibinfo{year}{2021}\natexlab{}.
\newblock \showarticletitle{An end-to-end differentiable framework for
  contact-aware robot design}.
\newblock \bibinfo{journal}{\emph{arXiv preprint arXiv:2107.07501}}
  (\bibinfo{year}{2021}).
\newblock


\bibitem[Zhang et~al\mbox{.}(2018)]%
        {zhang2018graphit}
\bibfield{author}{\bibinfo{person}{Yunming Zhang}, \bibinfo{person}{Mengjiao
  Yang}, \bibinfo{person}{Riyadh Baghdadi}, \bibinfo{person}{Shoaib Kamil},
  \bibinfo{person}{Julian Shun}, {and} \bibinfo{person}{Saman Amarasinghe}.}
  \bibinfo{year}{2018}\natexlab{}.
\newblock \showarticletitle{Graphit: A high-performance graph dsl}.
\newblock \bibinfo{journal}{\emph{Proceedings of the ACM on Programming
  Languages}} \bibinfo{volume}{2}, \bibinfo{number}{OOPSLA}
  (\bibinfo{year}{2018}), \bibinfo{pages}{1--30}.
\newblock


\bibitem[Zhao et~al\mbox{.}(2020)]%
        {zhao2020robogrammar}
\bibfield{author}{\bibinfo{person}{Allan Zhao}, \bibinfo{person}{Jie Xu},
  \bibinfo{person}{Mina Konakovi{\'c}-Lukovi{\'c}}, \bibinfo{person}{Josephine
  Hughes}, \bibinfo{person}{Andrew Spielberg}, \bibinfo{person}{Daniela Rus},
  {and} \bibinfo{person}{Wojciech Matusik}.} \bibinfo{year}{2020}\natexlab{}.
\newblock \showarticletitle{Robogrammar: graph grammar for terrain-optimized
  robot design}.
\newblock \bibinfo{journal}{\emph{ACM Transactions on Graphics (TOG)}}
  \bibinfo{volume}{39}, \bibinfo{number}{6} (\bibinfo{year}{2020}),
  \bibinfo{pages}{1--16}.
\newblock


\bibitem[Zilberstein(1996)]%
        {zilberstein1996using}
\bibfield{author}{\bibinfo{person}{Shlomo Zilberstein}.}
  \bibinfo{year}{1996}\natexlab{}.
\newblock \showarticletitle{Using anytime algorithms in intelligent systems}.
\newblock \bibinfo{journal}{\emph{AI magazine}} \bibinfo{volume}{17},
  \bibinfo{number}{3} (\bibinfo{year}{1996}), \bibinfo{pages}{73--73}.
\newblock


\end{thebibliography}
